\documentclass[11pt]{report}
\usepackage{upennstyle} 

\usepackage{import}
\usepackage{hyperref}       
\usepackage{url}            
\usepackage{booktabs}       
\usepackage{amsfonts}       
\usepackage{nicefrac}       
\usepackage{xcolor}         
\usepackage{graphicx}       
\usepackage{subcaption}     
\usepackage{textcomp}
\usepackage{amsmath}
\usepackage{amssymb}
\usepackage{multirow}
\usepackage{siunitx}
\usepackage{svg}
\usepackage{soul}
\usepackage{tikz}
\usetikzlibrary{shapes.misc}
\usetikzlibrary{arrows.meta}
\tikzset{cross/.style={cross out, draw=black, minimum size=2*(#1-\pgflinewidth), inner sep=0pt, outer sep=0pt},
cross/.default={4.5pt}}
\usepackage{enumitem}
\usepackage{algorithm}
\usepackage[noend]{algpseudocode}

\usetikzlibrary{shapes,arrows,positioning,fit,calc,backgrounds,shapes.geometric}

\definecolor{babyblue}{rgb}{0.06,0.58,0.97}
\definecolor{salmon}{rgb}{0.98,0.41,0.38}
\definecolor{darkergreen}{RGB}{0,170,0} 

\newcommand{\red}[1]{{\color{red}#1}}
\newcommand{\blue}[1]{{\color{blue}#1}}
\newcommand{\green}[1]{{\color{darkergreen}#1}}

\newcommand{\figref}[1]{Figure~\ref{fig:#1}}

\newcommand{\figrefs}[2]{Figures~\ref{fig:#1}--\ref{fig:#2}}
\newcommand{\figlabel}[1]{\label{fig:#1}} 
\newcommand{\tableref}[1]{Table~\ref{table:#1}}

\newcommand{\tablelabel}[1]{\label{table:#1}}
\newcommand{\algoref}[1]{Algorithm~\ref{algo:#1}}
\newcommand{\algolabel}[1]{\label{algo:#1}}
\newcommand{\chapterref}[1]{Chapter~\ref{chapter:#1}}
\newcommand{\chapterlabel}[1]{\label{chapter:#1}}
\newcommand{\sectionref}[1]{Section~\ref{section:#1}}
\newcommand{\sectionlabel}[1]{\label{section:#1}}
\newcommand{\appendixref}[1]{Section~\ref{appendix:#1}}
\newcommand{\appendixlabel}[1]{\label{appendix:#1}}
\newcommand{\equationref}[1]{Equation~\ref{equation:#1}}
\newcommand{\equationlabel}[1]{\label{equation:#1}}

\def\eqref#1{equation~\ref{#1}}

\def\1{\bm{1}}

\newcommand{\pointcloud}{P}
\newcommand{\pointcloudt}{P_t}
\newcommand{\pointcloudsub}[1]{P_{#1}}
\newcommand{\pointcloudtpone}{P_{t+1}}
\newcommand{\pointcloudtmone}{P_{t-1}}

\newcommand{\flow}{\hat{\mathcal{F}}}
\newcommand{\flowgt}{\mathcal{F}}

\newcommand{\flowttpone}{\flow_{t,t+1}}

\newcommand{\flowgtttpone}{\flowgt_{t,t+1}}

\newcommand{\poorparagraph}[1]{\textbf{{#1}.}}
\newcommand{\humanlabels}[1]{\underline{{#1}}}
\newcommand{\bgscale}[1]{\sigma(#1)}
\newcommand{\pointspeed}[1]{s(#1)}

\newcommand{\norm}[1]{\left\lVert #1 \right\rVert}
\newcommand{\flowforward}{\flow^+}
\newcommand{\flowrev}{\flow^-}
\newcommand{\chamferdistancenameraw}{TruncatedChamfer}
\newcommand{\chamferdistancename}{\textup{\chamferdistancenameraw{}}}
\newcommand{\chamferdistance}[2]{\chamferdistancename(#1, #2)}

\newcommand{\bevmaintextfontsize}{\fontsize{5pt}{5pt}\selectfont}
\DeclareCaptionFont{bevmaintextfont}{\bevmaintextfontsize}

\newcommand{\resulttablefontsize}{\fontsize{6.3pt}{6.3pt}\selectfont}
\DeclareCaptionFont{resulttablefont}{\resulttablefontsize}

\newcommand{\network}{\theta}

\newcommand{\dirforward}{\texttt{FWD}}
\newcommand{\dirbackward}{\texttt{BWD}}

\newcommand{\kstepname}{\textup{Euler}}
\newcommand{\kstep}[2]{\kstepname_{\network}\left(#1, #2\right)}

\makeatletter
\DeclareRobustCommand{\iscircle}{\mathord{\mathpalette\is@circle\relax}}
\newcommand\is@circle[2]{%
  \begingroup
  \sbox\z@{\raisebox{\depth}{$\m@th#1\bigcirc$}}%
  \sbox\tw@{$#1\square$}%
  \resizebox{!}{\ht\tw@}{\usebox{\z@}}%
  \endgroup
}
\makeatother

\newcommand{\tikzcircle}[2][black,fill=black]{\tikz[baseline=-0.5ex]\draw[#1,radius=#2] (0,0.03) circle ;}%

\newcommand{\tikzx}[2][black,fill=black]{\tikz[baseline=-0.5ex]\draw[#1] (0,0.03)  node[cross,rotate=0, line width=1.2pt] {}; }%

\usepackage[nonamebreak,round]{natbib}
\title{\MakeUppercase{Toward Scalable, Flexible Scene Flow for Point Clouds}}
\author{Kyle Vedder}
\gradgroup{Computer and Information Science} 
\date{2025} 
\supervisor{Eric Eaton}
\supervisortitle{Research Associate Professor of Computer and Information Science}
\gradchair{Anindya De, Assistant Professor of Computer and Information Science}
\committee{Dinesh Jayaraman, Assistant Professor of Computer and Information Science}
\committee{Kostas Daniilidis, Ruth Yalom Stone Professor of Computer and Information Science}
\committee{Camillo J. Taylor, Raymond S. Markowitz President's Distinguished Professor of Computer and Information Science}
\committee{Deva Ramanan, Professor at the Robotics Institute at Carnegie Mellon University}

\authorlegal{Kyle Clark Vedder} 

\dedication{Dedicated to the last generation of humans who must physically labor for their material needs.} 

\acknowledgement{I owe a lot of people for a lot of things. 

Most obviously, I am forever indebted to Ishan Khatri and Neehar Peri; I would not have finished any of this work without my collaboration with them. They were there every step of the way, and I am grateful for their support. I'm also deeply thankful for the larger Argo AI research family for their insights and strongly overlapping interests --- James Hays for taking me on as his research intern, Deva Ramanan for helping with framing and guidance through all of these projects, Nate Chodosh for his pioneering work in lidar scene flow. I'm also thankful for the people at Penn, who I appreciate for their facilitation of my work --- Dinesh Jayaraman for his guidance, both technical and non-technical; Eric Eaton, for admitting me and allowing me to work on the topics that I found interesting; Marcel Hussing and Long Le for the many interesting technical discussions.

I also would never have arrived here without my prior educational experiences. Thank you to Joydeep Biswas for the multiple years of hands-on robotics experience that have grounded me in focusing on real problems in the real world. Thank you to Arjun Guha, without whom I would never have been admitted to grad school. Thank you to Amy Prior, for teaching calculus in a way that's exciting and made me want to learn, and Norma Chico for making physics exciting.

I also stand on the shoulders of giants. Computer Science as a field has been upended over the last decade due to the deep learning revolution, built on the backs of foundational work by many great scientists including Hinton, Lecun, Bengio, and Schmidhuber. But it's not just my field that allows me to do what I do today; I, like many people, owe a great deal to the civilizational progress since the industrial revolution. We have abundant food, clean drinking water, cheap and highly available electricity, a wondrously complex computer manufacturing industry.

I believe end-to-end automation of resource production is inevitable, ushering in a new era for humanity. My hope is that my life's work may pay these gifts forward by accelerating this progress.} 

\abstract{Scene flow estimation is the task of describing 3D motion between temporally successive observations. This dissertation aims to build the foundation for building scene flow estimators with two important properties: they are \emph{scalable}, i.e.\ they improve with access to more data and computation, and they are \emph{flexible}, i.e.\ they work out-of-the-box in a variety of domains and on a variety of motion patterns without requiring significant hyperparameter tuning.

To do this, we address critical limitations of existing methods and evaluation protocols in the field. First, we propose ZeroFlow, a scalable and fully unsupervised approach that leverages the strengths of test-time optimization to generate high-quality pseudo-labels, which are then used to efficiently train feedforward networks. This distillation pipeline significantly improves computational efficiency, achieving state-of-the-art accuracy at orders of magnitude faster inference speeds without reliance on costly human annotations.

Next, we identify a systematic shortcoming in standard evaluation metrics, revealing that prior scene flow methods consistently fail to capture the motion of small or slowly moving objects, such as pedestrians or cyclists. To address this, we introduce Bucket Normalized Endpoint Error, a new class-aware and speed-normalized evaluation protocol that provides a more accurate and informative measure of estimator quality, particularly emphasizing performance on critical smaller objects.

We demonstrate the efficacy of our new evaluation by proposing TrackFlow, a surprisingly simple yet effective baseline that leverages the performance of a high-quality 3D detector. TrackFlow barely achieves state-of-the-art performance on existing metrics, but on our improved evaluation metric it becomes clear that TrackFlow is far stronger than other competitors.

Finally, we propose EulerFlow, an unsupervised method that significantly redefines scene flow estimation by estimating an ordinary differential equation over an entire sequence of observations, rather than just two successive observations. EulerFlow provides very strong flow estimates across diverse scenarios, and its simple ODE formulation works out-of-the-box on new domains and enables emergent capabilities including long-horizon 3D point tracking.

Collectively, these contributions represent significant advancements toward scalable, flexible, and robust scene flow estimation, laying groundwork for future research and practical deployments in motion understanding across a wide range of applications, from autonomous vehicles to robotics.} 

\begin{document}
\maketitle 
\setcounter{page}{2}

\makecopyright 

\makededication 

\makeacknowledgement 

\makeabstract
\tableofcontents




\begin{mainf} 

\chapter{\MakeUppercase{Introduction}}\chapterlabel{introduction}

This dissertation is focused on the problem of scene flow for point cloud representations. This may seem a peculiar and oddly specific problem formulation and input representation choice, but this setup is just one way to approach the far more general overarching goal of finding and scaling the right recipes to make computer systems that understand motion in the 3D world.

\section{Why Should We Care About Motion Understanding?}\sectionlabel{motivation}

\begin{figure}[htb]
    \centering
    \includegraphics[width=\textwidth]{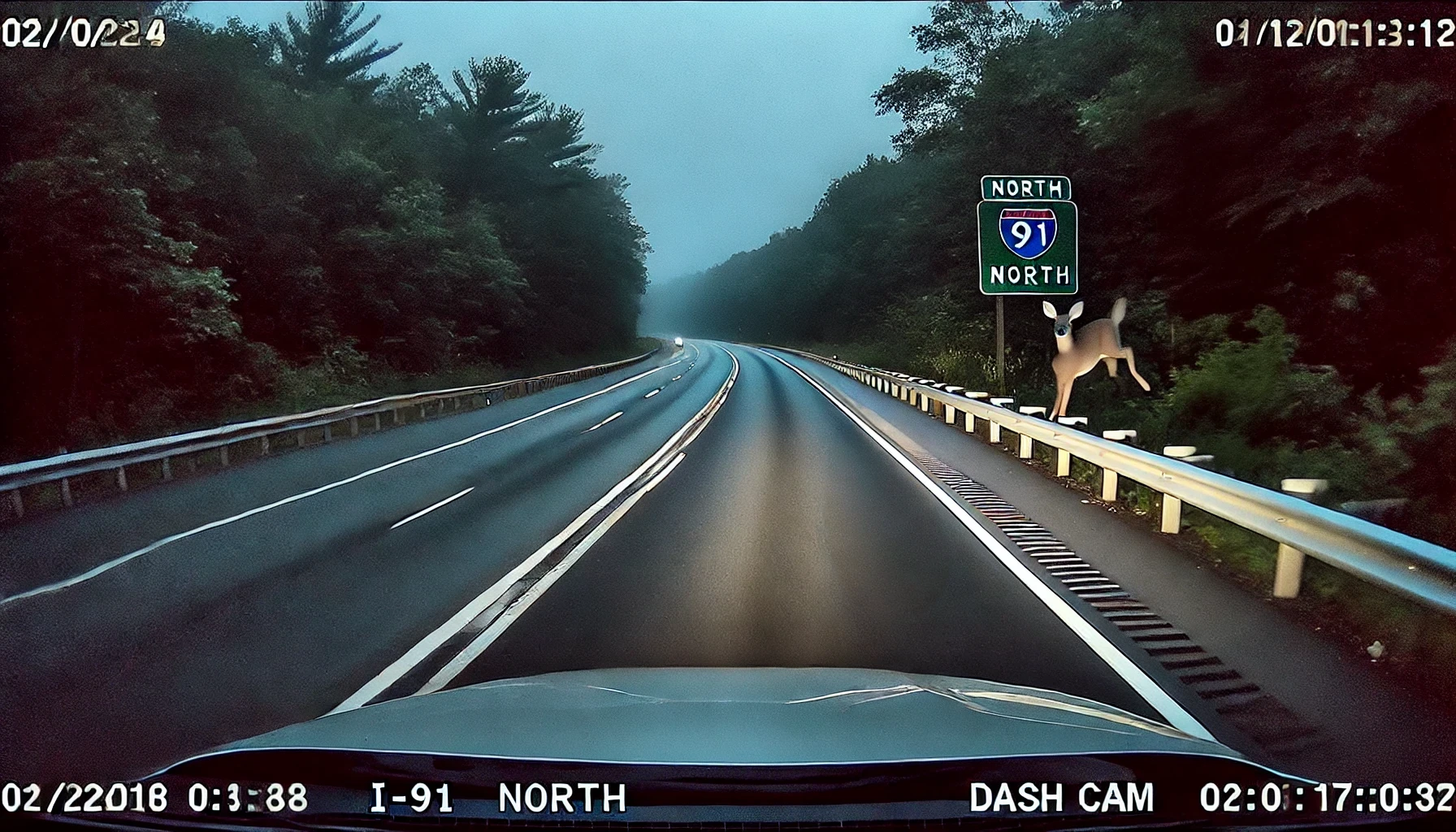}
    \caption{A deer leaping into the road, as imagined by DALL-E 3.}
    \label{fig:enter-label}
\end{figure}

It's just before dusk, and you're driving your car in the right lane up I-91 North along the boarder between Vermont and New Hampshire. It's a two lane highway with guard rails on either side, with a large, wooded median between you and I-91 South, with not another car in sight. Out of the right corner of your eye you sense something moving out of the trees and seemingly leap into the road. You have no idea what it is, but your instincts take over and you jolt into the left lane, narrowly missing the object. Afterwards, you realize it was a white-tailed deer.

None of this sense-plan-act loop happened at the level of object detection. You did not run a detector with ``deer'' in the taxonomy, estimate a bounding box for the deer, and then feed that representation into a tracker. Instead, you leveraged low-level motion cues from your sensor suite (eyes) and, based on your immediate estimate of the object's velocity, reacted to avoid a collision. Only afterwards were you able to understand the object's true extents and semantics as a deer.

This motivates motion understanding in Autonomous Vehicles, but it applies to all kinds of domains --- even domestic care tasks like carrying a loaded tray or stacking objects in a cabinet require systems to react to unexpected motion from potentially novel objects. Motion understanding of this nature is hardwired into the human brain and the brains of many other mammals, \citep{hubel1962receptive}, and so we take it for granted as a basic skill that happens subconsciously. However, this innate ability remains a challenge to computer vision systems, and motivates various lines of work on endowing these abilities to machines.

\section{What is Scene Flow and How is it Helping with Motion Understanding?}\sectionlabel{sceneflow}

In short, scene flow can be thought of as trying to provide dense 3D motion descriptions over short time horizons. But, to better contextualize this and make the problem more concrete, it is useful to first consider scene flow in the context of the problem out of which it was born: optical flow. 

Optical flow is classically formulated as the problem of estimating per-pixel image space motion vectors describing how an image at $t$ warps to form the image at $t+1$~\citep{gibson1950, gibson1966, gibson1977, HORN1981185, barron1994performance}; \figref{perceiver_sintel_example} provides an example input-output pair for optical flow on the classic Sintel dataset \citep{sintel}. While optical flow is able to describe object motion on the image plane, making it useful for applications like video compression \citep{9157366}, such descriptions are insufficient for tasks like obstacle avoidance of arbitrary objects, which requires an understanding of object motion in 3D.

\begin{figure}[htb!]
  \centering
  \begin{subfigure}[b]{0.32\textwidth}
    \centering
    \includegraphics[width=\textwidth]{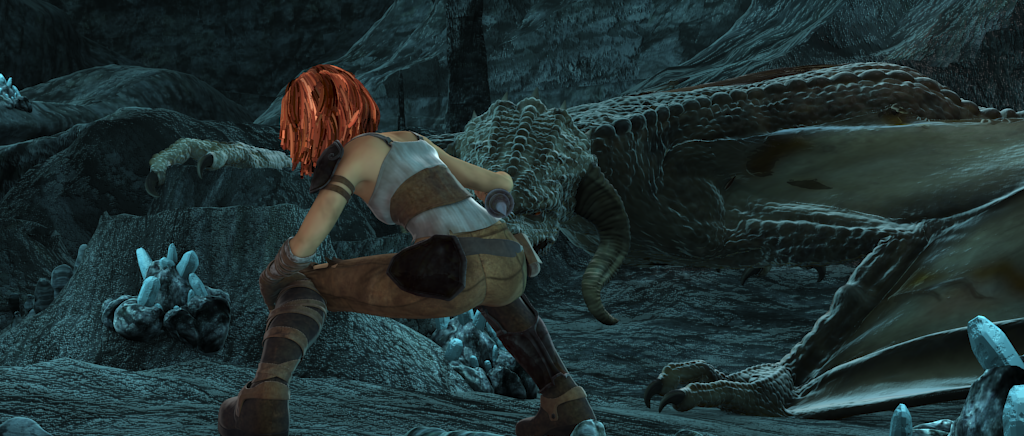}
    \caption{$t$ input image}
    \label{fig:sintel_input_1}
  \end{subfigure}
  \hfill
  \begin{subfigure}[b]{0.32\textwidth}
    \centering
    \includegraphics[width=\textwidth]{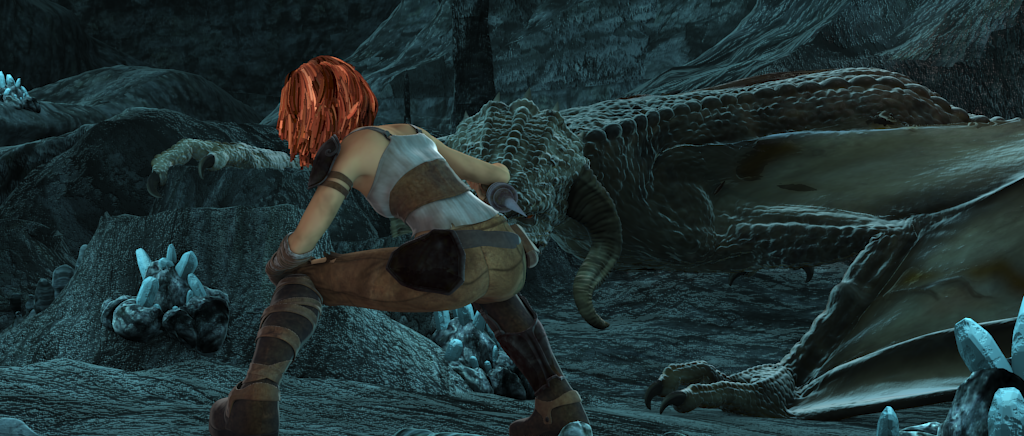}
    \caption{$t+1$ input image}
    \label{fig:sintel_input_2}
  \end{subfigure}
  \hfill
  \begin{subfigure}[b]{0.32\textwidth}
    \centering
    \includegraphics[width=\textwidth]{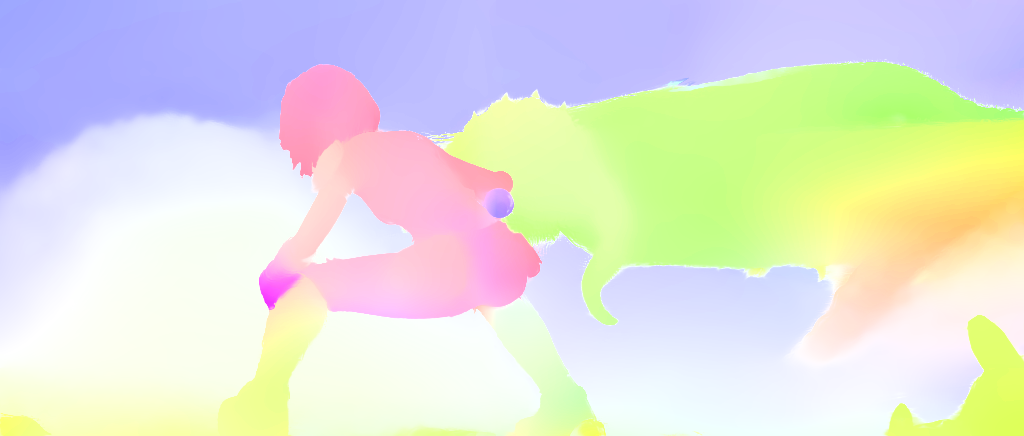}
    \caption{Optical Flow}
    \label{fig:sintel_optical_flow}
  \end{subfigure}
  \caption{An example of optical flow on the Sintel synthetic dataset \citep{sintel}. \figref{sintel_optical_flow} describes the image space motion of \figref{sintel_input_1} (at $t$) as it moves into the view of \figref{sintel_input_2} (at $t+1$); the color of each pixel describes flow direction, while intensity describes its magnitude, with white being $\vec{0}$.}
  \figlabel{perceiver_sintel_example}
\end{figure}

This motivated \cite{vedula1999} to pose the problem of scene flow, a 3D generalization of optical flow, where methods estimate a 3D motion vector for each pixel in the image. Significant follow up work focuses on stereo input or RGB + Depth input approaches~\citep{pons2003variational, pons2007multi, quiroga2014dense,jaimez2015motion,jaimez2015primal,menze2015object,ma2019deep,cao2019learning,ren2017cascaded,behl2017bounding,teed2021raft3d}, but these formulations are specific to images, motivating \cite{dewan2016rigid} to formulate it for arbitrary point clouds (\figref{birdflowexample}). 

\begin{figure}[htb]
    \centering
    \begin{minipage}{0.3\textwidth}
        \centering
        \includegraphics[width=\textwidth]{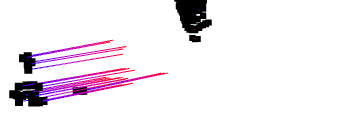}
        \caption*{(a)}
    \end{minipage}\hfill
    \begin{minipage}{0.3\textwidth}
        \centering
        \includegraphics[width=\textwidth]{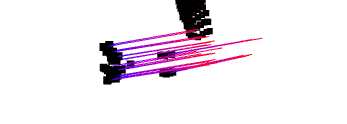}
        \caption*{(b)}
    \end{minipage}\hfill
    \begin{minipage}{0.3\textwidth}
        \centering
        \includegraphics[width=\textwidth]{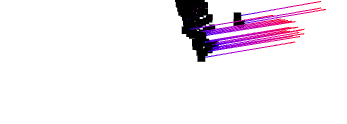}
        \caption*{(c)}
    \end{minipage}
    \caption{Three successive frames of scene flow on a point cloud containing a bird in the Argoverse 2 dataset \citep{argoverse2}. While birds are not labeled as part of the Argoverse 2 taxonomy, our method EulerFlow (\chapterref{eulerflow}, \cite{vedder2024eulerflow}) is able to extract its motion.}
    \figlabel{birdflowexample}
\end{figure}

Scene flow quality is characterized by the disagreement between estimated and ground truth flow vectors (\figref{flow_epe}). The workhorse for various evaluation metrics is Endpoint Error (EPE), the $L_2$ distance between the ground truth and estimated flow vector. In metric space, the unit on this error is a concrete distance (e.g.\ meters), but as we discuss in \chapterref{trackflow} and \cite{khatri2024trackflow}, EPE can also be computed in a unitless speed normalized space.

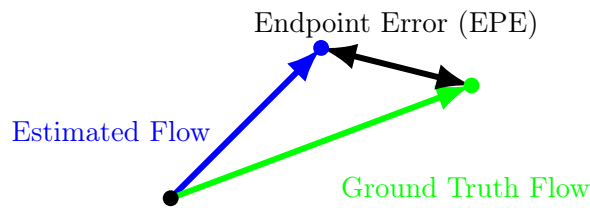
\begin{figure}[htb!]
  \centering
      \begin{tikzpicture}

        \coordinate (origin) at (0,0);
        \coordinate (estimated) at (2,2);
        \coordinate (groundtruth) at (4,1.5);
        
        \draw[thick,->,blue, line width=0.8mm, >=Latex] (origin) -- (estimated) node[midway,below left, xshift=-3mm, yshift=2mm] {\textcolor{blue}{Estimated Flow}};
        \draw[thick,->,green, line width=0.8mm, >=Latex] (origin) -- (groundtruth) node[midway,below right, xshift=1mm, yshift=-3mm] {\textcolor{green}{Ground Truth Flow}};
        
        \draw[thick,<->,black, line width=0.8mm, >=Latex] (estimated) -- (groundtruth) node[midway,above, yshift=2mm] {Endpoint Error (EPE)};
        
        \fill[black] (origin) circle (3pt);
        \fill[blue] (estimated) circle (3pt);
        \fill[green] (groundtruth) circle (3pt);
    
    \end{tikzpicture}
    \caption{Visual definition of Endpoint Error (EPE), the workhorse of scene flow evaluation.}
    \figlabel{flow_epe}
\end{figure}

Ground truth scene flow descriptions are difficult to obtain because scene flow is not trivially self-supervised (\figref{flow_vectors}) --- scene flow is \emph{not} correspondence matching, so motion descriptions must come from some exogenous oracular source. For this reason, synthetic datasets are popular (e.g. FlyingThings \citep{flyingthings}, Sintel \citep{sintel}, Kubric \citep{greff2021kubric}, Spring \citep{spring}, or PointOddessy \citep{zheng2023point}), as the data generation process also trivially provides motion descriptions. However, as discussed by \cite{chodosh2023}, these datasets do not represent realistic scan patterns and often have correspondences not found in realistic data --- real world lidar data tends to be sparse, particularly at range, requiring strong inductive biases or data driven priors to make sense of the scene. This motivates the use of real-world point cloud datasets generated using lidar sensors, such as KITTI \citep{kittisceneflow1, kittisceneflow2}, Argoverse~2 \citep{argoverse2}, NuScenes \citep{caesar2020nuscenes}, and Waymo Open \citep{waymoopen}, where amodal object bounding box tracks provided by human labelers can be used to extract rigid scene flow ground truth for lidar points.

\begin{figure}[htb!]
  \centering
  \includegraphics[width=0.5\textwidth]{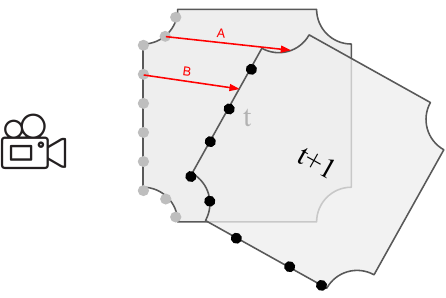}
    \caption{Scene flow is not just correspondence matching --- scene flow vectors describe where the point on an object at time $t$ will end up on the object at $t+1$. We illustrate this with ground truth flow vectors \emph{A} and \emph{B}. Flow vector \emph{A}, associated with a point in the upper left concave corner of the object at $t$, has no nearby observations at $t+1$ due to occlusion of the concave feature. The flow vector \emph{B}, associated with a point on the face of the object at $t$, does not directly match with any observed point on the object at $t+1$ due to observational sparsity. Thus, point matching between $t$ and $t+1$ alone is insufficient to generate ground truth flow.}
    \figlabel{flow_vectors}
\end{figure}

Scene flow, as typically formulated, is not an online forecasting task; given some series of existing observations about the world (e.g. $\pointcloudt$ and $\pointcloudtpone$), methods must post-hoc estimate an underlying model of the world that explains these observations. This post-hoc formulation has proved to be quite challenging, and from a computational perspective it cannot be any more challenging than an online forecasting setting\footnote{An online approach can always be used to solve the offline problem, but the opposite is not necessarily true.} similar to the motivating example of swerving to avoid a deer in \sectionref{motivation}. This motivates us to study the problem of scene flow to first develop systems that can produce a robust understanding post-hoc. While we hope that this process will generate new knowledge that enables researchers to better understand and tackle the general problem of motion understanding, even if no new knowledge is generated, the resulting high quality artifacts capable of doing post-hoc understanding system open up avenues for building self-supervised prediction system via student-teacher distillation, similar to the training pipeline presented in \chapterref{zeroflow} \citep{vedder2024zeroflow}.

\section{Why Are Point Clouds a Useful Input Representation?}\sectionlabel{pointclouds}

Sensors make observations of the world; digital cameras, stereo camera pairs, structured light sensors, and time-of-flight sensors all provide different means of observing the 3D world. While in practice these sensors often do not have a global shutter~\citep{chodosh2024simultaneousmapobjectreconstruction}, it is convenient to think of them as taking observations of the world at a single point in time. Successive observations with these sensors thus provide us a series of snapshots of the world which we then can interpret in order to understand how the state of the world has evolved in the immediate past and predict how it will evolve in the immediate future.

RGB cameras are an attractive input modality for this task because they are common, cheap, and there are many large scale internet video datasets. As we discuss in \sectionref{broaderrelatedwork}, this has spurred a recent renaissance in other motion understanding problem formulations like RGB point tracking and dynamic novel view synthesis. However, cameras as a modality suffer the significant limitation that they do not natively provide information about scene depth, requiring clever post-processing of the video to do stereo-matching between frames or data driven priors to estimate this depth.

This motivates other sensing modalities that can provide depth --- sensors like the Intel RealSense \citep{IntelD400SeriesDatasheet2024} that use a stereo camera pair with known intrinsics / extrinsics plus hardware level feature matching to estimate depth, or the ORBBEC Astra \citep{orbbec_astra} that projects structured light onto a scene and then uses its distortion to estimate depth. These depth cameras provide depth information as part of the regular grid structure of the image (often referred to as RGB+Depth or RGBD data), enabling processing methods to take advantage of this structure in processing (e.g.\ the concept of a neighborhood is well defined in image space~\citep{dewan2016rigid}). However, this regular structure makes it more challenging to place observations from one sensor into a different sensor's frame or into a consistent global frame. 

\begin{figure}[htb]
    \centering
    \includegraphics[width=0.5\linewidth]{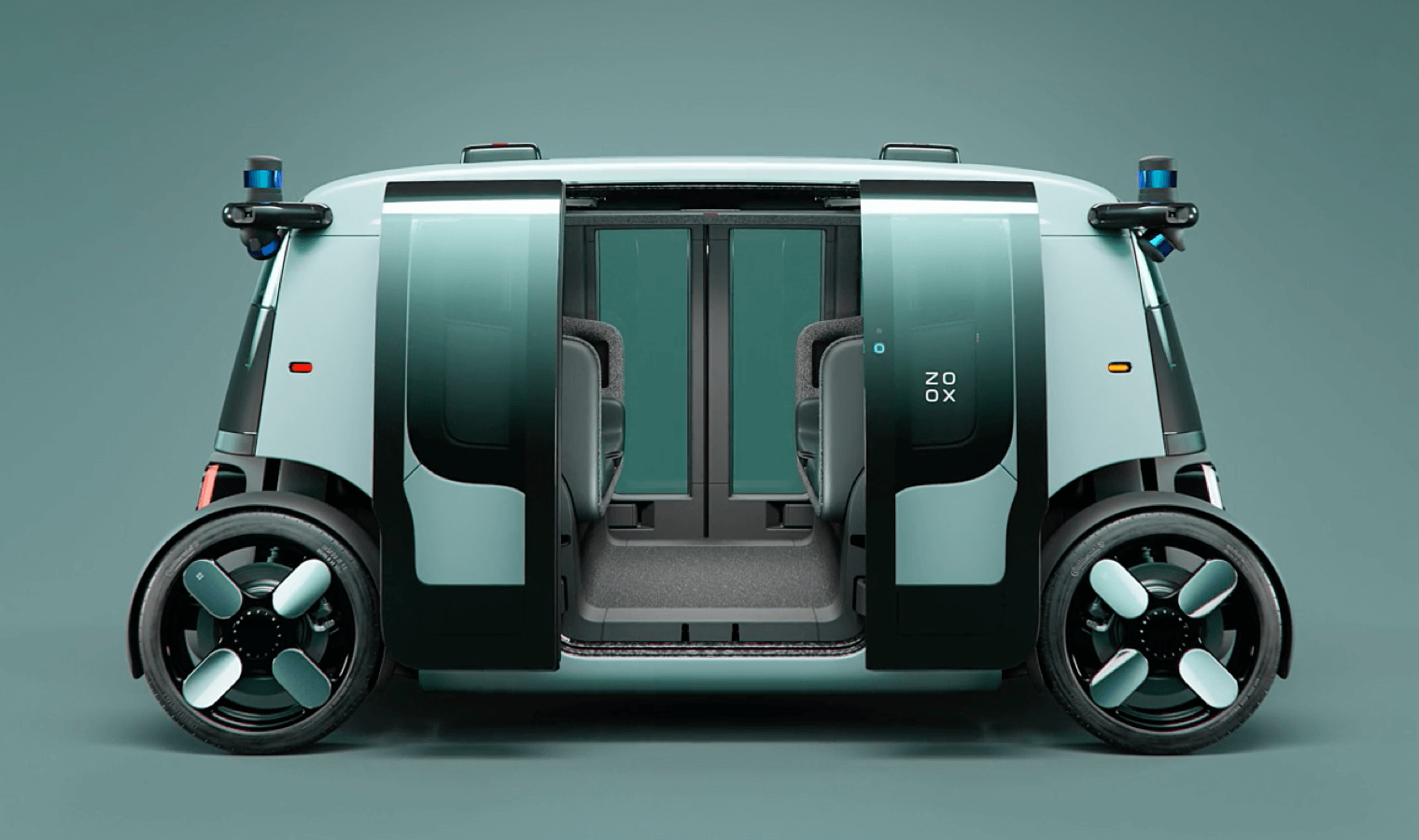}
    \caption{Zoox's Autonomous Vehicle features eight lidar sensors, two on each corner of the vehicle. Placing these sensors in a consistent global frame requires a representation that can readily handle arbitrary SE(3) transformations. Photo taken from \url{https://zoox.com/about}.}
    \figlabel{zooxcar}
\end{figure}

As an example of this need, the Autonomous Vehicle maker Zoox  has eight lidar sensors spread across the four corners of the vehicle (\figref{zooxcar}). In order to be able to process these observations in a consistent ego-vehicle frame, it is easier to represent depth returns as a collection of points in 3D space (a point cloud) that allows for the straightforward application of SE(3) transforms to take the points from the different sensors and put them in a single consistent frame for the car\footnote{There are also other reasons for lidar sensor to produce point clouds, e.g.\ the returns from the fixed beam lidar do not come back in a consistent order vertically, and the ordering flexibility inherent in the set definition of a point cloud means this sensor detail does not matter}. Frame transforms for point clouds are so straightforward that methods often treat the details as an afterthought in the point cloud processing literature, but in image-based methods, authors go to great lengths to develop methods that can handle the 3D structure in a consistent frame (e.g. the tremendous preprocessing typically required in the neural radiance field literature).

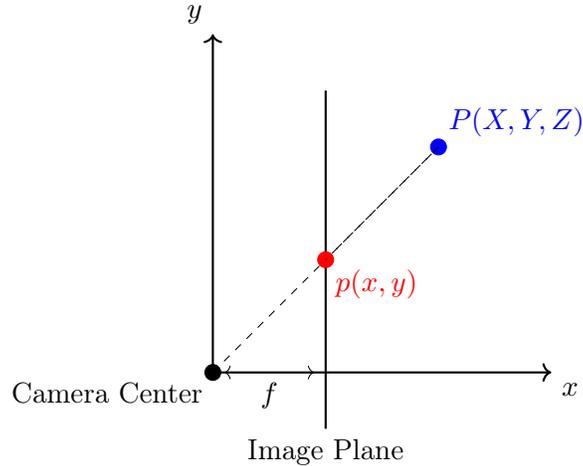
\begin{figure}[htb]
    \centering
    \begin{tikzpicture}[scale=1.5]
    \draw[thick,->] (0,0) -- (3,0) node[anchor=north west] {$x$};
    \draw[thick,->] (0,0) -- (0,3) node[anchor=south east] {$y$};

    \filldraw[black] (0,0) circle (2pt) node[anchor=north east] {Camera Center};

    \draw[thick] (1,2.5) -- (1,-0.5) node[below] {Image Plane};

    \filldraw[blue] (2,2) circle (2pt) node[anchor=south west] {$P(X, Y, Z)$};

    \draw[dashed] (0,0) -- (2,2);

    \filldraw[red] (1,1) circle (2pt) node[anchor=north west] {$p(x, y)$};
    
    \draw[dashed] (2,2) -- (1,1);

    \draw[<->] (0.1,0) -- (0.9,0) node[midway, below] {$f$};
\end{tikzpicture}
    \caption{A depth image can be straightforwardly converted to a point clouds: with access to metric depth and the focal length $f$, the image point $p(x, y)$ on the image plane can be converted to a 3D point $X, Y, Z$ in the camera coordinate system by projecting along the line of sight and scaling by the depth value.}
    \figlabel{imageprojection}
\end{figure}

Stated formally, a point cloud $\pointcloud$ is an unstructured set of 3D points (that may or may not have additional sensor-dependent information attached, such as return intensity, luminance, or color). A point cloud is a strictly more general representation than the RGBD representations discussed above; any RGBD image can be directly converted into a point cloud (assuming sensor intrinsics are known; \figref{imageprojection}), but the reverse is not true --- converting an arbitrary point cloud into an RGBD image will almost certainly result in missing grid values and aliasing across the regular discretized grid. This general definition also means that a point cloud, often presented in a list of points, is order invariant --- any reordering of this list is still the same point cloud --- and methods that handle point clouds often need to make architectural decisions to handle this \citep{qi2017pointnet}.

\begin{figure}[htb]
    \centering
    \includegraphics[width=0.5\textwidth]{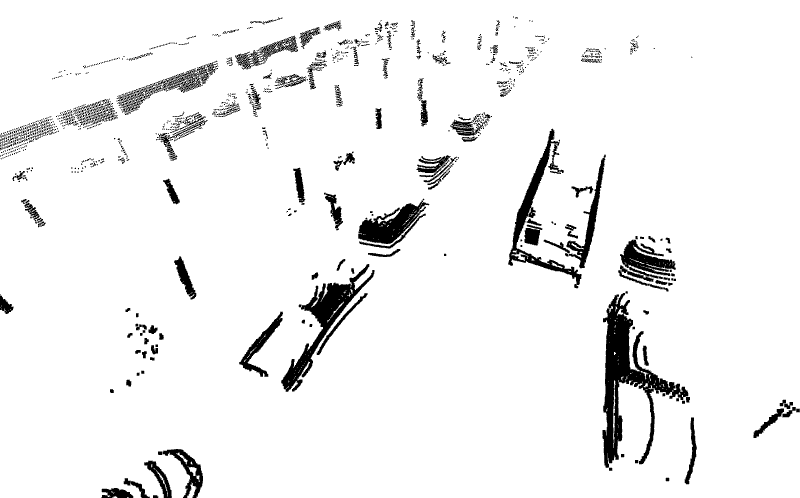}
    \caption{Lidar point cloud from the Argoverse 2 \citep{argoverse2} Autonomous Vehicle dataset after ground point removal.}
    \label{fig:enter-label}
\end{figure}


\section{How Does Prior Art Tackle Point Cloud Scene Flow? }\sectionlabel{priorart}

Prior art for point cloud scene flow broadly falls into one of two approaches, each with their own tradeoffs: test-time optimization methods that leverage hand-built inductive biases but require little to no labeled data, or supervised feedforward methods that improve with more data, but require in-domain labeled data.

\subsection{Feedforward Networks}

Techniques for learning with point clouds are fairly new, and the need for order invariance in their point input has produced highly varied architectures. Inspired by convolutional neural networks, FlowNet3D \citep{flownet3d} proposed operations like \emph{set conv} and \emph{set upconv} to process point clouds; however, even with custom CUDA kernels, learning with these operations is slow due to the heavy use of distance checks. 

Notably, FlowNet3D assumes point clouds have a fixed number of points, and runs all experiments with 20,000 randomly subsampled points from each observation. This is a significant limitation on real workloads  ---  in the autonomous vehicles domain, even after ground point removal, real lidar pointclouds contain a variable number of points with an average above 50,000 points (\figref{pointcloud_size}). This point cloud scale requirement motivated \cite{scalablesceneflow} to create FastFlow3D, a feedforward flow network based on PointPillars \citep{pointpillars}, a network architecture developed in the lidar detection literature that voxelizes 3D space and produces a vector per voxel (using a variant of PointNet \citep{qi2017pointnet}) to enable fixed size tensor processing.

While these methods are designed to learn data driven priors, they are severely limited by their dependence upon in-domain ground truth labels.
This motivated \cite{Mittal_2020_CVPR} to use use nearest neighbor and cycle consistency losses to supervise FlowNet3D, inspiring other feedforward self-supervisory schemes \citep{tishchenko2020self, wu2020pointpwc, baur2021slim, kittenplon2021flowstep3d, seflow}.

Importantly, despite this work on feedforward networks, their flow quality was typically poor; before we introduced the Scene Flow via Distillation pipeline in \chapterref{zeroflow}, state-of-the-art quality point cloud scene flow methods were test-time optimization based (e.g.\ NSFP \citep{nsfp}).

\subsection{Test-time Optimization}

\cite{dewan2016rigid} were the first to pose the scene flow problem for arbitrary unordered point clouds, proposing a test-time optimization approach that utilizes a combination of local surface feature description matching~\citep{tombari2010unique} and local rigidity assumptions. \cite{pontes2020scene} propose the graph Laplacian regularized method, inspired by non-rigid shape registration \citep{eisenberger2020smooth}, to perform ``as-rigid-as-possible'' scene flow, and show it can be used either as a test-time optimization objective or as a surrogate objective for learning. \cite{nsfp} build off this idea of regularization with Neural Scene Flow Prior (NSFP), which proposes optimizing a small ReLU multi-layer perceptron (a neural prior for scene flow, or ``neural scene flow prior'') against Chamfer Distance and cyclical consistency objectives, producing a strong, flexible scene flow method. 
\cite{objectdetectionmotion}, \cite{fastnsf}, and \cite{chodosh2023} further build upon NSFP's backbone, by preprocessing the scene with connected components or clustering, replacing Chamfer Distance with a faster-to-compute loss (Distance Transform, \cite{fastnsf}), or post-processing the flow vectors to produce rigid motion estimates~\citep{objectdetectionmotion, chodosh2023}.

Some methods also attempt to perform learning to aid in optimization. Rather than using hand engineered surface features like \citeauthor{dewan2016rigid}, \cite{ushani2017learning} learn a birds-eye-view (BEV) pillar similarity estimator and employ it in a similar optimization framework. Similarly, \cite{behl2019pointflownet} learn a representation to then optimize to describe non-ego motion, while works like \cite{gojcic2021weakly} and \cite{dong2022exploiting} take advantage of easier-to-acquire foreground / background masks, rather than full flow supervision, to train piece-wise representations to then optimize.

Before we introduced ZeroFlow, test-time optimization methods were state-of-the-art for quality. While these methods are powerful in that they do not need large amounts of labeled training data, they lacked the ability to improve their performance with access to more raw data, and their hand built priors were often engineered specifically for the autonomous vehicle domain. 

\subsection{Evaluation Protocols}

As is typical with the scene flow literature at this time, prior art evaluates against \emph{Average Endpoint Error} (Average EPE), i.e.\ the simple per-point EPE averaged across the entire scene (see \equationref{averageepedef} for a precise definition). \cite{chodosh2023} performs analysis of these methods on the commonly used synthetic dataset FlyingThings \citep{flyingthings} as well as real-world lidar scene flow datasets from autonomous vehicles, namely Argoverse~2 \citep{argoverse2}, NuScenes \citep{caesar2020nuscenes}, and Waymo Open \citep{waymoopen}, and shows that performance on synthetic datasets is \emph{negatively correlated} with performance on real-world datasets, driven by divergent structure and foreground / background balances (in autonomous vehicle datasets, roughly 85\% of points are from the background; \figref{fig:pointdistribution}). \citeauthor{chodosh2023} uses this motivation to introduce \emph{Threeway EPE}, a new evaluation protocol that uses the average of Average EPE computed for three disjoint buckets (\emph{Foreground Static}, \emph{Foreground Dynamic}, and \emph{Background Static}) in order to better evaluate method performance on movers (see \equationref{threewayepedef} for a precise definition). \citeauthor{chodosh2023} also show that in the autonomous vehicle domain where ego-motion information and high definition maps with ground plane information is readily available, ego-motion compensation and ground point removal preprocessing provides significant performance improvements to scene flow methods. This has motivated all modern scene flow methods in the autonomous vehicle domain to advantage of these preprocessing steps~\citep{vedder2024zeroflow, zhang2024deflow, seflow, khatri2024trackflow, lin2024icp, flow4d, vedder2024eulerflow}.

Importantly, Threeway EPE itself has significant shortcomings. Borne from the qualitative analysis of our results from \chapterref{zeroflow}, in \chapterref{trackflow} we show  that Threeway EPE's Foreground Dynamic bucket hides failures of smaller, slower objects like pedestrians, motivating evaluation protocols that take into account object class and speed.

\subsection{Key Takeaways}

Prior art in point cloud scene flow broadly fit on a spectrum between test-time optimization and supervised feed forward, with some works in the middle that try to learn representations to make optimization easier. As a fairly new subfield dealing with an unstructured input representation, there are many growing pains; approaches often feature strong assumptions like rigidity, take liberties like subsampling point clouds, and suffer from speed issues due to exotic or expensive custom operations. Prior art was making progress, but still exhibited a large gap between state-of-the-art and what we would consider ``good quality'' (\figref{teaserfigure}); this significantly limits its adoption and performance in would-be downstream applications like open-world object detection and tracking~\citep{objectdetectionmotion, Zhai2020FlowMOT3M, baur2021slim, huang2022accumulation, flowssl}.



\section{How Does Point Cloud Scene Flow Relate to Other Motion Understanding Formulations?}\sectionlabel{broaderrelatedwork}

In \sectionref{priorart} we discuss prior art in the point cloud scene flow literature; however, scene flow using point clouds is only way to approach the overarching goal of motion understanding. Thus, to provide context, in this section we first present work on other important motion description formulations like optical flow (\sectionref{opticalflow}), point tracking (\sectionref{pointtracking}), and trajectory estimation / dynamic video reconstruction (\sectionref{trajestimation}), which sit in other locations on the density / duration axes of motion description problem formulations (\tableref{methodtaxonomy}) and make different assumptions about input modalities and structure. We then close with a discussion of common themes across approaches and conjectures about the future (\sectionref{commonthemes}).

\newcommand{\methodsize}{\footnotesize}
\begin{table}[htb]
    \centering
    \renewcommand{\arraystretch}{1.5} 
    \begin{tabular}{|c|c|c|}
        \hline
        & \textbf{Single Frame} & \textbf{Long-Horizon} \\
        \hline
        \textbf{Sparse} & \methodsize{---} & \methodsize{Point Tracking (\sectionref{pointtracking})} \\
        \hline
        \textbf{Dense} & 
        \methodsize{\begin{tabular}{c}Scene Flow (\sectionref{priorart}) \\ Optical Flow (\sectionref{opticalflow})\end{tabular}} & 
        \methodsize{\begin{tabular}{c}Trajectory Estimation (\sectionref{trajestimation}) \\ Dynamic Video Reconstruction (\sectionref{trajestimation})\end{tabular}} \\
        \hline
    \end{tabular}
    \caption{Taxonomy of motion description problems.}
    \tablelabel{methodtaxonomy}
\end{table}

\subsection{Optical Flow}\sectionlabel{opticalflow}

As we discuss in \sectionref{sceneflow}, optical flow is the problem that gave birth to scene flow, with the key difference being it is formulated over images, with motion descriptions sitting only in the input image plane~\citep{gibson1950, gibson1966, gibson1977, HORN1981185, barron1994performance}. Optical flow estimates contain the addition of two separate pieces of information: how the camera is moving relative to the static background, as well as how other objects in the scene are moving. 

Due to the lack of depth information, decomposing motion into camera motion and non-ego motion is, in general, under-constrained; this has encouraged approaches to design and incorporate strong regularizing priors to generate plausible motion in practice. Historically, researchers have tackled optical flow with classical dense stereo correspondence algorithms \citep{beauchemin1995computation, scharstein2002taxonomy} combined with optimization techniques \citep{brox2010large, sundaram2010dense}. More recent methods learn optical flow from supervision \citep{dosovitskiy2015flownet, ilg2017flownet, li2022rigidflow, wang2024sea}, pretraining on many of the synthetic datasets found in the scene flow literature (e.g.\ FlyingThings \citep{flyingthings}, Sintel \citep{sintel}, and Spring \citep{spring}) and showing good performance on real world datasets (e.g.\ KITTI \citep{kittisceneflow1,kittisceneflow2}), but architectures remain bespoke, incorporating many of the priors found in more classical approaches. 

While some of these feedfoward networks are large, mostly unstructured networks like the point convolutional FlowNets  \citep{dosovitskiy2015flownet, ilg2017flownet} or the transformer-based FlowFormers \citep{huang2022flowformer, shi2023flowformer++}, current state-of-the-art methods tend to be highly structured \citep{wang2024sea}. For example, the popular Recurrent All-Pairs Field Transform (RAFT) \citep{teed2021raft} pipeline features three components: a learned feature encoder, an all-pairs correlation layer to fit a 4D volume between pixel pairs, and an iterative update operation to iteratively improve flow field estimates. While RAFT is optimized end-to-end, performing a multi-stage extract and associate algorithm, follow-up work like SEA-RAFT \citep{wang2024sea} takes this a step further with curriculum design, by first learning rigid motion descriptions in order to produce good initializations for further refinement when then trained on the full scene flow problem. RigidMask~\citep{rigidmask} takes a different  algorithmic perspective, computing motion estimates at the level of rigid foreground / background component of a scene rather than directly at the level of pixels.

\subsection{Point Tracking}\sectionlabel{pointtracking}

While optical flow computes dense image tracks over short time horizons (a single frame pair), point tracking takes the opposite approach, focusing on tracking (sparse) given points of interest over long time horizons (e.g. multi-second) in RGB videos.

\begin{figure}
    \centering
    \includegraphics[width=0.5\linewidth]{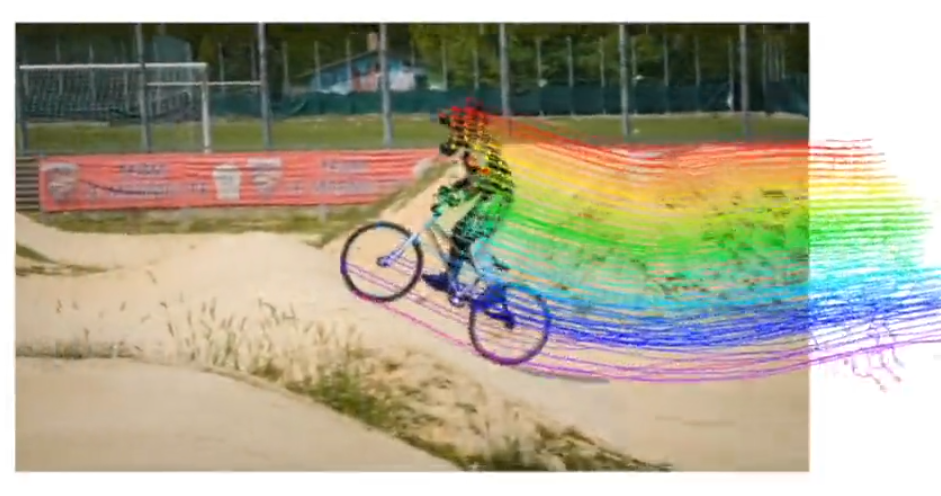}
    \caption{Example of point tracks from CoTracker3 \citep{karaev2024cotracker3}. Points were initialized on the rider in an earlier frame of the video and tracked to their current position as the camera panned to track the rider.}
    \label{fig:enter-label}
\end{figure}

Point tracking was arguably born out of \cite{sand2008particle}; the authors observe that motion description for video broadly falls into two categories: scene flow (as mentioned above) and feature extraction. At the time, these feature extractors like SIFT \citep{sift}, SURF \citep{surf}, ORB \citep{orb}, BRIEF \citep{brief}, and FAST \citep{fast} were designed to extract structure useful for downstream tasks like single image classification \citep{anthony2007image} or as a bag-of-words for association in multi-frame tasks Visual SLAM \citep{probabilisticrobotics}. While a well-tuned SLAM backend can do association between these features to estimate 3D scene structure and ego-motion, these overall features are quite noisy and are a far cry from well-associated point tracks.  This motivated \cite{sand2008particle} to pose the problem of ``particle video'', or what we call point tracking: representing a video as a series of points in image space that persist over many frames. Like with optical flow, the image-based input means that methods jointly estimate ego-motion and non-egomotion within the same point tracks. 

As with optical flow approaches, many modern learning-based point tracking approaches feature highly structured pipelines that implement a clear multi-stage algorithm. In general, approaches train on synthetic datasets like TAP-Vid-Kubric~\citep{doersch2022tapvid}, and implement a per-frame feature extract then a multistep trajectory optimize pipeline~\citep{harley2022particle, doersch2022tapvid, karaev2023cotracker}. A notable exception to this is OmniMotion~\citep{wang2023omnimotion}, which takes a significantly different approach, instead performing test-time optimization to model the scene using per-frame predictions against an existing scene flow method (RAFT, \cite{teed2021raft}); as part of this, it attempts to deal with camera ego-motion by learning a bijective mapping from local observations in camera frame at time $t$ into a shared consistent global coordinate frame as part of its reconstruction objectives.

\subsection{Dynamic Video Reconstruction and Trajectory Estimation}\sectionlabel{trajestimation}

\begin{figure}
    \centering
    \includegraphics[width=0.5\linewidth]{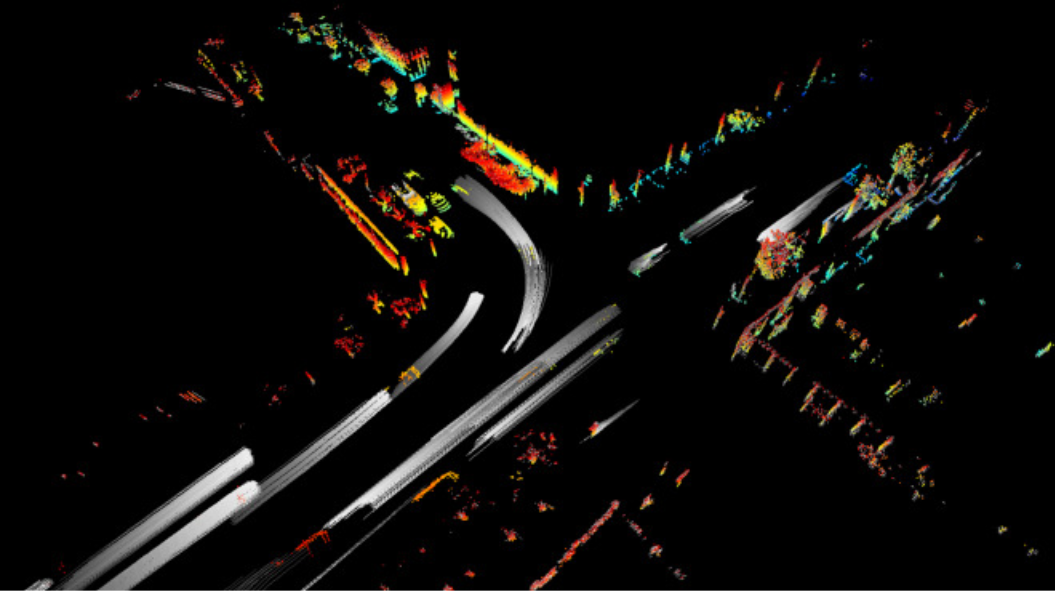}
    \caption{Example of estimated trajectories of several vehicles from Neural Trajectory Prior \citep{ntp} on the Argoverse \citep{argoverse2} Autonomous Vehicle dataset. Reproduced from \cite{ntp}.}
    \figlabel{ntpexample}
\end{figure}

While point tracking focuses on sparse, long-horizon image tracking via given points of interest, dynamic video reconstruction focuses on dense, long-horizon image tracking as part of 3D video reconstruction. Similarly, while scene flow focuses on dense, short-time horizon point tracks, trajectory estimation focuses on long-horizon point tracks.

As both are nascent problem formulations, exact inputs and outputs vary wildly. Some dynamic video reconstruction approaches like Shape-of-Motion \citep{som2024} take in a signal monocular RGB video and use monocular depth estimates to attempt a full 3D scene reconstruction plus tracking. Other approaches take in multiple camera views \citep{luiten2023dynamic}, or use directly sensed depth as an additional input \citep{yang2023emernerf}. Similarly, trajectory estimators take in a sequence of point clouds and estimate motion across them \citep{dynamicfusion, ntp, vidanapathirana2023mbnsf}.

Broadly speaking, these methods all employ a canonical frame definition and estimate a warp from the current observation frame to the canonical frame. This general strategy is extremely common, employed across many different categories of approaches from classical \citep{dynamicfusion} to modern NeRFs \citep{park2021nerfies, lin2024dynamic} and 3D Gaussian \citep{som2024, Wu_2024_CVPR} approaches. However, some notable exceptions differ from this; for example, EmerNeRF \citep{yang2023emernerf} chooses to estimate the scene using explicitly separate static and dynamic fields, with ray viewing angle changes enabling appearance changes  due to non-ego motion without explicit separate object dynamics modeling, and \cite{luiten2023dynamic} propose an online optimization algorithm that initializes the splatted gaussians and then updates the representation for each new observation.

This canonical frame definition assumption is something we address specifically in \chapterref{eulerflow} \citep{vedder2024eulerflow}; our differential equation formulation estimates the instantaneous velocity instead of point offsets, allowing us to extract point-to-point motion from arbitrary input and output timestamps without having to relate it to an intermediary third point.

\subsection{Common Themes Across Approaches}\sectionlabel{commonthemes}

By taking a unified view of scene flow and other problems, as we do in \tableref{methodtaxonomy}, trends start to emerge. Long-horizon methods tend to produce initial estimates by some means (selecting a canonical frame \citep{dynamicfusion,park2021nerfies, lin2024dynamic, luiten2023dynamic}, extracting image space features \citep{harley2022particle, doersch2022tapvid, doersch2023tapir}) and then perform several optimization loops atop to refine the estimate. Neural representations and Gaussian splats are both popular; Gaussians are far more explicit and traditional graphics pipelines can handle them efficiently \citep{luiten2023dynamic, yang2023deformable3dgs, duisterhof2024deformgs}, but require good initialization\footnote{For example, \cite{luiten2023dynamic} just use depth camera points get their Gaussian initializations and leave non-privileged initialization to future work.}, while neural priors are far more flexible but can be expensive to optimize and feature quirks like catastrophic forgetting \citep{tancik2022blocknerf, wu2024clnerf}. Synthetic datasets are starting to shine, both as part of a training curriculum  \citep{wang2024sea} as well as sources to train general feature extractors \citep{harley2022particle, doersch2022tapvid, doersch2023tapir} to produce good quality real world results. Strong, off-the-shelf pretrained models for problems like monocular depth estimation \citep{som2024} and segmentation \citep{zhang2024dynamic} also provide powerful new tools for constructing better scene understanding \citep{som2024}, with new foundational models like robust multi-view point cloud prediction \citep{dust3r_cvpr24} appearing from other parts of the computer vision literature.

I think this also lays out the path ahead for these problem formulations; methods tacking these various problem formulations will continue to march up and to the right on all problems shown in \tableref{methodtaxonomy}, and lines that are already starting to blur (e.g.\ \chapterref{eulerflow}'s EulerFlow \citep{vedder2024eulerflow} exhibits long horizon point tracking) will disappear completely. My prediction is that in 5 years, we will see methods that can consume casual video and produce high quality dynamic, fully 3D dynamic descriptions of structure that will serve as an optical flow method, a scene flow method, a point tracker, and a dynamic 3D video reconstruction method, and the differences in approaches to these problems will likely cease to exist shortly afterwards.


\section{Towards a Unified View: Formalizing Scene Flow for Point Clouds Beyond Two Frames}\sectionlabel{denselonghorizonformalisms}

In \sectionref{priorart} we laid out prior art in scene flow for point clouds and in \sectionref{broaderrelatedwork} 
 we connected it to related problems that fall elsewhere on the density / duration axes as shown in \tableref{methodtaxonomy}, closing with the conjecture that, in the future, all problems will move towards dense, long horizon tracking (\sectionref{commonthemes}).
 
 We now attempt to concretize this conjecture by laying out the classic formalization of scene flow (\sectionref{classicformalism}), followed by an encompassing formalism we propose as part of \chapterref{eulerflow} \citep{vedder2024eulerflow} that extends scene flow towards the long-horizon tracking paradigm, blurring the lines between motion description problems.

\subsection{The Classic Scene Flow for Point Clouds Formulation}\sectionlabel{classicformalism}

As originally formulated by \cite{dewan2016rigid}, given point clouds $\pointcloudt$ and $\pointcloudtpone$, scene flow is the task of estimating the flow $\flowgtttpone$, i.e.\ a residual vector for each point in $\pointcloudt$ that describes that point's change in position from $t$ to $t+1$. However, this framing is overly restrictive compared to real world needs; while $\pointcloudt$ and $\pointcloudtpone$ are the minimal observations needed to extract ridgid motion, there are not real-world problems that \emph{only} have access to two inputs. Consequently, we are starting to see ad-hoc appearances of more frames as input in the scene flow literature. As an example, \citet{liu2024selfsupervisedmultiframeneuralscene} and Flow4D~\citep{flow4d} use three ($\pointcloudtmone, \pointcloudt, \pointcloudtpone$) and  five input frames ($\pointcloudsub{t-3}, \ldots, \pointcloudsub{t+1}$) respectively to predict $\flowttpone$.

\subsection{Towards Long-horizon Scene Flow}\sectionlabel{towardslonghorizonsceneflowformalism}

In \chapterref{eulerflow} \citep{vedder2024eulerflow}, we extend scene flow estimation from between a single adjacent pair of frames ($\flowgtttpone$ for $\pointcloudt$ to $\pointcloudtpone$) to the full observation sequence ($\flowgt_{0,1}, \ldots, \flowgt_{N-1, N}$ for all adjacent frames in $(\pointcloud_0, \ldots, \pointcloud_N)$.

This formulation is strictly more general than the classic formulation in \sectionref{classicformalism} of only estimating $\flowgtttpone$\footnote{Any prior problem setup consuming $N \geq 2$ point clouds can be cast as $(\pointcloud_0, \ldots, \pointcloud_N)$, and all but $\flowgt_{N-1, N}$ can be ignored as output.}. While in principle this formulation can be solved by concatenating the output of two-frame scene flow methods, a full embrace of the formulation results in methods that produce a single, unified description of motion for an entire scene.

\section{Summary of Contributions}\sectionlabel{contributions}

Against the above backdrop, we set out to tackle several key research questions: 

\begin{itemize}
    \item \emph{How do we build flexible, scalable point cloud scene flow estimators that work across a variety of application domains?}
    \item \emph{How do we transition point cloud scene flow beyond two frames towards dense, long-horizon tracking?}
\end{itemize}

In \chapterref{zeroflow}, we tackle this by proposing a blueprint for scalable scene flow by distilling expensive unsupervised optimization methods into feedforward methods at scale \citep{vedder2024zeroflow}. In \chapterref{trackflow}, we then highlight the need to measure small object performance as part of evaluation to ensure we are making progress \citep{khatri2024trackflow}. In \chapterref{eulerflow}, we then propose a new test-time optimization approach that provides state-of-the-art performance out of the box on a variety of domains by producing motion estimates across the entire sequence of observations, also allowing for the extraction of long-horizon point trajectories \citep{vedder2024eulerflow}.

Below is a more detailed summary of each chapter and what I think are the right takeaways from their results.

\subsection{\chapterref{zeroflow}: \emph{ZeroFlow: Scalable Scene Flow via Distillation} \citep{vedder2024zeroflow}}

We need \emph{scalable} scene flow methods, i.e.\ methods that improve by adding more raw data and more parameters. As discussed in \sectionref{priorart}, when we started this project, scene flow methods were broadly either feed-forward supervised methods using human annotations (or from the synthetic dataset generator), or they were very expensive optimization methods. Worse, almost all of these methods did not run on full-size point clouds; they would downsample the point cloud to 20,000 or 8,196 points instead of processing the 50,000+ points in the (ground removed!) full point clouds. This is a \emph{critical} limitation, as it meant they are fundamentally unsuitable to detecting motion on all but the largest objects. This left us with only a couple optimization and feed-forward baseline scene flow methods that even tried to seriously solve the full scene flow problem.

ZeroFlow, the method we introduce in this chapter, is a very simple idea: distill one of the few (very) expensive optimization methods (Neural Scene Flow Prior~\citep{nsfp}) into one of the few feed-forward networks that could handle full-size point clouds (FastFlow3D~\citep{scalablesceneflow}). This was far more successful than we expected, and ZeroFlow was state-of-the-art on the Argoverse 2 2023 Self-Supervised Scene Flow Leaderboard (beating out the optimization teacher!). It was also 1,000x faster than the best optimization methods, and 1,000x cheaper to train than the human supervised methods.

While conceptually simple, ZeroFlow had several important take-home messages:

\begin{itemize}
    \item We have a working blueprint for scaling a scene flow method with data and compute
    \item At sufficient data scale, feed-forward networks will ignore uncorrelated noise in teacher pseudo-labels, enabling them to outperform the teacher
    \item With sufficient data scale, pseudolabel trained feed-forward networks can outperform human supervised methods with the exact same architecture.
\end{itemize}

\subsection{\chapterref{trackflow}: \emph{I Can't Believe It's Not Scene Flow!} \citep{khatri2024trackflow}}

After publishing ZeroFlow, we spent a long time looking at visualizations of its flow results to get a deeper understanding of its shortcomings. We realized it (and all of the baselines) systematically failed to describe most small object motion (e.g. Pedestrians, as shown in \figref{teaserfigure}). Worse, we were not alerted of these systematic failures by the standard metrics because, by construction, small objects are a very small fraction of the total points in a point cloud, and thus their error contribution was reduced to a rounding error compared to large objects in the overall average.

In order to better quantify this failure, we proposed a new metric, \emph{Bucket Normalized Scene Flow}, that reported error per class, and normalized these errors by point speed to report a \emph{percentage} of motion described --- it is clear that 0.5m/s error on a 0.5m/s walking pedestrian is far worse than 0.6m/s error on a 25m/s driving car.

To show that this wasn't an impossible gap to close, we proposed a very simple and crude supervised baseline, \emph{TrackFlow}, constructed by running an off-the-shelf 3D detector on each point cloud and then associating boxes across frames with a 3D Kalman filter to produce flow. Despite the crude construction without any scene flow specific training (hence the name, \emph{I Can't Believe It's Not Scene Flow!}), it was state-of-the-art by a slim margin on the old metrics but by an enormous margin on our new metric; it was the first method to describe more than 50\% of pedestrian motion correctly.

The key take-home messages are:

\begin{itemize}
    \item There was a huge performance gap to close between prior art and a qualitative notion of reasonable flow quality
    \item Prior standard metrics were broken, hiding this large gap
    \item This gap is closable, even with very simple methods, if we have proper guidance.
\end{itemize} 

\subsection{\chapterref{trackflow}: \emph{Argoverse 2 2024 Scene Flow Challenge}}

In order to push the field to close this gap, I hosted the \emph{Argoverse 2 2024 Scene Flow Challenge}\footnote{\url{https://www.argoverse.org/sceneflow}} as part of the CVPR 2024 \emph{Workshop on Autonomous Driving}, where methods were tasked with minimizing the mean normalized dynamic error of our new metric \emph{Bucket Normalized Scene Flow} and featured both a supervised and unsupervised track. Excitingly, ZeroFlow's distillation pipeline was  was featured in the winning unsupervised submissions ICP Flow \citep{lin2024icp}, demonstrating its general efficacy, but the most unexpected result was from the winner of the supervised track, Flow4D~\citep{flow4d}. Flow4D was able to \emph{halve} the error compared to the next best method, our baseline TrackFlow, \emph{only} using a novel feed forward architecture that was better able to learn general 3D motion cues, and it did \emph{not} need to employ any loss reweighting training tricks (e.g.\ class rebalancing, akin to CBGS \citep{cbgs}) which we expected out of winning methods given the class-aware nature of Bucket Normalized EPE.

Our key take-home messages:

\begin{itemize}
    \item Feed-forward architecture choice was a critically underexplored aspect of scene flow
    \item ZeroFlow and other prior work suffered from the use of FastFlow3D's architecture, which was ill-suited to describing small object motion.
\end{itemize}

\subsection{\chapterref{eulerflow}: \emph{Neural Eulerian Scene Flow Fields} \citep{vedder2024eulerflow}}

Under our new metric from \emph{I Can't Believe It's Not Scene Flow!} and with the Argoverse 2 2024 Scene Flow Challenge results, it became clear that ZeroFlow's poor performance was a crushing combination of a bad teacher (NSFP) and an impoverished student network (FastFlow3D). Flow4D represented a far superior student network, but the literature still lacked a superior teacher if we want to try the Scene Flow via Distillation pipeline again and get better results. This motivated us to design a high-quality, offline optimization method that, even if expensive, could describe the motion of small objects well.

To do this, we proposed \emph{EulerFlow}, a simple, unsupervised test-time optimization method that fits a neural flow volume to the \emph{entire} sequence of point clouds (\sectionref{towardslonghorizonsceneflowformalism}). This full sequence formulation, combined with multi-step optimization losses, results in extremely high quality unsupervised flow, allowing EulerFlow to capture state-of-the-art on the Argoverse 2 2024 Scene Flow Challenge leaderboard, beating out \emph{all} prior art, including \emph{all prior supervised methods}. EulerFlow also displayed a number of emergent capabilities: it is able to extract long tail, small object motion such as birds flying, and Euler integration across the neural prior exhibits 3D point tracking behavior across arbitrary time horizons.

The key take-home messages:

\begin{itemize}
    \item We now have a method to get extremely high quality unsupervised scene flow
    \begin{itemize}
        \item This is a prime candidate for a pseudolabel teacher
        \item This can be used to do long-tail object mining from motion cues
        \item This method works out-of-the-box on a wide variety of scenes, including indoor scenes
    \end{itemize}
    \item Multi-frame predictions were a critical factor to optimizing this representation; this likely has implications for loss design in feed-forward methods
\end{itemize}
 
\subsection{Publications, Artifacts, and Code}

When I started working on scene flow, there were no public model zoos, and the open source scene flow codebases available were a mess. I wrote the ZeroFlow codebase from scratch, which then turned into \texttt{SceneFlowZoo}\footnote{\url{https://github.com/kylevedder/SceneFlowZoo}} with several other baseline implementations.

As part of \emph{I Can't Believe It's Not Scene Flow!}, I also released a standalone dataloader and evaluation package which we used as the basis of the Argoverse 2 2024 Scene Flow Challenge. This codebase, \texttt{BucketedSceneFlowEval}\footnote{\url{https://github.com/kylevedder/BucketedSceneFlowEval}} is used by the model zoo, but is deep learning library agnostic (it produces everything in \texttt{numpy} arrays) and is thinly wrapped in the \texttt{SceneFlowZoo} codebase.

The following chapters are the presentation of the following papers:

\begin{itemize}
    \item \chapterref{zeroflow}: Kyle Vedder, Neehar Peri, Nathaniel Chodosh, Ishan Khatri, Eric Eaton, Dinesh Jayaraman, Yang Liu, Deva Ramanan, \& James Hays. ZeroFlow: Scalable Scene Flow via Distillation. \emph{Proceedings of the Twelfth International Conference on Learning Representations} (ICLR 2024)
    \item \chapterref{trackflow}: Ishan Khatri*, Kyle Vedder*, Neehar Peri, Deva Ramanan, \& James Hays. I Can't Believe It's Not Scene Flow! \emph{Proceedings of the European Conference on Computer Vision} (ECCV 2024)
    \item \chapterref{eulerflow}: Kyle Vedder, Neehar Peri, Ishan Khatri, Siyi Li, Eric Eaton, Mehmet Kocamaz, Yue Wang, Zhiding Yu, Deva Ramanan, \& Joachim Pehserl. Neural Eulerian Scene Flow Fields. To appear in \emph{Proceedings of the Thirteenth International Conference on Learning Representations} (ICLR 2025).
\end{itemize}


\chapter{\MakeUppercase{Scalable Scene Flow via Distillation}}\chapterlabel{zeroflow}

\newcommand{\ourpipeline}{SFvD}
\newcommand{\ourpipelinefull}{Scene Flow via Distillation}

\newcommand{\ourmethod}{ZeroFlow}
\newcommand{\xl}{XL}
\newcommand{\threex}{3X}
\newcommand{\fivex}{5X}
\newcommand{\twox}{2X}
\newcommand{\onex}{1X}
\newcommand{\ourmethodonex}{ZeroFlow \onex{}}
\newcommand{\ourmethodthreex}{ZeroFlow \threex{}}
\newcommand{\ourmethodfivex}{ZeroFlow \fivex{}}
\newcommand{\ourmethodtwox}{ZeroFlow \twox{}}
\newcommand{\ourmethodxlonex}{ZeroFlow \xl{} \onex{}}
\newcommand{\ourmethodxlthreex}{ZeroFlow \xl{} \threex{}}
\newcommand{\ourmethodfull}{Zero-Label Scalable Scene Flow}
\newcommand{\ourmethodfullunderlined}{\ourmethodfull{}}

Scene flow estimation is the task of describing the 3D motion field between temporally successive point clouds. State-of-the-art methods use strong priors and test-time optimization techniques, but require on the order of tens of seconds to process full-size point clouds, making them unusable as computer vision primitives for real-time applications such as open world object detection. Feedforward methods are considerably faster, running on the order of tens to hundreds of milliseconds for full-size point clouds, but require expensive human supervision. To address both limitations, we propose \emph{\ourpipelinefull{}}, a simple, scalable distillation framework that uses a label-free optimization method to produce pseudo-labels to supervise a feedforward model. Our instantiation of this framework, \emph{\ourmethod{}}, achieves \textbf{state-of-the-art} performance on the \emph{Argoverse~2 Self-Supervised Scene Flow Challenge} while using zero human labels by simply training on large-scale, diverse unlabeled data. At test-time, \ourmethod{} is over 1,000$\times$ faster than label-free state-of-the-art optimization-based methods on full-size point clouds (34 FPS vs 0.028 FPS) and over 1,000$\times$ cheaper to train on unlabeled data compared to the cost of human annotation (\$394 vs $\sim$\$750,000). To facilitate further research, we release our code, trained model weights, and high quality pseudo-labels for the Argoverse~2 and Waymo Open datasets at \texttt{\url{https://vedder.io/zeroflow}}.

\section{Overview}
Scene flow estimation is an important primitive for open-world object detection and tracking~\citep{objectdetectionmotion, Zhai2020FlowMOT3M, baur2021slim, huang2022accumulation, flowssl}. As an example, \citet{objectdetectionmotion} generates supervisory boxes for an open-world lidar detector via offline object extraction using high quality scene flow estimates from Neural Scene Flow Prior (NSFP) \citep{nsfp}. Although NSFP does not require human supervision, it takes tens of seconds to run on a single full-size point cloud pair. If NSFP were both high quality and real-time, its estimations could be directly used as a runtime primitive in the downstream detector instead of relegated to an offline pipeline. This runtime feature formulation is similar to \citet{Zhai2020FlowMOT3M}'s use of scene flow from FlowNet3D~\citep{flownet3d} as an input primitive for their multi-object tracking pipeline; although FlowNet3D is fast enough for online processing of subsampled point clouds, its supervised feedforward formulation requires significant in-domain human annotations. 


Broadly, these exemplar methods are representative of the strengths and weakness of their class of approach. Supervised feedforward methods use human annotations which are expensive to produce\footnote{At $\sim$\$0.10 / cuboid / frame, the Argoverse~2~\citep{argoverse2} \emph{train} split cost $\sim$\$750,000 to label; \ourmethod{}'s pseudo-labels cost \$394 at current cloud compute prices. See \appendixref{labelvspseudolabelcosts} for details.}; to amortize these costs, human annotations are typically done on consecutive observations, severely limiting the structural diversity of the annotated scenes (e.g.\ a 15 second sequence from an Autonomous Vehicle typically only covers a single city block). Due to costs and labeling difficulty, large-scale labels are also rarely even available outside of Autonomous Vehicle domains. Test-time optimization techniques circumvent the need for human labels by relying on hand-built priors, but they are too slow for online scene flow estimation\footnote{NSFP~\citep{nsfp} takes more than 26 seconds and Chodosh~\citep{chodosh2023} takes more than 35 seconds per point cloud pair on the Argoverse~2~\citep{argoverse2} train split. See \appendixref{labelvspseudolabelcosts} for details.}.

We propose \emph{\ourpipelinefull} (\ourpipeline{}), a simple, scalable distillation framework that uses a label-free optimization method to produce pseudo-labels to supervise a feedforward model. \ourpipeline{} generates a new class of scene flow estimation methods that combine the strengths of optimization-based and feedforward methods with the power of data scale and diversity to achieve fast run-time and superior accuracy without human supervision. We instantiate this pipeline into \emph{\ourmethodfull{}} (\ourmethod{}), a family of methods that, motivated by real-world applications, can process full-size point clouds while providing high quality scene flow estimates. 
We demonstrate the strength of \ourmethod{} on Argoverse~2 \citep{argoverse2} and Waymo Open \citep{waymoopen}, notably achieving \textbf{state-of-the-art} on the \emph{Argoverse~2 Self-Supervised Scene Flow Challenge} 
(\figref{tradeoff_curve}).

\begin{figure}[t]
  \centering
  \includegraphics[width=\linewidth]{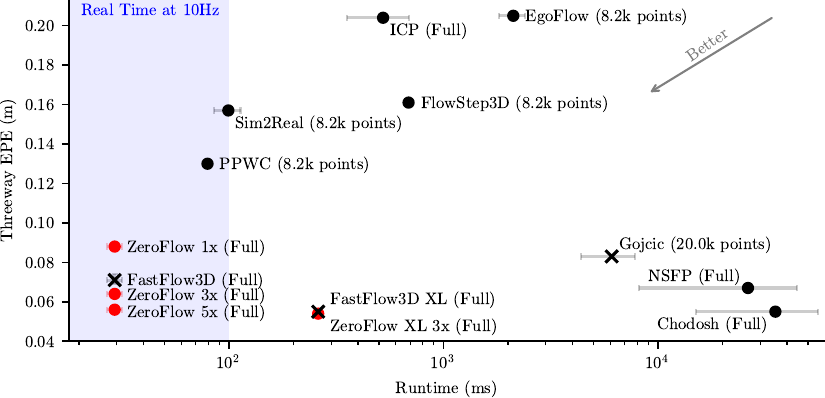}
  \caption{We plot the error and run-time of recent scene flow methods on the Argoverse~2 Sensor dataset~\citep{argoverse2}, along with the size of the point cloud prescribed in the method's evaluation protocol. Our method \textcolor{red}{\ourmethodthreex{}} (ZeroFlow trained on 3$\times$ pseudo-labeled data) outperforms its teacher (NSFP,~\cite{nsfp}) while running over 1,000$\times$ faster, and \textcolor{red}{\ourmethodxlthreex{}} (ZeroFlow with a larger backbone trained on 3$\times$ pseudo-labeled data) achieves \textbf{state-of-the-art}. Methods that use \emph{any} human labels are plotted with \tikzx{3.5pt}, and zero-label methods are plotted with \tikzcircle{3.5pt}.  
} 
  \figlabel{tradeoff_curve}
\end{figure}


Our primary contributions include: 
\begin{itemize}[leftmargin=*]
  \item We introduce a simple yet effective distillation framework, \emph{\ourpipelinefull{}} (\ourpipeline{}), which uses a label-free optimization method to produce pseudo-labels to supervise a feedforward model, allowing us to surpass the performance of slow optimization-based approaches at the speed of feedforward methods.
  \item Using \ourpipeline{}, we present \emph{\ourmethodfull{}} (\ourmethod{}), a family of 
 methods that produce fast, \textbf{state-of-the-art} scene flow on full-size clouds, with methods running over 1,000$\times$ faster than state-of-the-art optimization methods (29.33 ms for \ourmethodonex{} vs 35,281.4 ms for Chodosh) on real point clouds, while being over 1,000$\times$ cheaper to train compared to the cost of human annotations (\$394 vs $\sim$\$750,000).
  \item We release high quality flow pseudo-labels (representing 7.1 GPU-months of compute) for the popular Argoverse~2~\citep{argoverse2} and Waymo Open~\citep{waymoopen} autonomous vehicle datasets, alongside our code and trained model weights, to facilitate further research.
\end{itemize}

\section{Background and Related Work}\sectionlabel{background}

Given point clouds $\pointcloudt{}$ at time $t$ and $\pointcloudtpone{}$ at time $t+1$, scene flow estimators predict $\flowttpone{}$, a 3D vector for each point in $\pointcloudt$ that describes how it moved from $t$ to $t+1$ \citep{dewan2016rigid}. Performance is traditionally measured using the Endpoint Error (EPE) between the predicted flow $\flowttpone{}$ and ground truth flow $\flowgtttpone{}$ (\equationref{averageepedef}):
\begin{equation}
  \small
  \equationlabel{averageepedef}
  \textup{EPE}\left({\pointcloudt} \right) = \frac{1}{\norm{\pointcloudt}} \sum_{p \in \pointcloudt} \norm{\flowttpone{}(p) - \flowgtttpone{}(p)}_2.
\end{equation}

Unlike next token prediction in language~\citep{gpt} or next frame prediction in vision~\citep{weng2021inverting}, future observations do not provide ground truth scene flow (\figref{example_sceneflow}).
To simply evaluate scene flow estimates, ground truth motion descriptions must be provided by an oracle, typically human annotation of real data \citep{waymoopen, argoverse2} or the generator of synthetic datasets \citep{flyingthings,zheng2023point}.


\begin{figure}[t]
\centering
\begin{minipage}[l]{0.28\textwidth}
\centering
\includegraphics[trim=0cm 0 0cm 0cm, clip, width=1\linewidth]{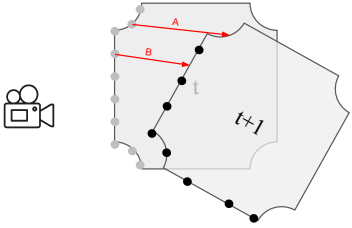}
\end{minipage} \hfill
\begin{minipage}[r]{0.67\textwidth}
\vspace{-1mm}
\hfill
    \caption{Scene flow vectors describe where the point on an object at time $t$ will end up on the object at $t+1$. In this example, ground truth flow vector \emph{A}, associated with a point in the upper left concave corner of the object at $t$ has no nearby observations at $t+1$ due to occlusion of the concave feature. The ground truth flow vector \emph{B}, associated with a point on the face of the object at $t$, does not directly match with any observed point on the object at $t+1$ due to observational sparsity. Thus, point matching between $t$ and $t+1$ alone is insufficient to generate ground truth flow.}
    \figlabel{example_sceneflow}
\end{minipage}
\end{figure}

Recent scene flow estimation methods either train feedforward methods via supervision from human annotations~\citep{flownet3d,behl2019pointflownet,tishchenko2020self,kittenplon2021flowstep3d,wu2020pointpwc,puy2020flot,li2021hcrf,scalablesceneflow,gu2019hplflownet,battrawy2022rms, 9856954}, perform human-designed test-time surrogate objective optimization over hand-designed representations~\citep{pontes2020scene,eisenberger2020smooth,nsfp,chodosh2023}, or learn from self-supervision from human-designed surrogate objectives~\citep{Mittal_2020_CVPR,baur2021slim,gojcic2021weakly,dong2022exploiting,li2022rigidflow}. 

Supervised feedforward methods are efficient at test-time; however, they require costly human annotations at train-time. Both test-time optimization and self-supervised feedforward methods seek to address this problem by optimizing or learning against label-free surrogate objectives, e.g.\ Chamfer distance~\citep{pontes2020scene}, cycle-consistency~\citep{Mittal_2020_CVPR}, and various hand-designed rigidity priors~\citep{dewan2016rigid,pontes2020scene,li2022rigidflow,chodosh2023,baur2021slim,gojcic2021weakly}. Self-supervised methods achieve faster inference by forgoing expensive test-time optimization, but do not match the quality of optimization-based methods~\citep{chodosh2023} and tend to require human-designed priors via more sophisticated network architectures compared to supervised methods~\citep{baur2021slim, gojcic2021weakly, kittenplon2021flowstep3d}. In practice, this makes them slower and more difficult to train. In contrast to existing work, we take advantage of the quality of optimization-based methods as well as the efficiency and architectural simplicity of supervised networks. Our approach, \ourmethod{}, uses label-free optimization methods~\citep{nsfp} to produce pseudo-labels to supervise a feedforward model~\citep{scalablesceneflow}, similar to methods used for distillation in other domains~\citep{yang2023adcnet}.

\section{ZeroFlow}

\begin{figure}[h]
  \centering
  \resizebox{\textwidth}{!}{
  \input{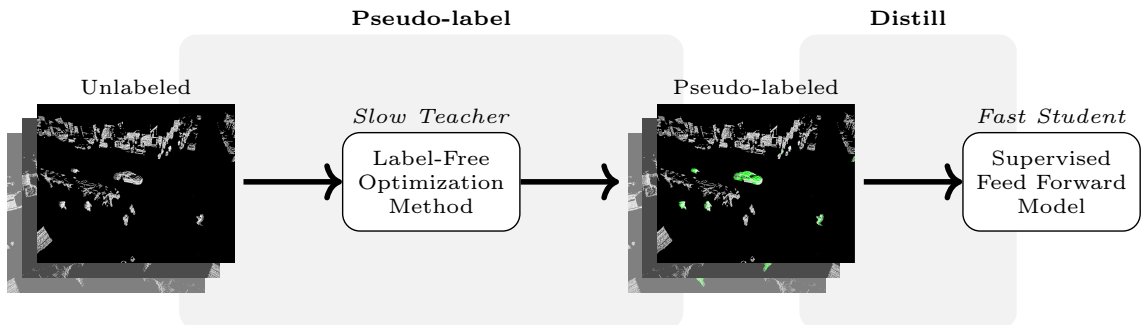}
  }
  \caption{The \emph{\ourpipelinefull{}} (\ourpipeline{}) framework, which describes a new class of scene flow methods that produce high quality, human label-free flow at the speed of feedforward networks.}
  \figlabel{method_diagram}
\end{figure}

We propose \emph{\ourpipelinefull{}} (\ourpipeline{}), a simple, scalable distillation framework that creates a new class of scene flow estimators by using a label-free optimization method to produce pseudo-labels to supervise a feedforward model (\figref{method_diagram}). While conceptually simple, efficiently instantiating \ourpipeline{} requires careful construction; most online optimization methods and feedforward architectures are unable to efficiently scale to full-size point clouds (\sectionref{scalability}).

Based on our scalability analysis, we propose \emph{\ourmethodfull{}} (\ourmethod{}), a family of scene flow models based on \ourpipeline{} that produces fast, \textbf{state-of-the-art} scene flow estimates for full-size point clouds without any human labels (\algoref{ourmethod}). \ourmethod{} uses Neural Scene Flow prior (NSFP)~\citep{nsfp} to generate high quality, label-free pseudo-labels on full-size point clouds (\sectionref{nsfp}) and FastFlow3D~\citep{scalablesceneflow} 
for efficient inference (\sectionref{fastflow3d}).

\subsection{Scaling \ourpipelinefull{} to Large Point Clouds}\sectionlabel{scalability}

Popular Autonomous Vehicle datasets including Argoverse~2~(\cite{argoverse2}, collected with dual Velodyne VLP-32 sensors) and Waymo Open~(\cite{waymoopen}, collected with a proprietary lidar sensor and subsampled) have full-size point clouds with an average of 52,000 and 79,000 points per frame, respectively, after ground plane removal (\appendixref{datasetdetails}, \figref{pointcloud_size}). For practical applications, sensors such as the Velodyne VLP-128 in dual return mode produce up to 480,000 points per sweep~\citep{velodyne128datasheet} and proprietary sensors at full resolution can produce well over 1 million points per sweep. Thus, scene flow methods must be able to process many points in real-world applications.

Unfortunately, most existing methods focus strictly on scene flow \emph{quality} for toy-sized point clouds, constructed by randomly subsampling full point clouds down to 8,192 points~\citep{jin2022deformation,tishchenko2020self,wu2020pointpwc,kittenplon2021flowstep3d,flownet3d,nsfp}. As we are motivated by real-world applications, we instead target scene flow estimation for the full-sized point cloud, making architectural efficiency of paramount importance. As an example of stark differences between feedforward architectures, FastFlow3D \citep{scalablesceneflow}, which uses a PointPillar-style encoder \citep{pointpillars}, can process 1 million points in under 100 ms on an NVIDIA Tesla P1000 GPU (making it real-time for a 10Hz lidar), while methods like FlowNet3D~\citep{flownet3d}
take almost 4 seconds to process the same point cloud.

We design our approach to efficiently process full-size point clouds. For \ourpipeline{}'s pseudo-labeling step, speed is less of a concern; pseudo-labeling each point cloud pair is offline and highly parallelizable. High-quality methods like Neural Scene Flow Prior (NSFP, \cite{nsfp}) require only a modest amount of GPU memory (under $3$GB) when estimating scene flow on point clouds with 70K points, enabling fast and low-cost pseudo-labeling using a cluster of commodity GPUs; as an example, pseudo-labeling the Argoverse 2 train split with NSFP is over 1,000$\times$ cheaper than human annotation (\appendixref{labelvspseudolabelcosts}). The efficiency of \ourpipeline{}'s student feedforward model \emph{is} critical, as it determines both the method's test-time speed and its training speed (faster training enables scaling to larger datasets), motivating models that can efficiently process full-size point clouds.

\subsection{Neural Scene Flow Prior is a Slow Teacher}\sectionlabel{nsfp}

Neural Scene Flow Prior (NSFP, \cite{nsfp}) is an optimization-based approach to scene flow estimation. Notably, it does not use ground truth labels to generate high quality flows, instead relying upon strong priors in its learnable function class (determined by the coordinate network's architecture) and optimization objective (\equationref{nsfploss}). Point residuals are fit per point cloud pair $\pointcloudt$, $\pointcloudtpone{}$ at test-time by randomly initializing two multi-layer perceptrons (MLPs); one to describe the forward flow $\flowforward$ from $\pointcloudt$ to $\pointcloudtpone{}$, and one to describe the reverse flow $\flowrev$ from $\pointcloudt{} + \flowttpone$ to $\pointcloudt{}$ in order to impose cycle consistency. The forward flow $\flowforward$ and backward flow $\flowrev$ are optimized jointly to minimize
\begin{equation}
\small
  \equationlabel{nsfploss}
  \chamferdistance{\pointcloudt{} + \flowforward}{\pointcloudtpone} + \chamferdistance{\pointcloudt{} + \flowforward + \flowrev}{\pointcloudt} \enspace ,
\end{equation}
where $\chamferdistancename$ is the standard Chamfer distance with per-point distances above 2 meters set to zero to reduce the influence of outliers.

NSFP is able to produce high-quality scene flow estimations due to its choice of coordinate network architecture and use of cycle consistency constraint. The coordinate network's learnable function class is expressive enough to fit the low frequency signal of residuals for moving objects while restrictive enough to avoid fitting the high frequency noise from $\chamferdistancename$, and the cycle consistency constraint acts as a local smoothness regularizer for the forward flow, as any shattering effects in the forward flow are penalized by the backwards flow. NSFP provides high quality estimates on full-size point clouds (\figref{tradeoff_curve}), so we select NSFP for \ourmethod{}'s pseudo-label step of \ourpipeline{}.

\subsection{FastFlow3D is a Fast Student}\sectionlabel{fastflow3d}
FastFlow3D~\citep{scalablesceneflow} is an efficient feedforward method that learns using human supervisory labels $\flowgtttpone{}$ and per-point foreground / background class labels. FastFlow3D's loss minimizes a variation of the EPE (\equationref{averageepedef}) that reduces the importance of annotated background points, thus minimizing

\begin{equation}
  \equationlabel{fastflowloss}
  \!\frac{1}{\norm{\pointcloudt}} \sum_{p \in \pointcloudt} \!\bgscale{p} \norm{\flowttpone{}(p) - \flowgtttpone{}(p)}_2
\end{equation}

where 

\begin{equation}
  \equationlabel{bgscale}
  \bgscale{p} = \begin{cases}
    1   & \text{if } p \in \text{Foreground} \\
    0.1 & \text{if } p \in \text{Background} \enspace .
  \end{cases}
\end{equation}

FastFlow3D's architecture is a PointPillars-style encoder~\citep{pointpillars}, traditionally used for efficient lidar object detection \citep{ Vedder2022sparsepointpillars}, that converts the point cloud into a birds-eye-view pseudoimage using infinitely tall voxels (pillars). This pseudoimage is then processed with a 4 layer U-Net style backbone. The encoder of the U-Net processes the $\pointcloudt$ and $\pointcloudtpone$ pseudoimage separately, and the decoder jointly processes both pseudoimages. A small MLP is used to decode flow for each point in $\pointcloudt{}$ using the point's coordinate and its associated pseudoimage feature.

As discussed in \sectionref{scalability}, FastFlow3D's architectural design choices make fast even on full-size point clouds. While most feedforward methods are evaluated using a standard toy evaluation protocol with subsampled point clouds, FastFlow3D is able to scale up to full resolution point clouds while maintaining real-time performance and emitting competitive quality scene flow estimates using human supervision, making it a good candidate for the distillation step of \ourpipeline{}. 

In order to train FastFlow3D using pseudo-labels, we replace the foreground / background scaling function (\equationref{bgscale}) with a simple uniform weighting ($\bgscale{\cdot} = 1$), which collapses to Average EPE; see \appendixref{speedscalingexperiments} for experiments with other weighting schemes. Additionally, we depart from FastFlow3D's problem setup in two minor ways: we delete ground points using dataset provided maps, a standard pre-processing step \citep{chodosh2023}, and use the standard scene flow problem setup of predicting flow between two frames (\sectionref{background}) instead of predicting future flow vectors in meters per second. \algoref{ourmethod} describes our approach, with details specified in \sectionref{methodperf}.

In order to take advantage of the unlabeled data scaling of \ourpipeline{}, we expand FastFlow3D to a family of models by designing a higher capacity backbone, producing \emph{FastFlow3D XL}. This larger backbone halves the size of each pillar to quadruple the pseudoimage area, doubles the size of the pillar embedding, and adds an additional layer to maintain the network's receptive field in metric space; as a result, the total parameter count increases from 6.8 million to 110 million.

\begin{algorithm}
\caption{\ourmethod{}}\algolabel{ourmethod}
\begin{algorithmic}[1]
\State $D \gets $ collection of unlabeled point cloud pairs \Comment{Training Data}
\For{$\pointcloudt, \pointcloudtpone \in D$}\Comment{Parallel \texttt{For}}
\State $\flowgtttpone \gets \textup{Teacher{NSFP}}(\pointcloudt, \pointcloudtpone)$ \Comment{\ourpipeline{} \emph{Pseudo-label} Step}
\EndFor
\For {epoch $\in$ epochs} 
\For{$\pointcloudt, \pointcloudtpone, \flowgtttpone \in D$} \Comment{\ourpipeline{}'s \emph{Distill} Step}
\State $l \gets \textup{\equationref{fastflowloss}}(\textup{StudentFastFlow3D}_\theta(\pointcloudt, \pointcloudtpone), \flowgtttpone)$
\State $\theta \gets \theta$ updated w.r.t.\ $l$
\EndFor
\EndFor
\end{algorithmic}
\end{algorithm}

\section{Evaluating ZeroFlow}\sectionlabel{experiments}

\ourmethod{} provides a family of fast, high quality scene flow estimators. In order to validate this family and understand the impact of components in the underlying \ourpipelinefull{} framework, we perform extensive experiments on the Argoverse~2~\citep{argoverse2} and Waymo Open~\citep{waymoopen} datasets. We compare to author implementations of NSFP~\citep{nsfp} and ~\citet{chodosh2023}, implement FastFlow3D~\citep{scalablesceneflow} ourselves (no author implementation is available), and use \citet{chodosh2023}'s implementations for all other baselines.

As discussed in \citet{chodosh2023}, downstream applications typically rely on good quality scene flow estimates for foreground points. Most scene flow methods are evaluated using average Endpoint Error (EPE, \equationref{averageepedef});  however, roughly 80\% of real-world point clouds are background, causing average EPE to be dominated by background point performance. To address this, we use the improved evaluation metric proposed by \citet{chodosh2023}, \emph{Threeway EPE}:
\begin{equation}
  \small
  \equationlabel{threewayepedef}
  \textup{Threeway EPE}(\pointcloudt) = \textup{Avg} \begin{cases}
      \textup{EPE}({p \in \pointcloudt:  p \in \textup{Background}}) & \textup{(Static BG)} \\
      \textup{EPE}({p \in \pointcloudt: p \in \textup{Foreground}\land \flowgtttpone{}(p) \leq 0.5 \textup{m/s}})  & \textup{(Static FG)} \\
      \textup{EPE}({p \in \pointcloudt: p \in \textup{Foreground}\land \flowgtttpone{}(p) > 0.5 \textup{m/s}})  & \textup{(Dynamic FG)}  \enspace .\\
  \end{cases}
\end{equation}

\subsection{How does \ourmethod{} perform compared to prior art on real point clouds?}\sectionlabel{methodperf}

The overarching promise of \ourmethod{} is the ability to build fast, high quality scene flow estimators that improve with the the availability of large-scale \emph{unlabeled} data. Does \ourmethod{} deliver on this promise? How does it compare to state-of-the-art methods?

To characterize the \ourmethod{} family's performance, we use Argoverse~2 to perform scaling experiments along two axes: dataset size and student size. For our standard size configuration, we use the Argoverse~2 Sensor \emph{train} split and the standard FastFlow3D architecture, enabling head-to-head comparisons against the fully supervised FastFlow3D as well as other baseline methods. For our scaled up dataset (denoted \emph{\threex} in all experiments), we use the Argoverse~2 Sensor \emph{train} split and concatenate a roughly twice as large set of unannotated frame pairs from the Argoverse~2 lidar dataset, uniformly sampled from its 20,000 sequences to maximize data diversity. For our scaled up student architecture (denoted \emph{\xl} in all experiments), we use the \xl{} backbone described in \sectionref{fastflow3d}. For details on the exact dataset construction and method hyperparameters, see \appendixref{datasetdetails}

\begin{table}[h]
  \caption{Quantitative results on the Argoverse~2 Sensor validation split using the evaluation protocol from \citet{chodosh2023}. The methods used in this paper, shown in the first two blocks of the table, are trained and evaluated on point clouds within a 102.4m $\times$ 102.4m area centered around the ego vehicle (the settings for the \emph{Argoverse~2 Self-Supervised Scene Flow Challenge}) . However, following the protocol of \citet{chodosh2023}, all methods report error on points in the 70m $\times$ 70m area centered around the ego vehicle. Runtimes are collected on an NVIDIA V100 with a batch size of 1 \citep{peri2023empirical}. FastFlow3D, \ourmethodonex{}, and \ourmethodthreex{} have identical feedforward architectures and thus share the same real-time runtime; FastFlow3D XL, \ourmethodxlonex{}, and \ourmethodxlthreex{} have identical feedforward architectures and thus share the same runtime.  Methods with an * have performance averaged over 3 training runs (see \appendixref{intertrainvariance} for details). Underlined methods require human supervision.}
  \centering
  \resulttablefontsize
  \setlength{\tabcolsep}{2.5pt}
  \resizebox{\textwidth}{!}{
  \begin{tabular}{lrr|r|rrrr}
 \toprule
 & \multicolumn{2}{c|}{Runtime (ms)} & Point Cloud     & Threeway & Dynamic & Static & Static\\
 & &                                 & Subsampled Size & EPE      & FG EPE  & FG EPE & BG EPE\\ \midrule
 \humanlabels{FastFlow3D}*~\citep{scalablesceneflow} & \multirow{3}{*}{${29.33\pm}$} & \multirow{3}{*}{${2.38}$} & Full Point Cloud & 0.071 & 0.186 & 0.021 & 0.006 \\
 \ourmethodonex{}* (Ours) & & & Full Point Cloud & 0.088 & 0.231 & 0.022 & 0.011 \\
  \ourmethodthreex{} (Ours) & & & Full Point Cloud & 0.064 & 0.164 & 0.017 & 0.011  \\
  \ourmethodfivex{} (Ours) & & & Full Point Cloud & \textbf{0.056} & 0.140 & 0.017 & 0.011  \\ \midrule
  \humanlabels{FastFlow3D XL}  & \multirow{3}{*}{${260.61\pm}$} & \multirow{3}{*}{${1.21}$} & Full Point Cloud & 0.055 & 0.139 & 0.018 & 0.007 \\
  \ourmethodxlonex{} (Ours) & & & Full Point Cloud & 0.070 & 0.178 & 0.019 & 0.013 \\
  \ourmethodxlthreex{} (Ours) & & & Full Point Cloud & \textbf{0.054} & 0.131 & 0.018 & 0.012 \\
 NSFP w/ Motion Comp~\citep{nsfp} & $26,285.0\pm$ & $18,139.3$ & Full Point Cloud & 0.067 & 0.131 & 0.036 & 0.034  \\
 Chodosh et al.~\citep{chodosh2023} & $35,281.4\pm$ & $20,247.7$ & Full Point Cloud & 0.055 & 0.129 & 0.028 & 0.008 \\ \midrule
 Odometry & \multicolumn{2}{c|}{---} & Full Point Cloud & 0.198 & 0.583 & 0.010 & 0.000  \\
 ICP~\citep{chen1992object} & $523.11\pm$ & $169.34$ & Full Point Cloud & 0.204 & 0.557 & 0.025 & 0.028  \\
 \humanlabels{Gojcic}~\citep{gojcic2021weakly} & $6,087.87\pm$ & $1,690.56$ & $20000$ & 0.083 & 0.155 & 0.064 & 0.032  \\
 Sim2Real~\citep{jin2022deformation} & $99.35\pm$ & $13.88$ & $8192$ & 0.157 & 0.229 & 0.106 & 0.137  \\
 EgoFlow~\citep{tishchenko2020self} & $2,116.34\pm$ & $292.32$ & $8192$ & 0.205 & 0.447 & 0.079 & 0.090  \\
 PPWC~\citep{wu2020pointpwc} & $79.43\pm$ & $2.20$ & $8192$ & 0.130 & 0.168 & 0.092 & 0.129  \\
 FlowStep3D~\citep{kittenplon2021flowstep3d} & $687.54\pm$ & $3.13$ & $8192$ & 0.161 & 0.173 & 0.132 & 0.176  \\
 \bottomrule
\end{tabular}
  }
  \tablelabel{argobigtable}
\end{table}

As shown in \tableref{argobigtable}, \ourmethod{} is able to leverage scale to deliver superior performance. While \ourmethodonex{} loses a head-to-head competition against the human-supervised FastFlow3D on both Argoverse~2 (\tableref{argobigtable}) and Waymo Open (\tableref{waymobigtable}), scaling the distillation process to additional unlabeled data provided by Argoverse 2 enables \ourmethodthreex{} to significantly surpass the performance of both methods just by training on more pseudo-labled data. \ourmethodthreex{} even surpasses the performance of its own teacher, NSFP, \emph{while running in real-time!} 

\ourmethod{}'s pipeline also benefits from scaling up the student architecture. We modify \ourmethod{}'s architecture with the much larger \xl{} backbone, and show that our \ourmethodxlthreex{} is able to combine the power of dataset and model scale to outperform all other methods, including significantly outperform its own teacher. Our simple approach achieves \textbf{state-of-the-art} on both the Argoverse~2 validation split and \textbf{\emph{Argoverse~2 Self-Supervised Scene Flow Challenge}}.




\begin{table}[h]
  \centering
  \caption{Quantitative results on Waymo Open using the evaluation protocol from \citet{chodosh2023}. Runtimes are scaled to approximate the performance on a V100~\citep{li2020towards}. Both FastFlow3D and \ourmethodonex{} have identical feedforward architectures and thus share the same runtime. Underlined methods require human supervision.}
  \resulttablefontsize
  \setlength{\tabcolsep}{3pt}
  \resizebox{\textwidth}{!}{
  \begin{tabular}{lrr|r|rrrr}
 \toprule
 & \multicolumn{2}{c|}{Runtime (ms)} & Point Cloud     & Threeway & Dynamic & Static & Static\\
 & &                                 & Subsampled Size & EPE      & FG EPE  & FG EPE & BG EPE\\ \midrule
 \ourmethodonex{} (Ours) & \multirow{2}{*}{$21.66\pm$} & \multirow{2}{*}{$0.48$} & Full Point Cloud & 0.092 & 0.216 & 0.015 & 0.045 \\
 \humanlabels{FastFlow3D}~\citep{scalablesceneflow} & & & Full Point Cloud & 0.078 & 0.195 & 0.015 & 0.024 \\
 \midrule
 Chodosh~\citep{chodosh2023} & $93,752.3\pm$ & $76,786.1$ & Full Point Cloud & \textbf{0.041} & 0.073 & 0.013 & 0.039 \\
 NSFP~\cite{nsfp} & $90,999.1\pm$ & $74,034.9$ & Full Point Cloud & 0.100 & 0.171 & 0.022 & 0.108 \\
 ICP~\citep{chen1992object} & $302.70\pm$ & $157.61$ & Full Point Cloud & 0.192 & 0.498 & 0.022 & 0.055 \\
 \humanlabels{Gojcic}~\cite{gojcic2021weakly} & $501.69\pm$ & $54.63$ & 20000 & 0.059 & 0.107 & 0.045 & 0.025 \\
 EgoFlow~\citep{tishchenko2020self} & $893.68\pm$ & $86.55$ & 8192 & 0.183 & 0.390 & 0.069 & 0.089 \\
 Sim2Real~\citep{jin2022deformation} & $72.84\pm$ & $14.79$ & 8192 & 0.166 & 0.198 & 0.099 & 0.201 \\
 PPWC~\citep{wu2020pointpwc} & $101.43\pm$ & $5.48$ & 8192 & 0.132 & 0.180 & 0.075 & 0.142 \\
 FlowStep3D~\citep{kittenplon2021flowstep3d} & $872.02\pm$ & $6.24$ & 8192 & 0.169 & 0.152 & 0.123 & 0.232\\
 \bottomrule
\end{tabular}
  }
  \tablelabel{waymobigtable}
\end{table}

\subsection{How does \ourmethod{} scale?}\sectionlabel{scalinglaws}

\sectionref{methodperf} demonstrates that \ourmethod{} can leverage scale to capture state-of-the-art performance. However, it's difficult to perform extensive model tuning for large training runs, so predictable estimates of performance as a function of dataset size are critical \citep{openai2023gpt4}. 
Does \ourmethod{}'s performance follow predictable scaling laws?

We train \ourmethod{} and FastFlow3D on sequence subsets / supersets of the Argoverse~2 Sensor train split. \figref{scaling_log} shows \ourmethod{} and FastFlow3D's validation Threeway EPE both decrease roughly logarithmically, and this trend appears to hold for \xl{} backbone models as well.

\begin{figure}[htb]
\centering
  \includegraphics[width=\textwidth]{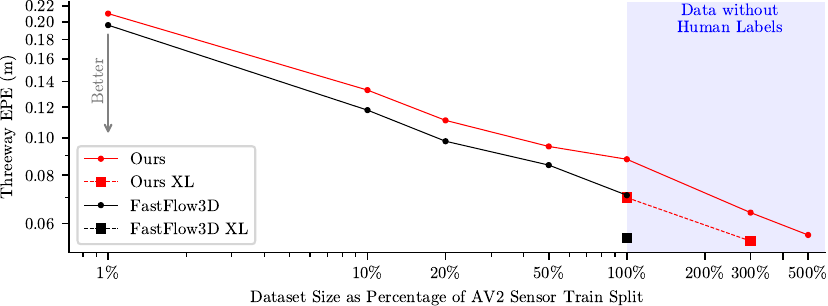}
\caption{Empirical scaling laws for \ourmethod{}. We report Argoverse 2 validation split Threeway EPE as a percentage of the Argoverse 2 \emph{train} split used, on a log$_{10}$-log$_{10}$ scale, trained to convergence. Threeway EPE performance of \ourmethod{} scales logarithmically with the amount of  training data.}
\figlabel{scaling_log}
\end{figure}

Empirically, \ourmethod{} adheres to predictable scaling laws that demonstrate more data (and more parameters) are all you need to get better performance. This makes \ourmethod{} a practical pipeline for building \emph{scene flow foundation models}~\citep{Bommasani2021FoundationModels} using the raw point cloud data that exists \emph{today} in the deployment logs of Autonomous Vehicles and other deployed systems.

\subsection{How does dataset diversity influence \ourmethod{}'s performance?}

In typical human annotation setups, a point cloud \emph{sequence} is given to the human annotator. The human generates box annotations in the first frame, and then updates the pose of those boxes as the objects move through the sequence, introducing and removing annotations as needed. This process is much more efficient than annotating disjoint frame pairs, as it amortizes the time spent annotating most objects in the sequence. This is why most human annotated training datasets (e.g. Argoverse~2 Sensor, Waymo Open) are composed of contiguous \emph{sequences}.
However, contiguous frames have significant structural similarity; in the 150 frames (15 seconds) of an Argoverse~2 Sensor sequence, the vehicle typically observes no more than a city block's worth of unique structure. 
\ourmethod{}, which requires \emph{zero} human labels, does not have this constraint on its pseudo-labels; NSFP run on non-sequential frames is no more expensive than NSFP run on non-sequential frames, enabling \ourmethod{} to train on a more diverse dataset. 
How does dataset diversity impact performance?

To understand the impact of data diversity, we train a version of \ourmethodonex{} and \ourmethodtwox{} \emph{only} on the diverse subset of our Argoverse~2 lidar data selected by uniformly sampling 12 frame pairs from each of the 20,000 unique sequences (\tableref{argodiversity}). 

\begin{table}[h]
  \centering
  \caption{Comparison between \ourmethod{} trained on Argoverse~2 Sensor dataset versus the more diverse, unlabeled Argoverse~2 lidar subset described in \sectionref{methodperf}. Diverse training datasets result in non-trivial performance improvements.}
  \scriptsize
  \setlength{\tabcolsep}{20pt}
  \begin{tabular}{lrrrr}
\toprule
& Threeway & Dynamic & Static & Static  \\
& EPE & FG EPE & FG EPE & BG EPE \\
\midrule
\humanlabels{FastFlow3D}*~\citep{scalablesceneflow} & $0.071$ & $0.186$ & $0.021$ & $0.006$  \\
\ourmethodonex{}  (AV2 Sensor Data)*
& $0.088$ & $0.231$ & $0.022$ & $0.011$  \\
\ourmethodonex{} (AV2 lidar Subset Data) 
& $0.082$ & $0.218$ & $0.018$ & $0.009$ \\
\ourmethodtwox{} (AV2 lidar Subset Data) 
& $0.072$ & $0.184$ & $0.022$ & $0.011$ \\
\bottomrule
\end{tabular}
  \tablelabel{argodiversity}
\end{table}

Dataset diversity has a non-trivial impact on performance; \ourmethod{}, by virtue of being able to learn across \emph{non-contiguous} frame pairs, is able to see more unique scene structure and thus learn to better to extract motion in the presence of the unique geometries of the real world.

 \subsection{How do the noise characteristics of \ourmethod{} compare to other methods?}\sectionlabel{noise_of_our_method}

 \ourmethod{} distills NSFP into a feedforward model from the FastFlow3D family. \sectionref{methodperf} highlights the \emph{average} performance of \ourmethod{} across Threeway EPE catagories, but what does the error \emph{distribution} look like?

 \begin{figure}[h!]
  \figlabel{endpoint_distribution}
  \centering
  \input{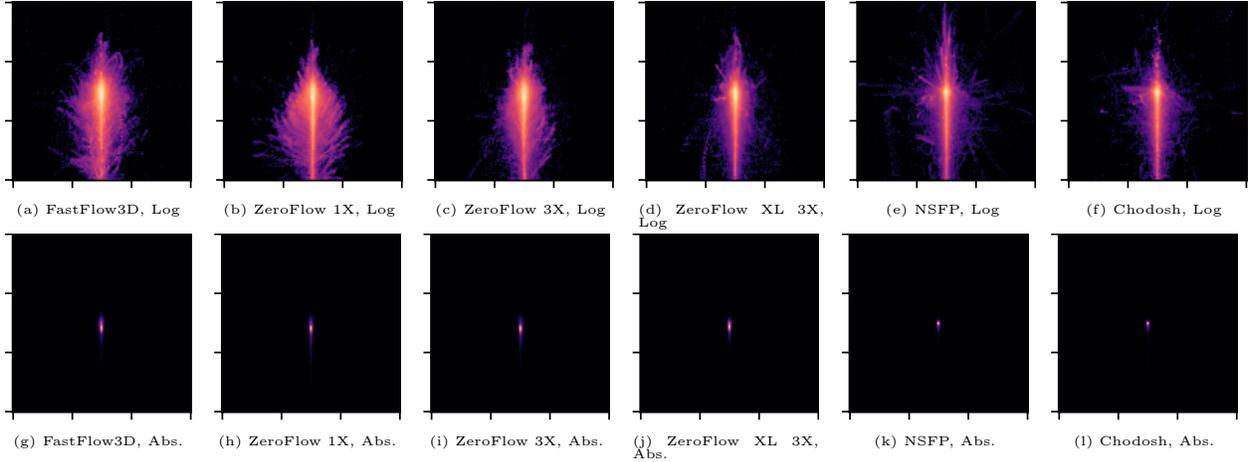}
  \caption{Normalized frame birds-eye-view heatmaps of endpoint residuals for Chamfer Distance, as well as the outputs for NSFP and Chodosh on moving points (points with ground truth speed above 0.5m/s). Perfect predictions would produce a single central dot. Top row shows the frequency on a $\log_{10}$ color scale, bottom row shows the frequency on an absolute color scale. Qualitatively, methods with better quantitative results have tighter residual distributions. See \appendixref{endpoint_errors_details} for details.}
  \figlabel{endpoint_distribution_test_time}
\end{figure}

 To answer this question, we plot birds-eye-view flow vector residuals of NSFP, Chodosh, FastFlow3D, and several members of the \ourmethod{} family on moving objects from the Argoverse~2 validation dataset, where the ground truth is rotated vertically and centered at the origin to present all vectors in the same frame (\figref{endpoint_distribution_test_time}; see \appendixref{endpoint_errors_details} for details on construction).
Qualitatively, these plots show that error is mostly distributed along the camera ray and distributional tightness ($\log_{10}$ plots) roughly corresponds to overall method performance.

Overall, these plots provide useful insights to practitioners and researchers, particularly for consumption in downstream tasks; as an example, open world object extraction~\citep{objectdetectionmotion} requires the ability to threshold for motion and cluster motion vectors together to extract the entire object. Decreased average EPE is useful for this task, but understanding the magnitude and \emph{distribution} of flow vectors is needed to craft good extraction heuristics.


\subsection{How does teacher quality impact \ourmethod{}'s performance?}\sectionlabel{chodoshpseudolabels}

As shown in \sectionref{methodperf} ~\citep{chodosh2023} has superior Threeway EPE over NSFP on both Argoverse~2 and Waymo Open. Can a better performing teacher lead a better version of \ourmethod{}?

To understand the impact of a better teacher, we train \ourmethod{} on Argoverse 2 using superior quality flow vectors from \citet{chodosh2023}, which proposes a refinement step to NSFP lablels to provide improvements to flow vector quality (\tableref{argochodosh}). \ourmethod{} trained on Chodosh refined pseudo-labels provides no meaningful quality improvement over NSFP pseudo-labels (as discussed in \appendixref{intertrainvariance}, the Threeway EPE difference is within training variance for \ourmethod{}). These results also hold for our ablated speed scaled version of \ourmethod{} in \appendixref{speedscalingexperiments}.

Since increasing the quality of the teacher over NSFP provides no noticeable benefit, can we get away with using a significantly faster but lower quality teacher to replace NSFP, e.g.\ the commonly used self-supervised proxy of \chamferdistancenameraw{}? 

To understand if NSFP is necessary, we train ZeroFlow on Argoverse~2 using pseudo-labels from the nearest neighbor, truncated to 2 meters as with \chamferdistancenameraw{}. The residual distribution of \chamferdistancenameraw{} is shown in \appendixref{endpoint_errors_details}, \figref{endpoint_distribution_nearest_neighbor}. \ourmethod{} trained on \chamferdistancenameraw{} pseudo-labels performs significantly worse than NSFP, motivating the use of NSFP as a teacher.
 
\begin{table}[h]
  \centering
  \caption{Comparison between \ourmethod{} trained on Argoverse 2 using NSFP pseudo-labels, \ourmethod{} using \citet{chodosh2023} pseudo-labels, and \ourmethod{} using \chamferdistancenameraw{}. Methods with an * have performance averaged over 3 training runs (see \appendixref{intertrainvariance} for details). The minor quality improvement of Chodosh pseudo-labels does not lead to a meaningful difference in performance, while the significant degradation of \chamferdistancenameraw{} leads to significantly worse performance.}
  \scriptsize
  \setlength{\tabcolsep}{15pt}
  \begin{tabular}{lrrrr}
\toprule
& Threeway & Dynamic & Static & Static \\
& EPE & FG EPE & FG EPE & BG EPE  \\
\midrule
\ourmethodonex{}  (NSFP pseudo-labels)* 
& $0.088$ & $0.231$ & $0.022$ & $0.011$  \\
\ourmethodonex{} (\citet{chodosh2023} pseudo-labels) 
& $0.085$ & $0.234$ & $0.018$ & $0.004$ \\
\ourmethodonex{} (\chamferdistancenameraw{} pseudo-labels) 
& $0.105$ & $0.226$ & $0.049$ & $0.040$ \\
\bottomrule
\end{tabular}
  \tablelabel{argochodosh}
\end{table}

\section{Exploring the Importance of Point Weighting}\appendixlabel{speedscalingexperiments}

In order to train FastFlow3D using pseudo-labels, we need a replacement $\bgscale{\cdot}$ semantics scaling function described in \equationref{bgscale}) because our pseudo-labels do not provide foreground / background semantics. In the main experiments, we use uniform scaling ($\bgscale{\cdot} = 1$).

\subsection{Can we design a better point weighting function for pseudo-labels?}\appendixlabel{speedscaling}

We propose a soft weighting based on pseudo-label flow magnitude: for the point $p$ in the pseudo-label flow $\flowgtttpone{}(p)$, where $\pointspeed{p}$ represents its speed in meters per second, we linearly interpolate the weight of $p$ between $0.1\times$ at 0.4~m/s and full weight at 1.0~m/s, i.e.\
\begin{equation}
  \small
  \bgscale{p} = \begin{cases}
    0.1      & \text{if } \pointspeed{p} < 0.4\text{ m/s} \\
    1.0      & \text{if } \pointspeed{p} > 1.0\text{ m/s} \\
    1.8s-0.8 & \text{o.w.}
  \end{cases}
  \equationlabel{pseudolabelscale}
\end{equation}
These thresholds are selected to down-weight approximately 80\% of points by $0.1\times$, with the other 20\% of points split between the soft  and full weight region\footnote{For Argoverse~2, exactly 78.1\% of points are downweighted, 11.8\% lie in the soft-weight region, and 10.1\% lie in the full weight region; for Waymo Open 80.0\% of points are downweighted, 7.9\% lie in the soft-weight region, and 12.1\% lie in the full-weight region respectively.}. In \tableref{argochodoshextended}, we show that our weighting scheme provides non-trivial improvements over uniform weighting (i.e.\ $\bgscale{\cdot} = 1$) for \ourmethodonex{}; however, it actually hurts performance for \ourmethodthreex{}.

\begin{table}[bt]
  \centering
  \caption{Comparison between \ourmethod{} trained on Argoverse 2 using NSFP pseudo-labels and \ourmethod{} using \citet{chodosh2023} pseudo-labels using both uniform and speed scaled point weighting. Methods with an * have performance averaged over 3 training runs (see \appendixref{intertrainvariance} for details).}
  \scriptsize
  \setlength{\tabcolsep}{8pt}
  \begin{tabular}{lrrrr}
\toprule
& Threeway & Dynamic & Static & Static\\
& EPE & FG EPE & FG EPE & BG EPE \\
\midrule
\ourmethodonex{} (\equationref{pseudolabelscale}, NSFP pseudo-labels)* & $0.084$ & $0.217$ & $0.023$ & $0.011$\\
\ourmethodonex{} (\equationref{pseudolabelscale}, \citet{chodosh2023} pseudo-labels) & $0.086$ & $0.227$ & $0.019$ & $0.011$  \\
\ourmethodonex{}  (NSFP pseudo-labels)* 
& $0.088$ & $0.231$ & $0.022$ & $0.011$  \\
\ourmethodonex{} (\citet{chodosh2023} pseudo-labels)
& $0.085$ & $0.234$ & $0.018$ & $0.004$  \\ \midrule
\ourmethodxlthreex{} & $0.053$ & $0.131$ & $0.018$ & $0.011$ \\
\ourmethodxlthreex{} (\equationref{pseudolabelscale}) & $0.056$ & $0.139$ & $0.017$ & $0.011$ \\
\bottomrule
\end{tabular}
  \tablelabel{argochodoshextended}
\end{table}

\subsection{How much of FastFlow3D's performance is due to its semantic point weighting?}\sectionlabel{semanticsmatter}

Unlike \ourmethod{}, FastFlow3D \emph{can} use human foreground / background point labels to upweight the flow importance of foreground points (\sectionref{fastflow3d}, \equationref{bgscale}). To understand the impact of this weighting, we train FastFlow3D with two modified losses; rather than scaling using semantics as described in \equationref{bgscale}, we uniformly weight all points ($\bgscale{\cdot} = 1$) or our speed based weighting (\equationref{pseudolabelscale}).

\begin{table}[h]
  \centering
  \caption{Comparison between \ourmethod{}, FastFlow3D, and the ablated {FastFlow3D with uniform scaling ($\bgscale{\cdot} = 1$)} trained on Argoverse~2. The performance of FastFlow3D with Uniform Scaling and our speed scaling (\equationref{pseudolabelscale}) are nearly identical to \ourmethod{}'s performance.
  Methods with an * have performance averaged over 3 training runs (see \appendixref{intertrainvariance} for details).
  Underlined methods require human supervision.}
  \scriptsize
  \setlength{\tabcolsep}{10pt}
  \begin{tabular}{lrrrr}
    \toprule
& Threeway & Dynamic & Static  & Static  \\
& EPE      & FG EPE  & FG EPE  & BG EPE  \\ \midrule
    \ourmethodonex{}* (Ours)& $0.088$  & $0.231$ & $0.022$ & $0.011$   \\
    \humanlabels{FastFlow3D ($\bgscale{\cdot} = 1$)}               & $0.081$  & $0.220$ & $0.018$ & $0.006$ \\
    \humanlabels{FastFlow3D (\equationref{pseudolabelscale})} & $0.081$ & $0.224$ & $0.018$ & $0.002$ \\
    \humanlabels{FastFlow3D}*~\citep{scalablesceneflow} & $0.071$  & $0.186$ & $0.021$ & $0.006$ \\ 
    \bottomrule
  \end{tabular}
  \tablelabel{argoablation}
\end{table}

 As shown in \tableref{argoablation}, the performance of FastFlow3D ($\bgscale{\cdot} = 1$) and (\equationref{pseudolabelscale}) degrades more than halfway to \ourmethod{}'s performance.

This raises the question: why is the performance improvement of semantic weighting larger than the improvement of our unsupervised moving point weighting scheme (\appendixref{speedscaling})? We conjecture that not only does semantic weighting provide increased loss on moving objects, it implicitly teaches the network to recognize the structure of objects themselves. For example, with \equationref{bgscale} scaling, end-point error on a stationary pedestrian is significantly higher than static background points, incentivizing the network to learn to detect the point \emph{structure} common to pedestrians, even if immobile, to perfect the predictions on those points.

\subsection{Characterizing inter-training run final performance variance for \ourmethod{} and FastFlow3D}\appendixlabel{intertrainvariance}

On Argoverse 2, Threeway EPE difference between \ourmethod{} and the human supervised FastFlow3D is $1.6$cm (\tableref{argobigtable}); how much of this gap can be attributed to training variance between runs? To answer this question, we train \ourmethod{} and FastFlow3D from scratch 3 times each. \ourmethod{} is trained on the same Argoverse 2 NSFP pseudo-labels (\tableref{argoretrain}), resulting in a mean Threeway EPE of $0.088$m with error of $0.003$m ($0.3$cm) in either direction, and FastFlow3D is trained on the Argoverse 2 human labels (\tableref{argoretrainfastflow}), resulting in a mean Threeway EPE of $0.071$m with error under $0.003$m ($0.3$cm) in either direction.

To contextualize the scale of this variance, the underlying Velodyne VLP-32 sensors used to collect the Argoverse 2 are only certified to $\pm 3$ cm of error~\citep{lasersystemcharacterization} (an order of magnitude greater than the deviation from the mean train performance for \ourmethod{}), and this entirely neglects additional sources of noise introduced from other real world effects such as empirical ego motion compensation.



\begin{table}[h]
  \centering
  \caption{Performance of \ourmethod{} over 3 train runs on the same NSFP pseudo-labels.}
  \scriptsize
  \setlength{\tabcolsep}{3pt}
  \begin{tabular}{lrrrr}
    \toprule
& Threeway & Dynamic & Static  & Static     \\
& EPE      & FG EPE  & FG EPE  & BG EPE  \\ \midrule
    \ourmethodonex{} Run \#1 & $0.085$  & $0.224$ & $0.021$ & $0.011$   \\
    \ourmethodonex{} Run \#2 & $0.088$  & $0.231$ & $0.022$ & $0.010$\\
    \ourmethodonex{} Run \#3 & $0.091$  & $0.240$ & $0.023$ & $0.011$ \\
    \midrule
    \ourmethodonex{} Average & $0.088$  & $0.231$ & $0.022$ & $0.011$ \\
    \bottomrule
  \end{tabular}
  \tablelabel{argoretrain}
\end{table}

\begin{table}[h]
  \centering
  \caption{Performance of \ourmethod{} ablated with point scaling  (\equationref{pseudolabelscale}) over 3 train runs on the same NSFP pseudo-labels.}
  \scriptsize
  \setlength{\tabcolsep}{3pt}
  \begin{tabular}{lrrrr}
    \toprule
& Threeway & Dynamic & Static  & Static    \\
& EPE      & FG EPE  & FG EPE  & BG EPE   \\ \midrule
    \ourmethodonex{} (\equationref{pseudolabelscale}) Run \#1 & $0.083$  & $0.214$ & $0.023$ & $0.011$    \\
    \ourmethodonex{} (\equationref{pseudolabelscale}) Run \#2 & $0.083$  & $0.215$ & $0.024$ & $0.011$    \\
    \ourmethodonex{} (\equationref{pseudolabelscale}) Run \#3 & $0.085$  & $0.222$ & $0.022$ & $0.011$    \\
    \midrule
    \ourmethodonex{} (\equationref{pseudolabelscale}) Average & $0.084$  & $0.217$ & $0.023$ & $0.011$    \\
    \bottomrule
  \end{tabular}
  \tablelabel{argoretrainneweq}
\end{table}

\begin{table}[h]
  \centering
  \caption{Performance of FastFlow3D over 3 train runs on the Argoverse 2 human labels.}
  \scriptsize
  \setlength{\tabcolsep}{3pt}
  \begin{tabular}{lrrrr}
    \toprule
& Threeway & Dynamic & Static  & Static \\
& EPE      & FG EPE  & FG EPE  & BG EPE \\ \midrule
    FastFlow3D Run \#1 & $0.070$  & $0.181$ & $0.020$ & $0.006$   \\
    FastFlow3D Run \#2 & $0.071$  & $0.186$ & $0.021$ & $0.007$ \\
    FastFlow3D Run \#3 & $0.073$  & $0.191$ & $0.023$ & $0.006$ \\
    \midrule
    FastFlow3D Average & $0.071$  & $0.186$ & $0.021$ & $0.006$ \\
    \bottomrule
  \end{tabular}
  \tablelabel{argoretrainfastflow}
\end{table}

\subsection{Characterizing how \ourmethod{}'s performance evolves during training}\appendixlabel{epochs_error}

Threeway EPE breaks down performance into three categories: \emph{Foreground Dynamic}, \emph{Foreground Static}, and \emph{Background}. How does \ourmethod{}'s performance evolve during training?

To understand this, we plot \ourmethodonex{} and \ourmethodthreex{} in \figref{threeway_epe_breakdown}. Both methods converge to their final background performance almost immediately, and most of the improvements seen in the final Threeway EPE stem from improvements in Foreground Dynamic (\figref{fg_dynamic_performance}). The impact of additional data is also made clear early in training, as \ourmethodthreex{} has significantly lower Threeway EPE by epoch 15 than \ourmethodonex{}.

\begin{figure}[htbp!]

  \centering
  \begin{subfigure}[T]{0.48\textwidth}
    \centering
    \includegraphics{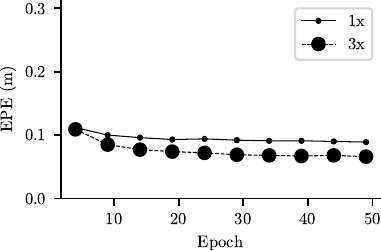}
    \caption{Threeway EPE}
    \figlabel{}
  \end{subfigure}
  \hfill
  \begin{subfigure}[T]{0.48\textwidth}
    \centering
    \includegraphics{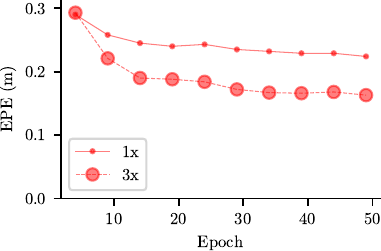}
    \caption{Foreground Dynamic}
    \figlabel{fg_dynamic_performance}
  \end{subfigure}

  \begin{subfigure}[T]{0.48\textwidth}
    \centering
    \includegraphics{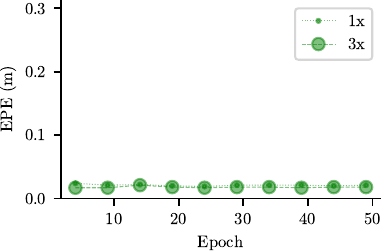}
    \caption{Foreground Static}
    \figlabel{}
  \end{subfigure}
  \hfill
  \begin{subfigure}[T]{0.48\textwidth}
    \centering
    \includegraphics{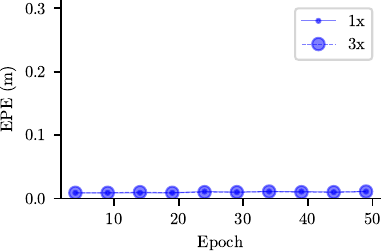}
    \caption{Background}
    \figlabel{}
  \end{subfigure}

  \caption{Performance of \ourmethodonex{} and \ourmethodthreex{} on the Argoverse~2 \emph{val} split by training epoch. Both methods converge to their final background performance almost immediately, and most of the improvements seen in the final Threeway EPE stem from improvements in Foreground Dynamic (\figref{fg_dynamic_performance}).}
  \figlabel{threeway_epe_breakdown}
\end{figure}

\subsection{Estimating Human Labeling versus Pseudo-labeling costs}\appendixlabel{labelvspseudolabelcosts}

NSFP pseudolabeling of the Argoverse 2 {train} split (700 sequences of 150 frames) required a total of 753 hours of NVidia Turing generation GPU time. At September, 2023 Amazon Web Services EC2 prices, a single \texttt{g4dn.xlarge}, equipped with a single NVidia Tesla T4, costs \$0.526 per hour\footnote{\url{https://aws.amazon.com/ec2/pricing/on-demand/}}, for a total cost of \$394 to pseudo-label. By comparison, at an estimated \$0.10 per frame per cuboid (no public cost statements exist for production quality AV dataset labels, but this the standard price point within the industry), Argoverse 2's train split has an average of 75 cuboids per frame~\citep{argoverse2}, for a total cost on the order of \$787,500 to human annotate.

\subsection{Details on Endpoint Residuals}\appendixlabel{endpoint_errors_details}

The process of constructing these endpoint residual plots is shown in \figref{endpoint_residual_construction}. For moving points (points with a ground truth flow vector magnitude >0.5m/s), the raw points (\figref{raw_points}) are transformed into a standard frame with the ground truth vector pointing up and the endpoint at the center of the plot (\figref{standard_frame}), and the residual endpoints are accumulated (\figref{error_dots}). Residual plots for baselines, as well as their unrotated counterparts, are shown in \figref{endpoint_distribution_baselines}.

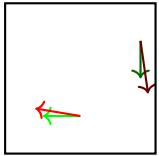
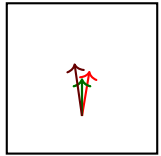
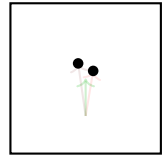
\begin{figure}
    \centering
    \begin{subfigure}[T]{0.30\textwidth}
    \centering
    \begin{tikzpicture}
    \definecolor{lightgreen}{rgb}{0,1,0}
    \definecolor{darkgreen}{rgb}{0,0.4,0}
    \definecolor{lightred}{rgb}{1,0,0}
    \definecolor{darkred}{rgb}{0.4,0,0}

    \draw[->, lightgreen, thick] (0.5,0) -- (0,0); 
    \draw[->, lightred, thick] (0.5,0) -- (-0.1,0.1);

    \draw[->, darkgreen, thick] (1.3,1) -- (1.3,0.5); 
    \draw[->, darkred, thick] (1.3,1) -- (1.4,0.3);
    
    \draw[thick] (-0.5,-0.5) rectangle (1.5,1.5);
    \end{tikzpicture}
    \caption{Raw Points}
    \figlabel{raw_points}
    \end{subfigure}
    \hfill
    \begin{subfigure}[T]{0.30\textwidth}
    \centering
    \begin{tikzpicture}
    \definecolor{lightgreen}{rgb}{0,1,0}
    \definecolor{darkgreen}{rgb}{0,0.4,0}
    \definecolor{lightred}{rgb}{1,0,0}
    \definecolor{darkred}{rgb}{0.4,0,0}
    \draw[->, lightgreen, thick] (0.5,0) -- (0.5,0.5);
    
    \draw[->, lightred, thick] (0.5,0) -- (0.6,0.6);

    \draw[->, darkgreen, thick] (0.5,0) -- (0.5,0.5);

    \draw[->, darkred, thick] (0.5,0) -- (0.4,0.7);
    
    \draw[thick] (-0.5,-0.5) rectangle (1.5,1.5);
    \end{tikzpicture}
    \caption{Standard Frame}
    \figlabel{standard_frame}
    \end{subfigure}
     \hfill
    \begin{subfigure}[T]{0.30\textwidth}
    \centering
    \begin{tikzpicture}
    \definecolor{lightgreen}{rgb}{0,1,0}
    \definecolor{darkgreen}{rgb}{0,0.4,0}
    \definecolor{lightred}{rgb}{1,0,0}
    \definecolor{darkred}{rgb}{0.4,0,0}
    \draw[->, lightgreen, thick, opacity=0.1] (0.5,0) -- (0.5,0.5);
    
    \draw[->, lightred, thick, opacity=0.1] (0.5,0) -- (0.6,0.6);

    \draw[->, darkgreen, thick, opacity=0.1] (0.5,0) -- (0.5,0.5);

    \draw[->, darkred, thick, opacity=0.1] (0.5,0) -- (0.4,0.7);

    \fill[black] (0.6,0.6) circle (2pt);

    \fill[black] (0.4,0.7) circle (2pt);
    
    \draw[thick] (-0.5,-0.5) rectangle (1.5,1.5);
    \end{tikzpicture}
    \caption{Error Dots}
    \figlabel{error_dots}
    \end{subfigure}
    
    \caption{Process for constructing the endpoint residual plots. The raw points (\figref{raw_points}) are transformed into a standard frame with the ground truth vector pointing up and the endpoint at the center of the plot (\figref{standard_frame}), and the residual endpoints are accumulated (\figref{error_dots}).}
    \figlabel{endpoint_residual_construction}
\end{figure}

\begin{figure}[htb!]

  \centering
  \begin{subfigure}[T]{0.22\textwidth}
    \centering
    \includegraphics{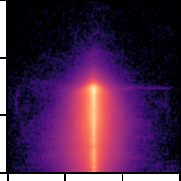}
    \caption{Nearest Neighbor, Log, Rotated}
    \figlabel{endpoint_distribution_nearest_neighbor}
  \end{subfigure}
  \hfill
  \begin{subfigure}[T]{0.22\textwidth}
    \centering
    \includegraphics{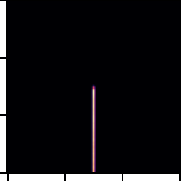}
    \caption{$\vec{0}$ Flow, Log, Rotated}
    \figlabel{endpoint_distribution_odom}
  \end{subfigure}
  \hfill
  \begin{subfigure}[T]{0.22\textwidth}
    \centering
    \includegraphics{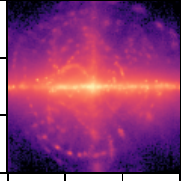}
    \caption{Nearest Neighbor, Log, Unrotated}
    \figlabel{endpoint_distribution_nearest_neighbor_unrotated}
  \end{subfigure}
  \hfill
  \begin{subfigure}[T]{0.22\textwidth}
    \centering
    \includegraphics{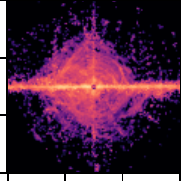}
    \caption{$\vec{0}$ Flow, Log, Unrotated}
    \figlabel{endpoint_distribution_odom_unrotated}
  \end{subfigure}

  \begin{subfigure}[T]{0.22\textwidth}
    \centering
    \includegraphics{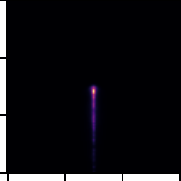}
    \caption{Nearest Neighbor, Abs, Rotated}
    \figlabel{endpoint_distribution_nsfp_unrotated}
  \end{subfigure}
  \hfill
  \begin{subfigure}[T]{0.22\textwidth}
    \centering
    \includegraphics{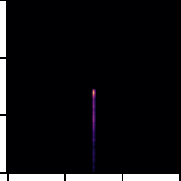}
    \caption{$\vec{0}$ Flow, Abs, Rotated}
    \figlabel{endpoint_distribution_nsfp_unrotated}
  \end{subfigure}
  \hfill
  \begin{subfigure}[T]{0.22\textwidth}
    \centering
    \includegraphics{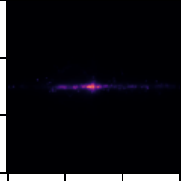}
    \caption{Nearest Neighbor, Abs, Unrotated}
    \figlabel{endpoint_distribution_nearest_neighbor_unrotated}
  \end{subfigure}
  \hfill
  \begin{subfigure}[T]{0.22\textwidth}
    \centering
    \includegraphics{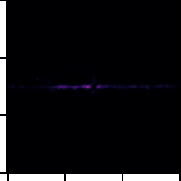}
    \caption{$\vec{0}$ Flow, Abs, Unrotated}
    \figlabel{endpoint_distribution_odom_unrotated}
  \end{subfigure}

  \caption{Birds-eye-view heatmap of endpoint residuals for na\"ive flow methods of predicting flow (Nearest Neighbor and $\vec{0}$ Flow on all points) for non-background points moving above 0.5m/s in the raw coordinate frame of the ground truth labels. Brighter color indicates more points in each bin. Perfect labels would produce a single central dot. Distance between ticks is 1 meter. Top row shows frequency on a log color scale to display error distribution shape. Bottom row shows frequency on an absolute color scale to display centroid. Left half shows results in the rotated ground truth coordinate frame. Right half shows results in the unrotated ground truth coordinate frame.}
  \figlabel{endpoint_distribution_baselines}
\end{figure}

\section{Discussion}\sectionlabel{conclusion}

Our scene flow approach, \ourmethodfull{} (\ourmethod{}), produces fast, state-of-the-art scene flow \emph{without human labels} via our conceptually simple distillation pipeline. 

But, more importantly, we present the first  practical pipeline for building \emph{scene flow foundation models}~\citep{Bommasani2021FoundationModels} using the raw point cloud data that exists \emph{today} in the deployment logs at Autonomous Vehicle companies and other deployed robotics systems. Foundational models in other domains like language~\citep{gpt3,openai2023gpt4} and vision~\citep{segmentanything,r3m} have enabled significant system capabilities with little or no additional domain-specific fine-tuning~\citep{wang2023voyager,ma2022vip,ma2023liv}. We posit that a scene flow foundational model will enable new systems that can leverage high quality, general scene flow estimates to robustly reason about object dynamics even in foreign or noisy environments.

\poorparagraph{Limitations and Future Work} \ourmethod{} inherits the biases of its pseudo-labels. Unsurprisingly, if the pseudo-labels consistently fail to estimate scene flow for certian objects, our method will also be unable to predict scene flow for those objects; however, further innovation in model architecture, loss functions, and pseudo-labels may yield better performance. In order to enable further work on \ourpipelinefull{}-based methods, we release\footnote{\url{https://vedder.io/zeroflow}} our code, trained model weights, and NSFP flow pseudo-labels, representing $3.6$ GPU-months for Argoverse 2 and $3.5$ GPU-months for Waymo Open.

\section{Argoverse 2 and Waymo Open Dataset Configuration Details}\appendixlabel{datasetdetails}

\poorparagraph{Argoverse 2} The Sensor dataset contains 700 training and 150 validation sequences. Each sequence contains 15 seconds of 10Hz point clouds collected using two Velodyne VLP-32s mounted on the roof of a car.
As part of the training protocol for \ourmethod{}, FastFlow3D, and NSFP w/ Motion Compensation, we perform ego compensation, ground point removal, and restrict all points to be within a 102.4m $\times$ 102.4m area centered around the ego vehicle, resulting in point clouds with an average of 52,871 points (\figref{argo_pointcloud_size}). The point cloud $\pointcloudtpone{}$ is centered at the origin of the ego vehicle's coordinate system and $\pointcloudt{}$ is projected into $\pointcloudtpone{}$'s coordinate frame. For \ourmethod{} and FastFlow3D, the PointPillars encoder uses $0.2$m$\times 0.2$m pillars, with all architectural configurations matching \citep{scalablesceneflow}. For NSFP w/ Motion Compensation, we use the same architecture and early stopping parameters as the original method~\citep{nsfp}. For FastFlow3D and the FastFlow3D student architecture of \ourmethod{}, we train to convergence (50 epochs) with an Adam~\citep{kingma2014adam} learning rate of \SI{2e-6} and batch size $64$. For FastFlow3D XL and the FastFlow3D XL student architecture of \ourmethod{} (\ourmethodxlonex{}, \ourmethodxlthreex{}), we train to convergence (10 epochs) with the same optimizer settings and a batch size $12$. For \ourmethodthreex{} and and \ourmethodxlthreex{}, we train on an additional 240,000 unlabeled frame pairs (roughly twice the size as the Argoverse~2 Sensor \emph{train} split), constructed by selecting 12 frame pairs at uniform intervals from the 20,000 sequences of the Argoverse~2 lidar dataset. For all other methods in \tableref{argobigtable}, we use the implementations provided by \citet{chodosh2023}, which follow ground removal and ego compensation protocols from their respective papers.

\poorparagraph{Waymo Open} The dataset contains 798 training and 202 validation sequences. Each sequence contains 20 seconds of 10Hz point clouds collected using a custom lidar mounted on the roof of a car.
We use the same preprocessing and training configurations used on Argoverse 2; after ego motion compensation and ground point removal, the average point cloud has 79,327 points (\figref{waymo_pointcloud_size}).

As shown in \figref{pointcloud_size}, Argoverse 2~\citep{argoverse2} and Waymo Open~\citep{waymoopen} are significantly larger than the 8,192 point subsampled point clouds used by prior art.

\begin{figure}[h]
  \begin{subfigure}[T]{0.48\textwidth}
    \centering
    \includegraphics{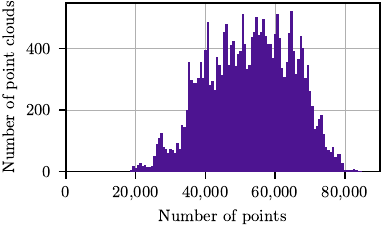}
    \caption{Distribution of point cloud sizes in the Argoverse~2 Sensor \emph{val} split: $\mu = 52,871.6; \sigma = 12,227.2$.}
    \figlabel{argo_pointcloud_size}
  \end{subfigure}
  \hfill
  \begin{subfigure}[T]{0.48\textwidth}
    \centering
    \includegraphics{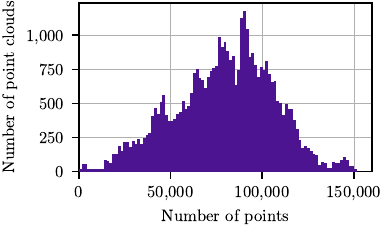}
    \caption{Distribution of point cloud sizes in the Waymo Open \emph{val} split: $\mu = 79,327.8; \sigma = 27,182.1$.}
    \figlabel{waymo_pointcloud_size}
  \end{subfigure}
  \caption{Point cloud size distributions for the \emph{val} set of the Argoverse~2 Sensor~\citep{argoverse2} and Waymo Open~\citep{waymoopen} datasets after ground removal and clipped to a 102.4m $\times$ 102.4m box around the ego vehicle.}
  \figlabel{pointcloud_size}
\end{figure}

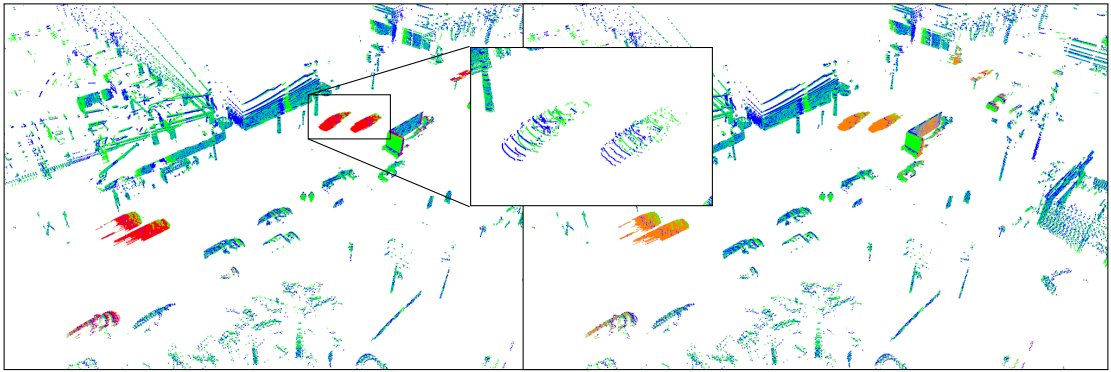
\begin{figure}[t]
  \centering
  \begin{tikzpicture}
    \setlength{\fboxsep}{0.0pt}
    \node[anchor=south west,inner sep=0] (image) at (0,0) {\fbox{\includegraphics[width=0.47\textwidth]{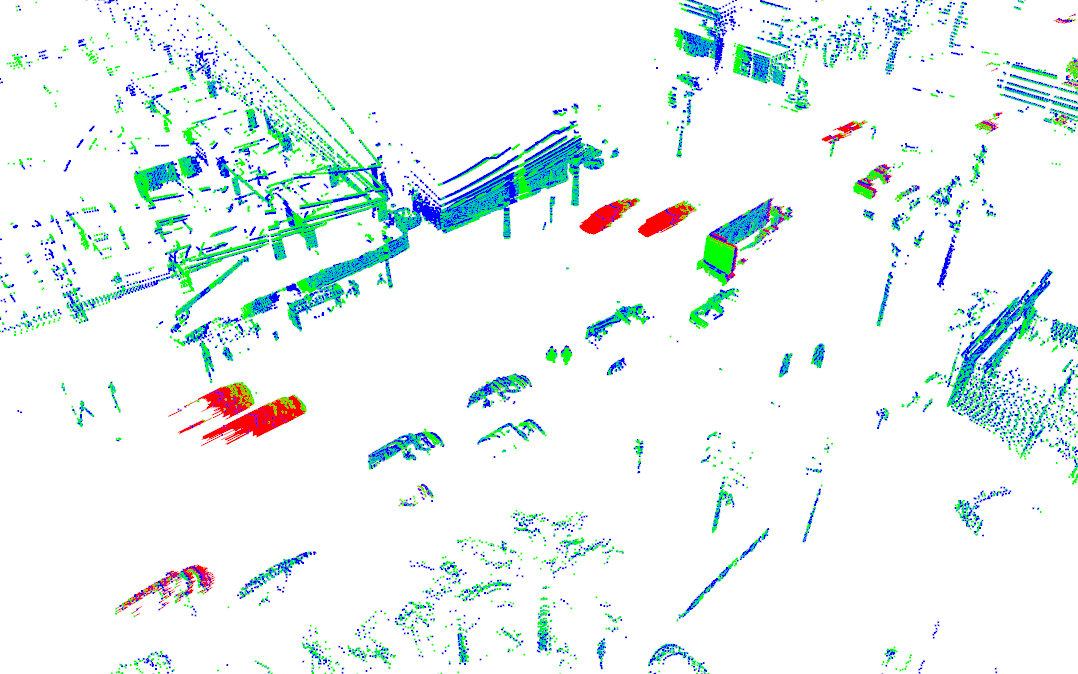}}};
    \node[anchor=south west,inner sep=0] (image2) at (6.9,0) {\fbox{\includegraphics[width=0.47\textwidth]{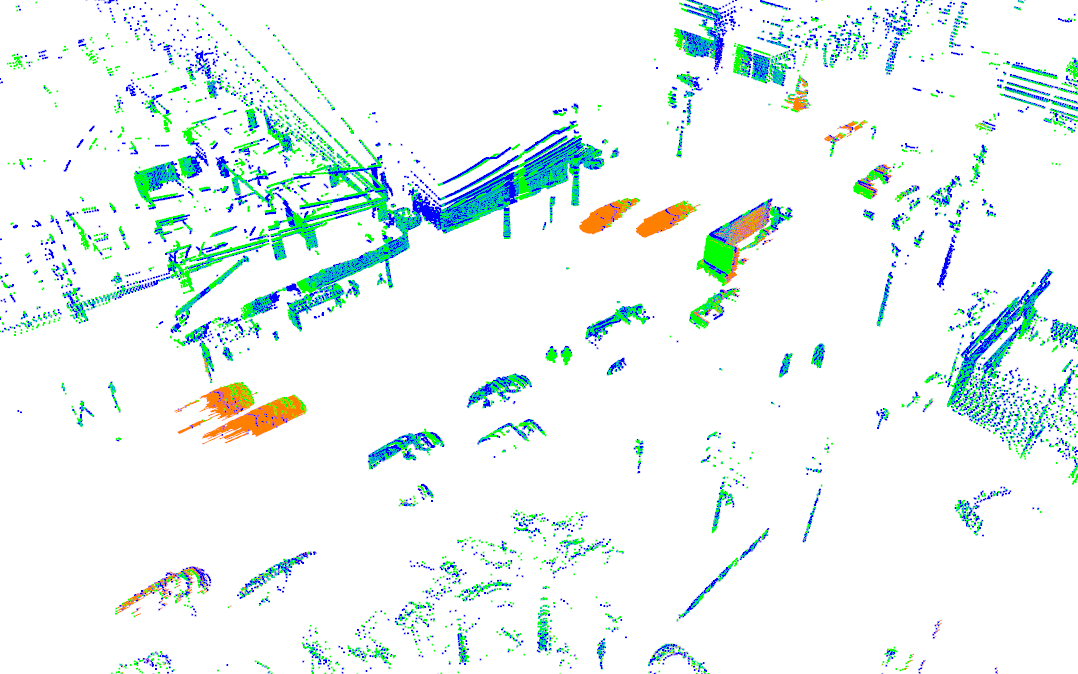}}};
    \begin{scope}[x={(image.south east)},y={(image.north west)}]
      \coordinate (zoom_area_south_west) at (0.52,0.63);
      \coordinate (zoom_area_north_east) at (0.66,0.75);
      \coordinate (zoom_area_north_west) at (0.52,0.75);

      \draw[black] (zoom_area_south_west) rectangle (zoom_area_north_east);

      \node[inner sep=0,right=-1.6cm of image, yshift=0.8cm] (zoomed_image) {{\fbox{\includegraphics[width=3.2cm]{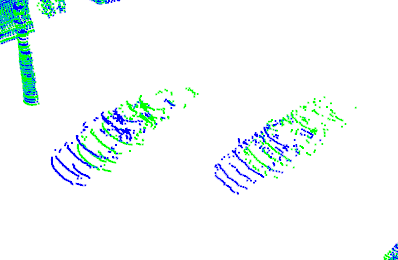}}}};

      \draw[black] (zoom_area_south_west) -- (zoomed_image.south west);
      \draw[black] (zoom_area_north_west) -- (zoomed_image.north west);
    \end{scope}
  \end{tikzpicture}
  \caption{Scene flow estimation of two consecutive point clouds sampled 100 ms apart (green and blue, respectively) on Argoverse 2~\citep{argoverse2}. \textbf{Left:} Ground truth scene flow annotations in red. These annotations are derived from the motion of amodal bounding boxes. \textbf{Right:} \ourmethod{}'s scene flow estimates estimates in orange, which closely match with the ground truth.
  }
  \figlabel{sceneflowexample}
\end{figure}

\section{FAQ}

\subsection{Our method is ``just'' a combination of existing methods using standard distillation. Where does the novelty come in?}

Michael Black argues that ``the simplicity of an idea is often confused with a lack of novelty when exactly the opposite is often true.''~\citep{reviewerguidelines}.
Indeed, we think our novelty comes from the fact that our simple and post-hoc obvious pipeline produces surprisingly good results; our simple pipeline need only consume more raw data to improve and capture state-of-the-art over expensive human supervision while using the same feedforward model architectures. 

\subsection{What are the fundamental insights from this chapter? What new knowledge was generated?}

Beyond producing a useful artifact, our straight-forward pipeline shows that simply training a supervised model with imperfect pseudo-labels can \emph{exceed} the performance of perfect human labels on substantial fraction of the data. We think this is itself surprising, but we also think it has highly impactful implications for the problem of scene flow estimation: \emph{point cloud quantity and diversity is more important than perfect flow label quality for training feedforward scene flow estimators}.

We also think this statement and our empirical scaling laws~(\sectionref{scalinglaws}) lead directly to actionable advice for practitioners at Autonomous Vehicle companies and other organizations with a large trove of diverse point cloud data: \emph{scaling \ourmethod{} on this large scale data will net a significantly better scene flow estimator than expensive human supervision would using a 1,000$\times$ larger budget}.

In addition to insights, we also present a novel scene flow estimation analysis technique. To our knowledge, the residual plots in \sectionref{noise_of_our_method} are the first attempt at visualizing the residual \emph{distribution} of scene flow estimators. We think these plots provide useful insights to practitioners and researchers, particularly for consumption in downstream tasks; as an example, open world object extraction~\citep{objectdetectionmotion} requires the ability to threshold for motion and cluster motion vectors together to extract the entire object. Decreased average EPE is useful for this task, but understanding the \emph{distribution} of flow vectors is needed to craft good extraction heuristics.

\chapter{\MakeUppercase{Evaluating Real World Scene Flow}}\chapterlabel{trackflow}

%

\newcommand{\ourframework}{Scene Flow via Tracking}
\newcommand{\oureval}{Bucket Normalized EPE}

\renewcommand{\ourmethod}{TrackFlow}
\newcommand{\ourmethodbevfusion}{\ourmethod{}BEVF}

\newcommand{\zerovec}{\overrightarrow{0}}

\newcommand{\tinier}{\fontsize{2pt}{2pt}\selectfont}

In \chapterref{zeroflow} we presented ZeroFlow, an unsupervised scene flow method built via scaling up a distillation pipeline  training pipeline to achieve state-of-the-art performance on \emph{Threeway EPE}, a commonly used scene flow evaluation protocol developed by \cite{chodosh2023}. In this work, we start by asking a simple question: what does ZeroFlow's actual scene flow output look like across diverse object types and diverse scenes? What about other strong methods?

Our analysis reveals that ZeroFlow, and other strong scene flow methods, broadly fail to describe the motion of small objects, and existing evaluation protocols hide this failure by averaging error in metric space over many points. To address this limitation, this chapter we propose \emph{\oureval{}}, a new class-aware and speed-normalized evaluation protocol that better contextualizes error comparisons between object types of vastly different sizes that move at vastly different speeds (e.g. cars vs pedestrians). In addition, we propose \emph{\ourmethod{}}, a frustratingly simple supervised scene flow baseline that combines a high-quality 3D object detector (trained using standard class re-balancing techniques) with a simple Kalman filter-based tracker. Notably, \emph{\ourmethod{}} achieves state-of-the-art performance on existing metrics and shows large improvements over prior work on our Bucket Normalized EPE metric. Our results highlight that scene flow evaluation must be class and speed aware, and supervised scene flow methods should consider point-level class imbalances. Our evaluation toolkit and code is available at \texttt{\url{https://github.com/kylevedder/BucketedSceneFlowEval}}.

\section{Overview}

{\bf Status Quo.} Standard scene flow metrics suggest that existing methods can estimate motion to centimeter-level accuracy. 
For example, ZeroFlow XL 5x~\citep{vedder2024zeroflow} achieves 
an average Threeway EPE~\citep{chodosh2023} of only 4.9 centimeters (1.9 inches) and a Dynamic EPE (averaged over points moving faster than 0.5 m/s)  of 11.7 centimeters (4.6 inches). Notably, these errors are relatively small compared to the scale of cars and pedestrians, implying that current scene flow methods produce high quality flow. On the scale of cars and people, these \emph{feel} like tiny errors and seem to imply that current scene flow methods are high quality.

\begin{figure}[h!]
\centering
\begin{subfigure}{.32\textwidth}
  \centering
  \includegraphics[width=\linewidth]{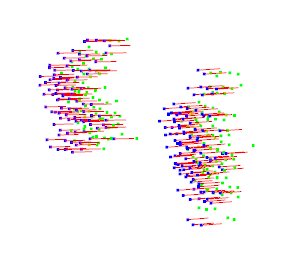}
  \caption{Ground Truth}
  \label{fig:sub5}
\end{subfigure}%
\begin{subfigure}{.32\textwidth}
  \centering
  \includegraphics[width=\linewidth]{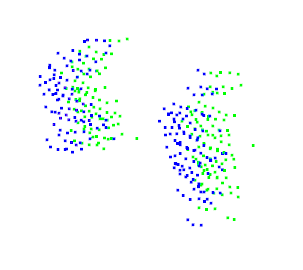}
  \caption{FastFlow3D~\citep{scalablesceneflow}}
  \label{fig:sub2}
\end{subfigure}
\begin{subfigure}{.32\textwidth}
  \centering
  \includegraphics[width=\linewidth]{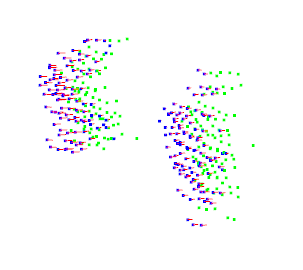}
  \caption{DeFlow~\citep{zhang2024deflow}}
  \label{fig:sub3}
\end{subfigure}

\begin{subfigure}{.32\textwidth}
  \centering
  \includegraphics[width=\linewidth]{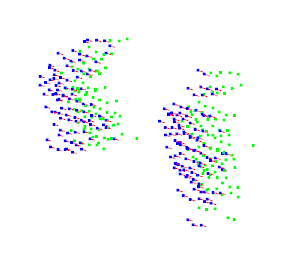}
  \caption{NSFP~\citep{nsfp}}
  \label{fig:sub4}
\end{subfigure}%
\begin{subfigure}{.32\textwidth}
  \centering
  \includegraphics[width=\linewidth]{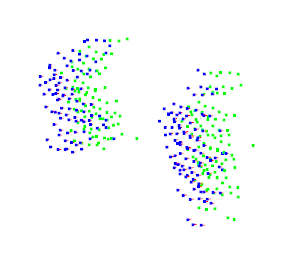}
  \caption{ZeroFlow XL 5x~\citep{vedder2024zeroflow}}
  \label{fig:sub1}
\end{subfigure}%
\begin{subfigure}{.32\textwidth}
  \centering
  \includegraphics[width=\linewidth]{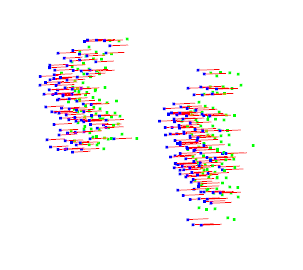}
  \caption{\textbf{\ourmethod{} (Ours)}}
  \label{fig:sub6}
\end{subfigure}
\caption{We visualize an example of two pedestrians (walking from \blue{left} to \green{right}), cherry-picked to have unusually high density lidar returns, making it particularly easy to estimate flow. We expect that state-of-the-art scene flow methods should work well in this case, but find that all prior art fails catastrophically. Notably, \ourmethod{} is the only method to estimate flow for these pedestrians.}
\figlabel{teaserfigure}
\end{figure} 

{\bf \oureval{}.}  We visualize flow predictions from several state-of-the-art supervised (FastFlow3D~\citep{scalablesceneflow}, DeFlow~\citep{zhang2024deflow}) and unsupervised (NSFP~\citep{nsfp}, ZeroFlow~\citep{vedder2024zeroflow}) approaches and find that all methods underestimate flow for small objects (\figref{teaserfigure}) with fewer lidar points (e.g. pedestrians and bicyclists). 
Surprisingly, existing scene flow metrics do not highlight such failure cases on these safety-critical categories because small objects only make up a tiny fraction of the dynamic points in a scene (\figref{fig:pointdistribution}).
To address this limitation, we propose \emph{\oureval{}}, a new evaluation protocol that allows us to directly measure performance disparities across classes of different sizes and speed profiles. Specifically, \emph{\oureval{}} evaluates the \emph{percentage} of described motion, allowing us to normalize comparisons between objects moving at different speeds. 
Our proposed evaluation metric takes inspiration from \emph{mean Average Precision} (mAP), a metric commonly used to evaluate object detectors. Notably, unlike existing scene flow metrics, mAP equally weights the performance of large common objects like cars and small rare objects like strollers. Therefore, state-of-the-art 3D object detectors use data augmentation and class re-balancing techniques~\citep{cbgs} to perform well on both common and rare classes. 

{\bf \ourmethod{}.} Based on this observation, we propose \ourmethod{}, a frustratingly simple baseline that generates scene flow estimates using rigid transformations to describe point-level motion within a 3D object track.  Specifically, we run a state-of-the-art 3D object detector~\citep{wang2023le3de2e} followed by a simple 3D Kalman filter-based tracker~\citep{Weng2020_AB3DMOT} to generate object trajectories. 
Despite its simplicity,  \emph{\ourmethod{}} achieves state-of-the-art performance on the standard metric of Threeway EPE. Even more importantly, \ourmethod{} significantly outperforms prior art on our \oureval{}, reducing error by 10\% across all classes, and 20\% on pedestrians in particular. Importantly, our simple baseline's state-of-the-art performance is an indictment of existing supervised scene flow methods. We argue that utilizing (well established) class re-balancing techniques can improve performance on rare safety-critical categories in real-world datasets, and evaluating scene flow methods using class and speed-aware metrics more closely reflects real-world performance. 

{\bf Contributions}. In this chapter we present three primary contributions.
\begin{enumerate}
  \item We highlight the qualitative failure of state-of-the-art scene flow methods on safety-critical categories like pedestrians and bicycles.
  \item We introduce \emph{\oureval{}}, a new evaluation protocol that allows us to quantify this qualitative failure on small objects.
  \item We propose \emph{\ourmethod{}}, a frustratingly simple baseline that achieves state-of-the-art performance on the standard metric of Threeway EPE. Additionally, it significantly outperforms prior art on our class-aware \emph{\oureval{}} metric.
\end{enumerate}

\section{Related Work}

\subsection{Scene Flow Datasets and Ground Truth}\sectionlabel{sceneflowdatasets}
Unlike next token prediction in language~\citep{gpt} or next frame prediction in vision~\citep{weng2021inverting}, scene flow is not na\"ively self-supervised: future observations do not provide ground truth scene flow. Therefore, ground truth motion descriptions must be provided by an oracle, typically from human annotators for real data~\citep{kittisceneflow1, kittisceneflow2, waymoopen, argoverse2, caesar2020nuscenes} or a data generator for synthetic datasets~\citep{flyingthings,zheng2023point}. For real world datasets (typically from the autonomous vehicle domain) human annotations are provided in the form of 3D bounding boxes and tracks for every object in the scene~\citep{chodosh2023}. Consequently, the generated ground truth flow is assumed to be rigid, even in the case of non-rigid motion like pedestrian gaits.

\subsection{Evaluating Scene Flow}
Given point clouds $\pointcloudt{}$ and $\pointcloudtpone{}$, scene flow estimators predict $\flowttpone{}$,  a 3D vector per point in $\pointcloudt$ that describes its motion from $t$ to $t+1$ \citep{dewan2016rigid}. Performance is typically measured using Average Endpoint Error (EPE) which is the $L_2$ norm between the predicted ($\flowttpone$) and ground truth flow ($\flowgtttpone$), as in \equationref{averageepedef}.

\begin{equation}
  \small
  \equationlabel{averageepedef}
  \textup{Average EPE}\left({\pointcloudt} \right) = \frac{1}{\norm{\pointcloudt}} \sum_{p \in \pointcloudt} \norm{\flowttpone{}(p) - \flowgtttpone{}(p)}_2.
\end{equation}

In real-world scenes, most points belong to the static background. Consequently, simply computing Average EPE (\equationref{averageepedef}) over all points is dominated by background points. 
In order to separately measure non-ego dynamics, Chodosh et al.~\citep{chodosh2023} introduces \emph{Threeway EPE}, which computes a mean over the Average EPE for three disjoint classes of points: \emph{Foreground Dynamic} (points inside bounding box labels moving greater than 0.5m/s), \emph{Foreground Static} (points inside bounding box labels moving less than 0.5m/s), and \emph{Background Static}. In \sectionref{eval} we extend Threeway EPE to consider different class and speed profiles, producing \emph{Bucket Normalized EPE}.

\subsection{3D Object Detection and 
Tracking}\sectionlabel{detectorrelatedwork}

Object detectors have advanced techniques for training with imbalanced datasets. Notably, modern object detectors use carefully designed losses to mitigate foreground-vs-background imbalances in proposal generation, and data augmentation strategies to train with long-tailed taxonomies. Existing methods address imbalanced foreground-vs-background region proposals using Focal Loss~\citep{focalloss} to upweight the importance of foreground regions. More recently, 3D object detectors use class-balanced sampling~\citep{cbgs} and copy-paste augmentation to upsample and rebalance the distribution of examples per class. In addition, state-of-the-art 3D object detectors take advantage of multi-modal data to improve detection \citep{vora2020pointpainting, peri2022towards, ma2023long} of small and rare categories. Since many state-of-the-art tracking algorithms \citep{Weng2020_AB3DMOT} follow the tracking-by-detection paradigm, improving detection quality also significantly improves tracking performance.

\section{\emph{\oureval{}}: Small Objects (Should) Matter in Scene Flow}\sectionlabel{eval}

As shown in \figref{teaserfigure} (and further in \figref{morequalitativeone}), existing scene flow methods consistently struggle to describe the motion of safety-critical objects like pedestrians. However, these failures are not captured by Threeway EPE \emph{because} these objects are small and have few points. Specifically, Threeway EPE's \emph{Foreground Dynamic} category is dominated by large, common objects with many points like cars and other vehicles. As shown in \figref{fig:pointdistribution}, 15\% of all points are from cars or other vehicles (dominating \emph{Foreground Dynamic}'s Average EPE), while fewer than 1\% of points are from pedestrians and other vulnerable road users (VRUs). 

\begin{figure}[t]
\centering
\includegraphics{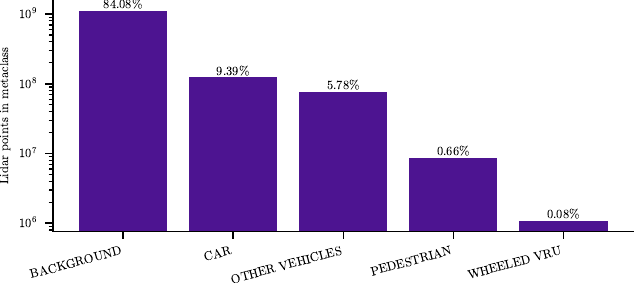}
\caption{Number of points from each semantic meta-class for Argoverse 2's \emph{val} split. Although \texttt{PEDESTRIAN} instances are common, they contribute less than 1\% of the total number of points owing to their small instance size relative to \texttt{CAR} and \texttt{OTHER VEHICLES}. Number of points (Y axis) shown on a log scale.}
\figlabel{fig:pointdistribution}
\end{figure}

Additionally, Threeway EPE fails to account for large differences in speed  across objects. For example, a 0.5~m/s estimation error on a car moving 20~m/s is negligible (<2.5\%), while a 0.5~m/s estimation error on a pedestrian moving 0.5~m/s fails to describe 100\% of the pedestrian's motion. However, Threeway EPE treats both estimation errors equally.

We address these two limitations with our \emph{\oureval{}} metric. First, our proposed metric breaks down the object distribution using a taxonomy that human labelers have deemed important (similar to \emph{mean Average Precision}~\citep{cocodataset}, see \appendixref{semanticsfree} for discussion on semantics-free evaluation). Second, our proposed metric allows us to contextualize the percentage of object motion being described by normalizing for the speed of the object, allowing us to directly compare performance across object categories.


{
\setlength{\tabcolsep}{1.4em}

\begin{table}[t]
\centering
\scalebox{0.99}{
\begin{tabular}{lcc}
\toprule
Class & Static (Avg EPE) & Dynamic (Norm EPE) \\
\midrule
\texttt{BACKGROUND} & 0.002402 & - \\
\texttt{CAR} & 0.018442 & 0.182092 \\
\texttt{OTHER VEHICLES} & 0.081475 & 0.312882 \\
\texttt{PEDESTRIAN} & 0.052842 & 0.396849 \\
\texttt{WHEELED VRU} & 0.062573 & 0.257647 \\
\bottomrule
\end{tabular}
}
\vspace{5mm}
\caption{\ourmethod{}'s \oureval{} on the Argoverse 2 \emph{test} split. Similar to Threeway EPE, we breakdown our evaluation into static and dynamic buckets. However, we also further breakdown performance by meta-categories and normalize by speed to compare performance disparities on safety-critical categories. TrackFlow is able to capture most dynamic car motion (lower is better), but performs considerably worse on other vehicles and pedestrian.}
\tablelabel{tab:tracktorflowresults}
\end{table}
}

We implement our class-aware and speed-normalized metric by accumulating every point into a class-speed matrix (e.g.\ \tableref{speedclassmatrix}) based on its ground truth speed and class, recording an Average EPE as well as a per-bucket average speed. To summarize these results, we report two numbers per class:

\begin{itemize}
    \item \emph{Static EPE}, taken directly from the Average EPE of the first speed bucket for that class (i.e.\ the first column of \appendixref{ourevalstructure}, \tableref{speedclassmatrix})
    \item \emph{Dynamic Normalized EPE}, computed from a mean over the Normalized EPE ($\frac{\textup{Average EPE}}{\textup{average speed}}$) of each non-empty speed bucket (i.e.\ an average across the Normalized EPEs of the second column onwards in \tableref{speedclassmatrix})
\end{itemize}

\begin{table}[h!]
\centering
\begin{tabular}{l|ccccc}
\toprule
Class & \multicolumn{5}{c}{Speed Columns} \\
\midrule 
 & 0-0.4m/s & 0.4-0.8m/s & 0.8-1.2m/s & ... & 20-$\infty$m/s \\
\midrule
\texttt{BACKGROUND} & - & - & - & - & - \\
\texttt{CAR} & - & - & - & - & - \\
\texttt{OTHER VEHICLES} & - & - & - & - & - \\
\texttt{PEDESTRIAN} & - & - & - & - & - \\
\texttt{WHEELED VRU} & - & - & - & - & - \\
\bottomrule
\end{tabular}
\caption{Example of \oureval{}'s class-speed matrix.}
\tablelabel{speedclassmatrix}
\end{table}

\emph{Dynamic Normalized EPE} measures the fraction of motion {\em not} described by the estimated flow vectors across the entire speed spectrum. A method that only predicts ego motion (e.g.\ $\zerovec$ after ego-motion compensation) will achieve 1.0 Dynamic Normalized EPE, and a method that perfectly describes all motion will have 0.0 Dynamic Normalized EPE. Methods may achieve errors greater than 1.0 by predicting errors with a magnitude greater than the average speed. For example, a method that describes the negative vector of true motion will get exactly 2.0 Dynamic Normalized EPE (every bucket's Average EPE will be exactly 2$\times$ the magnitude of the average speed). The range of Dynamic Normalized EPE is between 0 (perfect) and $\infty$, and is undefined for buckets without any points. After normalization, Dynamic Normalized EPE can be directly compared across classes.

We provide an example per-class performance breakdown in \tableref{tab:tracktorflowresults} for \ourmethod{} (\sectionref{method}). Results can be further summarized into a single tuple of \emph{mean Static EPE} and \emph{mean Dynamic Normalized EPE} by taking a mean across classes (similar to \emph{mean Average Precision}~\citep{cocodataset}). \ourmethod{} has a mean Static EPE of 0.076277 and a mean Dynamic Normalized EPE of 0.287368. We rank methods according to their mean Dynamic Normalized EPE. 

\textbf{\oureval{} Without Semantics.} While this makes sense to use semantics to break down the object distribution when they are available, this is not a fundamental requirement for \oureval{}. To demonstrate this, we break down Argoverse 2's bounding boxes by \emph{size} instead of semantics. We group the ground truth boxes into one of three volume based clusters: \texttt{SMALL}: $<9.5m^3$, \texttt{MEDIUM}:  $\ge 9.5m^3 \land <40m^3$, or \texttt{LARGE}: $\ge40m^3$. As shows in \figref{semanticsfree}, this distribution breakdown still highlights the poor performance of prior art on small objects. 

In \sectionref{eval} we present \oureval{} with the object distribution broken down by semantic classes. 

\begin{figure}[h!]
\centering
\vspace{-1em}
\begin{subfigure}[b]{0.32\textwidth}
    \centering
    \includegraphics[width=\textwidth]{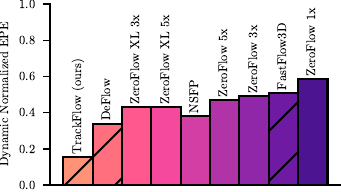}
    \caption{\texttt{SMALL}}
\end{subfigure}%
\begin{subfigure}[b]{0.32\textwidth}
    \centering
    \includegraphics[width=\textwidth]{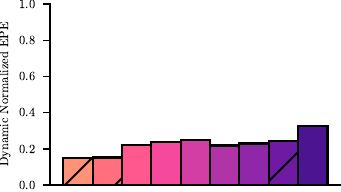}
    \caption{\texttt{MEDIUM}}
\end{subfigure}
\begin{subfigure}[b]{0.32\textwidth}
    \centering
    \includegraphics[width=\textwidth]{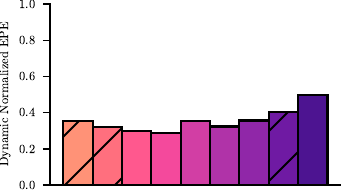}
    \caption{\texttt{LARGE}}
\end{subfigure}%
\caption{\oureval{} using ground truth size based clustering.}
\figlabel{semanticsfree}
\end{figure}

\section{\emph{\ourmethod{}}: \ourframework{}}\sectionlabel{method}

 To highlight the failure of current supervised scene flow methods on smaller objects, we propose \emph{\ourframework{}}, a simple framework that uses bounding box track motion from off-the-shelf 3D detectors and trackers to generate scene flow estimates (\figref{fig:detectplustrack}).
 We instantiate \emph{\ourframework{}} with LE3DE2E~\citep{wang2023le3de2e}\footnote{LE3DE2E~\citep{wang2023le3de2e} is the winning method from the \emph{Argoverse 2 2023 3D Detection, Tracking and Forecasting challenge}~\citep{peri2022towards, peri2023empirical, Peri_2022_CVPR}.}, a state-of-the-art 3D detector, and AB3DMOT~\citep{Weng2020_AB3DMOT}, a Kalman filter-based 3D tracker. As shown in \sectionref{experiments}, \emph{\ourmethod{}} achieves state-of-the-art performance on Threeway EPE and beats all prior art by a large margin on \oureval{}.

\begin{figure}[t]
    \centering
    \includegraphics[width=\textwidth]{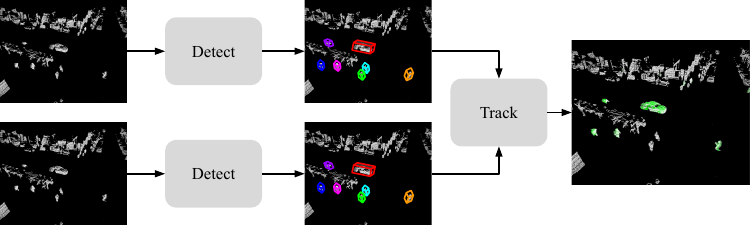}
    \caption{Overview of the \emph{\ourframework{}} framework. Our proposed framework generates scene flow estimates using rigid transformations to describe point-level motion within a 3D object track.}
    \figlabel{fig:detectplustrack}
\end{figure}

\ourframework{} works well in practice because it mimics the ground truth flow annotation process. Specifically, ground truth flow is generated using rigid transforms to describe point-level motion within ground truth 3D object tracks
(\sectionref{sceneflowdatasets}). Therefore, a perfect 3D detector and tracker will achieve perfect flow. However, the power of \ourmethod{} is not just derived from its use of bounding boxes; it also greatly benefits from recent advances in class-imbalanced learning~\citep{cbgs}. As discussed in \sectionref{detectorrelatedwork}, modern detectors are trained with a variety of data augmentation techniques to achieve high precision and recall on all semantic class. \ourmethod{} leverages the strength of modern 3D detectors to significantly outperform prior art on pedestrians and other small objects.

Interestingly, we find that the \ourframework{} framework performs best when using detectors tuned to a low confidence threshold. Typically, detectors are optimized to only predict high confidence boxes (0.7 - 0.9) to minimize the number of false positives. However, our method works best when setting the confidence threshold lower (0.2 for \ourmethod{}) to increase recall. Specifically, we find that detectors with higher recall and more accurate heading estimation are better suited for \ourframework{}. We explore detector choice and ablate the impact of confidence thresholds further in \sectionref{detectorquality}.

\section{TrackFlow as a Scene Flow Method and a Measuring Device}\sectionlabel{experiments}

In this section, 
we compare \ourmethod{} against state-of-the-art supervised and unsupervised scene flow methods like FastFlow3D~\citep{scalablesceneflow}, DeFlow~\citep{zhang2024deflow}, NSFP~\citep{nsfp}, and ZeroFlow~\citep{vedder2024zeroflow} on the Argoverse 2 benchmark~\citep{argoverse2}\footnote{All evaluations are performed with a maximum radius of 35m from the ego vehicle to maintain consistency with Chodosh et al.~\citep{chodosh2023}.}.

\begin{figure}[t]
\centering
\begin{subfigure}[b]{0.49\textwidth}
    \includegraphics{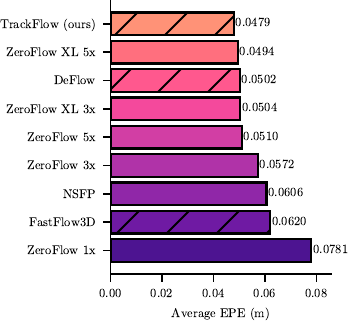}  
    \caption{Threeway EPE}
    \figlabel{fig:threewayepesub}
\end{subfigure}%
\begin{subfigure}[b]{0.49\textwidth}
    \includegraphics{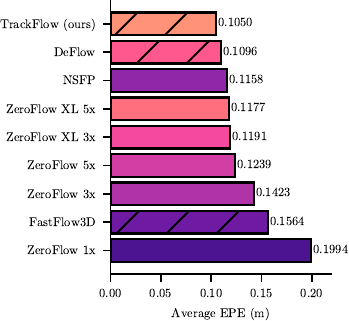}    
    \caption{Threeway EPE's \emph{Foreground Dynamic}}
    \figlabel{fig:threewayepedynamicsub}
\end{subfigure}%
\caption{\emph{Threeway EPE} and \emph{Threeway EPE's Foreground Dynamic} performance of recent supervised and unsupervised scene flow  methods on Argoverse 2's \emph{test} split. Supervised methods shown with hatching. Lower is better. Method color is consistent between plots. We find that all recent methods achieve 5cm error on Threeway EPE, suggesting that these approaches work well in-the-wild. However, this number hides the failure of these methods to describe small object motion.}
\figlabel{fig:threewayepe}
\end{figure}

\begin{figure}[t]
\centering
\begin{subfigure}[b]{0.49\textwidth}
    \centering
    \includegraphics{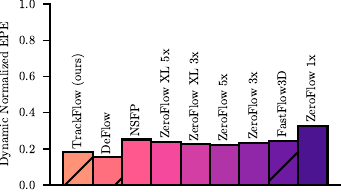}
    \caption{\texttt{CAR}}
    \figlabel{fig:car}
\end{subfigure}%
\begin{subfigure}[b]{0.49\textwidth}
    \centering
    \includegraphics{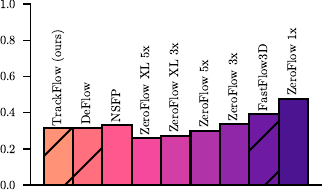}
    \caption{\texttt{OTHER VEHICLES}}
    \figlabel{fig:other-vehicles}
\end{subfigure}
\begin{subfigure}[b]{0.49\textwidth}
    \centering
    \includegraphics{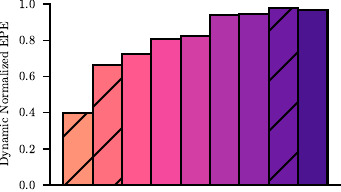}
    \caption{\texttt{PEDESTRIAN}}
    \figlabel{fig:pedestrian}
\end{subfigure}%
\begin{subfigure}[b]{0.49\textwidth}
    \centering
    \includegraphics{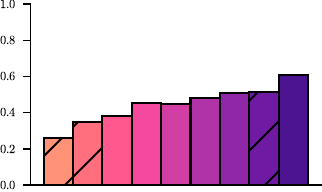}
    \caption{\texttt{WHEELED VRU}}
    \figlabel{fig:wheeled-vru}
\end{subfigure}
\caption{Per meta-class Dynamic Normalized EPE of recent supervised and unsupervised scene flow estimation methods on Argoverse 2's \emph{test} split. Supervised methods shown with hatching. Lower is better. Method color and position is consistent between plots. \ourmethod{} significantly improves over prior work on both pedestrian and wheeled VRUs. Notably, \oureval{} quantitatively demonstrates significant method performance differences not highlighted in Threeway EPE.}
\figlabel{metacatagorydynamic}
\end{figure}

\begin{figure}[t]
\centering
\includegraphics{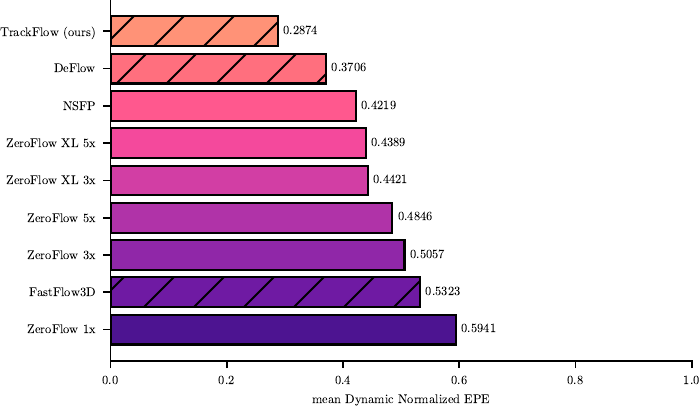}
\caption{Average Dynamic Normalized EPE of recent supervised and unsupervised scene flow estimation methods on Argoverse 2's \emph{test} split. Supervised methods shown with hatching. Lower is better. Our simple baseline achieves state-of-the-art performance, suggesting that supervised scene flow methods should embrace point-level class re-balancing.}
\figlabel{fig:meandynamicepe}
\end{figure}

{
\setlength{\tabcolsep}{1.4em}

\begin{table}[t]
\centering
\scalebox{0.99}{
\begin{tabular}{lcc}
\toprule
Class & Static (Avg EPE) & Dynamic (Norm EPE) \\
\midrule
\texttt{BACKGROUND} & \green{-0.000228} & - \\
\texttt{CAR} & \red{+0.039049} & \red{+0.117944} \\
\texttt{OTHER VEHICLES} & \red{+0.009013} & \red{+0.224830} \\
\texttt{PEDESTRIAN} & \red{+0.007187} & \red{+0.224250} \\
\texttt{WHEELED VRU} & \green{-0.025889} & \red{+0.151373} \\
\bottomrule
\end{tabular}
}
\vspace{5mm}
\caption{Relative \oureval{} performance of \ourmethodbevfusion{} compared to \ourmethod{}, on the Argoverse 2's \emph{test} split. Increases in error (worse) are shown with a + in \red{red}, and decreases in error (better) are shown with a - in \green{green}. \ourmethod{}'s absolute results are shown in \tableref{tab:tracktorflowresults}.BEVFusion only has 2\% lower mAP than LE3DE2E on the AV2 detection leaderboard, but performs significantly worse than \ourmethod{} on Dynamic Normalized EPE.}
\tablelabel{tab:bevfusionresults}
\end{table}
}

{
\setlength{\tabcolsep}{3.35em}

\begin{table}[htb]
\centering
\scalebox{0.99}{
\begin{tabular}{cc}
\hline
Confidence & Mean Dynamic Norm EPE \\ \hline
0.1        & 0.4816             \\
0.2        & 0.4643             \\
0.3        & 0.6008             \\
0.4        & 0.8176             \\ \hline
\end{tabular}
}
\vspace{5mm}
\caption{Mean Dynamic Bucketed EPE values for TrackFlowBEVF using various confidence thresholds for the detector. Lower confidences with higher recall significantly improve Dynamic Norm EPE performance.}
\tablelabel{tab:bevfusionconf}
\end{table}
}

\begin{figure}[h!]
    \centering
    \includegraphics[width=\textwidth]{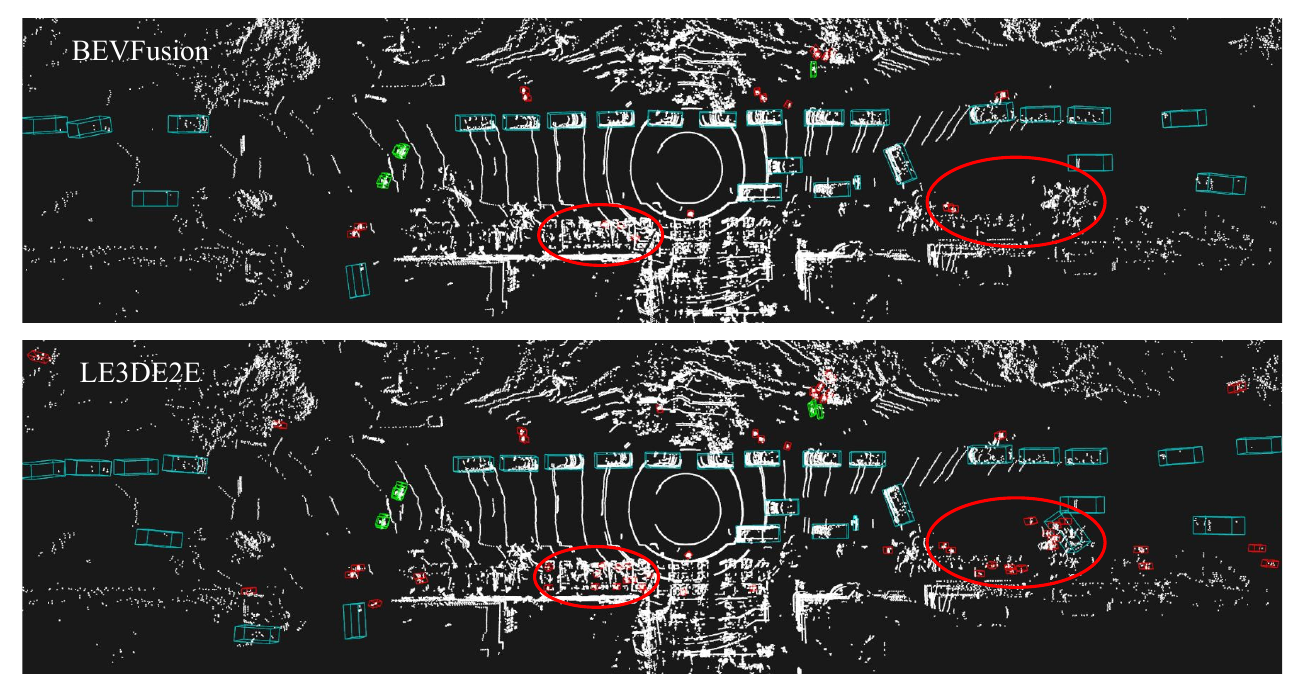}
    \caption{A qualitative comparison of the recall of BEVFusion and LE3DE2E. LE3DE2E has much higher recall, allowing it to pick out pedestrians BEVFusion missed (circled in red), and better quality box heading estimates. Both detectors are using a confidence threshold of 0.2.}
    \figlabel{fig:bevfusionrecall}
    \vspace{-2em}
\end{figure}

\subsection{\emph{\ourmethod{}} Achieves SOTA Performance on Threeway EPE}

\ourmethod{} is state-of-the-art on Threeway EPE (\figref{fig:threewayepesub}) on the Argoverse 2 benchmark \citep{argoverse2}, achieving an overall reduction of 0.0015m (0.15cm, or 1.5mm) over the next best method, ZeroFlow XL 5x. Notably, this improvement can be attributed to significantly better Threeway EPE's \emph{Dynamic Foreground} (\figref{fig:threewayepedynamicsub}). Is this performance difference meaningful? 

Based on our reduction of 1.5mm on Threeway EPE (about 4$\times$ the thickness of a human fingernail), it would seem that \ourmethod{} is only an incremental improvement over prior art. However, \ourmethod{} qualitatively outperforms prior work on important small objects such as pedestrians (\figref{teaserfigure}, \figref{morequalitativeone}). As shown in the next section, our proposed evaluation protocol \emph{\oureval{}} makes it quantitatively clear that \ourmethod{} performs significantly better on safety-critical categories like pedestrians and VRUs.


\subsection{\emph{\oureval{}} Highlights Failures on Small Objects}

Evaluating existing state-of-the-art methods on our class-aware, speed-normalized evaluation, \oureval{}, makes it clear that \ourmethod{} meaningfully outperforms prior art (\figref{fig:meandynamicepe}) --- \ourmethod{} correctly describes almost 10\% additional total motion across meta-classes compared to  DeFlow~\citep{zhang2024deflow}. This difference in dynamic performance becomes even more clear when broken down by meta-class: \figref{metacatagorydynamic} shows that \ourmethod{} is the only method able to describe more than 50\% of pedestrian motion, beating DeFlow~\citep{zhang2024deflow} by more than  20\% (\figref{fig:pedestrian}), a $1.5\times$ improvement. Similarly, other state-of-the-art methods like NSFP~\citep{nsfp} and ZeroFlow XL 5x~\citep{vedder2024zeroflow} describe less than 30\% and 20\% of pedestrian motion, respectively.

\oureval{} allows practitioners to effectively compare performance between methods that were almost indistinguishable under Threeway EPE. For example, if you only care about flow performance on cars, DeFlow out-performs all other methods including \ourmethod{} (\figref{fig:car}), while ZeroFlow XL 5x out-performs all other methods on larger vehicles (\figref{fig:other-vehicles}). 

\subsection{What Makes a Good Detector for \ourmethod{}?}\sectionlabel{detectorquality}

As discussed in \sectionref{method}, we tune \emph{\ourframework{}} with a low confidence threshold to maximize recall. What makes a good detector for \ourmethod{}?


\begin{figure}[htbp]
\centering
\begin{subfigure}{.16\textwidth}
  \centering
  \includegraphics[width=\linewidth]{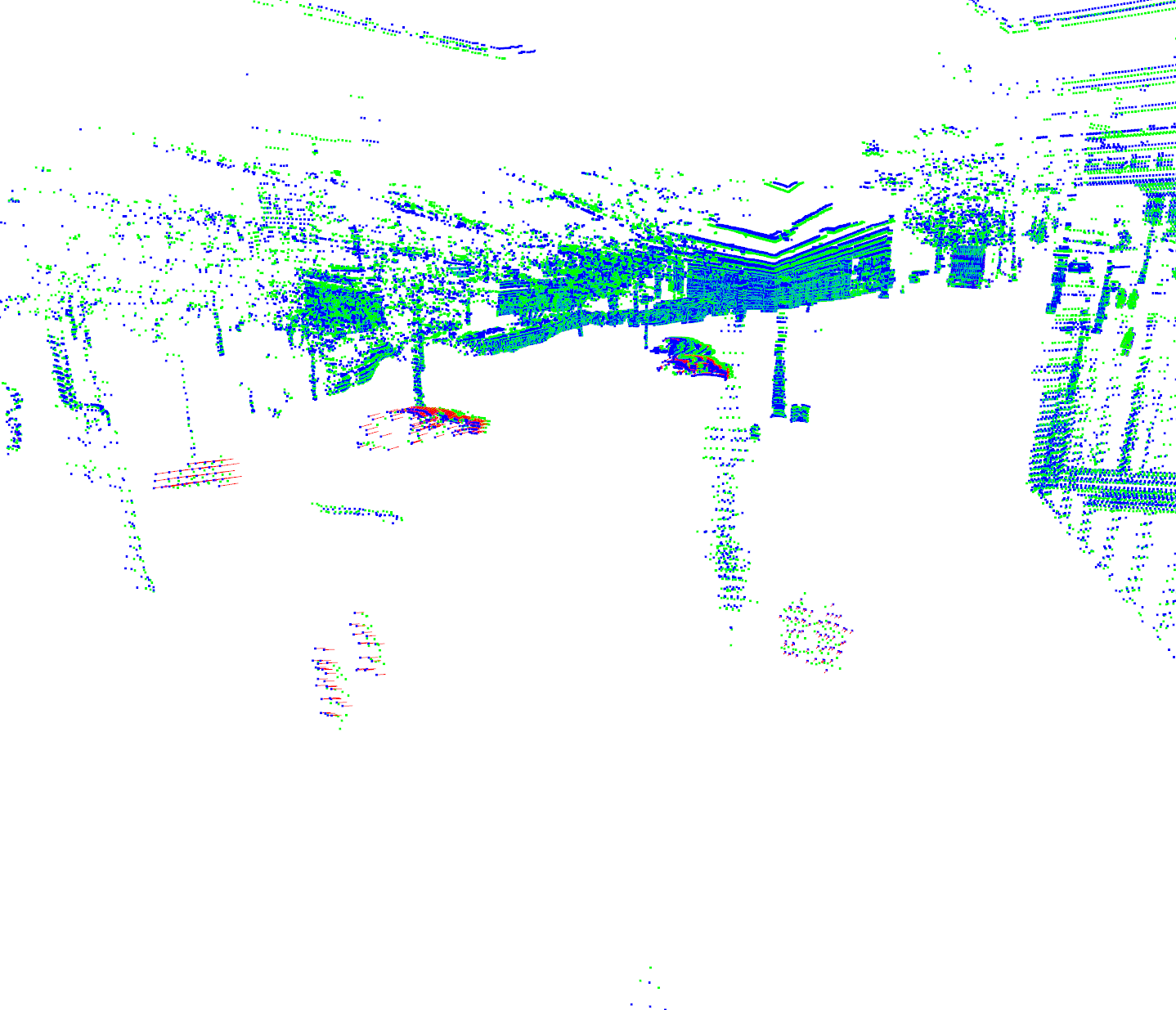}
  \includegraphics[width=\linewidth]{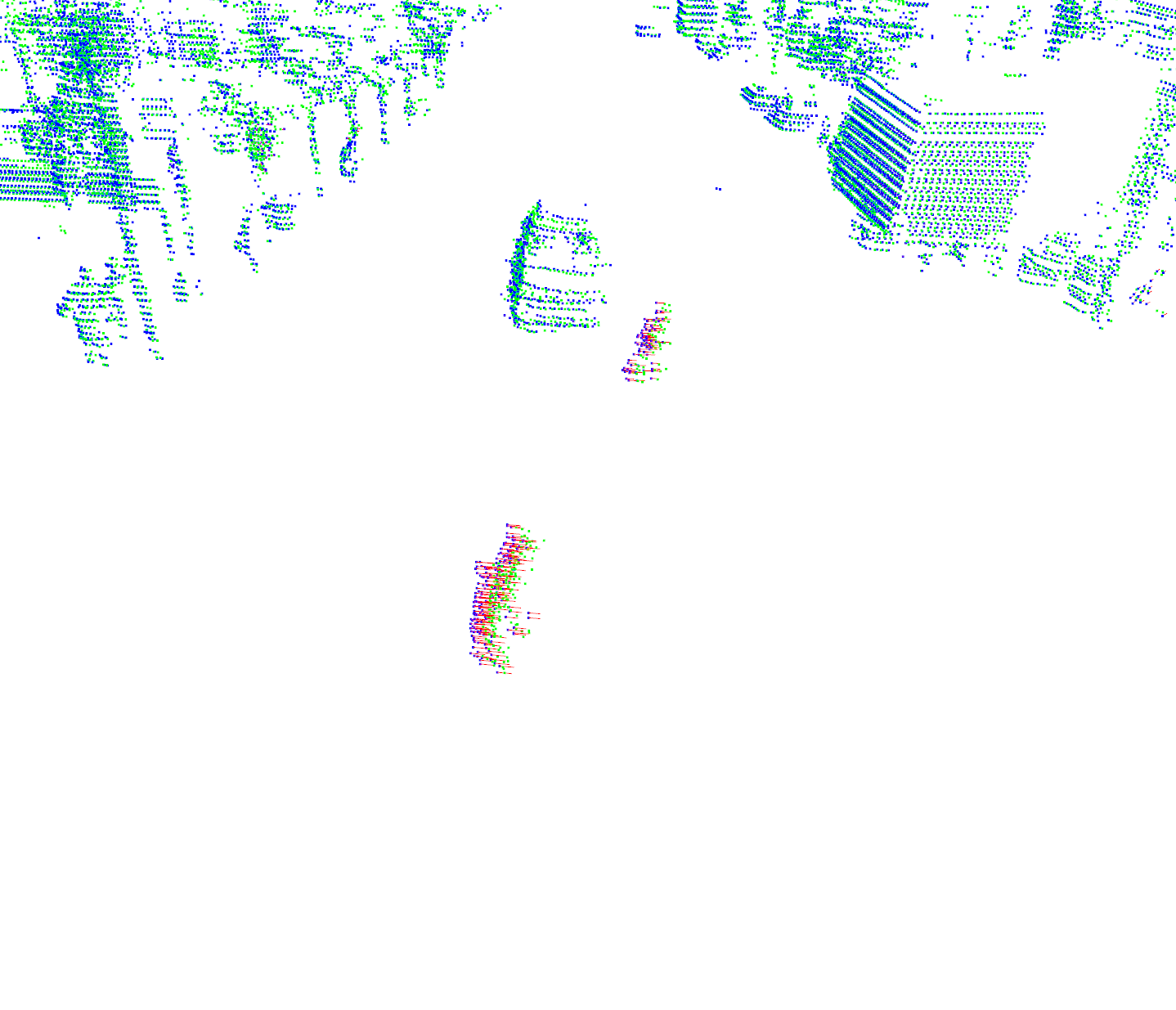}
  \includegraphics[width=\linewidth]{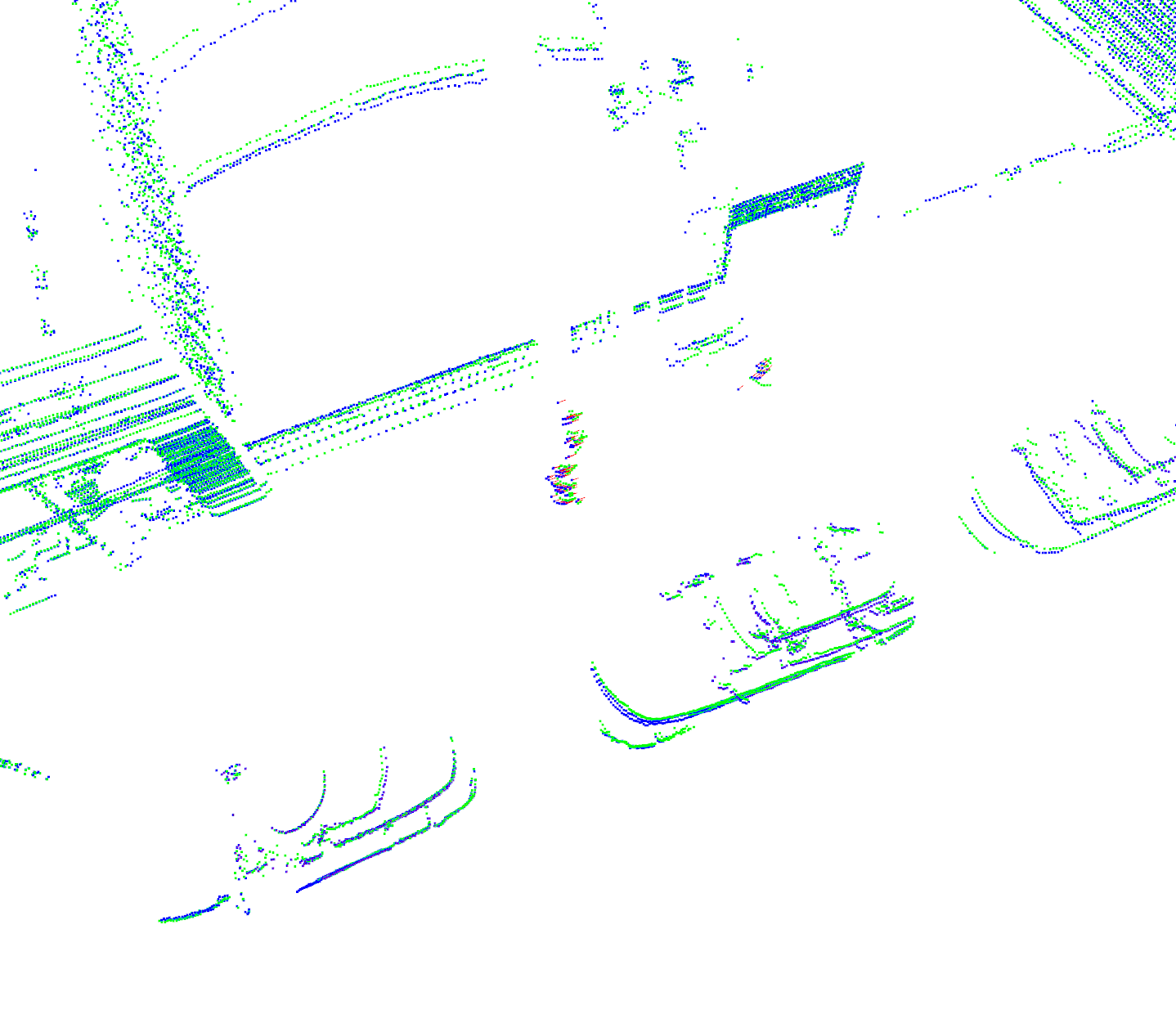}
  \includegraphics[width=\linewidth]{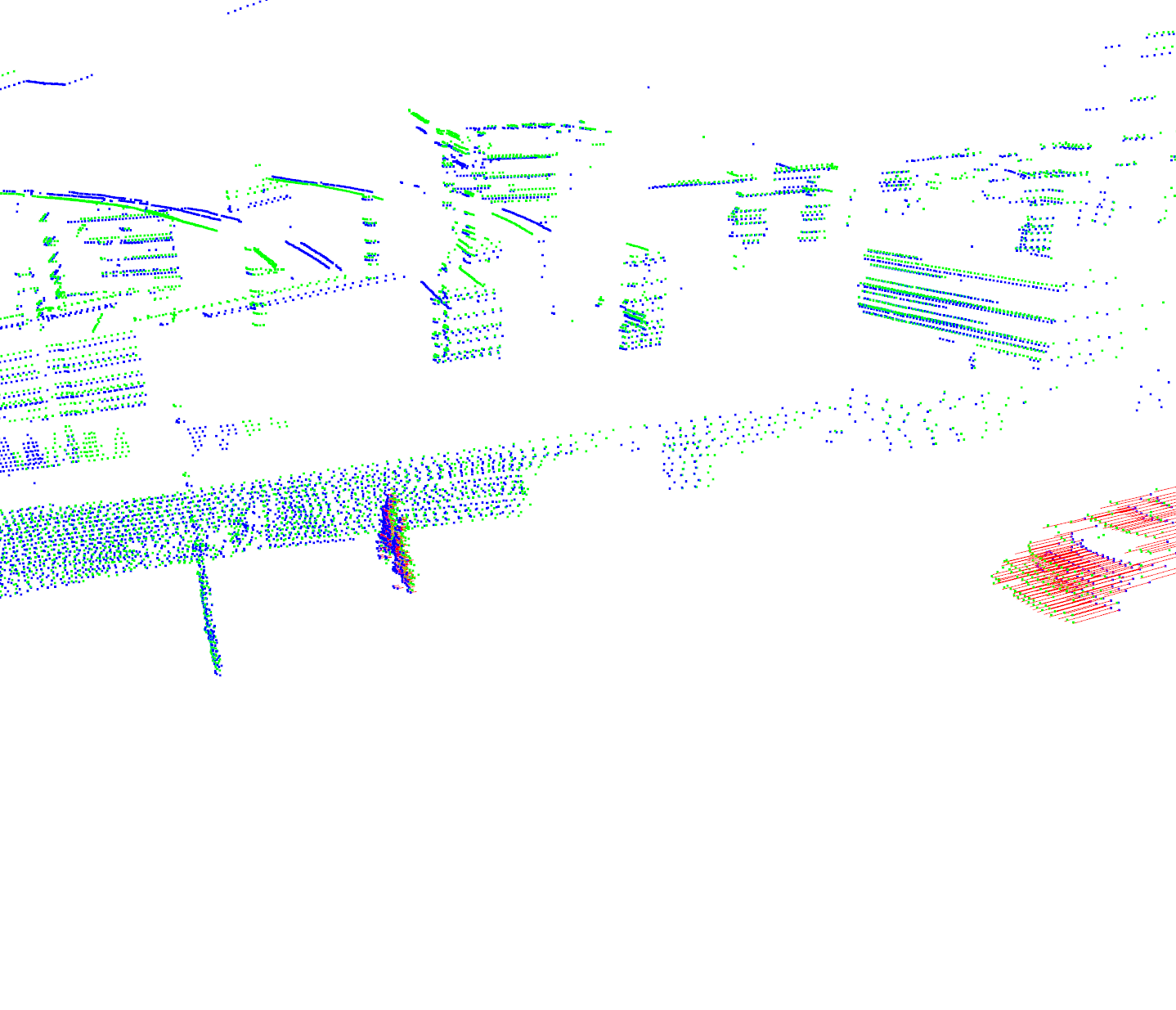}
  \includegraphics[width=\linewidth]{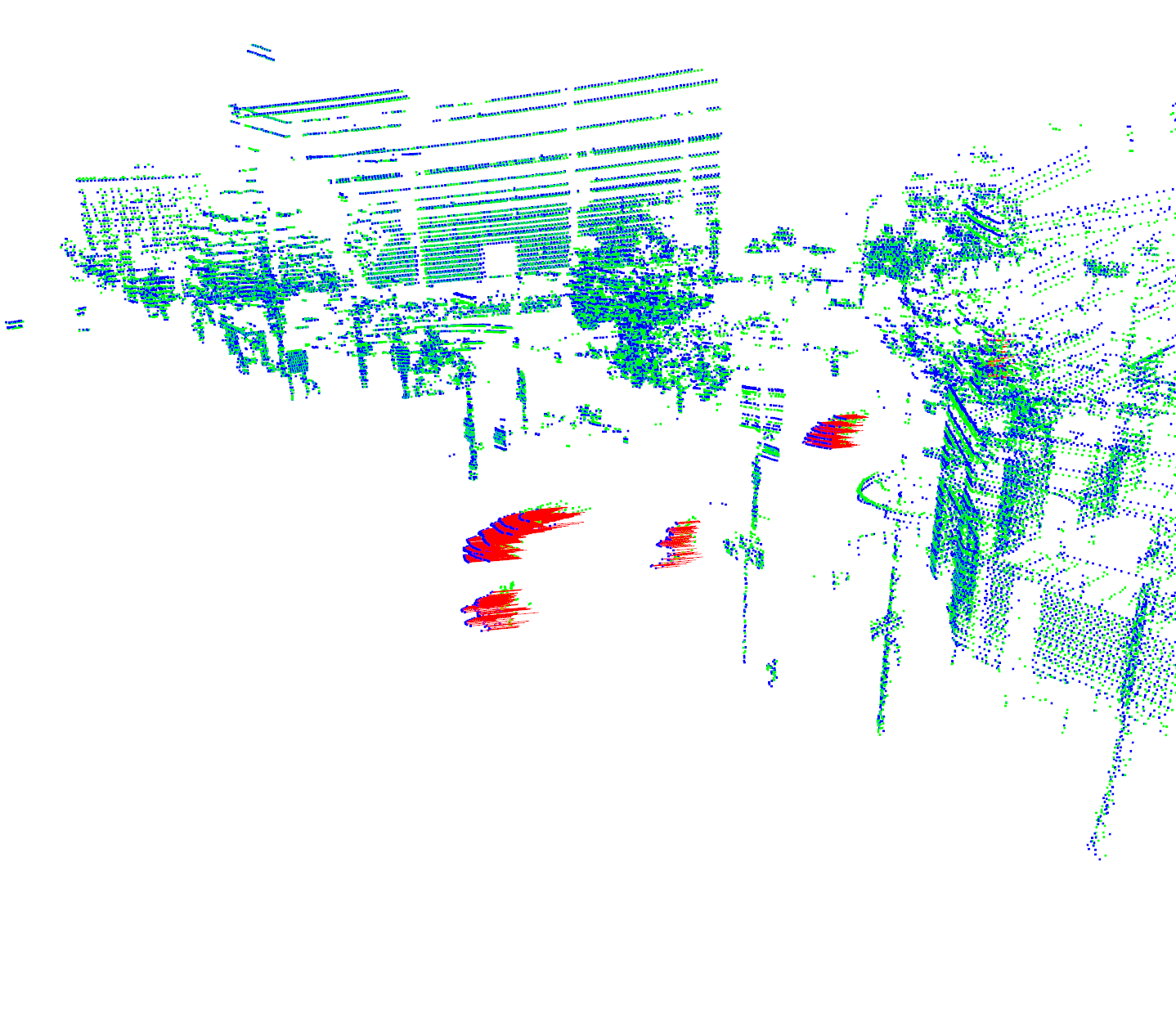}
  \includegraphics[width=\linewidth]{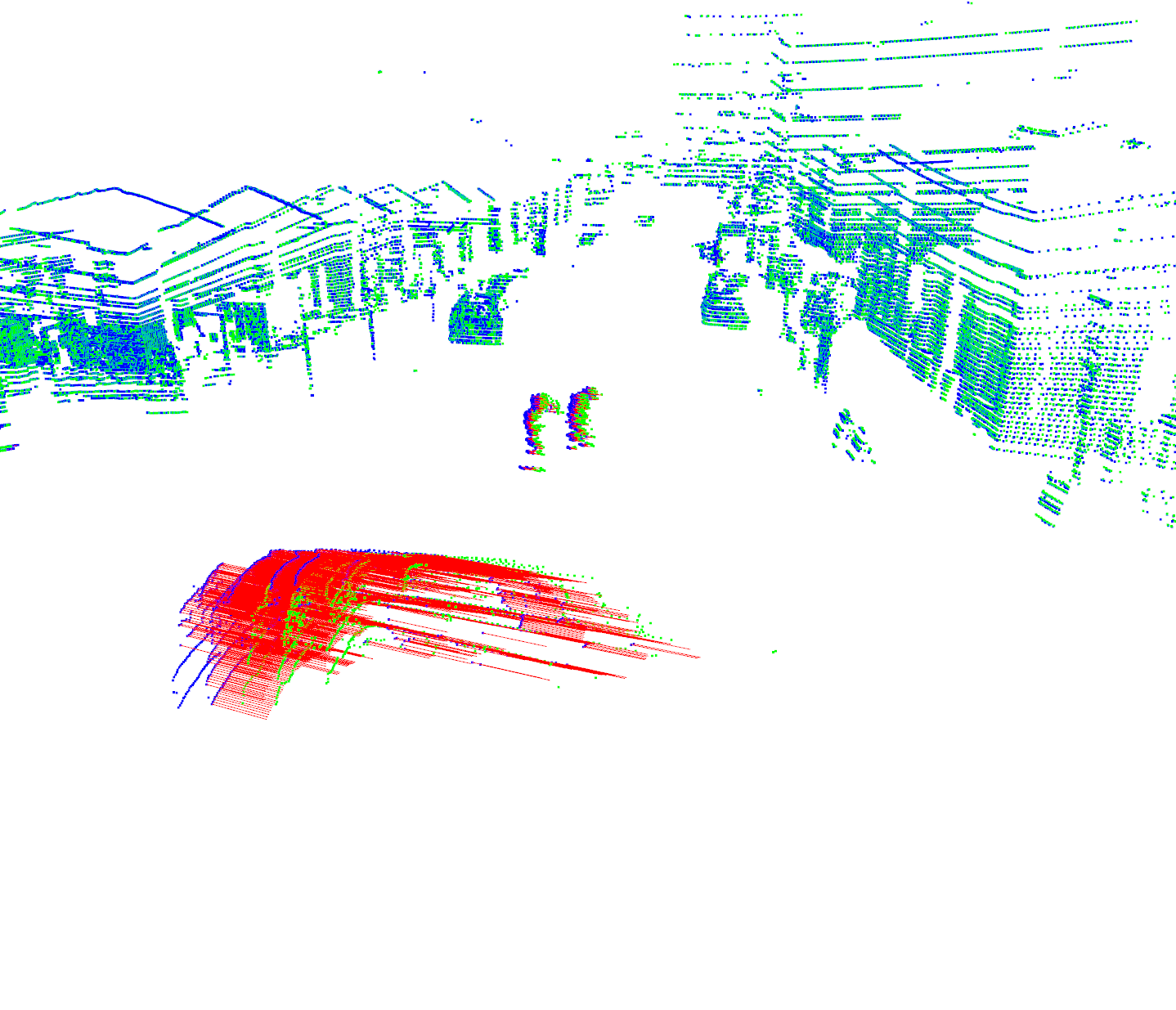}
  \caption{\tinier{Ground Truth}}
  \label{fig:sub5}
\end{subfigure}%
\begin{subfigure}{.16\textwidth}
  \centering
  \includegraphics[width=\linewidth]{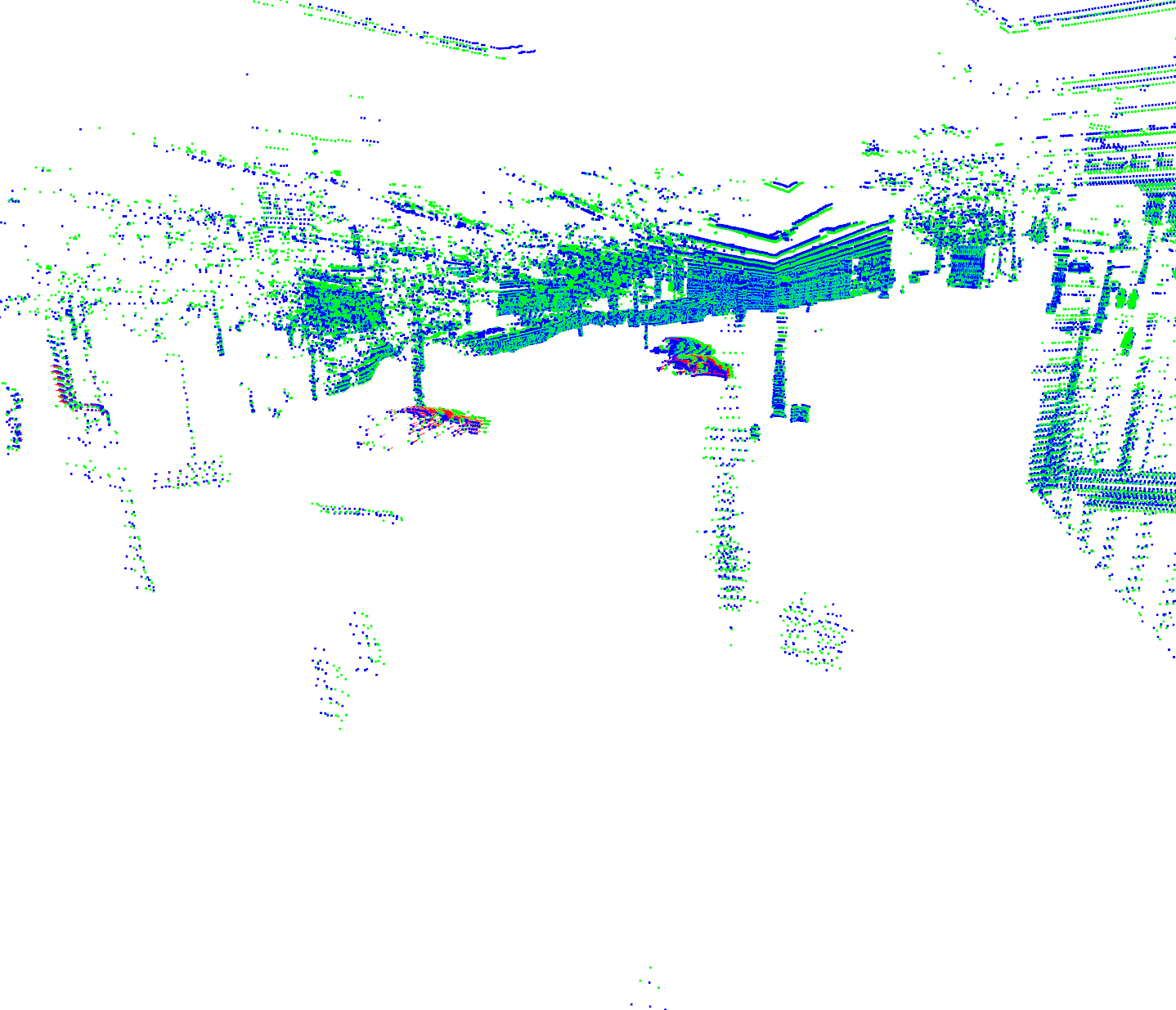}
  \includegraphics[width=\linewidth]{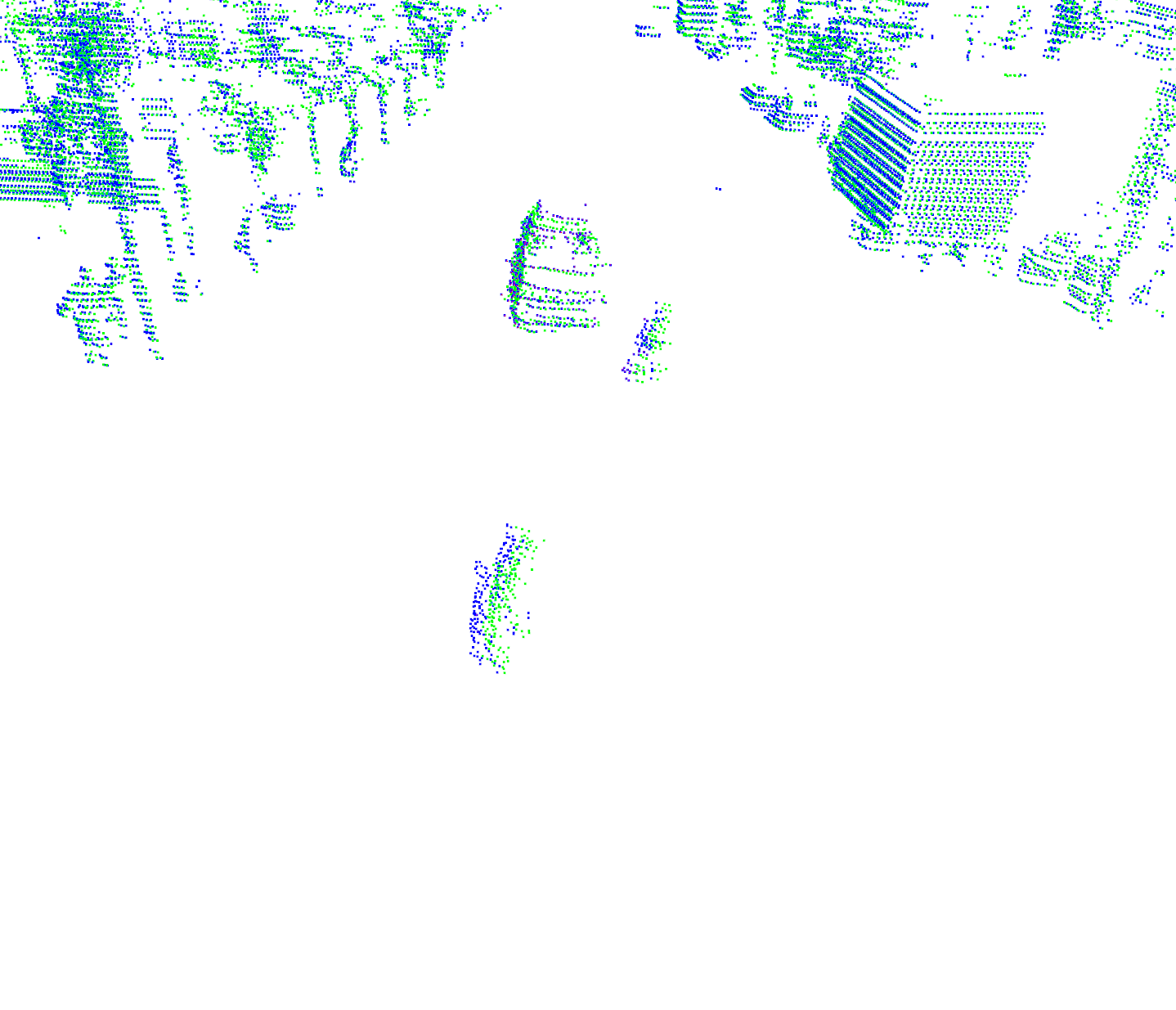}
  \includegraphics[width=\linewidth]{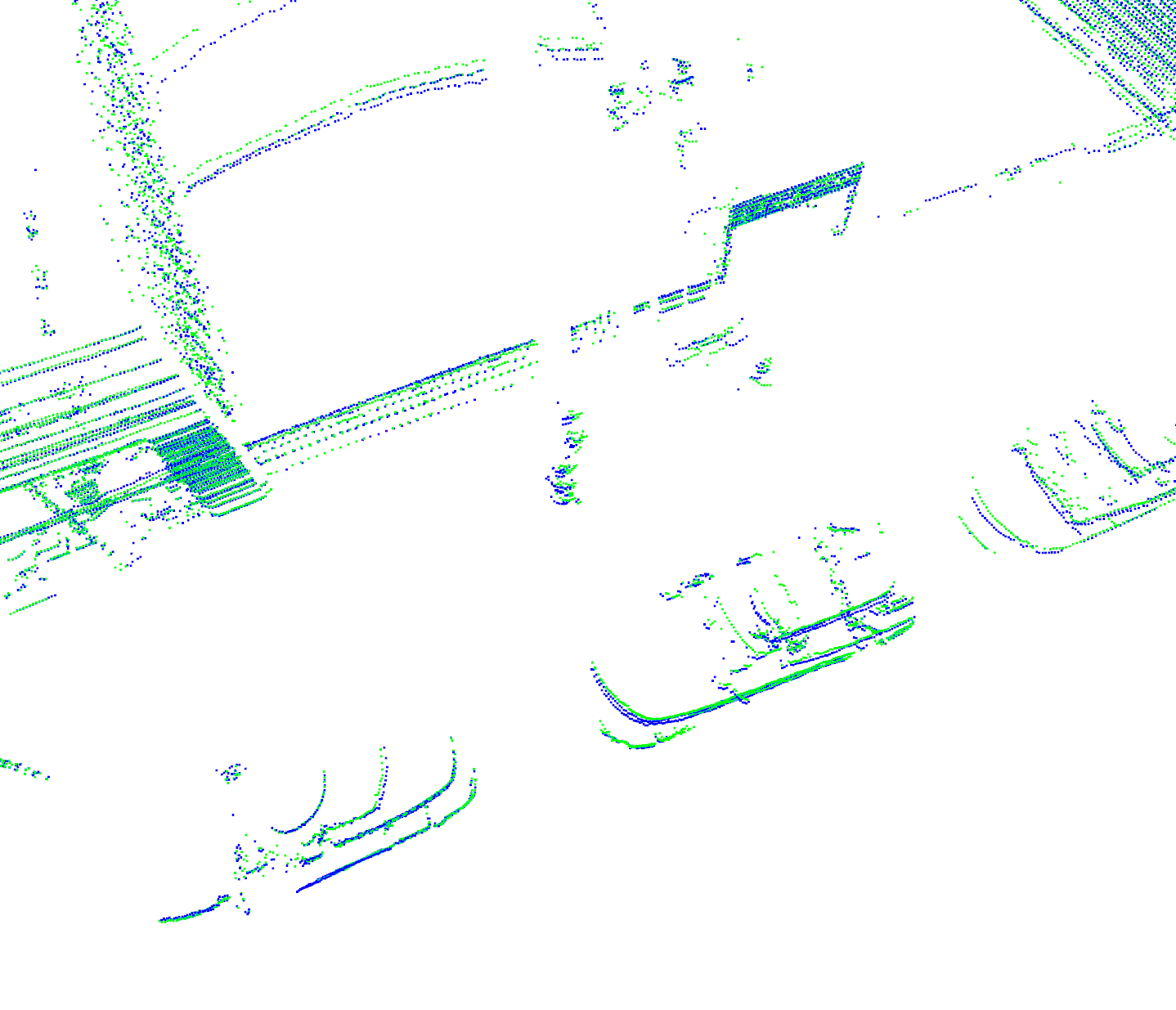}
  \includegraphics[width=\linewidth]{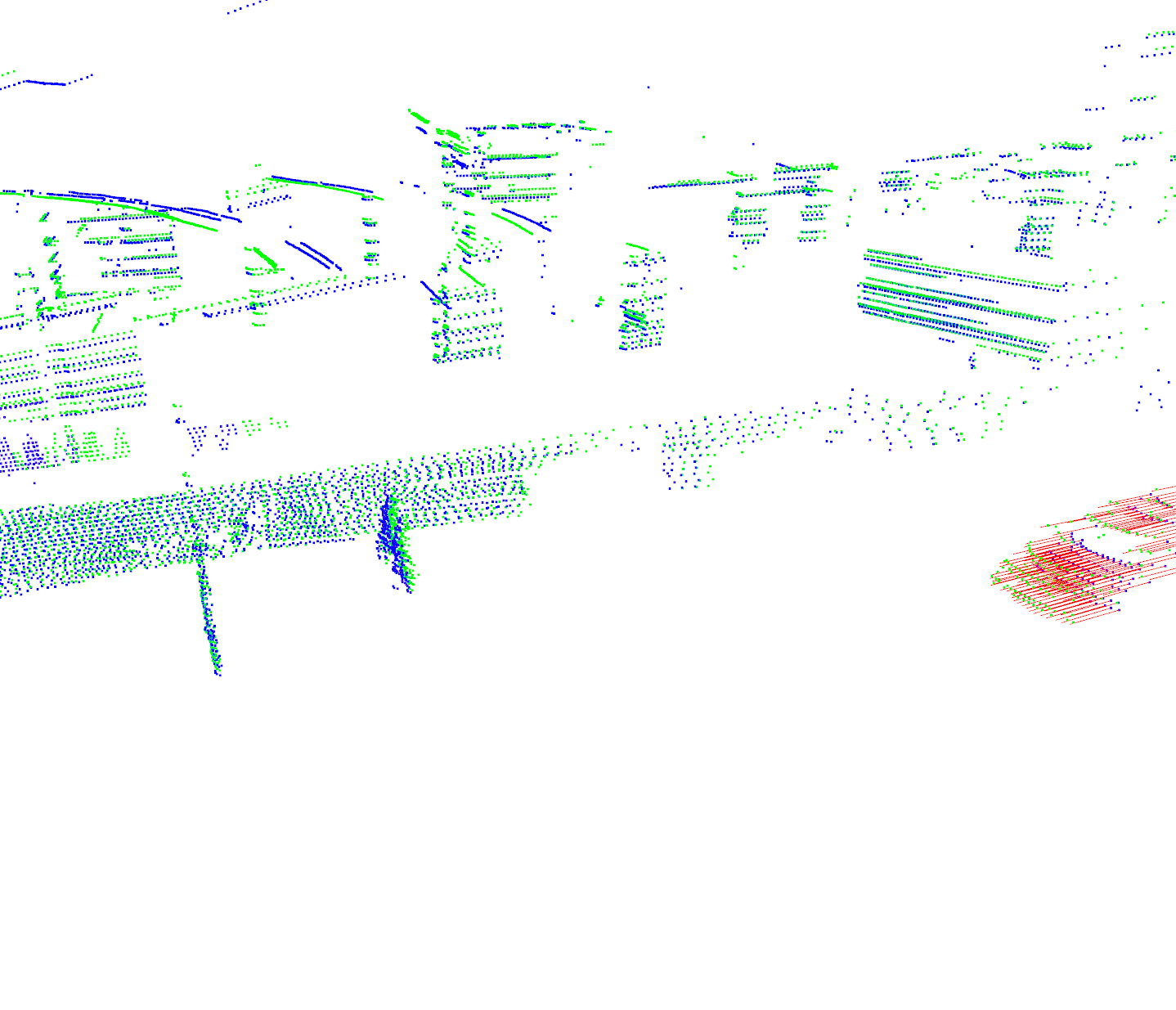}
  \includegraphics[width=\linewidth]{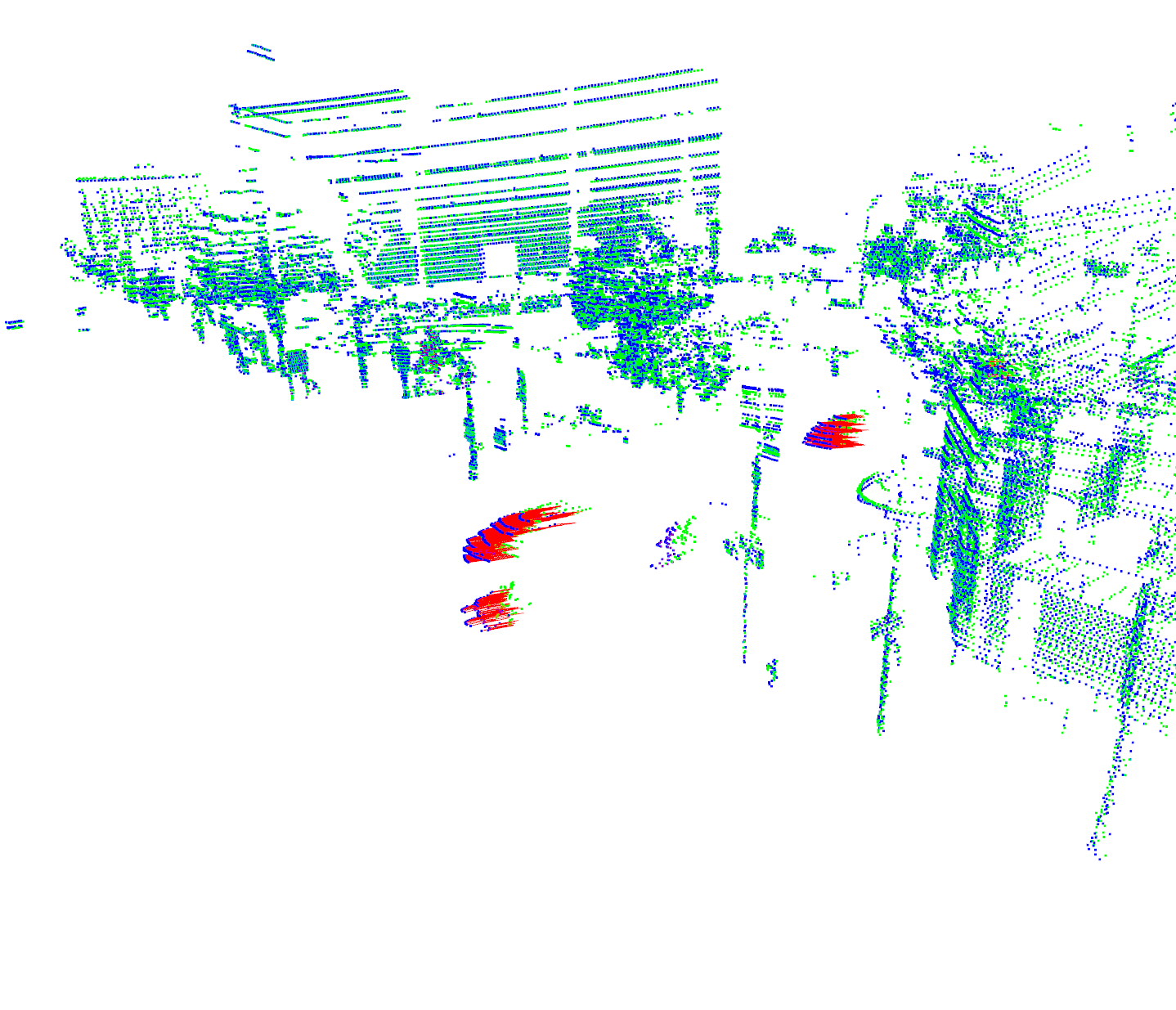}
  \includegraphics[width=\linewidth]{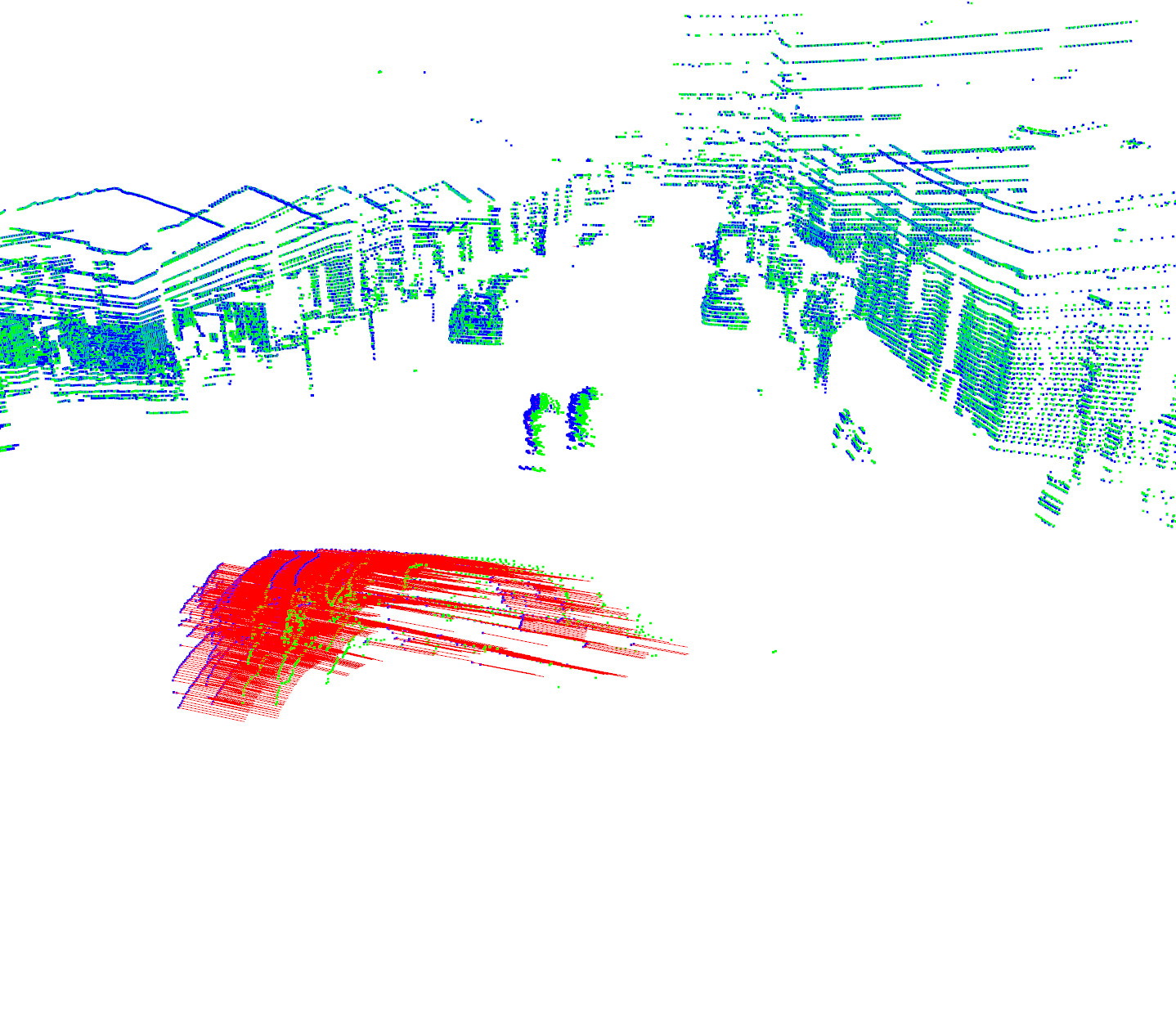}
  \caption{\tinier FastFlow3D}
  \label{fig:sub2}
\end{subfigure}
\begin{subfigure}{.16\textwidth}
  \centering
  \includegraphics[width=\linewidth]{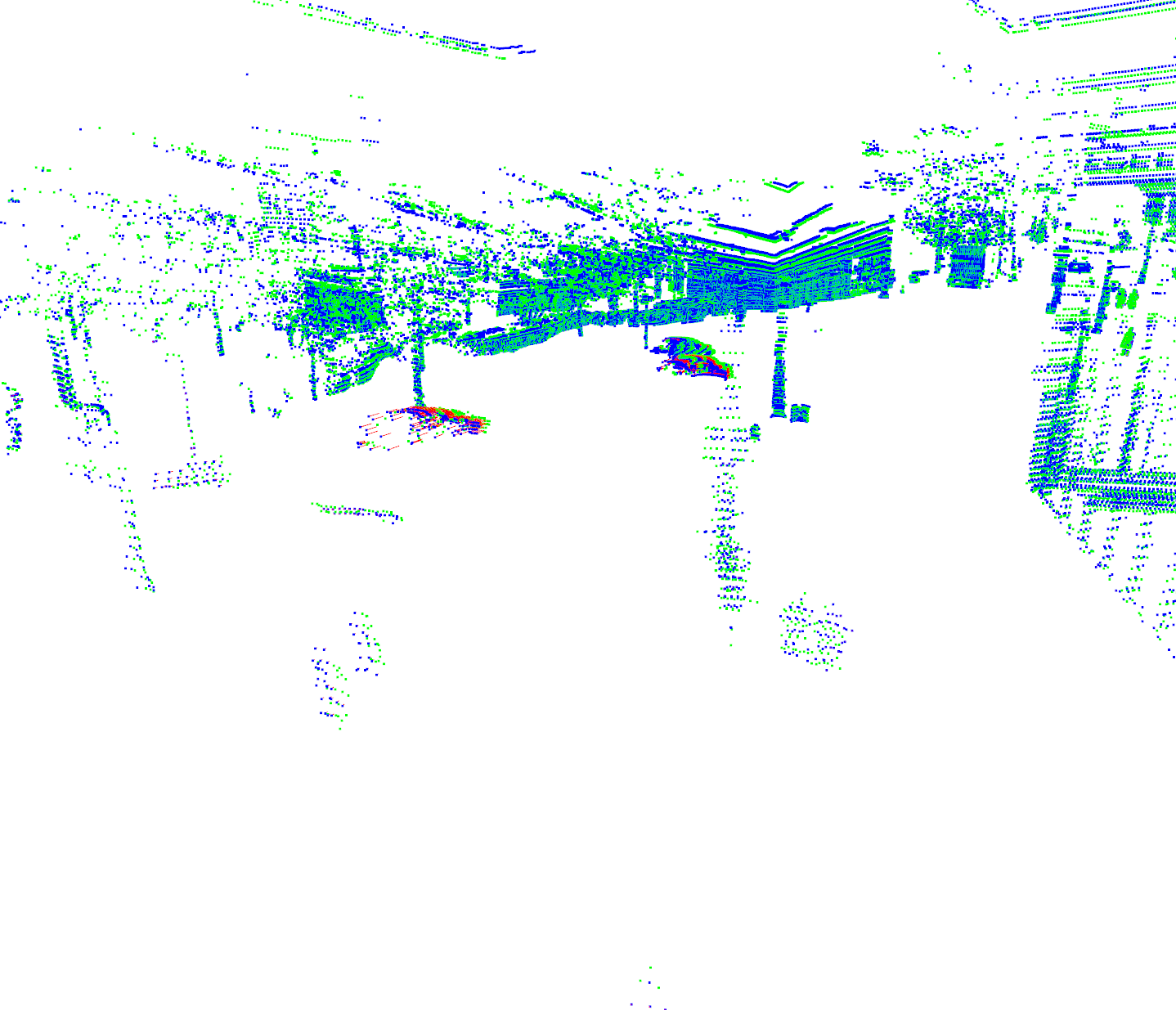}
  \includegraphics[width=\linewidth]{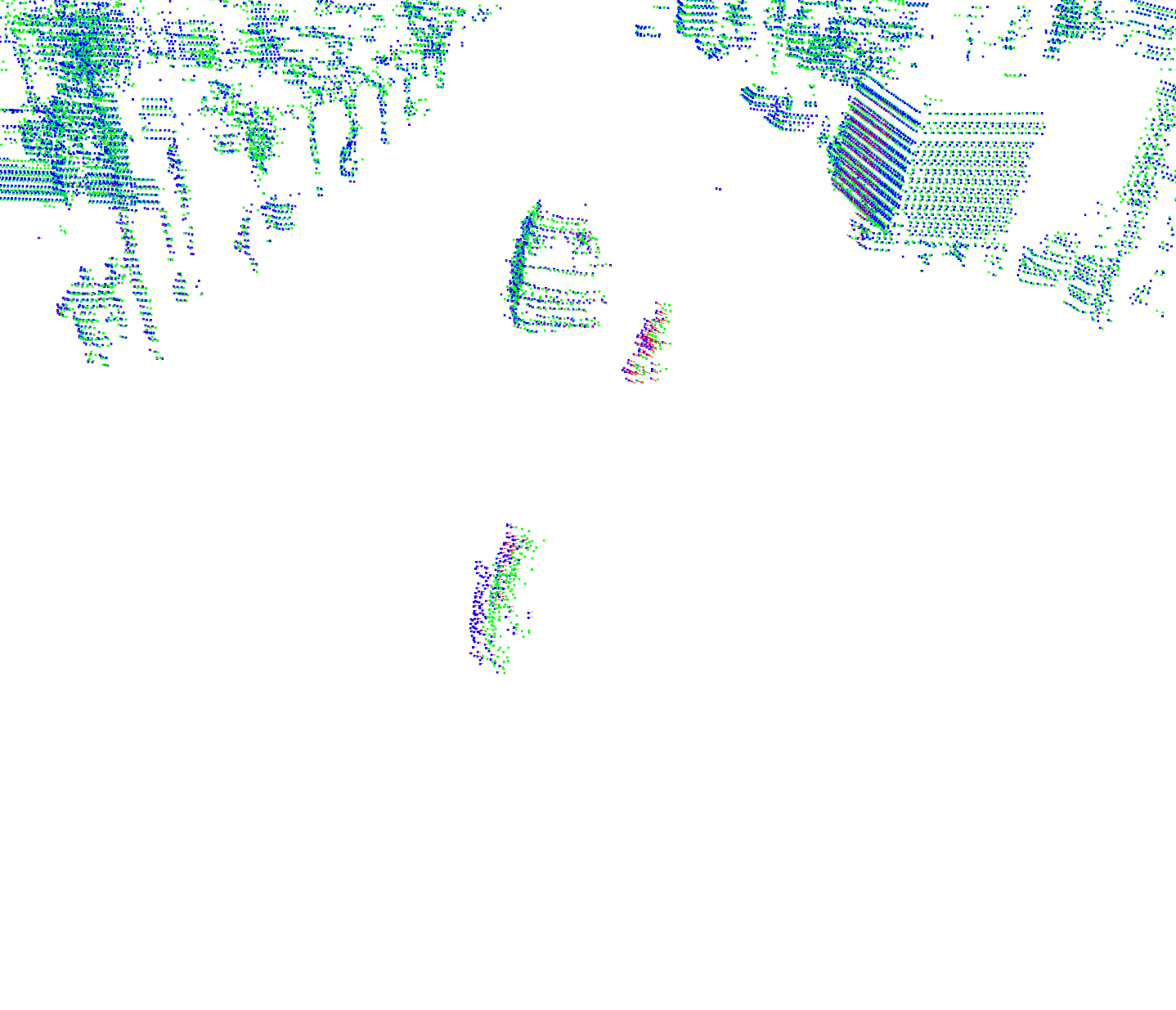}
  \includegraphics[width=\linewidth]{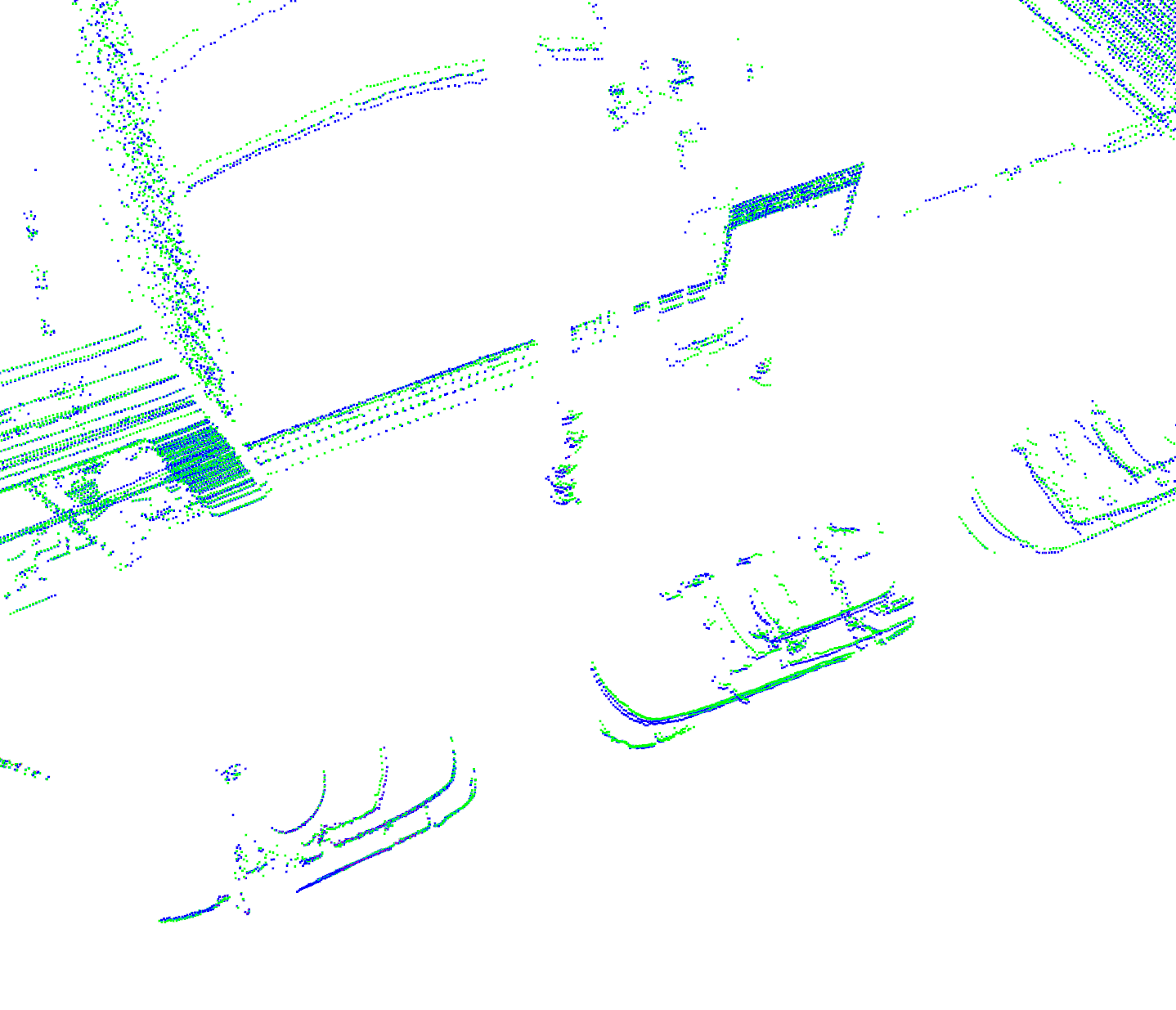}
  \includegraphics[width=\linewidth]{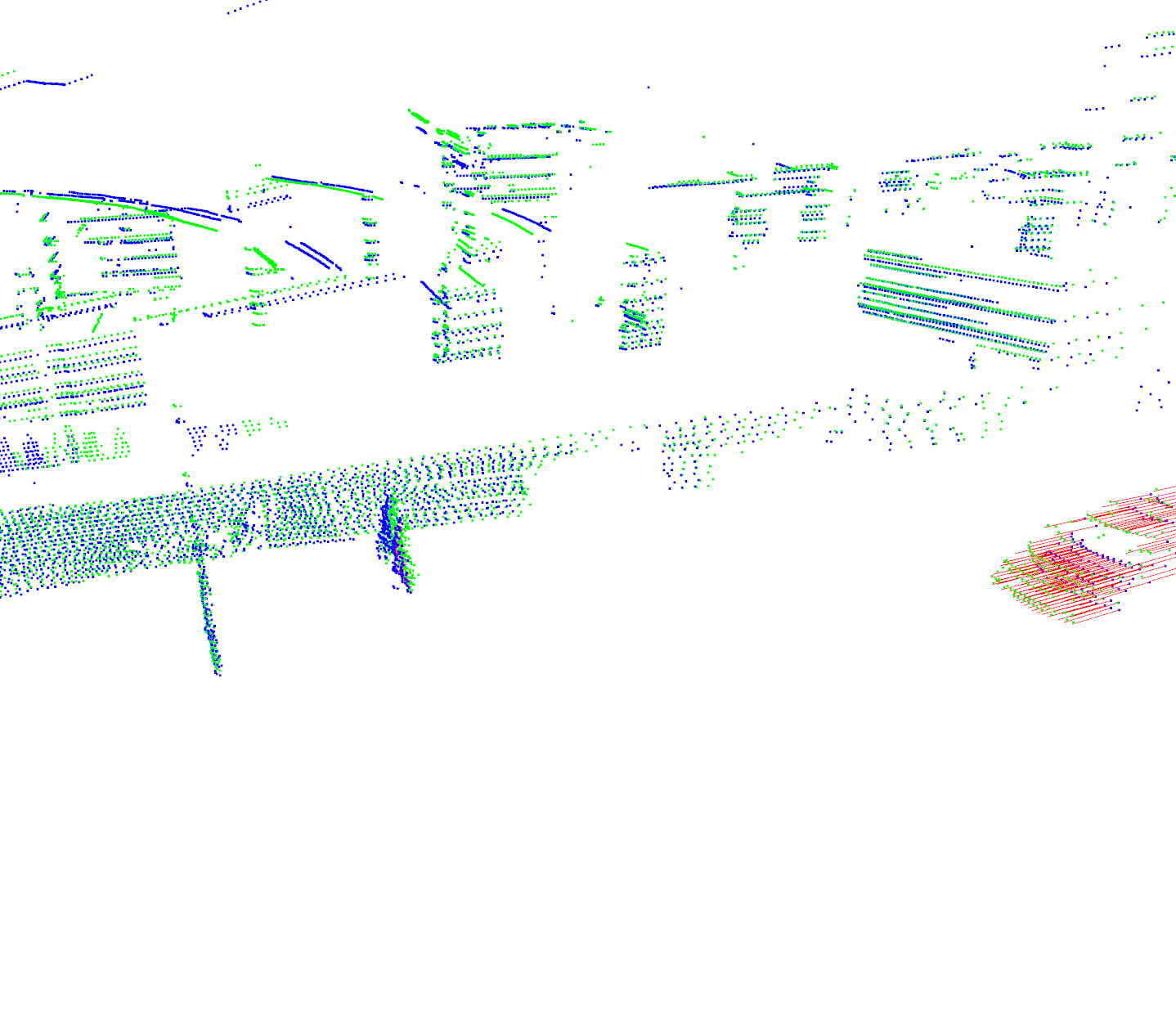}
  \includegraphics[width=\linewidth]{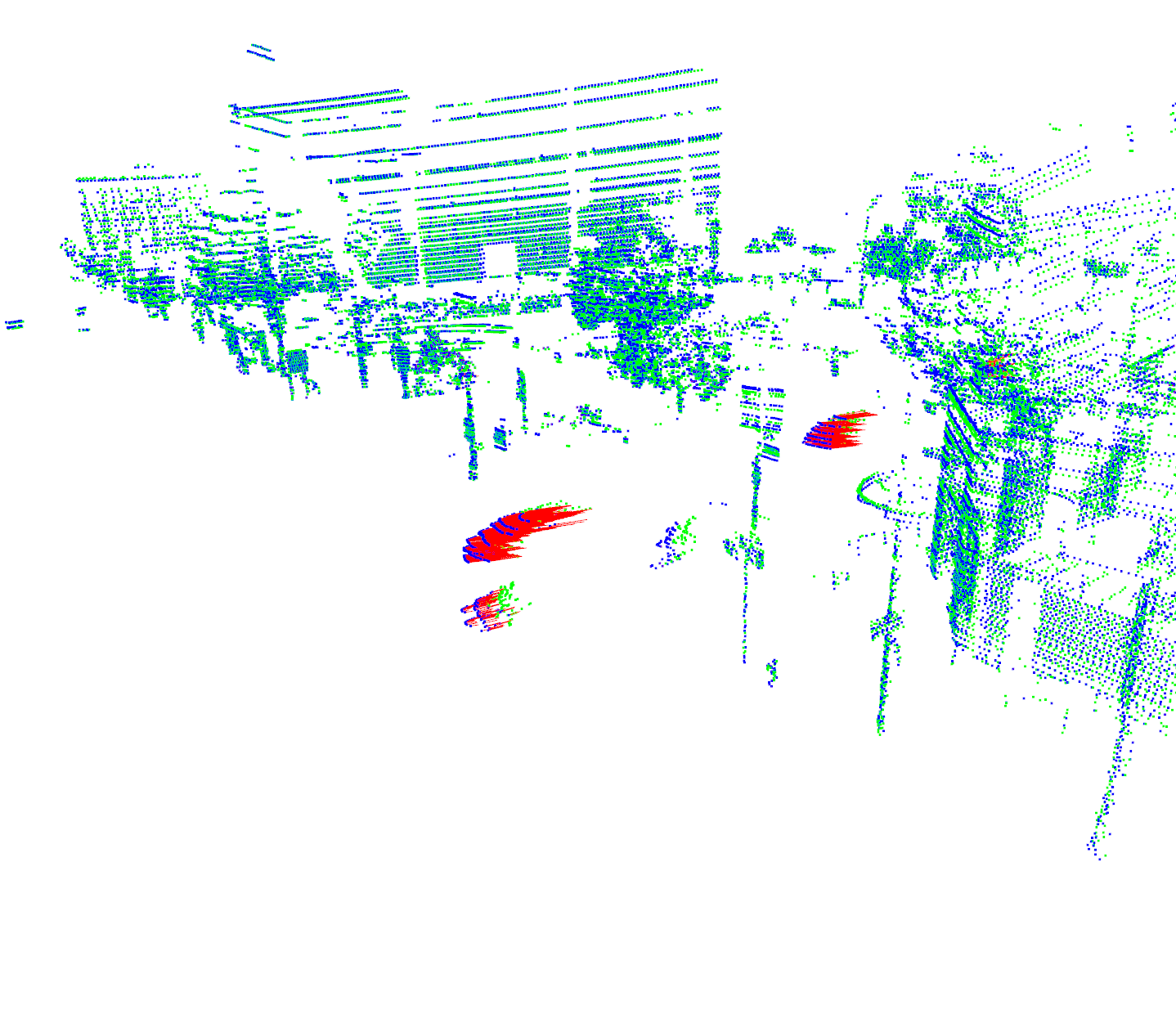}
  \includegraphics[width=\linewidth]{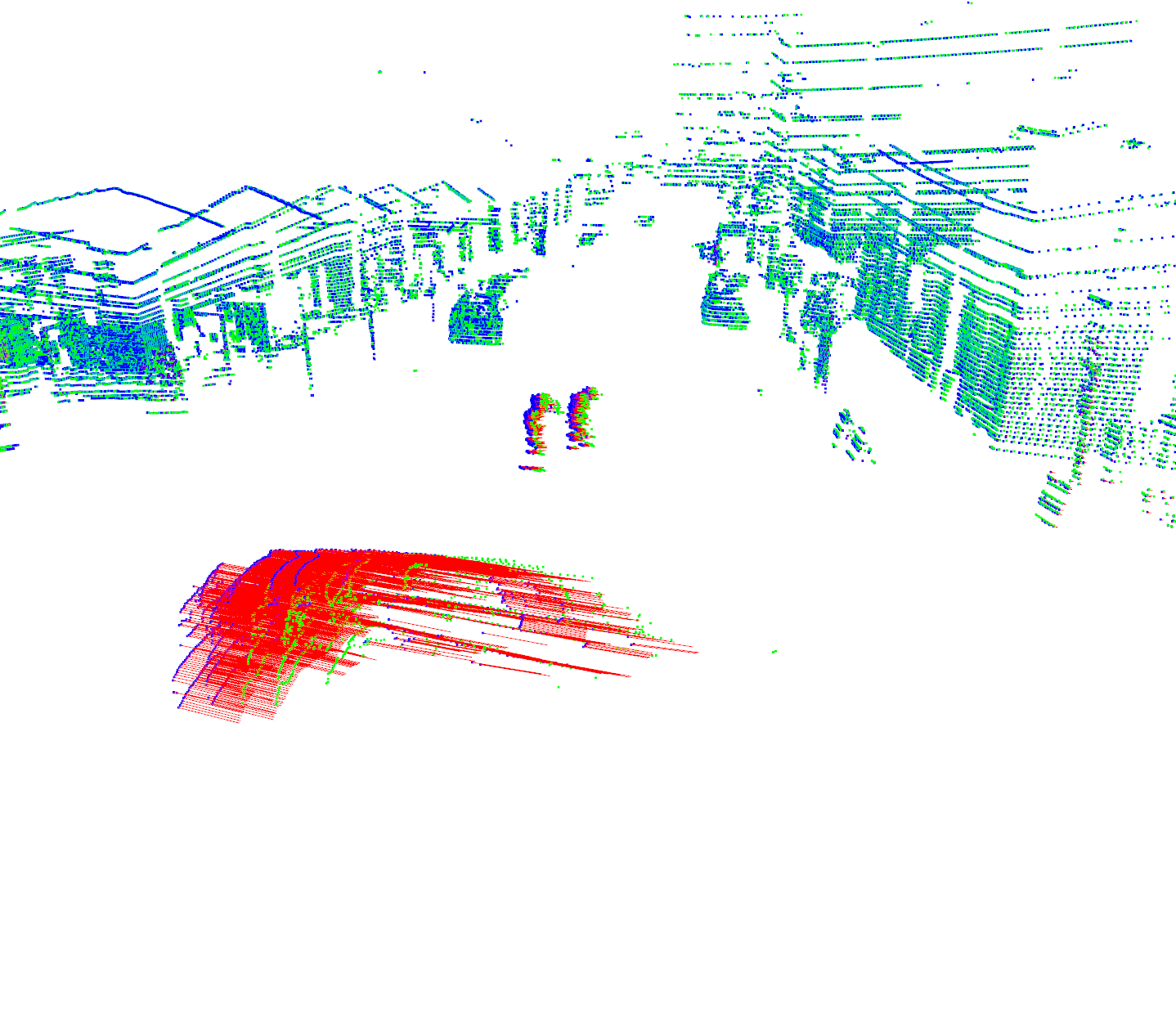}
  \caption{\tinier DeFlow}
  \label{fig:sub3}
\end{subfigure}
\begin{subfigure}{.16\textwidth}
  \centering
  \includegraphics[width=\linewidth]{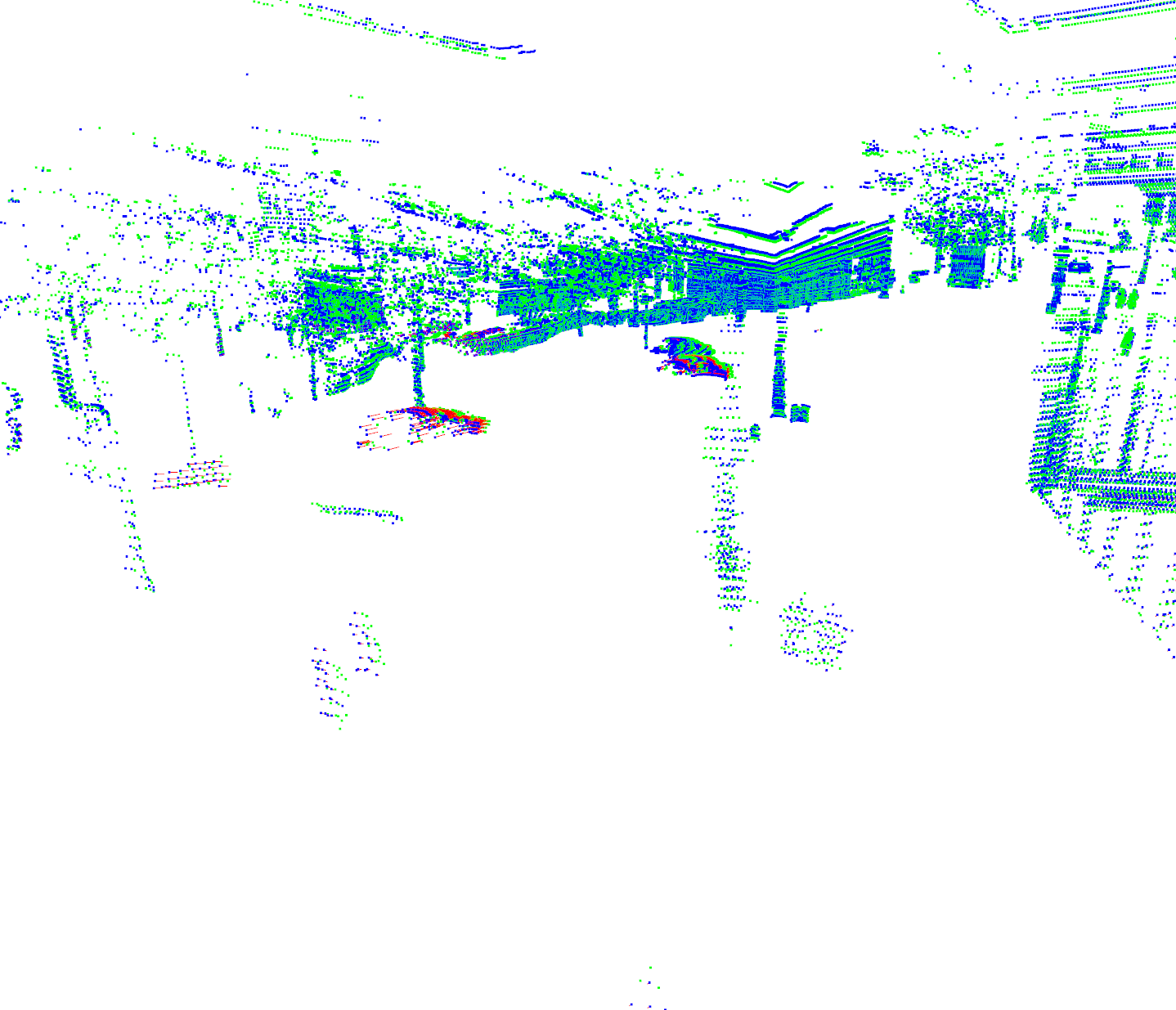}
  \includegraphics[width=\linewidth]{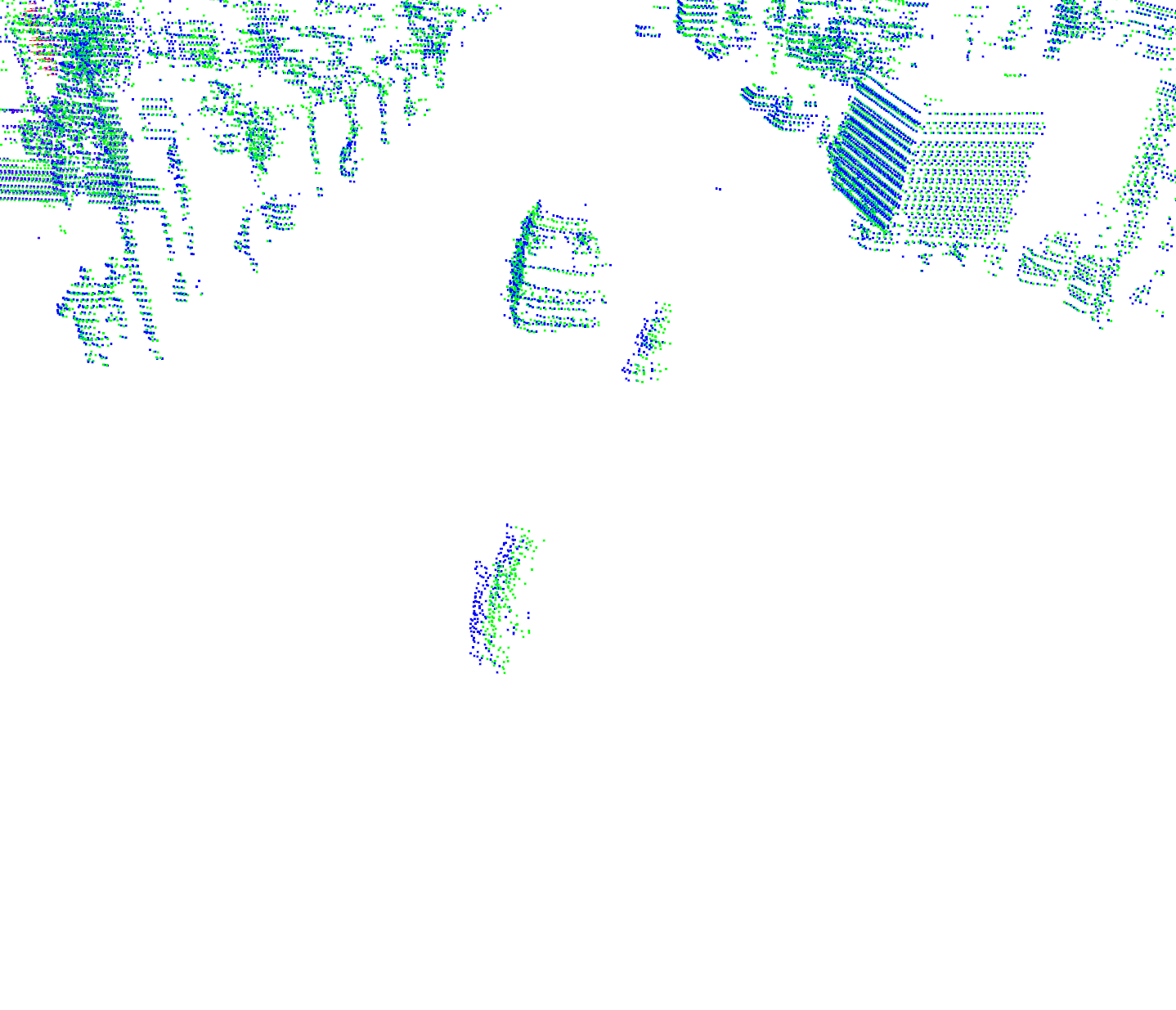}
  \includegraphics[width=\linewidth]{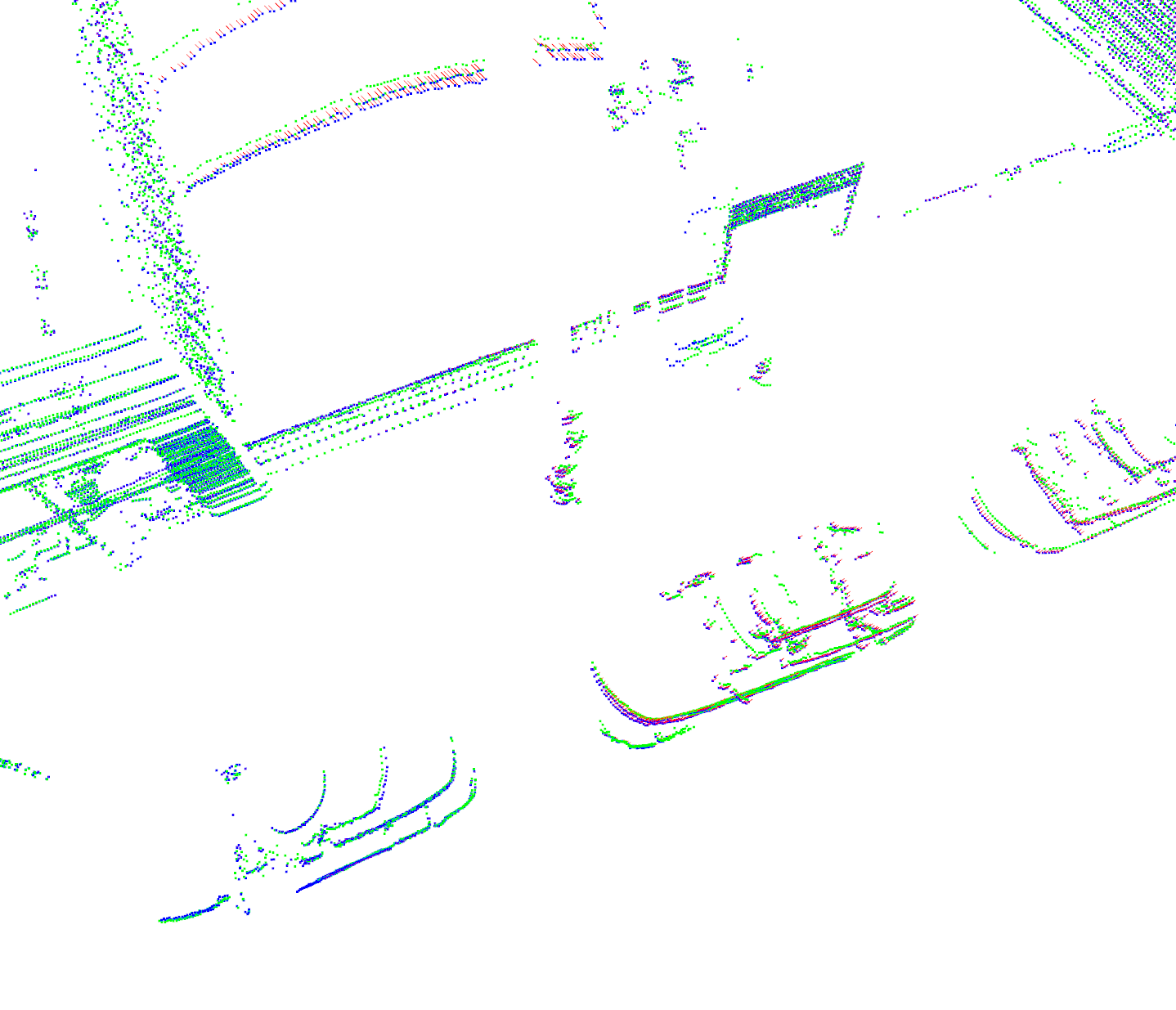}
  \includegraphics[width=\linewidth]{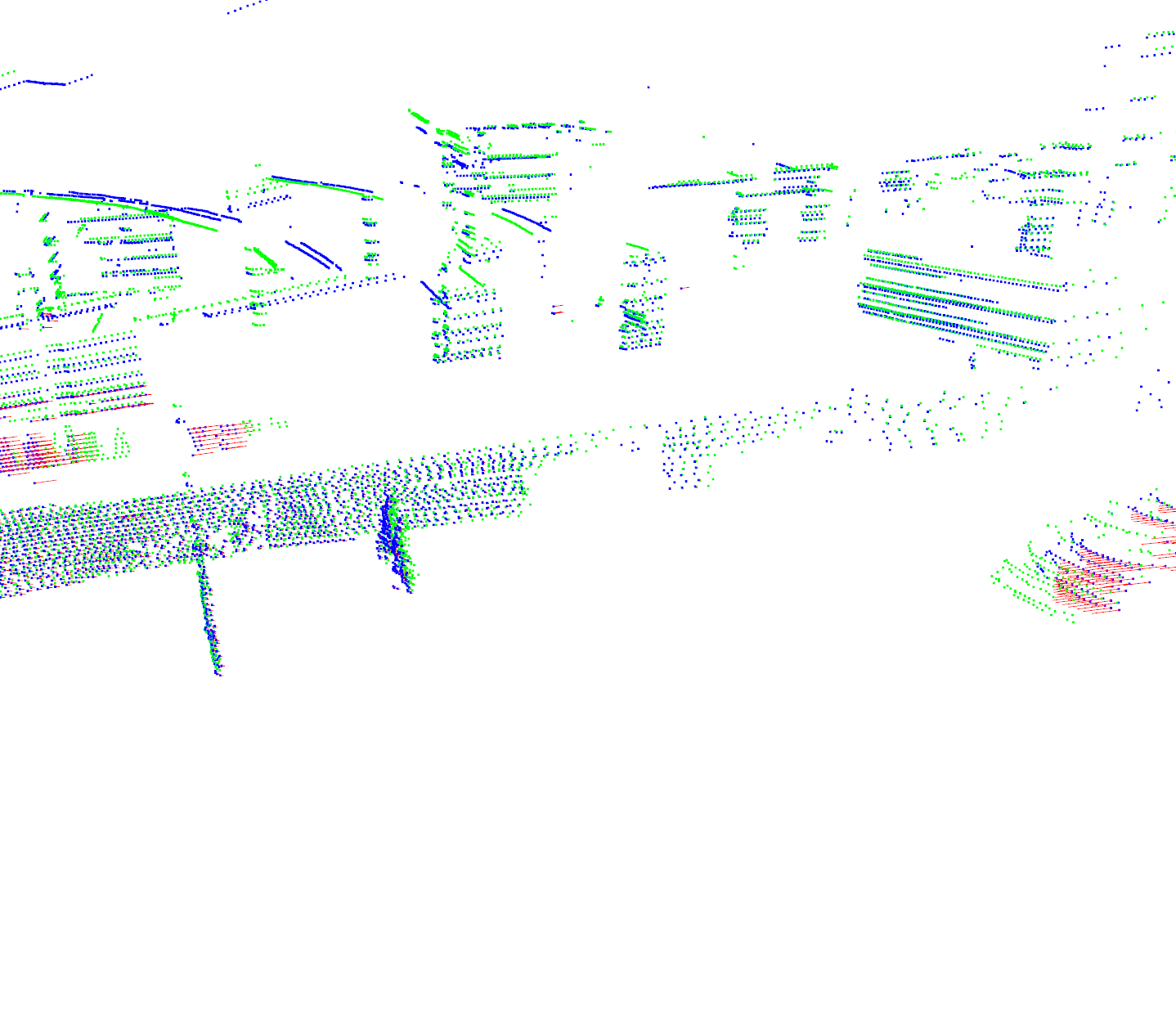}
  \includegraphics[width=\linewidth]{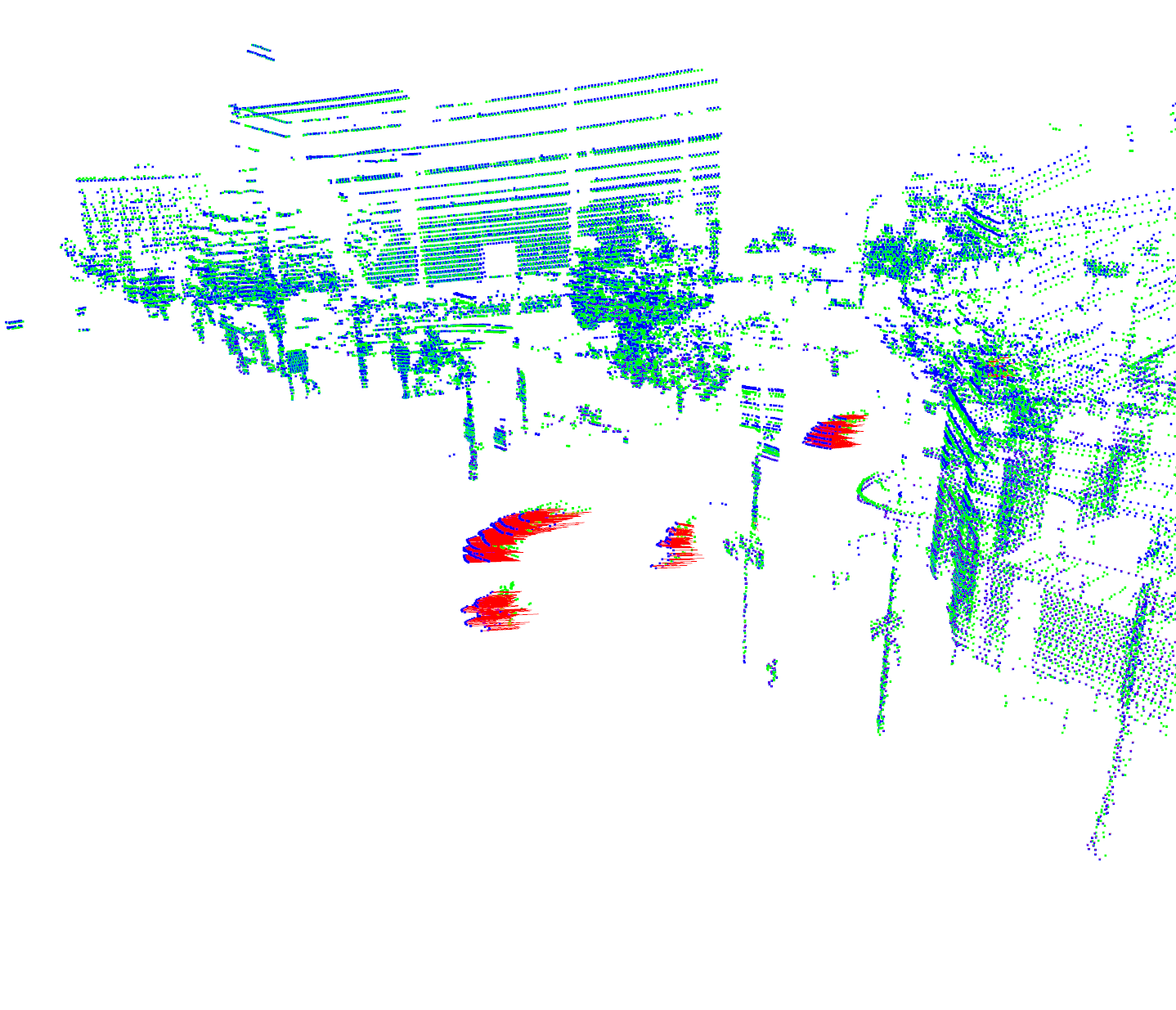}
  \includegraphics[width=\linewidth]{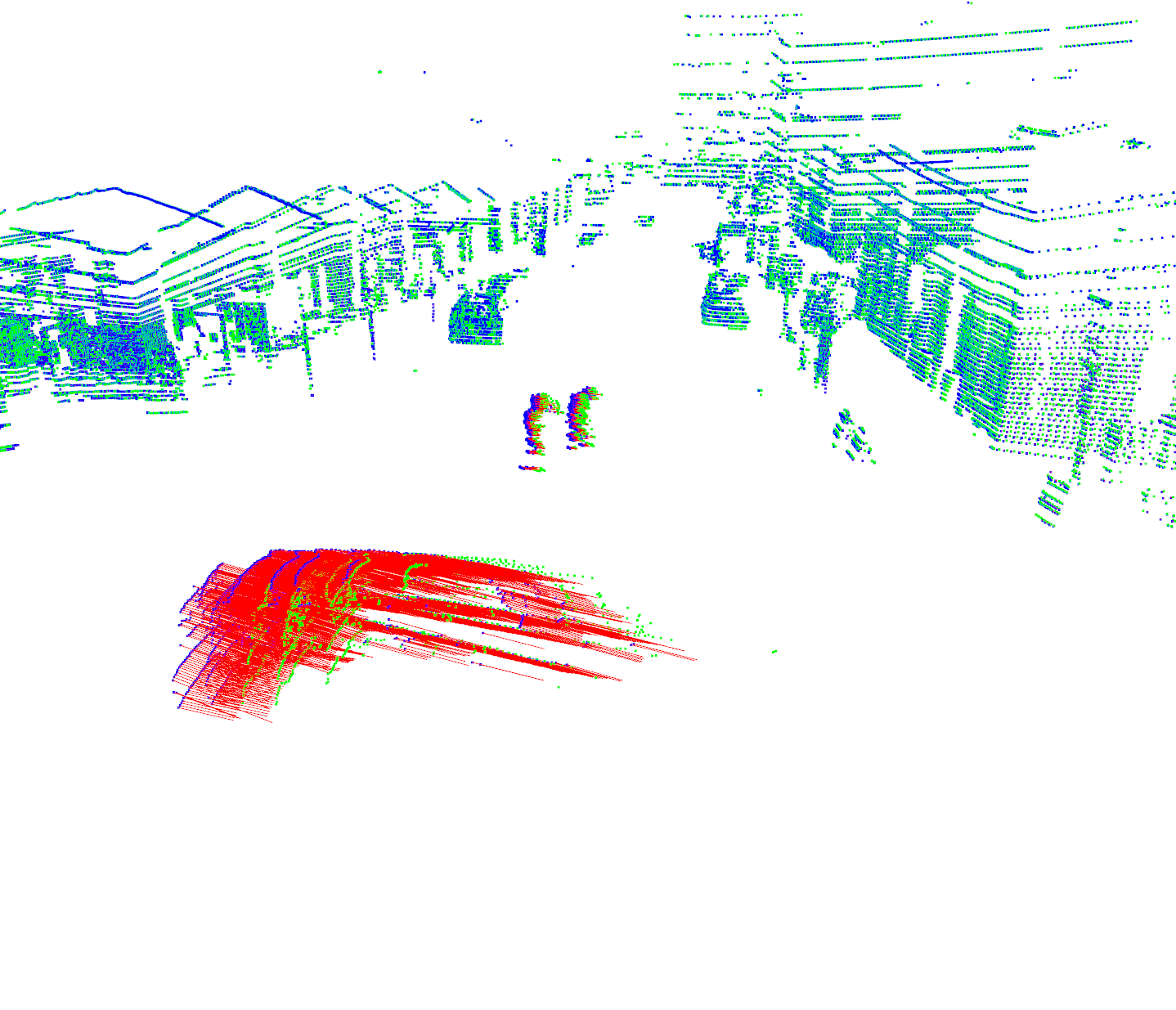}
  \caption{\tinier NSFP}
  \label{fig:sub4}
\end{subfigure}%
\begin{subfigure}{.16\textwidth}
  \centering
  \includegraphics[width=\linewidth]{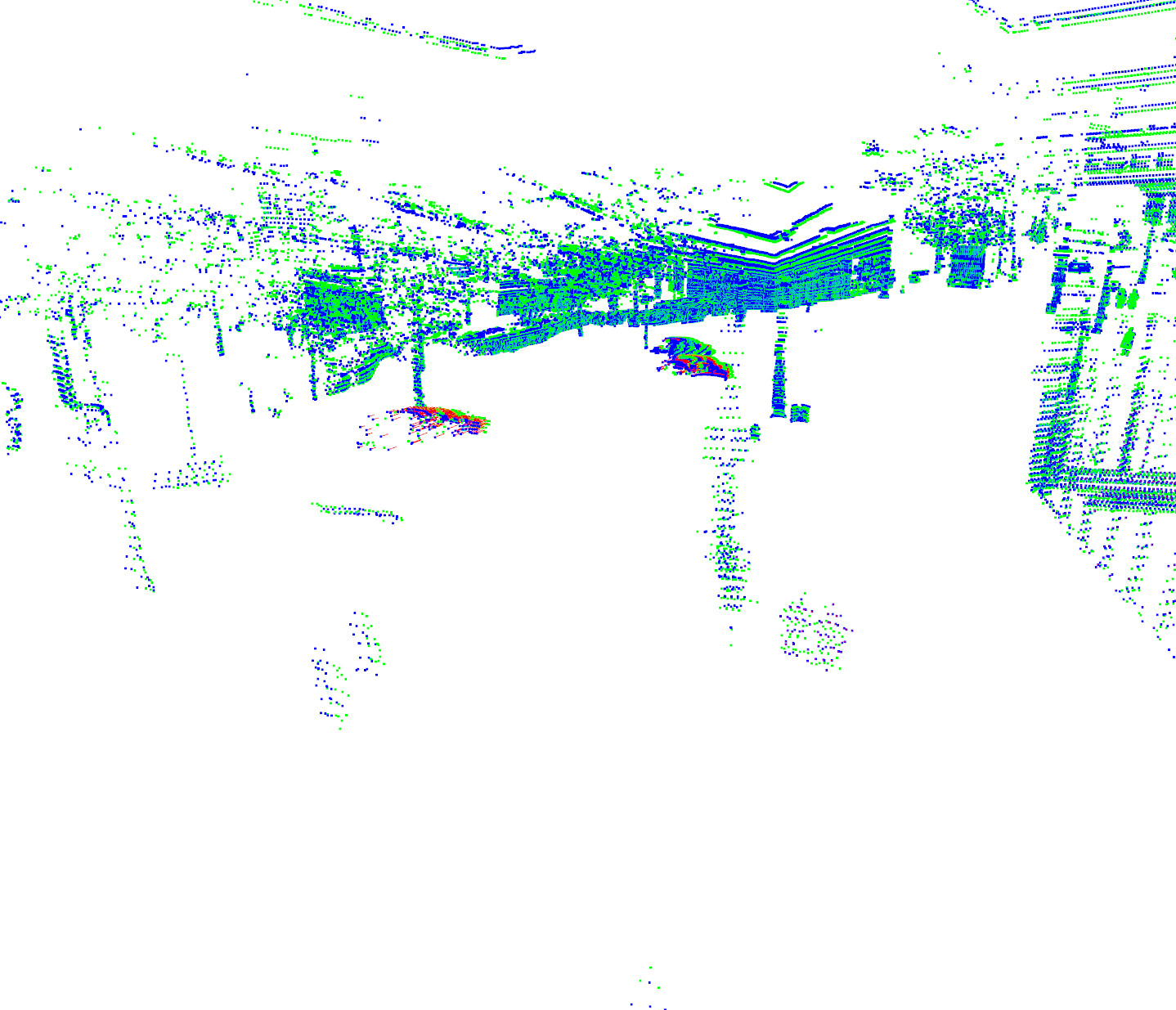}
  \includegraphics[width=\linewidth]{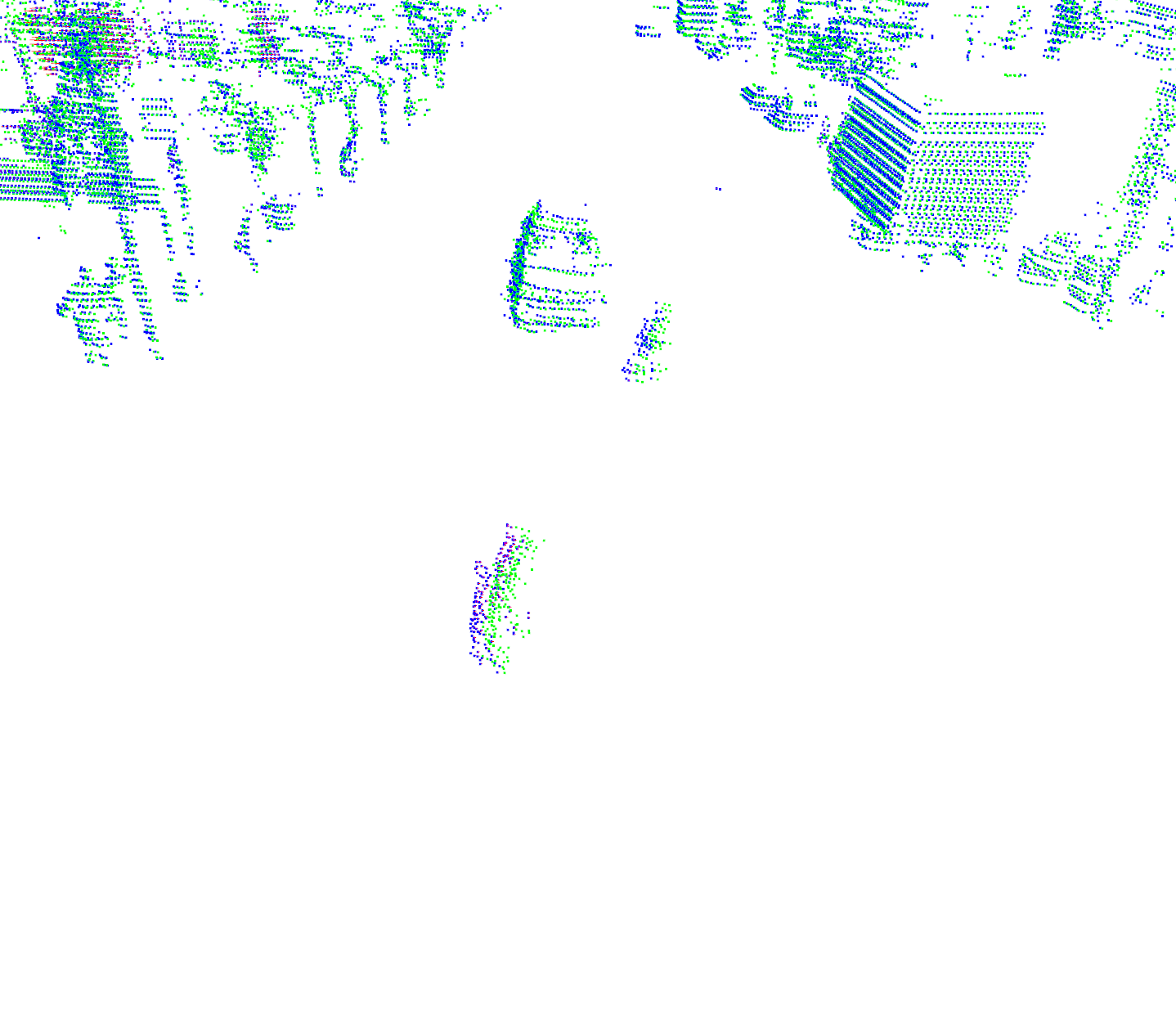}
  \includegraphics[width=\linewidth]{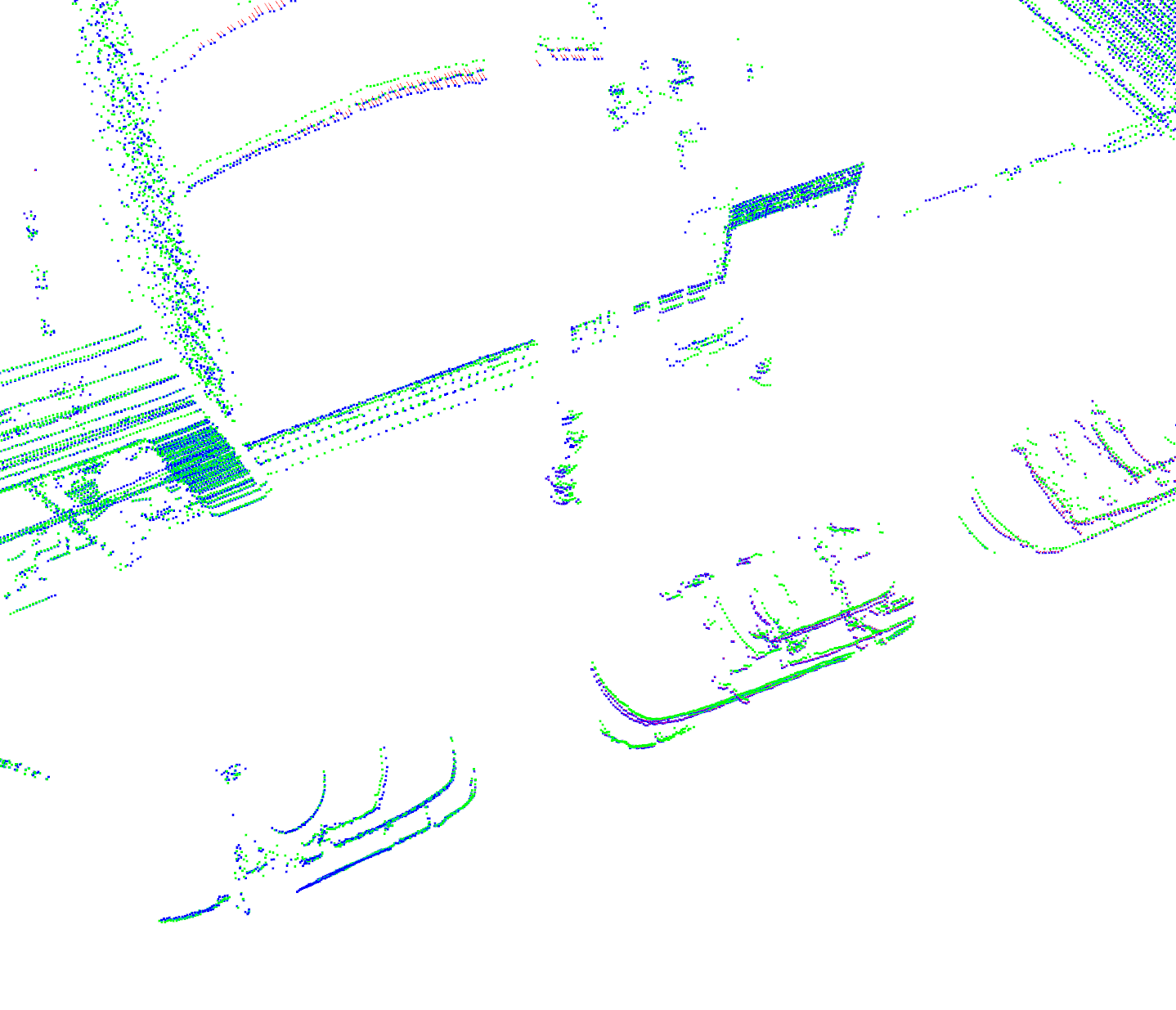}
  \includegraphics[width=\linewidth]{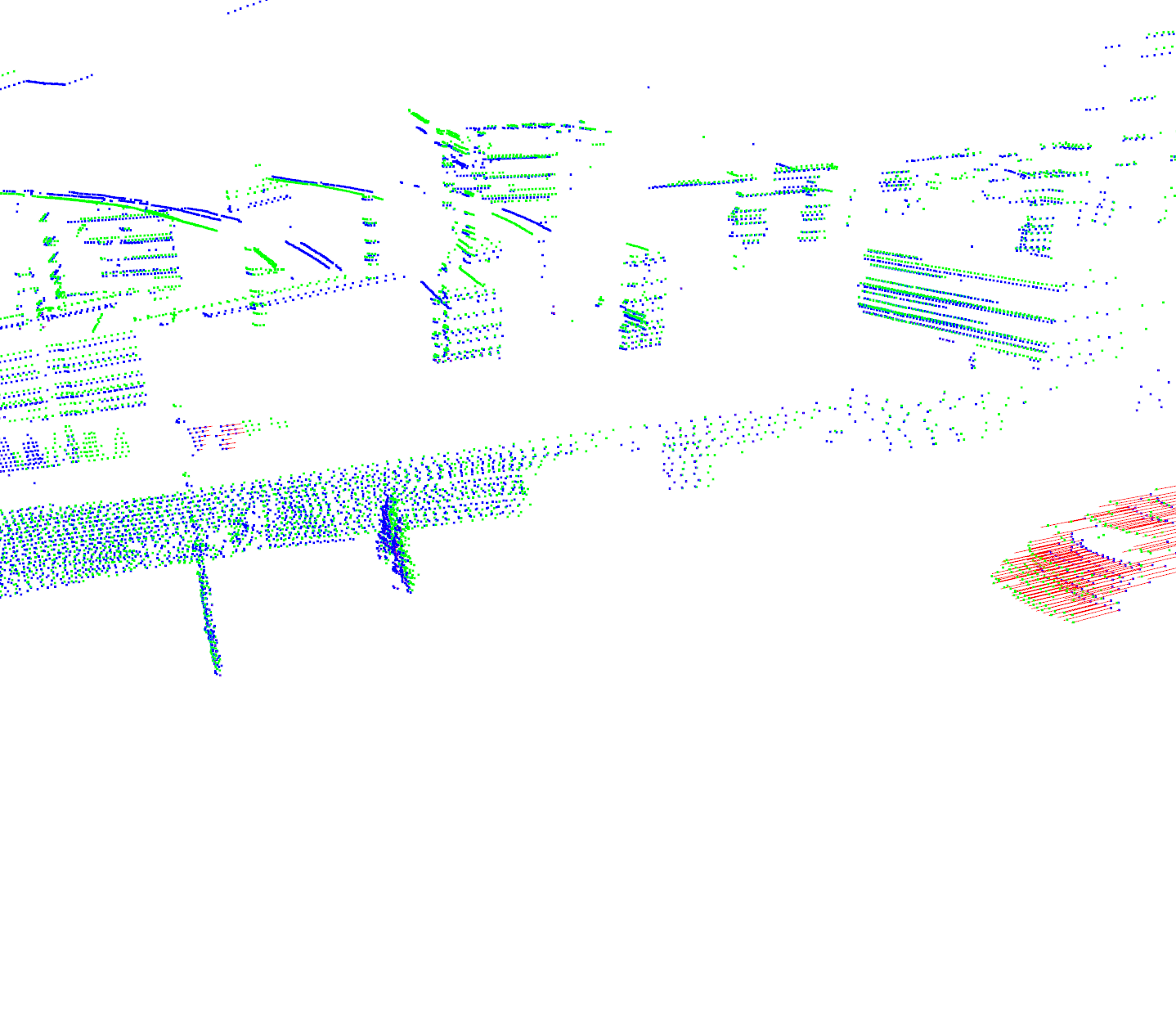}
  \includegraphics[width=\linewidth]{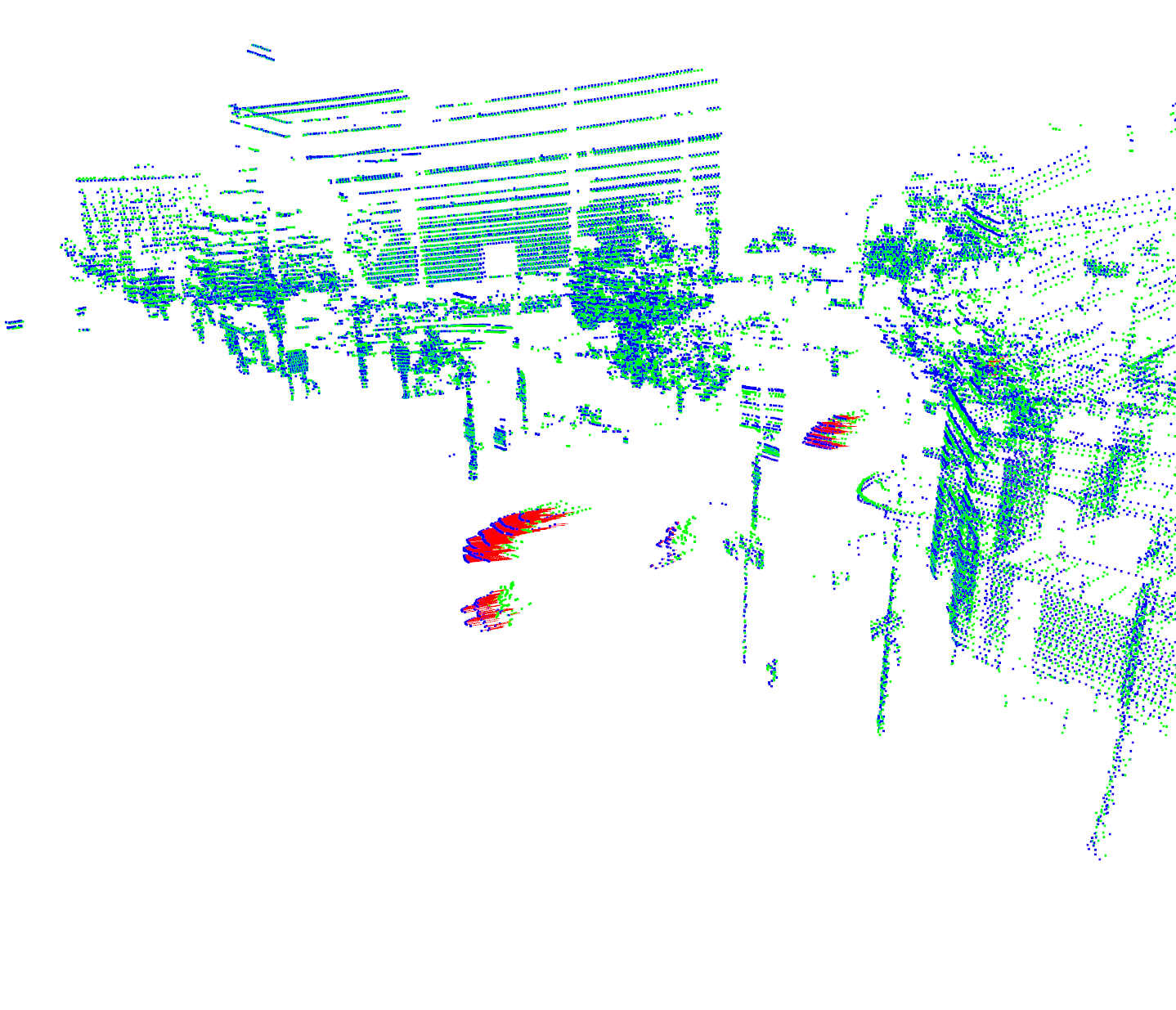}
  \includegraphics[width=\linewidth]{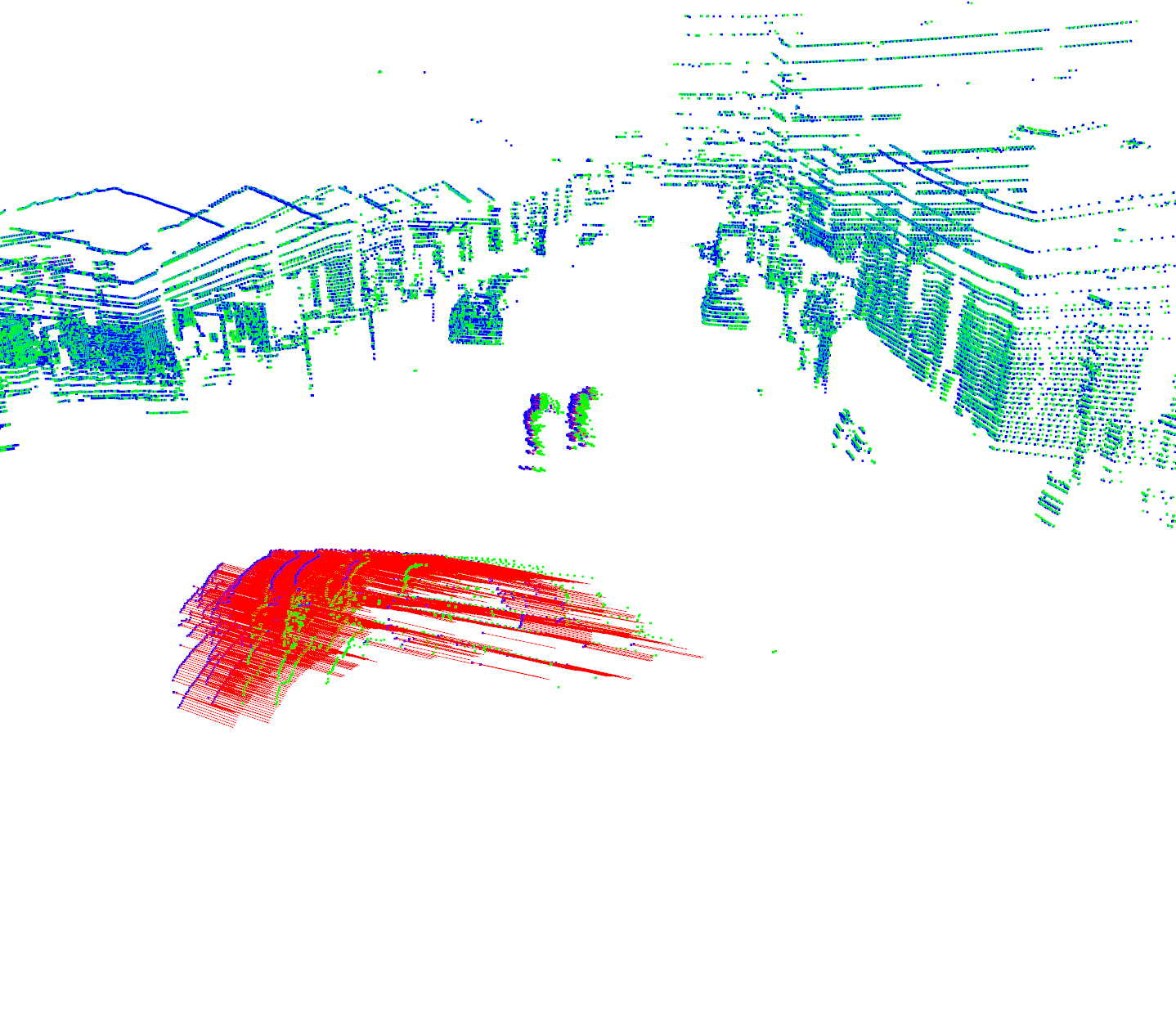}
  \caption{\tinier ZeroFlow}
  \label{fig:sub1}
\end{subfigure}%
\begin{subfigure}{.16\textwidth}
  \centering
  \includegraphics[width=\linewidth]{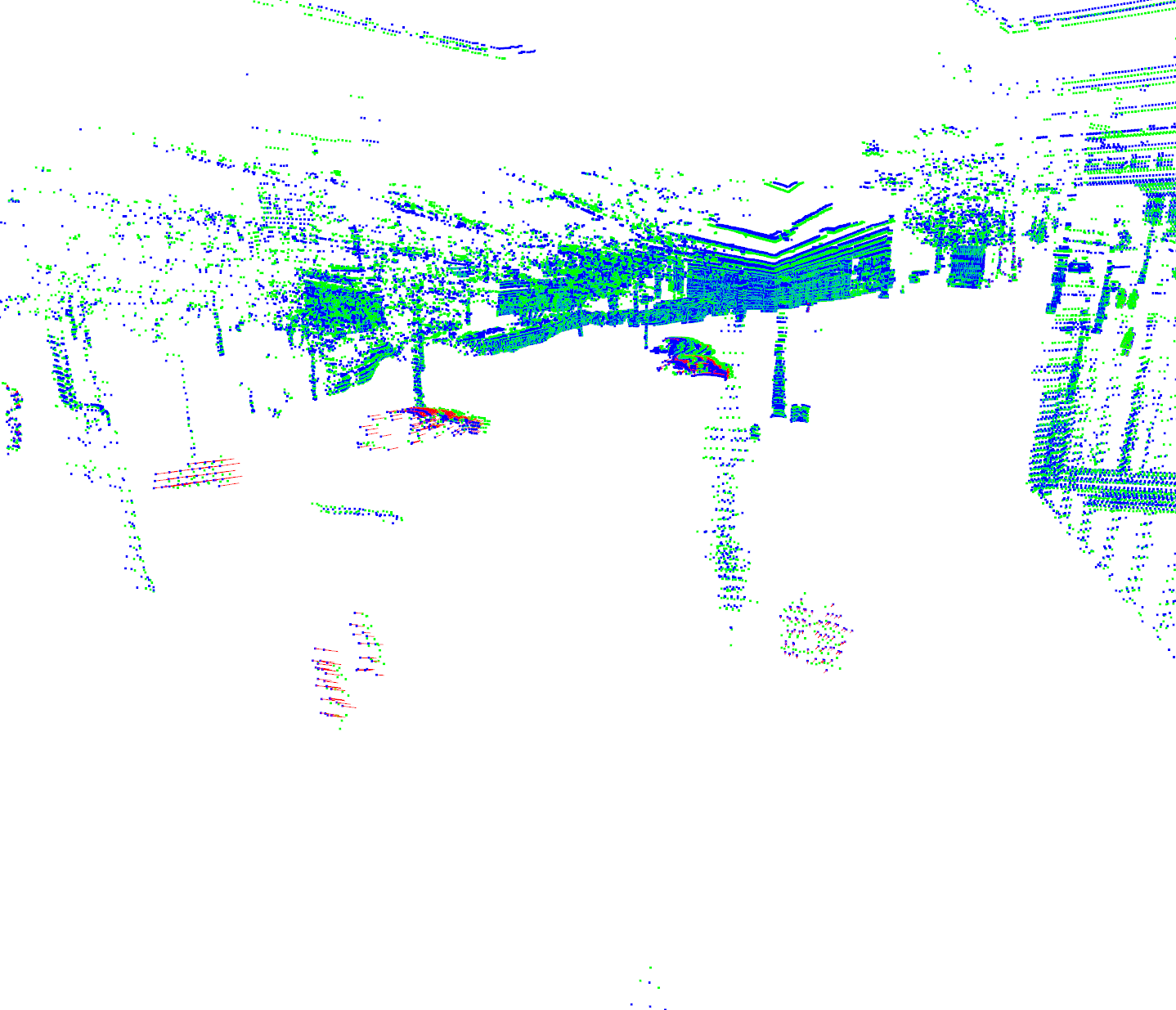}
  \includegraphics[width=\linewidth]{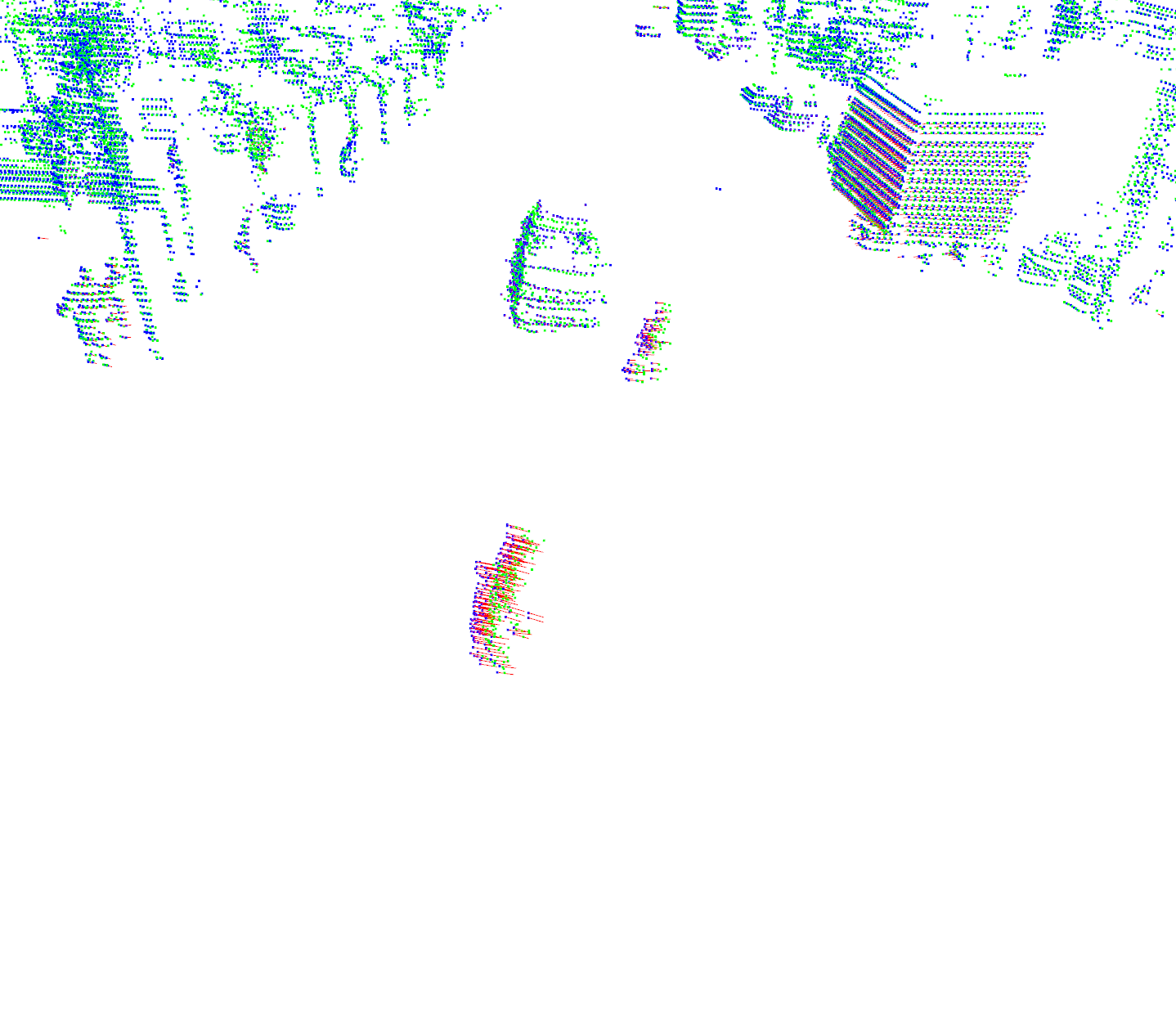}
  \includegraphics[width=\linewidth]{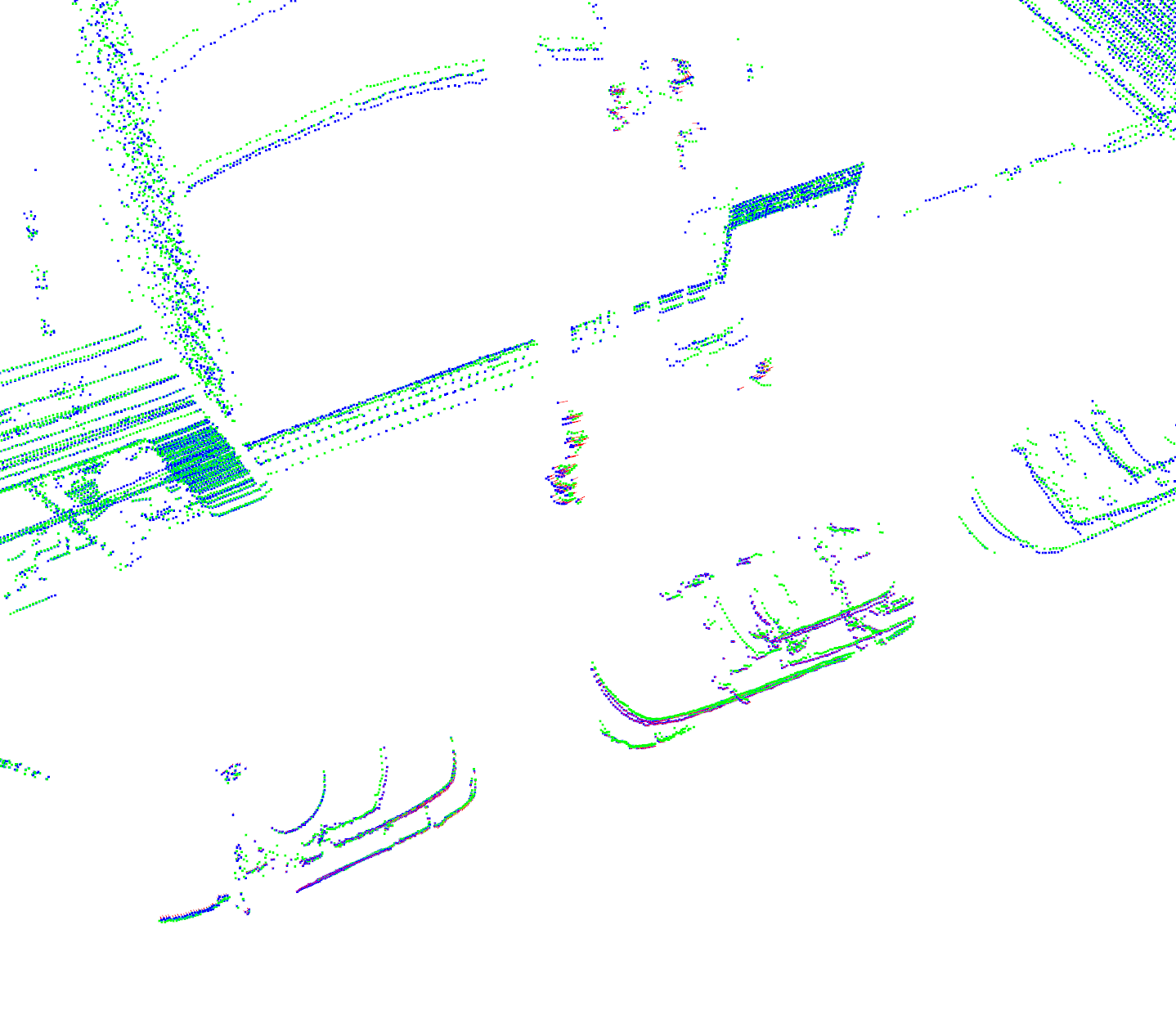}
  \includegraphics[width=\linewidth]{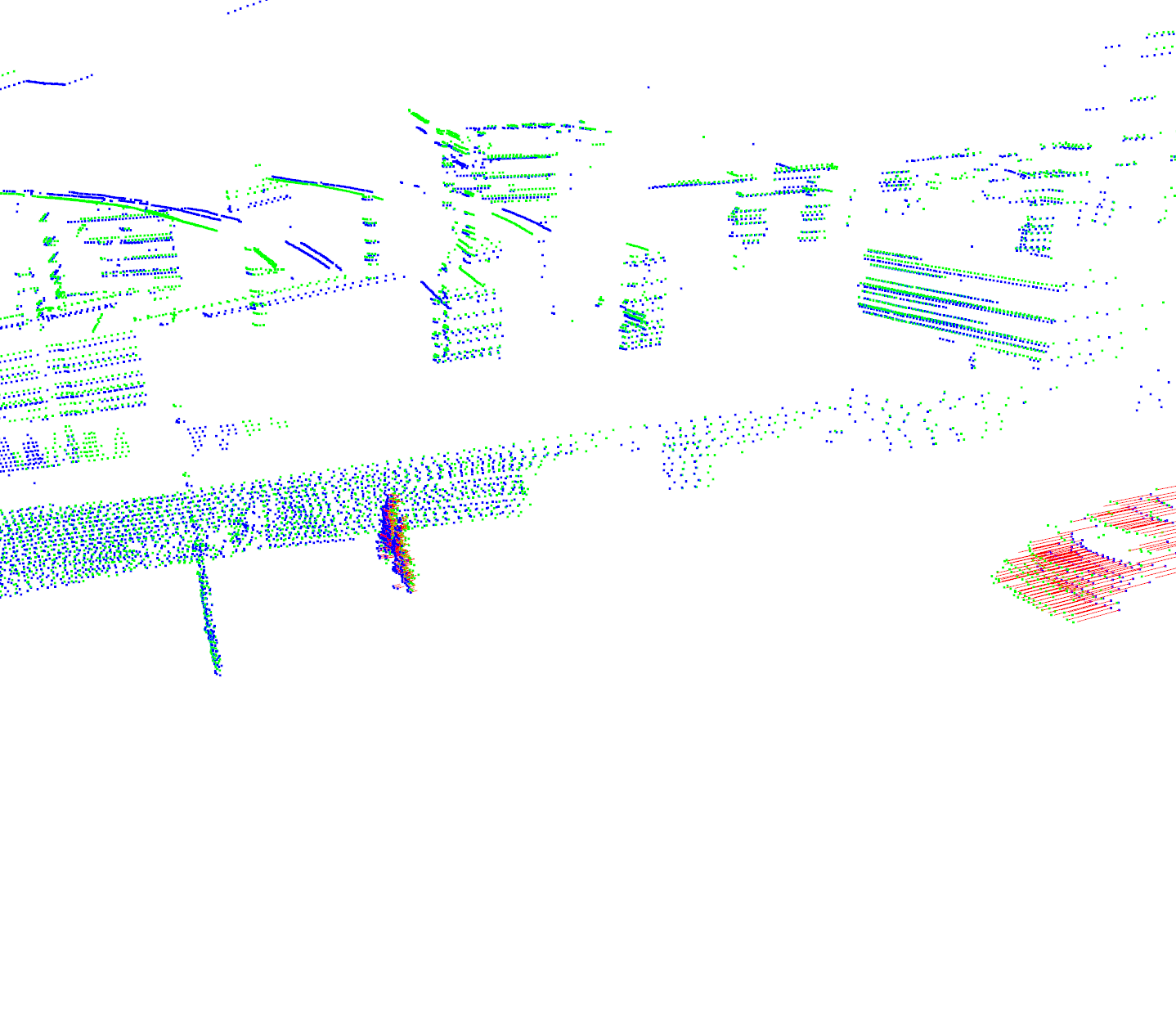}
  \includegraphics[width=\linewidth]{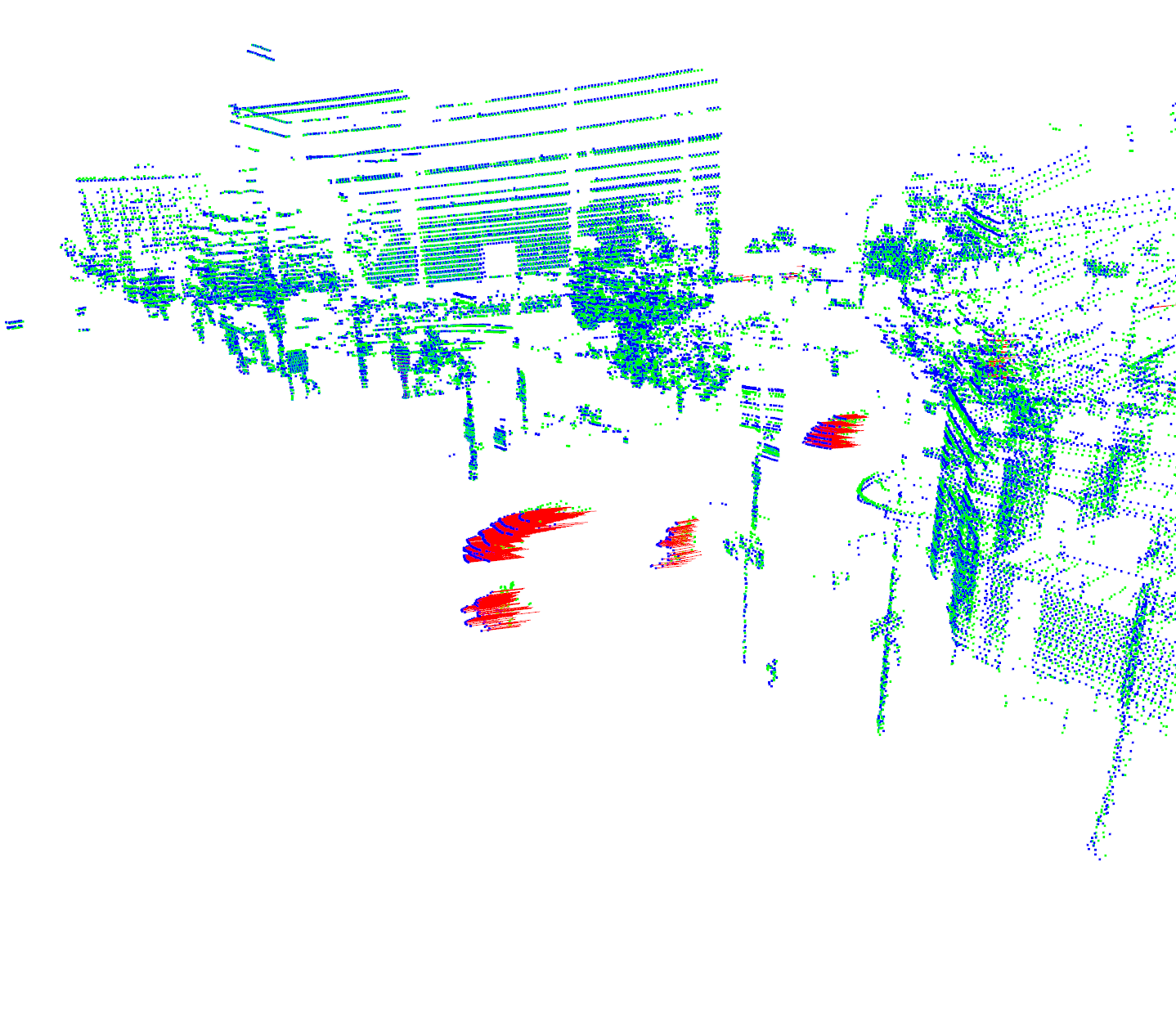}
  \includegraphics[width=\linewidth]{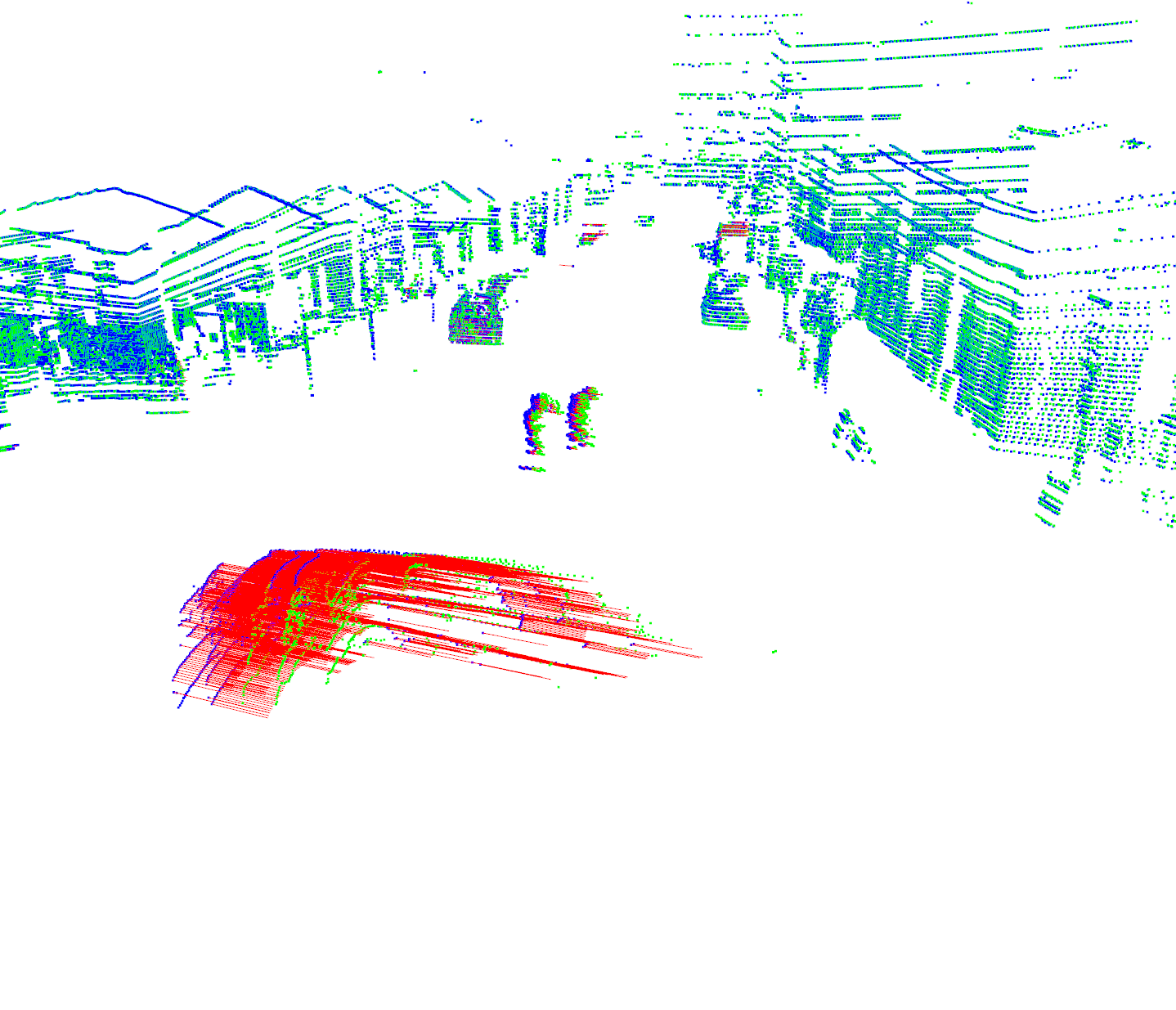}
  \caption{\tinier \textbf{\ourmethod{}}}
  \label{fig:sub6}
\end{subfigure}
\captionsetup{singlelinecheck=off}
\caption[foo bar]{Visualizations of different methods on diverse scenes in Argoverse 2. Each method is estimating flow from the \blue{blue} to the \green{green} point cloud.
\begin{enumerate}[label=Row \arabic*:,leftmargin=4em]
    \item Two pedestrians in left foreground with cars moving in the background. \ourmethod{} is the only method able to describe the pedestrian motion.
    \item Three pedestrians walking across an intersection in front of a stationary car. DeFlow is able to capture the furthest pedestrian, but only \ourmethod{} is able to capture the motion of all three. \ourmethod{} also falsely estimates motion of the moving box truck in the background.
    \item Top view of pedestrians walking down the sidewalk between a building and several cars parked in the street. \ourmethod{} is the only method able to describe the pedestrian motion.
    \item Pedestrians walking down the sidewalk next to a moving car.  \ourmethod{} is the only method able to describe the pedestrian motion.
    \item Two bicyclists riding across an intersection next to driving cars. Most methods are able to capture the training bicyclists and the moving cars, but only NSFP and \ourmethod{} are able to capture the lead bicyclist.
    \item Two pedestrians walk across an intersection while a car drives parallel to them. All methods capture the car motion, but only DeFlow, NSFP, and \ourmethod{} capture most of the pedestrian motion. \ourmethod{} also falsely estimates motion of one of the parked cars far down the street in the background.
\end{enumerate}
}
\figlabel{morequalitativeone}
\end{figure}

We ablate the impact of detector quality on  \ourmethod{} by replacing LE3DE2E ~\citep{wang2023le3de2e} with BEVFusion~\citep{liu2022bevfusion}. We call this new approach \emph{\ourmethodbevfusion{}}. BEVFusion only has 2\% lower mAP than LE3DE2E on the AV2 detection leaderboard\footnote{BEVFusion~\citep{liu2022bevfusion} was second on the \emph{Argoverse 2 2023 3D Detection, Tracking and Forecasting challenge}~\citep{peri2022towards, peri2023empirical, Peri_2022_CVPR}.}, but we find that \ourmethodbevfusion{} performs significantly worse than \ourmethod{}, with 10\% to 22\% drops in performance on Dynamic Normalized EPE (\tableref{tab:bevfusionresults}).

This significant degradation is the result of BEVFusion's poor recall at low confidence thresholds (\tableref{tab:bevfusionconf}). In contrast, LE3DE2E has very high recall at low thresholds (\figref{fig:bevfusionrecall}), producing many candidate boxes for pedestrians in the scene. BEVFusion's false negatives are extremely costly to \ourmethodbevfusion{}, as they result in $\zerovec$ flow estimates that miss 100\% of each false positive pedestrian's motion.

More broadly, a good detector for \ourframework{} is not necessarily one with a high mAP; it is one with very high recall and accurate heading estimates. Notably, these error characteristics enable the tracker to reject false positives. We believe this interaction between the detector and tracker is an important yet subtle point --- two detectors may have the same mAP, but dramatically different performance in our \ourframework{} framework.

\section{Argoverse 2 2024 Scene Flow Challenge}\sectionlabel{sceneflowchallenge}

\begin{figure}[htb]
\centering
\includegraphics{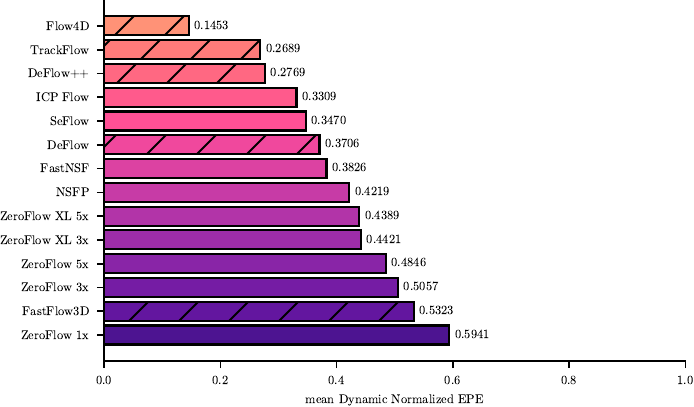}
\caption{mean Dynamic Normalized EPE of submissions to \emph{the Argoverse 2 2024 Scene Flow Challenge} on Argoverse 2's \emph{test} split. Supervised methods shown with hatching. Lower is better.}
\figlabel{competitionmeandynamic}
\end{figure}
\begin{figure}[h!]
\centering
\begin{subfigure}[b]{0.49\textwidth}
    \centering
    \includegraphics{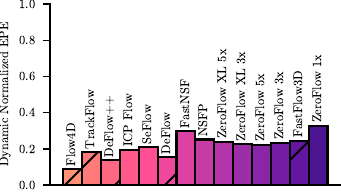}
    \caption{\texttt{CAR}}
    \figlabel{fig:competitioncar}
\end{subfigure}%
\begin{subfigure}[b]{0.49\textwidth}
    \centering
    \includegraphics{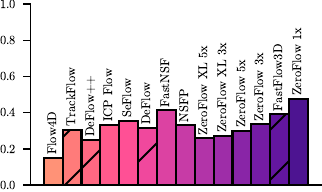}
    \caption{\texttt{OTHER VEHICLES}}
    \figlabel{fig:competitionother-vehicles}
\end{subfigure}
\begin{subfigure}[b]{0.49\textwidth}
    \centering
    \includegraphics{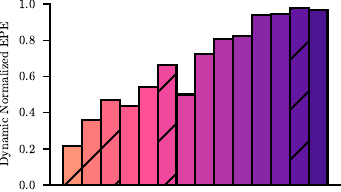}
    \caption{\texttt{PEDESTRIAN}}
    \figlabel{fig:competitionpedestrian}
\end{subfigure}%
\begin{subfigure}[b]{0.49\textwidth}
    \centering
    \includegraphics{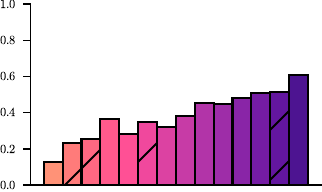}
    \caption{\texttt{WHEELED VRU}}
    \figlabel{fig:competitionwheeled-vru}
\end{subfigure}
\caption{Per meta-class Dynamic Normalized EPE of submissions to \emph{the Argoverse 2 2024 Scene Flow Challenge} on Argoverse 2's \emph{test} split. Supervised methods shown with hatching. Lower is better. Method color and position is consistent between plots.}
\figlabel{competitionmetacatagorydynamic}
\end{figure}

\oureval{} was the basis for the \emph{Argoverse 2 2024 Scene Flow Challenge}\footnote{Full details about the competition can be found at \url{http://argoverse.org/sceneflow}}. The competition featured two tracks: a supervised track, and an unsupervised track, with \ourmethod{} serving as a baseline in the supervised track. Leaderboards for both tracks are ranked by minimum \emph{mean Dynamic} component of \oureval{}.

Notably, Flow4D~\citep{flow4d} significantly improved over all prior supervised methods, halving the dynamic error of \ourmethod{}. Interestingly, Flow4D does not feature any class-aware loss features, instead focusing on architectural improvements over FastFlow3D~\citep{scalablesceneflow}-based architectures (e.g. ZeroFlow~\citep{vedder2024zeroflow}, DeFlow~\citep{zhang2024deflow}). Unsupervised scene flow methods also saw meaningful improvements; ICP-Flow~\citep{lin2024icp} significantly outperformed FastNSF~\citep{fastnsf}, the best performing unsupervised baseline, closely followed by SeFlow~\citep{seflow}.

\section{Discussion}

In this chapter, we highlight that current scene flow methods consistently fail to describe motion on pedestrians and other small objects. We demonstrate that current standard evaluation metrics hide this failure and present \oureval{}, a new class-aware, speed normalized evaluation protocol, to quantify this failure. In addition, we present \ourmethod{}, a frustratingly simple supervised scene flow baseline that achieves state-of-the-art on Threeway EPE and \oureval{}. We argue that current evaluation protocols fail to reveal performance across the distribution of safety-critical objects, and do not contextualize absolute errors in the context of an object's speed. Moreover, we highlight that class and speed aware evaluation is important \emph{even if a method has zero human supervision}. Importantly, we cannot expect any method to meaningfully generalize to the long tail of unknown objects if it cannot provide high quality motion descriptions on a known set of objects. Lastly, \ourmethod{} outperforms prior art by a wide margin because it leverages recent advances in class imbalanced learning. Our approach highlights that supervised scene flow methods should adopt many of the lessons learned by the detection community to properly address class and point imbalances. 

\subsection{TrackFlow's Limitations as a Scene Flow Method}

\ourmethod{} only predicts rigid flow for objects within LE3DE2E's fixed taxonomy because it uses a closed-world bounding box based detector. However, as discussed in \appendixref{faq}, these limitations can be addressed with a different detector architectures, and is not a fundamental limitations of the \ourframework{} framework.

\section{FAQ}\appendixlabel{faq}

\subsection{\ourmethod{} is \emph{just} a tracking method}

Yes, \ourmethod{} is a tracking method applied to the scene flow problem. The state-of-the-art performance of \ourmethod{} suggests that \ourframework{} is a fruitful area of exploration for future work on supervised scene flow.

\subsection{\ourmethod{} uses bounding boxes and thus can only estimate rigid flow --- what does this paper have to say about non-rigid scene flow?}

It's true that \ourmethod{} operates on the level of bounding boxes, but as we discuss in \sectionref{sceneflowdatasets}, public real-world datasets derive motion annotations from bounding box tracks. If non-rigid labels were available, one could train a detector to also regress keypoints (or use an off-the-shelf pretrained method~\citep{yang2023unipose}) and track across those keypoints.

\subsection{\ourmethod{} uses bounding boxes from a detector --- does this mean it cannot detect open-set objects?}

\ourmethod{} uses a class-aware object detector as its bounding box proposer. However, the \ourframework{} framework does not require class annotations -- nothing prevents the use of a class agnostic open world bounding box proposer, either trained like FasterRCNN's RPN~\citep{fastrcnn,fasterrcnn}, Object Localization Network~\citep{kim2021oln}, or via geometric priors~\citep{pmlr-v205-huang23b}.

\subsection{Our metric is ``just'' Threeway EPE extended to multiple classes and multiple speed buckets with normalization, and our method ``just'' combines a detector and tracker. Where is the novelty in this idea?}

The ideas presented in this paper are simple and post-hoc obvious, but serve to highlight catastrophic failures currently overlooked in existing approaches. 


\chapter{\MakeUppercase{Neural Eulerian Scene Flow Fields}}\chapterlabel{eulerflow}

\newcommand{\etal}{et al.}

\renewcommand{\ourpipeline}{SFvODE}
\renewcommand{\ourpipelinefull}{Scene Flow via ODE}

\renewcommand{\ourmethod}{EulerFlow}

\newcommand{\writingnote}[1]{{\color{red} #1}}

\newcommand{\websitelink}{\href{https://vedder.io/eulerflow}{\texttt{vedder.io/eulerflow}}}



\newcommand{\captionfontsize}{\fontsize{7}{7}\selectfont}

In \chapterref{zeroflow} we presented \emph{Scene Flow via Distillation}, a framework to use expensive, unsupervised optimization methods to scale up pseudo-labeling for training feedforward neural networks to estimate scene flow, and concretely instantiated it with ZeroFlow, where we used Neural Scene Flow Prior \cite{nsfp} as the unsupervised pseudo-labeling teacher and FastFlow3D \citep{scalablesceneflow} as the feedforward supervised student. In \chapterref{trackflow}, we then propose a new evaluation metric that reveals, among other things, that \emph{all} scene flow methods, supervised or unsupervised, struggle with small objects. The results from the Argoverse 2 2024 Scene Flow Challenge on this evaluation metric (\sectionref{sceneflowchallenge}) reveal significant innovation in feedfoward architectures\footnote{Flow4D \citep{flow4d}, the winning supervised method, masively outperformed FastFlow3D \citep{scalablesceneflow}, the student used in ZeroFlow, despite using the exact same training loss, indicating its architecture is clearly superior and feedfoward architecture clearly limited the performance of ZeroFlow.}, but unsupervised methods still perform relatively poorly: on pedestrians, the winning unsupervised scene flow method, ICP Flow \citep{lin2024icp}, still has over 40\% dynamic error (\figref{fig:competitionpedestrian}).




This motivates the need for an unsupervised method that can produce extremely high quality scene flow estimates in the offline setting. To pursue this, we reframe scene flow as the task of estimating a continuous space-time ODE that describes motion for an entire observation sequence, represented with a neural prior. Our method, \emph{\ourmethod{}}, optimizes this neural prior estimate against several multi-observation reconstruction objectives, enabling high quality scene flow estimation via pure self-supervision on real-world data. \ourmethod{} works out-of-the-box without tuning across multiple domains, including large-scale autonomous driving scenes and dynamic tabletop settings. Remarkably, \ourmethod{} produces high quality flow estimates on small, fast moving objects like birds and tennis balls, and exhibits emergent 3D point tracking behavior by solving its estimated ODE over long-time horizons. On the Argoverse 2 2024 Scene Flow Challenge, \ourmethod{} outperforms \emph{all} prior art, surpassing the next-best \emph{unsupervised} method by more than $2.5\times$, and even exceeding the next-best \emph{supervised} method by over 10\%. Interactive scene visualizations are available at \websitelink{}

\setcounter{footnote}{0} 
\section{Overview}


\begin{figure}[h!]
    \centering
    \begin{subfigure}[b]{0.359\textwidth}
        \centering
        \includegraphics[width=\textwidth]{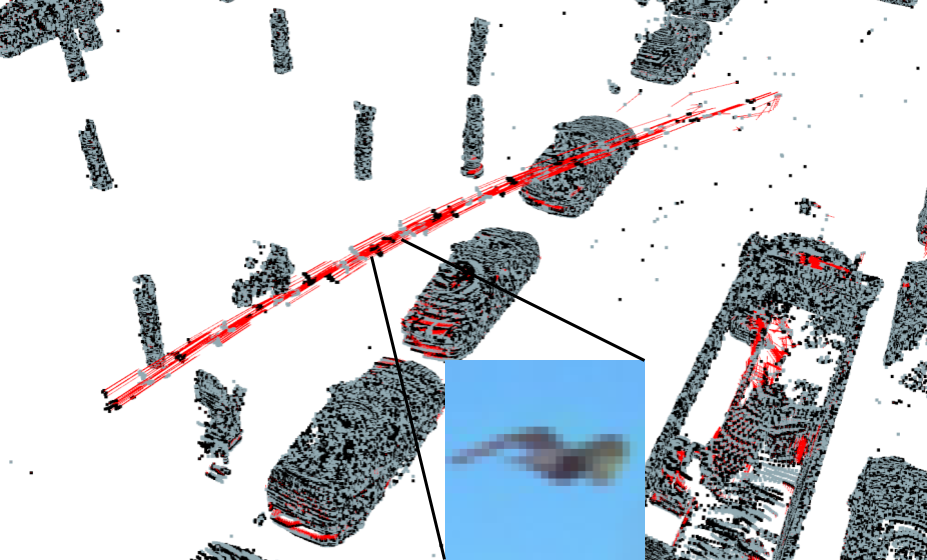}
        \caption{\captionfontsize{} Small object motion extraction...}
        \figlabel{teasersceneflow}
    \end{subfigure}
    \hfill
    \begin{subfigure}[b]{0.248\textwidth}
        \centering
        \includegraphics[width=\textwidth]{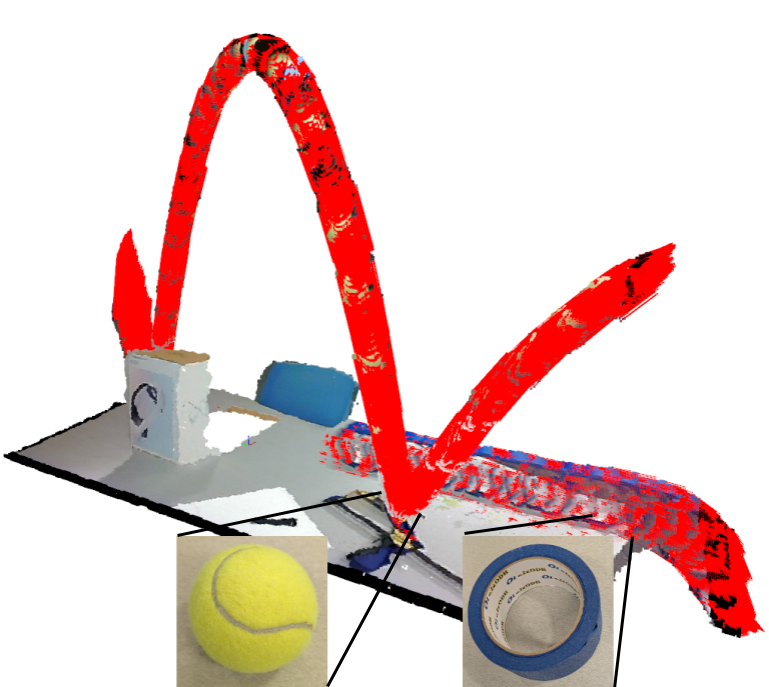}
        \caption{\captionfontsize{} ...in diverse scenes...}
        \figlabel{teaserbounce}
    \end{subfigure}
    \hfill
    \begin{subfigure}[b]{0.359\textwidth}
        \centering
        \includegraphics[width=\textwidth]{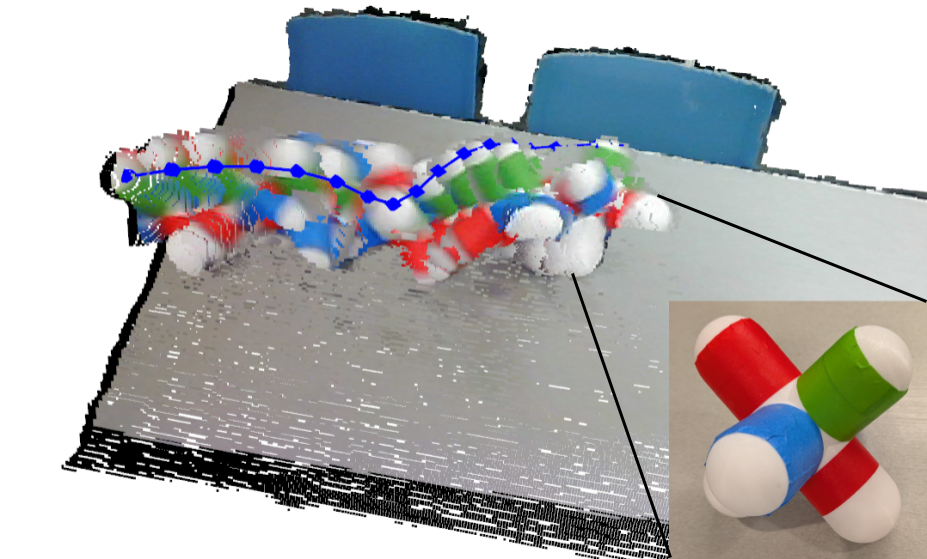}
        \caption{\captionfontsize{} ...with emergent 3D point tracking behavior!}
        \figlabel{teaserjack}
    \end{subfigure}
    \caption{\ourmethod{} is able to capture the motion of small, fast moving objects with few lidar points, such a bird flying in front of an autonomous vehicle (\figref{teasersceneflow}). \ourmethod{}'s flexibility allows it to estimate scene flow for fast-moving table top objects \emph{without additional hyperparameter tuning} (\figref{teaserbounce}). \ourmethod{}'s ODE estimate exhibits emergent 3D point tracking behavior without explicit long-horizon supervision (\figref{teaserjack}). Note that point clouds are shown in color for visualization purposes only; RGB is not used during optimization.}
    \figlabel{teaser}
\end{figure}

\newcommand{\seqlencaptionfontsize}{\fontsize{7}{7}\selectfont}

\begin{figure}[t]
\centering
\begin{subfigure}[b]{0.24\textwidth}
    \centering
    \includegraphics[width=\textwidth]{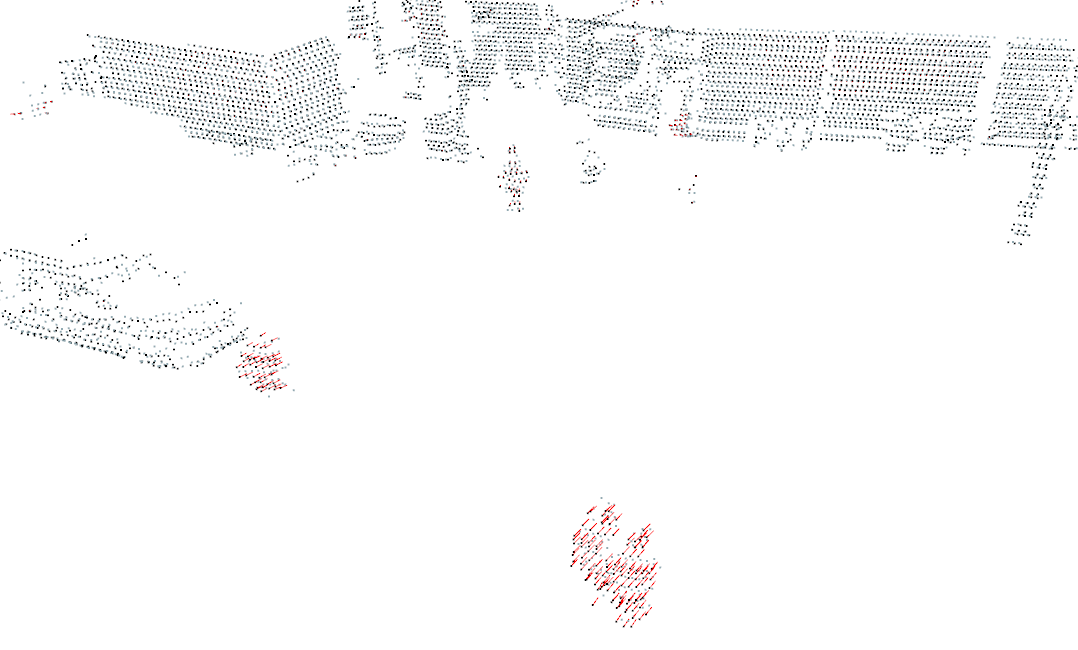}
    \caption{\seqlencaptionfontsize \ourmethod{} (Two Frame)}
    \figlabel{flyingbirdgigachad_twoframe}
\end{subfigure}%
\begin{subfigure}[b]{0.24\textwidth}
    \centering
    \includegraphics[width=\textwidth]{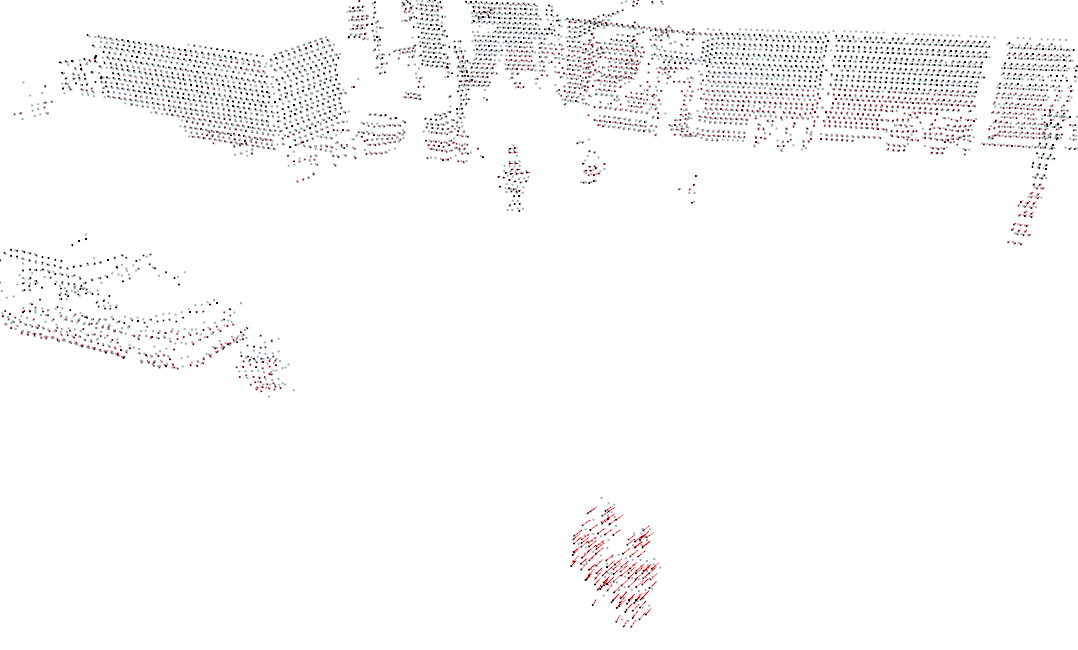}
    \caption{\seqlencaptionfontsize Fast NSF (Two Frame)}
    \figlabel{flyingbirdfastnsf_twoframe}
\end{subfigure}%
\begin{subfigure}[b]{0.24\textwidth}
    \centering
    \includegraphics[width=\textwidth]{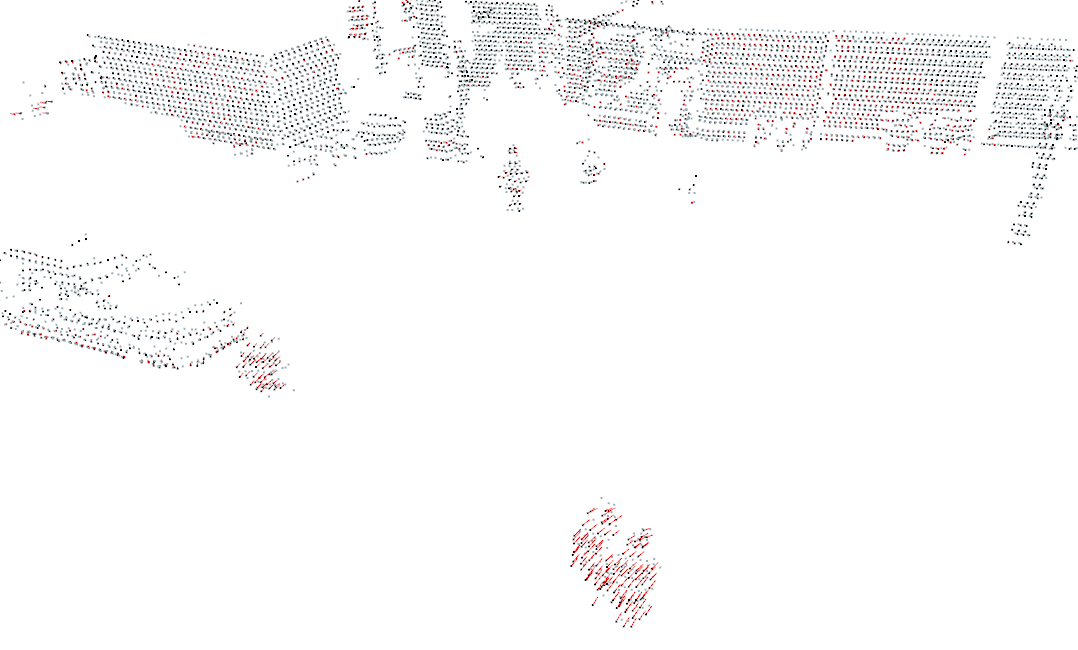}
    \caption{\seqlencaptionfontsize \citeauthor{liu2024selfsupervisedmultiframeneuralscene} (Two Frame)}
    \figlabel{flyingbirdliuetal_twoframe}
\end{subfigure}%
\begin{subfigure}[b]{0.24\textwidth}
    \centering
    \includegraphics[width=\textwidth]{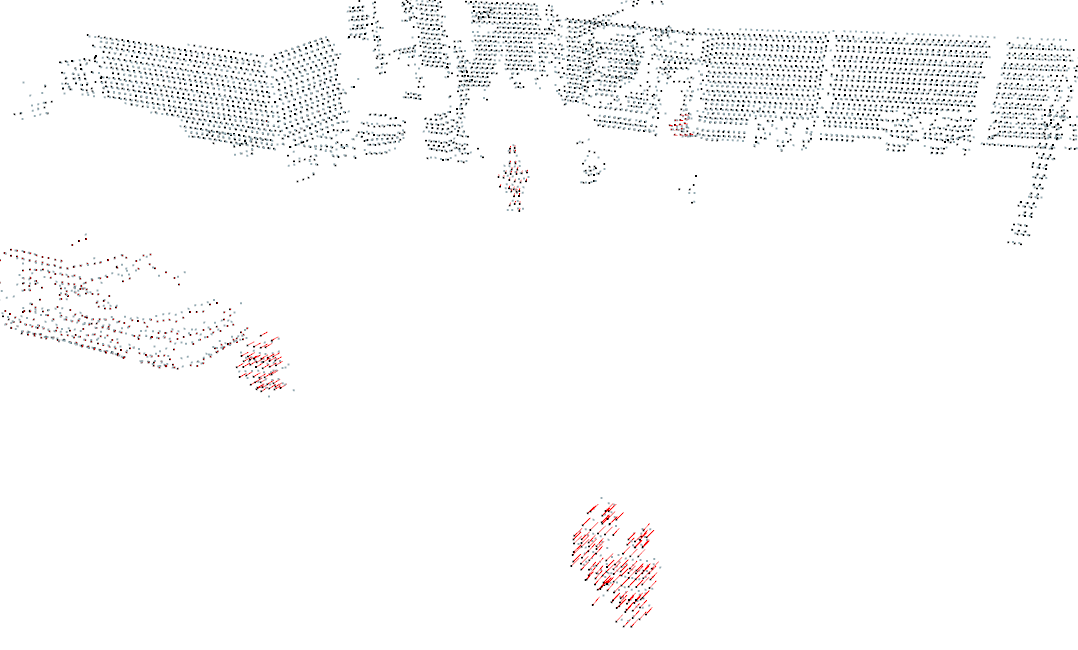}
    \caption{\seqlencaptionfontsize Ground Truth (Two Frame)}
    \figlabel{flyingbirdgroundtruth_twoframe}
\end{subfigure}

\begin{subfigure}[b]{0.24\textwidth}
    \centering
    \includegraphics[width=\textwidth]{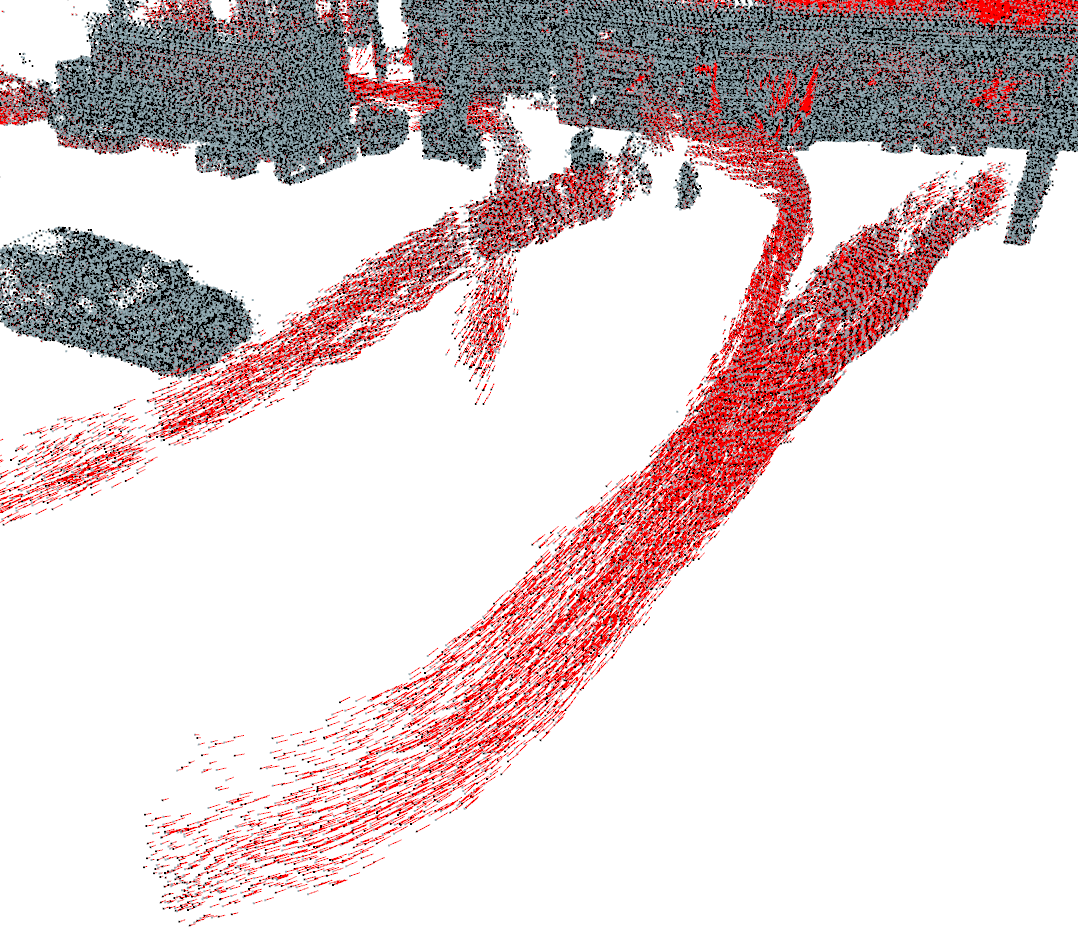}
    \caption{\seqlencaptionfontsize \ourmethod{} (Full)}
    \figlabel{flyingbirdgigachad_fullsequence}
\end{subfigure}%
\begin{subfigure}[b]{0.24\textwidth}
    \centering
    \includegraphics[width=\textwidth]{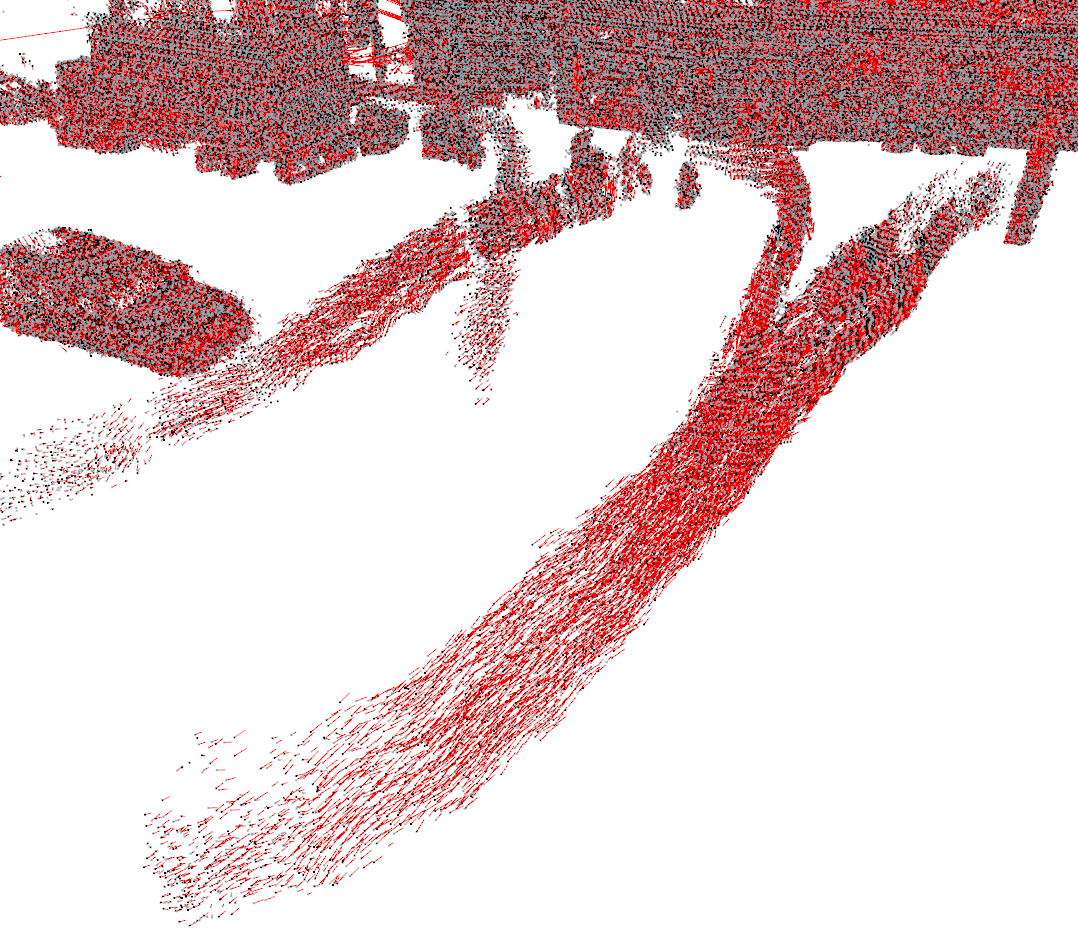}
    \caption{\seqlencaptionfontsize Fast NSF (Full)}
    \figlabel{flyingbirdfastnsf_fullsequence}
\end{subfigure}%
\begin{subfigure}[b]{0.24\textwidth}
    \centering
    \includegraphics[width=\textwidth]{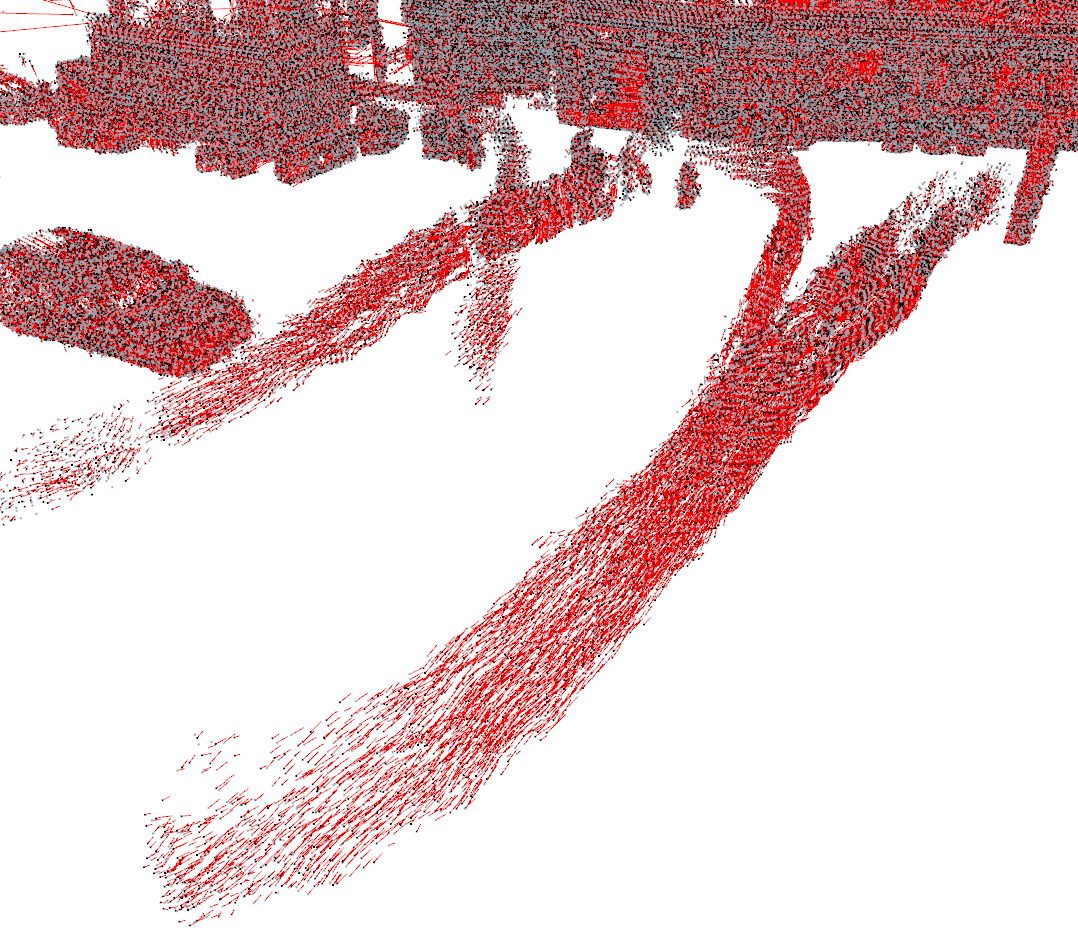}
    \caption{\seqlencaptionfontsize \citeauthor{liu2024selfsupervisedmultiframeneuralscene} (Full)}
    \figlabel{flyingbirdliuetal_fullsequence}
\end{subfigure}%
\begin{subfigure}[b]{0.24\textwidth}
    \centering
    \includegraphics[width=\textwidth]{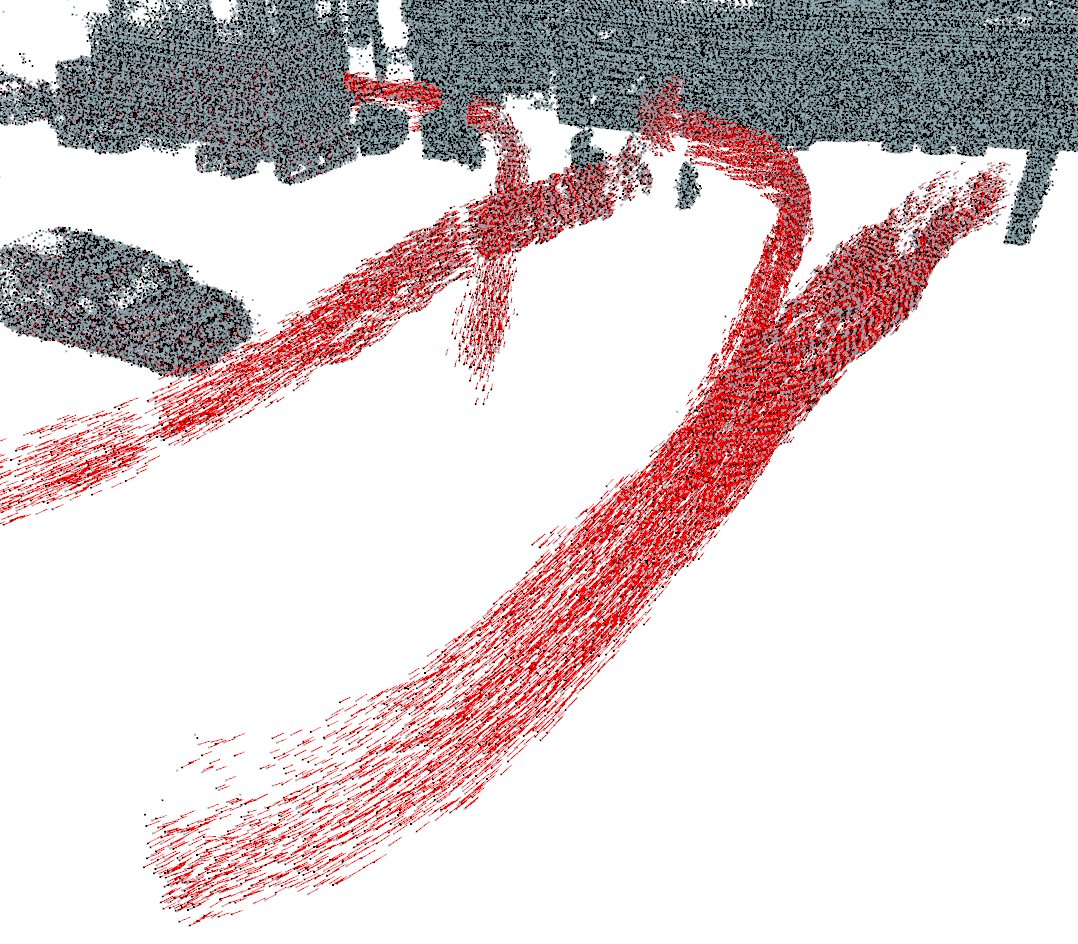}
    \caption{\seqlencaptionfontsize Ground Truth (Full)}
    \figlabel{flyingbirdgroundtruth_fullsequence}
\end{subfigure}

\caption{ We visualize the same scene as depicted in the motivational figure for \chapterref{trackflow} (\figref{teaserfigure}), featuring five pedestrians crossing the street in front of a stopped car. As we motivated in \chapterref{trackflow}, this scene is cherrypicked to have unusually high density lidar returns, making it particularly easy to estimate flow. \figrefs{flyingbirdgigachad_twoframe}{flyingbirdgroundtruth_twoframe} depict a two-frame flow visualization of \ourmethod{} and several strong baselines. Notably, only visualizing flow over two frames makes it difficult to distinguish flow quality. In contrast,  \figrefs{flyingbirdgigachad_fullsequence}{flyingbirdgroundtruth_fullsequence} depict flow vectors over the full sequence, making differences in quality clear; for example, \ourmethod{} is the only one without artifacts on the stopped car.}
\figlabel{walkingpeds}
\end{figure}

\poorparagraph{\ourpipelinefull{}} In \figref{walkingpeds}, visual assessment of scene flow quality is much easier in an accumulated global frame; while incomplete due to an implicit time axis, these accumulated flow vectors allow viewers to imagine how positions in 3D space evolve over \emph{many} timesteps, and compare that to predicted flows. This imagination of scene flow as continuous motion over large time intervals motivates us to model scene flow as an Ordinary Differential Equation (ODE) that describes the scene's instantaneous motion across continuous position and time; this is the same formalism used to represent fluid flow or heat flow in classical mechanics. Scene flow estimation then becomes the task of estimating this ODE. We can straightfowardly represent this ODE estimate with a neural prior~\citep{nsfp} and optimize it against scene flow surrogate objectives, both over single frame pairs and extended \emph{across arbitrary time intervals}, unlocking new optimization objectives that produce better quality estimates. We formalize this in \sectionref{ourpipeline} and propose the \emph{\ourpipelinefull{}} framework.

\poorparagraph{\ourmethod{}} We instantiate \ourpipelinefull{} with standard point cloud distance objectives like Chamfer Distance to create \emph{\ourmethod{}}. Notably, \ourmethod{} outperforms \emph{all} prior art, supervised or unsupervised, on the Argoverse~2 2024 Scene Flow Challenge and Waymo Open Scene Flow benchmark. It outperforms prior \emph{unsupervised} methods by a large margin ($>2.5\times$ mean dynamic error reduction), and is able to capture small, fast moving objects, including those outside of labeled taxonomies (e.g.\ the flying bird in \figref{teasersceneflow}). Compared to the unsupervised methods we used in \chapterref{zeroflow}, the gap is even larger: \ourmethod{} is over $3\times$ better on dynamic error than Neural Scene Flow Prior \citep{nsfp}, the teacher used for pseudolabeling, and the largest variant of ZeroFlow (ZeroFlow XL 5x). Similarly, compared to our strong supervised baseline TrackFlow we proposed in \chapterref{trackflow}, \ourmethod{} is over $2\times$ better on dynamic error.

Due to its simplicity, \ourmethod{} is able to provide high quality scene flow out-of-the-box on real-world data for other important domains such as dynamic tabletop settings (\figref{teaserbounce}) \emph{without} domain-specific tuning. Finally, we show that simple ODE solving techniques like Euler integration can be used to extract 3D point tracking behaviors (\figref{teaserjack}), which serves as both an exciting emergent behavior as well as a mechanism for visualizing and interpreting the quality of the continuous ODE estimate.

In this chapter we present four primary contributions:

\begin{list}{$\bullet$}{\setlength{\leftmargin}{10pt}} 
  \setlength{\itemsep}{0pt}   
  \setlength{\parskip}{0pt}   
    \item We propose \emph{\ourpipelinefull{}} (\ourpipeline{}), a reframing of scene flow estimation as the task of fitting a ODE that describes the change of continuous positions over continuous time, unlocking a new class of surrogate objectives that enable better scene flow estimates.
    \item We instantiate \ourpipeline{} with \emph{\ourmethod{}}, a flexible \textbf{unsupervised} scene flow method that achieves \textbf{state-of-the-art} performance on the Argoverse~2 2024 Scene Flow Challenge, \textbf{beating all prior supervised and unsupervised methods}, including the methods we propose in \chapterref{zeroflow} by more than $3\times$ and the method we propse in \chapterref{trackflow} by more than $2\times$.
    \item We study \ourmethod{} and show its strong performance is derived from its ability to optimize its ODE estimate against the full sequence of observations over arbitrary time horizons.
    \item We show that \ourmethod{}'s simple, flexible formulation allows it to run unmodified on a variety of domains, with emergent capabilities like 3D point tracking behavior.
\end{list}

\section{Background and Related Work}\sectionlabel{relatedwork}

\poorparagraph{Input / Output Formulation of Scene Flow} \cite{dewan2016rigid}'s choice to formulate scene flow using \emph{only} two input frames is arbitrary; it's the minimal information needed to extract rigid motion, but there are not real-world problems constrained to \emph{only} have access to two frames. Indeed, using five or ten frames of past observations is standard practice in the 3D detection literature~\citep{cbgs, vedder2022sparse, peri2022futuredet,peri2023empirical, surveyofpersonrecog}, and multi-frame formulations have started to appear in the scene flow literature: \citet{liu2024selfsupervisedmultiframeneuralscene} and Flow4D~\citep{flow4d} use three ($\pointcloudtmone, \pointcloudt, \pointcloudtpone$) and  five input frames ($\pointcloudsub{t-3}, \ldots, \pointcloudsub{t+1}$) respectively to predict $\flowttpone$. As we discuss in \sectionref{ourpipeline}, rather than just using more observations to estimate flow for a single frame pair, we formulate scene flow as a joint estimation problem: given the full observation sequence $\left( \pointcloud_0, \ldots,\pointcloud_N \right)$, we estimate \emph{all} flows $\flow_{0,1}, \ldots, \flow_{N-1,N}$ between \emph{all} adjacent observations.

\newcommand{\nsfpforward}{\network}
\newcommand{\nsfpbackward}{\network'}

\poorparagraph{Neural Scene Flow Prior} 
\citet{nsfp} propose Neural Scene Flow Prior (NSFP), a widely adopted unsupervised scene flow approach. NSFP uses the inductive bias of the smooth, restricted learnable function class of two ReLU MLP coordinate networks (8 hidden layers of 128 neurons); $\nsfpforward$ to estimate forward flow and $\nsfpbackward$ to estimate backwards flow, minimizing 

\begin{equation}
\small
  \equationlabel{nsfploss}
  \chamferdistance{\pointcloudt{} + \nsfpforward\left(\pointcloudt{}\right)}{\pointcloudtpone} + \norm{\pointcloudt{} + \nsfpforward\left(\pointcloudt{}\right) + \nsfpbackward\left(\pointcloudt{} + \nsfpforward\left(\pointcloudt{}\right) \right) - \pointcloudt}_2 \enspace ,
\end{equation}

where $\chamferdistancename$ is defined as the standard $L_2$ Chamfer distance, but with per-point distances above 2 meters set to zero in order to reduce the influence of outliers. NSFP is optimized for at most 1000 steps with early stopping.



\poorparagraph{Motion Beyond Two Frames} 
 \citet{ntp} tackles the adjacent problem of estimating 3D point \emph{trajectories} over 25 frames with Neural Trajectory Prior (NTP) by jointly optimizing three separate ReLU MLP neural priors: 1) a sinusoidal embedded, time conditioned, 25 frame trajectory basis estimator ($\textup{embed}(t) \mapsto 256 \times 25 \times 3$ tensor, where $256$ is the dimension of the trajectory basis), 2) a coordinate network bottleneck encoder, and 3) a bottleneck decoder to estimate a per-point linear combination over the learned trajectories. Trajectories are optimized for both a one-frame lookahead $L_2$ Chamfer loss and a cyclic consistency loss over the entire velocity space trajectory.

\poorparagraph{Deformation in Reconstruction} Nerfies~\citep{park2021nerfies} and DynamicFusion~\citep{dynamicfusion} estimate a deformation field to warp a canonical frame to explain the observed frame. While capable of describing small motions, these methods require a canonical frame that contains all of the relevant geometry to deform; however, in large, highly dynamic scenes like autonomous driving, there is often no frame that contains all moving objects. By comparison, \ourpipelinefull{} does not assume the existence of a canonical frame, instead only describing how the scene changes.

\begin{figure}[tbh]
    \centering
\includegraphics[width=\textwidth]{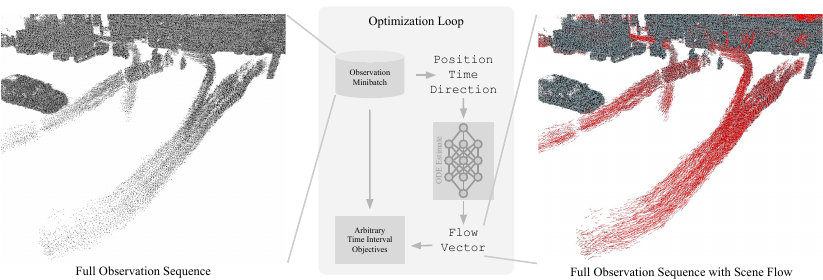}
    \caption{Overview of our \emph{\ourpipelinefull{}} framework, which estimates a ODE across the entire observation sequence by optimizing against multi-frame objectives. This ODE estimate is represented with a neural prior \citep{nsfp}, providing a flexible, general representation for describing position-time motion.}
    \figlabel{method}
\end{figure}

\section{\ourpipelinefull{}}\sectionlabel{ourpipeline}

Prior art in scene flow consume various number of frames $\left(\pointcloud_{t-N}, \ldots, \pointcloudtpone\right)$, but these methods are ultimately only tasked with estimating flow vectors between $\pointcloudt$ and $\pointcloudtpone$. We instead pose the problem of estimating a time-conditioned flow field that describes motion for \emph{all} adjacent point clouds $\pointcloudt, \pointcloudtpone$ in the entire sequence $\left( \pointcloud_0, \ldots,\pointcloud_N \right)$. To do this, rather than describing scene flow as positional change over a fixed interval ($\flowgtttpone$ are residual vectors over the interval $t$ to $t+1$) as we did in \sectionref{relatedwork}, we can instead express these changes as a differential equation that describes positional change over \emph{continuous} time.

\begin{figure}[ht]
    \centering
    \begin{tikzpicture}[scale=1.2]

        \node at (-3.5, 4.5) {\textbf{Eulerian View}};
        
        \draw[step=1cm,gray,very thin] (-5,1) grid (-2,4);
        
        \draw[->,thick] (-4.5,3.5) -- (-4.5+0.5,3.5+0.3);
        \draw[->,thick] (-3.5,3.5) -- (-3.5+0.2,3.5+0.6);
        \draw[->,thick] (-2.5,3.5) -- (-2.5+0.3,3.5+0.2);
        
        \draw[->,thick] (-4.5,2.5) -- (-4.5+0.4,2.5+0.4);
        \draw[->,thick] (-3.5,2.5) -- (-3.5+0.6,2.5+0.1);
        \draw[->,thick] (-2.5,2.5) -- (-2.5+0.3,2.5+0.5);
        
        \draw[->,thick] (-4.5,1.5) -- (-4.5+0.2,1.5+0.4);
        \draw[->,thick] (-3.5,1.5) -- (-3.5+0.5,1.5+0.3);
        \draw[->,thick] (-2.5,1.5) -- (-2.5+0.4,1.5+0.4);
        
        \node at (4, 4.5) {\textbf{Lagrangian View}};
        
        \draw[->,thick,blue] (2,1) .. controls (2.5,2) and (3,3) .. (4,4);
        \draw[->,thick,green] (2.5,1) .. controls (3,2) and (4,3) .. (5,4);
        \draw[->,thick,red] (3,1) .. controls (3.5,2) and (4.5,3) .. (6,4);

        \filldraw[blue] (2,1) circle (1pt) node[anchor=south] {\tiny$A_0$};
        \filldraw[green] (2.5,1) circle (1pt) node[anchor=south] {\tiny$B_0$};
        \filldraw[red] (3,1) circle (1pt) node[anchor=south] {\tiny$C_0$};

        \filldraw[blue] (4,4) circle (1pt) node[anchor=north] {\tiny$A_t$};
        \filldraw[green] (5,4) circle (1pt) node[anchor=north] {\tiny$B_t$};
        \filldraw[red] (6,4) circle (1pt) node[anchor=north] {\tiny$C_t$};

    \end{tikzpicture}
    \caption{Comparison of Eulerian and Lagrangian descriptions of 2D flow. An Eulerian view characterizes a flow field via instantaneous velocities at many different points, while a Lagrangian view characterizes a flow field via trajectories of many different particles across time. Both approaches are valid ways of describing an underlying flow field, and with sufficient characterization one view can be readily converted to another.}
    \figlabel{euler_vs_lagrange}
\end{figure}
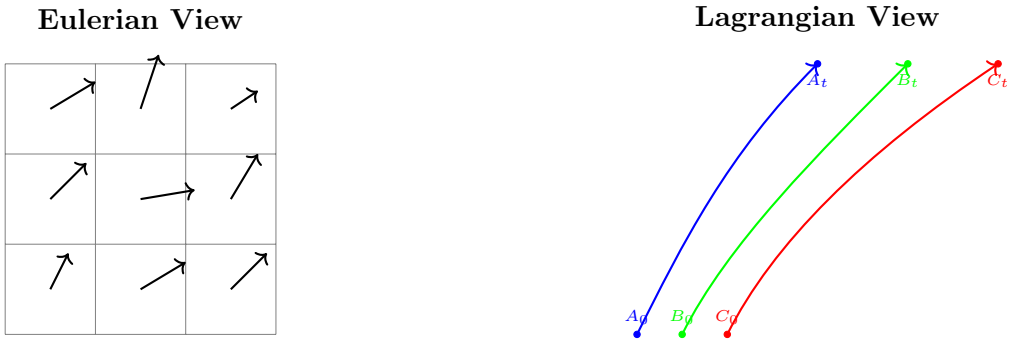

Formally, given a scene, let $L(x_0,y_0,z_0,t)$ be the Lagrangian view of the scene's true flow field, i.e.\ a continuous function that, based on a canonical frame at time $0$, describes the true position of the canonical frame particle $x_0,y_0,z_0$ at some other time $t$. As we discuss in \sectionref{relatedwork}, this Lagrangian view is common in the the deformable reconstruction literature, and the requirement for a canonical frame definition means these approaches struggle to describe scenes where not all moving structure is located in a single frame (e.g.\ objects pop in and out of the scene at different times).

To break this canonical frame dependence, we choose to take an Eulerian view of the flow field, i.e.\ $F = \frac{d L}{d t}$, which describes the velocity of a query point at some arbitrary time. As we show in our derivation in \appendixref{derivation}, this formulation forms an ODE such that, when estimating a point's trajectory from $t$ to $t'$, we don't also need to know its position in some other canonical frame; instead, we can just set the initial conditions of the ODE at $t$ to $x_t, y_t, z_t$ and utilize an off-the-shelf ODE solver (e.g. Euler integration) to extract flow from $t$ to $t'$, expressed as $E(x_t, y_t, z_t, t, t')$.

Grounding this in the scene flow estimation problem, we do not know $F$; however, we can represent an estimate of $F$ with a neural prior $\network$ ($F \approx \network$), and optimize $\network$ against optimization objectives. This framing, which we formalize into the \emph{\ourpipelinefull{}}  framework (\ourpipeline{}; \figref{method}), allows $\network$ to benefit from constructive interference between objectives, as well as enables us to formulate objectives over arbitrarily long time horizons, unlocking high quality estimates.

\section{\ourmethod{}}

\emph{\ourpipelinefull{}} proposes a framework where the neural prior $\network$ represents an estimate of the Eulerian flow field $F$  (i.e. $F \approx \network$); however, it does not prescribe the optimization objectives for $\network$. Thus, we instantiate {\ourpipelinefull{}} with \emph{\ourmethod{}}, a point cloud only scene flow method\footnote{Visualizations shown in color only for better viewing; color is not used during optimization. However, 
\ourmethod{} can use RGB-based monodepth estimates to compute flow on the resulting point cloud (\appendixref{monodepth})} with reconstruction and cyclic consistency objectives across the entire sequence of observations.

As we show in \equationref{eulertrajectory} (\appendixref{neuralpriorrollout}), we can use $\network$'s Eulerian flow field estimate to extract an estimated point trajectory from $x_t, y_t, z_t$ at $t$ to some future location at time $t'$ via Euler integration over $\network$ without requiring a canonical frame definition, i.e. $E_\network(x_t, y_t, z_t, t, t')$. By extracting point trajectories for every point in $\pointcloudt$ using $E_\network$, we can form a traditional scene flow estimate of $\flowgtttpone$, but we can also estimate flow to arbitrary future or prior timesteps (e.g.\ $\flowgt_{t,t+2}$ or $\flowgt_{t, t-1}$). This unlocks powerful multi-frame reconstruction objectives: we can now pose reconstruction surrogate objectives between \emph{any} two point clouds in our observation sequence, not just adjacent point clouds $\pointcloudt$ and $\pointcloudtpone$. Similarly, we can straightforwardly pose cyclic consistency objectives by composing $\flowgt_{t,t+1}$ and $\flowgt_{t + 1, t}$. Formally, for flow extraction for $\pointcloudt$ from $t$ to $t+k$ (for any $k \in \mathbb{Z}$), we define

\begin{equation}
\kstep{\pointcloudt}{k} = \forall p \in \pointcloudt: E_\network(p_{xt}, p_{yt}, p_{zt}, t, t+k)\enspace ,
\end{equation}

enabling us to pose $\network$'s optimization objective $\forall \pointcloudt \in \left( \pointcloud_0, \ldots,\pointcloud_N \right)$ across the window of size $W$

\begin{equation}\equationlabel{gigachadloss}
    \arg\min_{\network} \sum{
    \begin{array}{l}
        \forall k \in \{-W, \ldots, W\} \setminus \{0\}:  \chamferdistance{\kstep{\pointcloudt}{k}}{\pointcloudsub{t+k}}\\
        \alpha \norm{\kstep{\kstep{\pointcloudt}{1}}{-1} - \pointcloudt}_2
        %
    \end{array}
    } \enspace .
\end{equation}

\equationref{gigachadloss} gives rise to our method's name of \emph{\ourmethod{}}; in order to optimize $\network$, our estimate of the Eulerian flow field $F$, we perform Euler integration to extract point cloud flow estimates as part of reconstruction losses. \ourmethod{} is far simpler than prior art, requiring only a single optimization loop over a single neural prior $\network$ compared to NSFP's two priors $\network$ and $\network'$. Our neural prior $\network$ is a straightforward extension to NSFP's coordinate network prior. Like with NSFP, $\chamferdistancename$ is defined as the standard $L_2$ Chamfer distance with per-point distances below 2 meters. As we show in \sectionref{experiments}, \ourmethod{}'s simple ODE estimation formulation across multiple observations produces high quality flow, and solving this ODE over arbitrary time spans unlocks emergent point tracking behavior.

\subsection{Implementation details}\appendixlabel{implementationdetails}

Our neural prior $\network$ is a straightforward extension to NSFP's coordinate network prior\footnote{Hyperparameters (e.g.\ filter width of 128) of NSFP's prior are kept fixed, except for depth (\sectionref{mlpdepth}).}; however, instead of taking a 3D space vector (positions $X, Y, Z \in \mathbb{R}$) as input, we encode a 5D space-time-direction vector: positions $X,Y, Z, \in \mathbb{R}$, sequence normalized time $t \in [-1, 1]$ (i.e.\ the point cloud time scaled to this range), and direction $d \in \{\dirbackward = -1, \dirforward = 1\}$. This simple encoding scheme enables description of arbitrary regions of the ODE, allowing for the ODE to be queried at frequencies different from the sensor frame rate. Euler integration enables simple implementation of multi-step forward, backward, and cyclic consistency losses without extra bells and whistles. For efficiency, we use Euler integration with $\Delta{}t$ set as the time between observations for our ODE solver, enabling support for arbitrary sensor frame rates, and set the cycle consistency balancing term $\alpha = 0.01$ and optimization window $W=3$ for all experiments.

\section{Deriving \ourmethod{}'s ODE}\appendixlabel{derivation}\sectionlabel{derivation}

\subsection{Formulating the ODE}

Imagine we want some function that, given a (possibly moving) particle in some canonical frame (i.e.\ time $0$), can describe its location at an arbitrary future time $t$, i.e.\ a Lagrangian description of motion (\figref{euler_vs_lagrange}).

\begin{equation}
  L(x_0, y_0, z_0, t ) = x_t, y_t, z_t  \enspace .
\end{equation}

For sake of notation to access $x_t, y_t, z_t$ individually, we can define 

\begin{align}
  L_x(x_0, y_0, z_0, t) &= x_t\\  
  L_y(x_0, y_0, z_0, t) &= y_t\\
  L_z(x_0, y_0, z_0, t) &= z_t \enspace .
\end{align}

Similarly, we can imagine wanting to define the instantaneous velocity of particles across the field through time, i.e.\ a Eulerian description of motion (\figref{euler_vs_lagrange}). Thus, formally, let $F(x, y, z, t)$ describe the instantaneous velocity of a point $x,y,z$ at some arbitrary time $t$, i.e.\

\begin{equation}
  \frac{d L(x_0, y_0, z_0, t)}{d t} = \frac{d L}{d t} = \left(\frac{d L_x}{d t}, \frac{d L_y}{d t},\frac{d L_z}{d t}\right) = F(x_t , y_t, z_t, t)  \enspace .
\end{equation}

$F$ is defined in terms of the total derivative of $L$ with respect to $t$, as $x_0, y_0, z_0$ are initial conditions that do not vary with time (i.e. $\frac{dL}{dt} = \frac{\partial L}{\partial t} + \frac{\partial L}{\partial x_0} \frac{d x_0}{d t} + \frac{\partial L}{\partial y_0} \frac{d y_0}{d t} + \frac{\partial L}{\partial z_0} \frac{d z_0}{d t} = \frac{\partial L}{\partial t}$, as $\frac{dx_0}{dt} = \frac{dy_0}{dt} = \frac{dz_0}{dt} = 0$). Thus we can we can exactly define $L$ recursively in terms of the initial conditions and $F$, i.e.

\begin{equation}
  L(x_0, y_0, z_0, t) = \left({x_0, y_0, z_0}\right) + \int_0^t F\left(L_x(x_0, y_0, z_0, \tau), L_y(x_0, y_0, z_0, \tau), L_z(x_0, y_0, z_0, \tau), \tau\right)d\tau   
\end{equation}

or, more compactly,

\begin{equation}
  L(x_0, y_0, z_0, t) = \left({x_0, y_0, z_0}\right) + \int_0^t F(x_\tau, y_\tau, z_\tau, \tau)d\tau \enspace .
  \equationlabel{ldefinitionzerotot}
\end{equation}

Our function $L$ can thus be defined as a multi-dimensional ODE in terms of $F$ with initial conditions $x_0, y_0, z_0$. 

\subsection{Arbitrary start and end times from the Eulerian formulation}\appendixlabel{arbitrarystartandend}

In the above derivation, $L$ requires that a moving point be defined in terms of a canonical frame defined at time $0$, as is common in the deformation in reconstruction literature. However, the Eulerian formulation has no such requirement, allowing us to select arbitrary start and end times across different point queries. To showcase this, using $F$ we can extract the trajectory of a particle at $t$ across the range $[t,t']$ starting at $x_t, y_t, z_t$ simply by changing the range of the integral in the expression of \equationref{ldefinitionzerotot}, i.e.

\begin{equation}
 E(x_t, y_t, z_t, t, t') = \left({x_t, y_t, z_t}\right) + \int_{t}^{t'} F(x_\tau, y_\tau, z_\tau, \tau)d\tau \enspace .  
  \equationlabel{rangeshifttrick}
\end{equation}

While $E$ and $L$ appear similar on their face, $E$ is strictly more flexible than $L$. In principle you could choose to redefine $L$ to use $t$ as the time for your canonical frame, but this is \emph{global} choice; you cannot do this on a per-query basis. However, with $E$'s  Eulerian framing, we can extract a different point's trajectory from the entirely different range $t^\dagger$ to $t^\ddagger$ (i.e.\ $E(x_{t^\dagger}, y_{t^\dagger}, z_{t^\dagger}, t^\dagger, t^\ddagger)$) without concern for a canonical frame definition. It need not even be the case that $t < t'$; indeed, this extraction works even if $t > t'$, i.e.\ extracting the backwards trajectory through time.

\subsection{Euler Integration to approximately solve the ODE}

If $F$ is of arbitrary form and we want to compute the concrete values of $L$, we cannot exactly compute the continuous integral from $0$ to $t$; we must approximate this with finite differences. Thus, we split the time range $0$ to $t$ into $k$ steps; each step is of size $\frac{t}{k}$. Thus, we can again define $L$ via recursion, but this time explicitly,

\begin{align}
  L(x_0, y_0, z_0, 0) &= \left({x_0, y_0, z_0}\right) \\
  L(x_0, y_0, z_0, \tau + \frac{t}{k}) &\approx L(x_0, y_0, z_0, \tau) + \frac{t}{k} \cdot F(x_{\tau}, y_{\tau}, z_{\tau}, \tau) \enspace, 
\end{align}

or directly without recursion,

\begin{equation}
    L(x_0, y_0, z_0, t) \approx \left({x_0, y_0, z_0}\right) + \sum_{n=1}^k \frac{t}{k} \cdot F(x_{n \frac{t}{k}}, y_{n \frac{t}{k}}, z_{n \frac{t}{k}}, n \frac{t}{k}) \enspace .
    \equationlabel{expliciteuler}
\end{equation}

This finite difference solving approach is Euler integration.

\subsection{Estimating the flow field with \ourmethod{}'s neural prior}\appendixlabel{neuralpriorrollout}

For a given scene, we do not have access to $L$ or $F$ directly; these are are the \emph{true} functions that unique characterize the underlying motion of the scene that we are trying to estimate. For \ourmethod{}, represent our estimate of the scene's flow field $F$ with a neural prior, $\network$, i.e. 

\begin{equation}
F(x,y,z,t) \approx \network(x,y,z,t) \enspace .
\equationlabel{eulerflownetworkequiv}
\end{equation}

Thus

\begin{equation}
    L(x_0, y_0, z_0, t) \approx \left({x_0, y_0, z_0}\right) + \sum_{n=1}^k \frac{t}{k} \cdot \network{}\left(x_{n \frac{t}{k}}, y_{n \frac{t}{k}}, z_{n \frac{t}{k}}, n \frac{t}{k}\right) \enspace .
    \equationlabel{expliciteulernetworkequiv}
\end{equation}

and, using the arbitrary start and end definition from \appendixref{arbitrarystartandend}, with $k$ steps from the range $t$ to $t'$ and $\delta = \frac{t' - t}{k}$

\begin{equation}
    E(x_t, y_t, z_t, t, t') \approx E_{\network}(x_t, y_t, z_t, t, t') = \left({x_t, y_t, z_t}\right) + \sum_{n=1}^{k} \delta \cdot \network{}(x_{n \delta + t}, y_{n \delta + t}, z_{n \delta + t}, n \delta + t) \enspace .
    \equationlabel{eulertrajectory}
\end{equation}

This formulation makes \ourmethod{} highly flexible, enabling optimization practically to improve $\network$'s estimate of $F$ with objectives that take either an Eulerian view (directly on $\network$ via \equationref{eulerflownetworkequiv}) or a Lagrangian view (on point rollouts for arbitrary start and end ranges via \equationref{eulertrajectory}).

\section{Experiments}\sectionlabel{experiments}

In order to validate \ourmethod{}'s construction and better understand the impact of its design choices, we perform extensive experiments on the Argoverse~2 \citep{argoverse2} and Waymo Open~\citep{waymoopen} autonomous vehicle datasets. We compare against open source implementations of FastNSF~\citep{fastnsf}, \citeauthor{liu2024selfsupervisedmultiframeneuralscene}, NSFP~\citep{nsfp}, FastFlow3D~\citep{scalablesceneflow}, and variants of ZeroFlow introduced in \chapterref{zeroflow}~\citep{vedder2024zeroflow} provided by the ZeroFlow model zoo\footnote{\url{https://github.com/kylevedder/SceneFlowZoo}, from \citet{vedder2024zeroflow}.}, a third-party implementation of NTP~\citep{ntp} from \citeauthor{vidanapathirana2023mbnsf}, and Argoverse 2 2024 Scene Flow Challenge leaderboard submission results from the authors of Flow4D~\citep{flow4d}, TrackFlow~\citep{khatri2024trackflow}, DeFlow++/DeFlow~\citep{zhang2024deflow}, ICP Flow~\citep{lin2024icp}, and SeFlow~\citep{seflow}. As discussed in \citeauthor{khatri2024trackflow} and used in the Argoverse~2 2024 Scene Flow Challenge, methods are ranked by their speed normalized \emph{mean Dynamic Normalized EPE} as introduced in \chapterref{trackflow}, \sectionref{eval}. 

\poorparagraph{Implementation Details} To showcase the flexibility of \ourmethod{} without hyperparameter tuning, for all quantitative experiments we run with a neural prior of depth 8 (NSFP's default depth), except for our submission to the Argoverse 2 2024 Scene Flow Challenge (\sectionref{performance}) where, based on our depth ablation study on the val split (\sectionref{mlpdepth}), we set the depth of the neural prior to 18. As discussed in NTP's original paper \citep{ntp} and confirmed by our experiments, NTP struggles to converge beyond 25 frames, so we only compare against it in a 20 frame settings. As is typical in the scene flow literature~\citep{chodosh2023}, we perform ego compensation and ground point removal on both Argoverse~2 and Waymo Open using the dataset provided map and ego pose.

\subsection{How does \ourmethod{} compare to prior art on real data?}\sectionlabel{performance}
\ourmethod{} achieves \textbf{state-of-the-art} performance on the \emph{Argoverse 2 2024 Scene Flow Challenge} leaderboard.
Despite being unsupervised, \ourmethod{} \textbf{surpasses \emph{all} prior art, supervised or unsupervised}, including Flow4D~\citep{flow4d}\footnote{Flow4D is the winner of the 2024 Argoverse 2 Scene Flow Challenge supervised track.} and ICP~Flow~\citep{lin2024icp}\footnote{ICP~Flow is the winner of the 2024 Argoverse 2 Scene Flow Challenge unsupervised track.}. Notably, \ourmethod{} achieves $< 2.5\times$ lower error mean Dynamic EPE than ICP~Flow and beats Flow4D by over 10\%.

\begin{figure}[htb]
    \centering
    \includegraphics{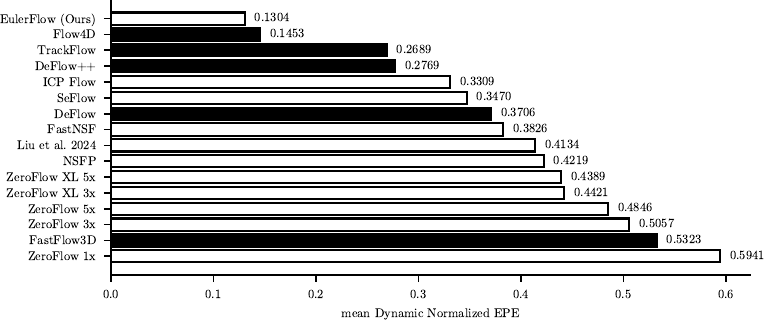}
    \caption{Mean Dynamic Normalized EPE of \ourmethod{} compared to prior art on the Argoverse~2 2024 Scene Flow Challenge test set. \ourmethod{} is state-of-the-art, beating all supervised (shown in black) and unsupervised (shown in white) methods. Lower is better.}
    \figlabel{argodynamicepe}
\end{figure}

\begin{figure}[htb]
    \centering
    \includegraphics{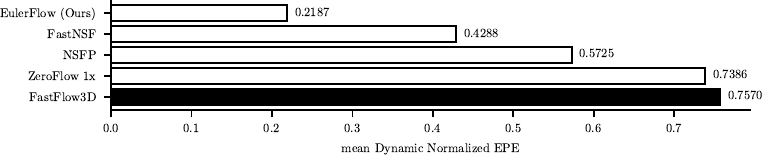}
    \caption{Mean Dynamic Normalized EPE of \ourmethod{} compared to prior art on the Waymo Open validation set. \ourmethod{} is state-of-the-art, beating all supervised (shown in black) and unsupervised (shown in white) methods. Lower is better.}
    \figlabel{waymodynamicepe}
\end{figure}

\ourmethod{}'s dominant performance also holds on Waymo Open~\citep{waymoopen}; we compare against several popular methods (\figref{waymodynamicepe}), and \ourmethod{} again out-performs the baselines by a wide margin, more than halving the error over the next best method. 

\subsection{What contributes to \ourmethod{}'s state-of-the-art performance?}

\begin{figure}[htb]
\centering
\begin{subfigure}[b]{0.49\textwidth}
    \centering
    \includegraphics{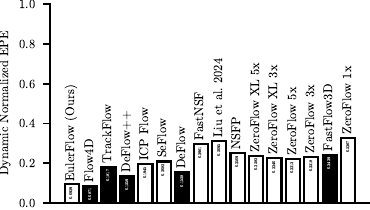}
    \caption{\texttt{CAR}}
    \figlabel{fig:car}
\end{subfigure}%
\begin{subfigure}[b]{0.49\textwidth}
    \centering
    \includegraphics{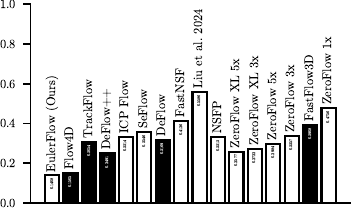}
    \caption{\texttt{OTHER VEHICLES}}
    \figlabel{fig:other-vehicles}
\end{subfigure}
\begin{subfigure}[b]{0.49\textwidth}
    \centering
    \includegraphics{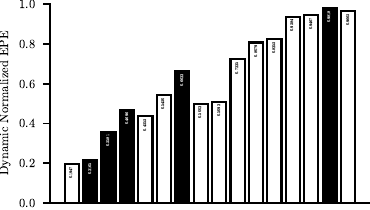}
    \caption{\texttt{PEDESTRIAN}}
    \figlabel{fig:pedestrian}
\end{subfigure}%
\begin{subfigure}[b]{0.49\textwidth}
    \centering
    \includegraphics{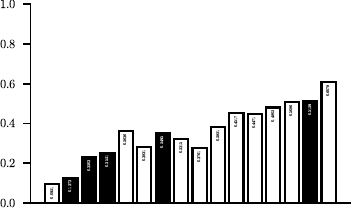}
    \caption{\texttt{WHEELED VRU}}
    \figlabel{fig:wheeled-vru}
\end{subfigure}
\caption{Per class Dynamic Normalized EPE of \ourmethod{} compared to prior art on the Argoverse~2 2024 Scene Flow Challenge test set. Supervised methods shown in black, unsupervised methods shown in white. Methods are ordered left to right by increasing mean Dynamic Normalized EPE. Lower is better. }
\figlabel{argometacatagorydynamic}
\end{figure}

\begin{figure}[h!]
\centering
\begin{subfigure}[b]{0.32\textwidth}
    \centering
    \includegraphics{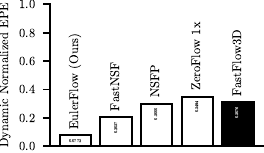}
    \caption{\texttt{VEHICLE}}
    \figlabel{fig:waymovehicle}
\end{subfigure}%
\begin{subfigure}[b]{0.32\textwidth}
    \centering
    \includegraphics{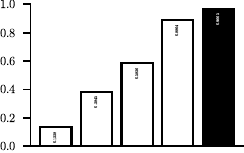}
    \caption{\texttt{CYCLIST}}
    \figlabel{fig:waynocyclist}
\end{subfigure}%
\begin{subfigure}[b]{0.32\textwidth}
    \centering
    \includegraphics{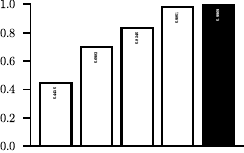}
    \caption{\texttt{PEDESTRIAN}}
    \figlabel{fig:waymopedestrian}
\end{subfigure}
\caption{Per class Dynamic Normalized EPE of \ourmethod{} compared to prior art on the Waymo Open validation set. Supervised methods shown in black, unsupervised methods shown in white. Methods are ordered left to right by increasing mean Dynamic Normalized EPE. Lower is better.}
\figlabel{waymometacatagorydynamic}
\end{figure}

We find that \ourmethod{}'s lower mean Dynamic EPE can be attributed to better performance on smaller objects. On Argoverse~2, compared to Flow4D, \ourmethod{}'s improves on \texttt{WHEELED VRU} (\figref{fig:wheeled-vru}), a small, rare, fast moving class. Compared to ICP~Flow, \ourmethod{}'s improves on all classes (at least halving the error on every class!), with the largest improvements coming from the smaller and harder to detect objects \texttt{PEDESTRAIN} and \texttt{WHEELED VRU} (\figrefs{fig:pedestrian}{fig:wheeled-vru}). On Waymo Open, the same story holds; the most dramatic performance improvements come from the small object classes of \texttt{CYCLIST} and \texttt{PEDESTRIAN} (\figref{waymometacatagorydynamic}).

These results are consistent with our qualitative visualizations. \figref{flyingbird} shows \ourmethod{} is able to cleanly extract the motion of a bird flying past the ego vehicle. Euler integration using \ourmethod{}'s ODE, starting at the bird's takeoff position and ending when it loses lidar returns, produces emergent 3D point tracking behavior on the bird through its trajectory (\figref{birdtracking}), further demonstrating the quality of \ourmethod{}'s model of the scene's motion. 

\begin{figure}[htb]
\centering
\begin{subfigure}[b]{0.49\textwidth}
    \centering
    \includegraphics[width=\textwidth]{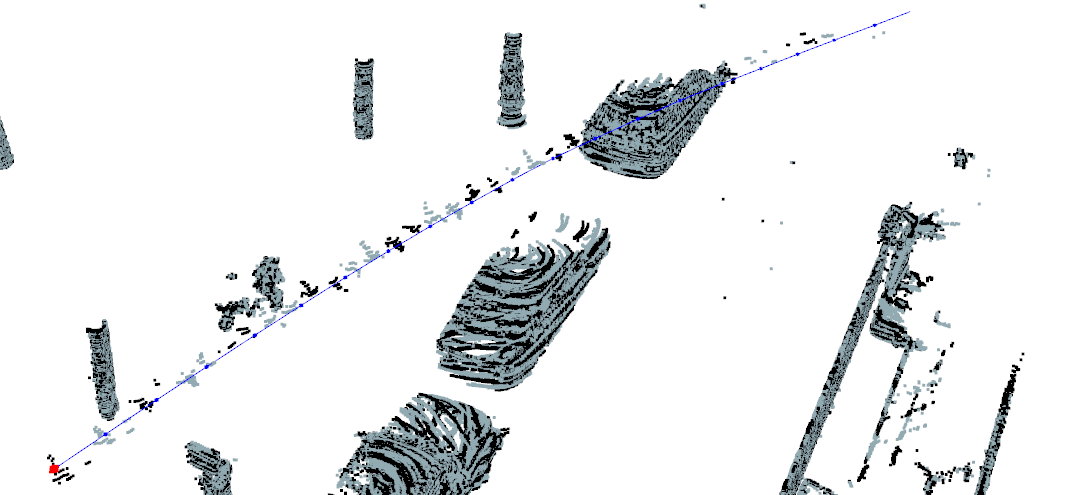}
    \caption{Bird trajectory via Euler integration from takeoff}
    \figlabel{birdtrajectory}
\end{subfigure}%
\begin{subfigure}[b]{0.49\textwidth}
    \centering
    \includegraphics[width=\textwidth]{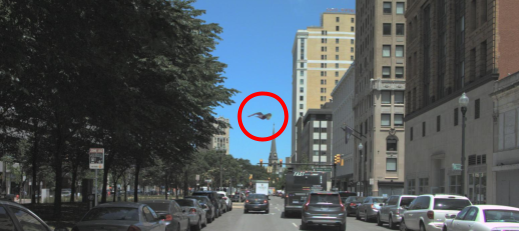}
    \caption{Bird being tracked}
    \figlabel{birdrgb}
\end{subfigure}
    \caption{\ourmethod{} is able to track the bird over 20 frames by performing Euler integration starting from takeoff until it loses all point cloud lidar returns.}
    \figlabel{birdtracking}
\end{figure}



\subsubsection{How does observation sequence length impact \ourmethod{}?}\sectionlabel{sequencelength}

 \begin{figure}[htb]
     \centering
    \includegraphics{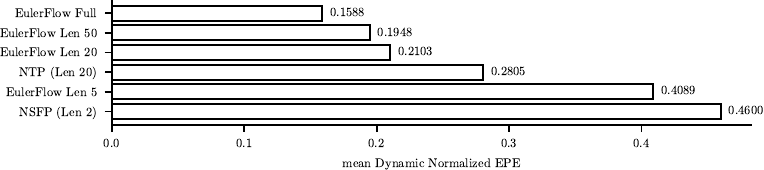}
     \caption{Mean Dynamic Normalized EPE of \ourmethod{} for various sequence lengths on the Argoverse 2 val split, compared against representative baselines. These results demonstrate that \ourmethod{} improves with sequence length; however, at a sequence length of 20, our method significantly outperforms NTP, suggesting that \ourmethod{}'s performance cannot solely be attributed to longer sequence modeling. }
     \figlabel{sequencelength}
 \end{figure}

As we discuss in \sectionref{ourpipeline}, \ourmethod{} benefits from constructive interference from ODE estimation over many observations. Does this sufficiently explain \ourmethod{}'s performance?
\figref{sequencelength} shows the performance of \ourmethod{} at length 5, 20, 50, and full sequence (roughly 160 frames) compared to NSFP and NTP at length 20. \ourmethod{} sees clear continual improvements as the number of frames increases without signs of saturation. However, sequence length alone does not explain \ourmethod{}'s performance; even at the same sequence length of 20, \ourmethod{} demonstrates significantly better performance than NTP.

\subsubsection{How do multi-frame optimization objectives impact \ourmethod{}?}

\begin{figure}[htb]
    \centering
    \includegraphics{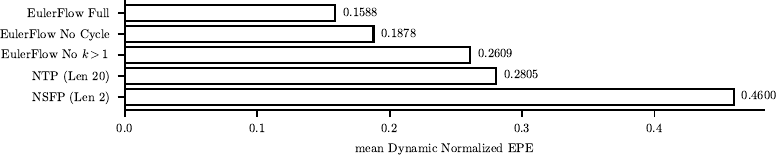}
    \caption{Mean Dynamic Normalized EPE of \ourmethod{} for various losses on the Argoverse 2 val split, compared against representative baselines. These results demonstrate that \ourmethod{}'s multi-observation optimization objectives significantly improve overall performance.}
    \figlabel{lossablations}
\end{figure}

\equationref{gigachadloss} outlines two major components of \ourmethod{}'s loss: multi-frame Euler integration for Chamfer Distance reconstruction, and cycle consistency. \figref{lossablations} compares \ourmethod{} without more than one integration step (No $k>1$) and without cycle consistency rollouts (No Cycle) to better understand the impact of these components. Ablating multi-step Euler integrated rollouts results in significant degredation, as they are a strong forcing function to have consistent, smooth flow volumes; indeed, despite consuming the entire sequence, \ourmethod{} (No $k>1$) is only slightly better than NTP with a sequence length of 20. These results highlight the power of multi-step rollouts and their potential as a objective for other test-time optimization methods and feedforward methods.

\subsubsection{How does the capacity of the neural prior impact \ourmethod{}?}\sectionlabel{mlpdepth}

\citeauthor{nsfp} ablate the capacity of  NSFP's neural prior to characterize underfitting and overfitting to optimization objective noise, ultimately selecting a depth of 8. \ourmethod{}'s neural prior is structured similarly; however, NSFP is fitting a single snapshot in time, while \ourmethod{} is fitting an entire ODE over significant time intervals. Intuitively, one would expect that full sequence modeling would benefit from greater network capacity.

To evaluate this, we perform a sweep of \ourmethod{}'s network depth on the Argoverse 2 validation split (\figref{depthablation}). While \ourmethod{} with NSFP's default of depth 8 performs well on our Argoverse 2 evaluations (0.1\% worse than the supervised state-of-the-art Flow4D), we see that performance improves as the neural prior's depth increases until depth 18 (indicating underfitting), where we start to see degradation (indicating overfitting to noise). Based on these results our Argoverse 2 2024 Scene Flow Challenge leaderboard submission uses a depth 18 neural prior (\figref{argodynamicepe}).

\begin{figure}[htb]
    \centering
    \includegraphics{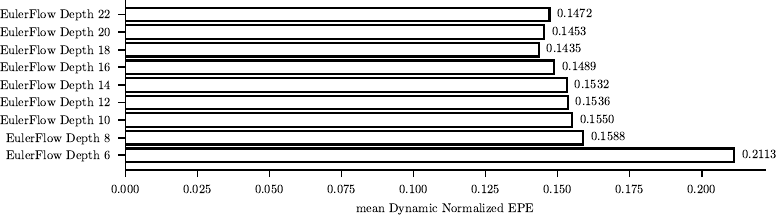}
    \caption{Mean Dynamic Normalized EPE of \ourmethod{} on the Argoverse 2 val split for different neural prior capacities. Shallow networks underfit the ODE, while deeper networks overfit to noise in the optimization objectives.}
    \figlabel{depthablation}
\end{figure}

\begin{figure}[htb]
\centering
\begin{subfigure}[b]{0.32\textwidth}
    \centering
    \includegraphics[width=\textwidth]{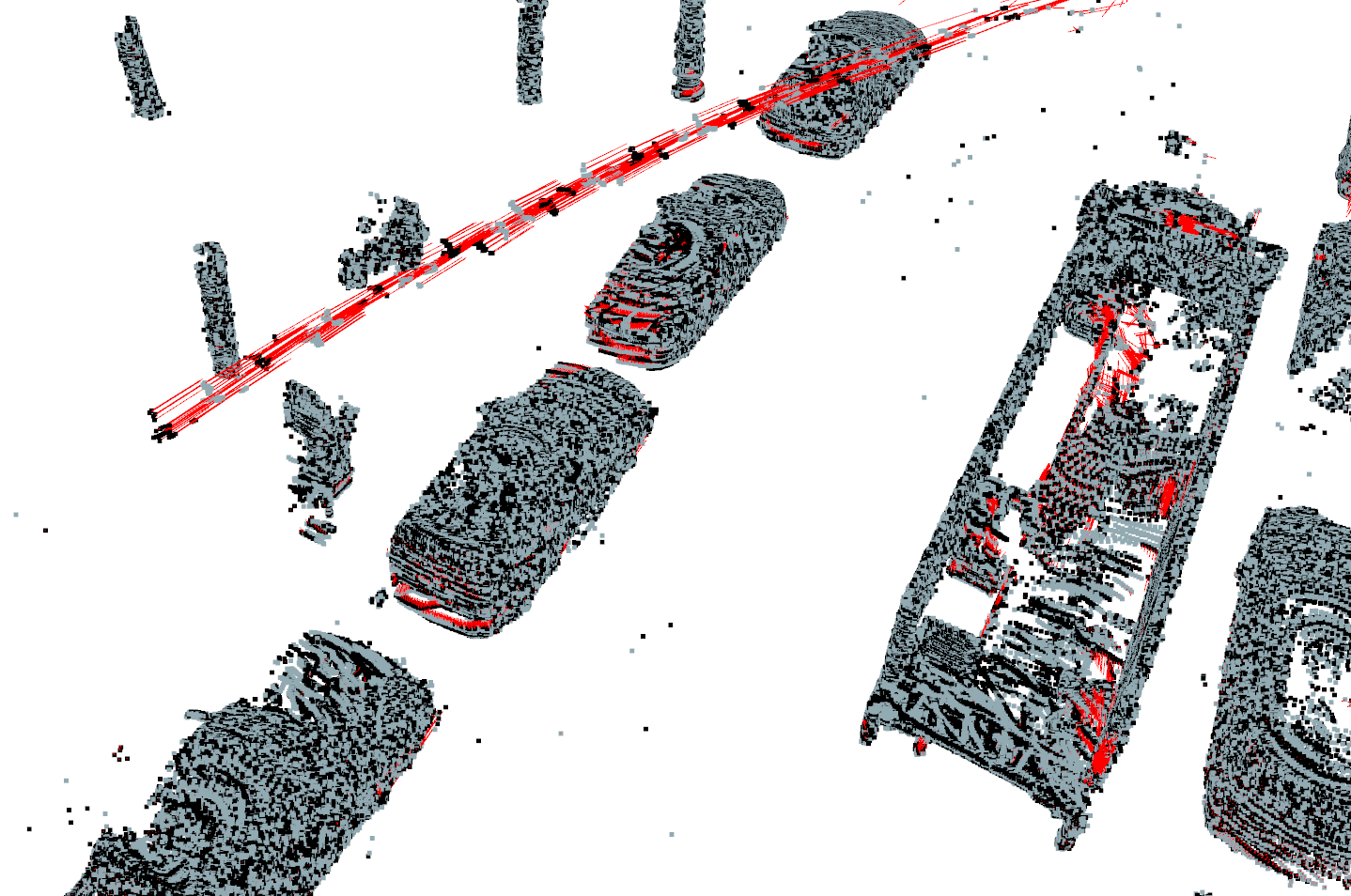}
    \caption{\ourmethod{} (Ours)}
    \figlabel{flyingbirdgigachad}
\end{subfigure}%
\begin{subfigure}[b]{0.32\textwidth}
    \centering
    \includegraphics[width=\textwidth]{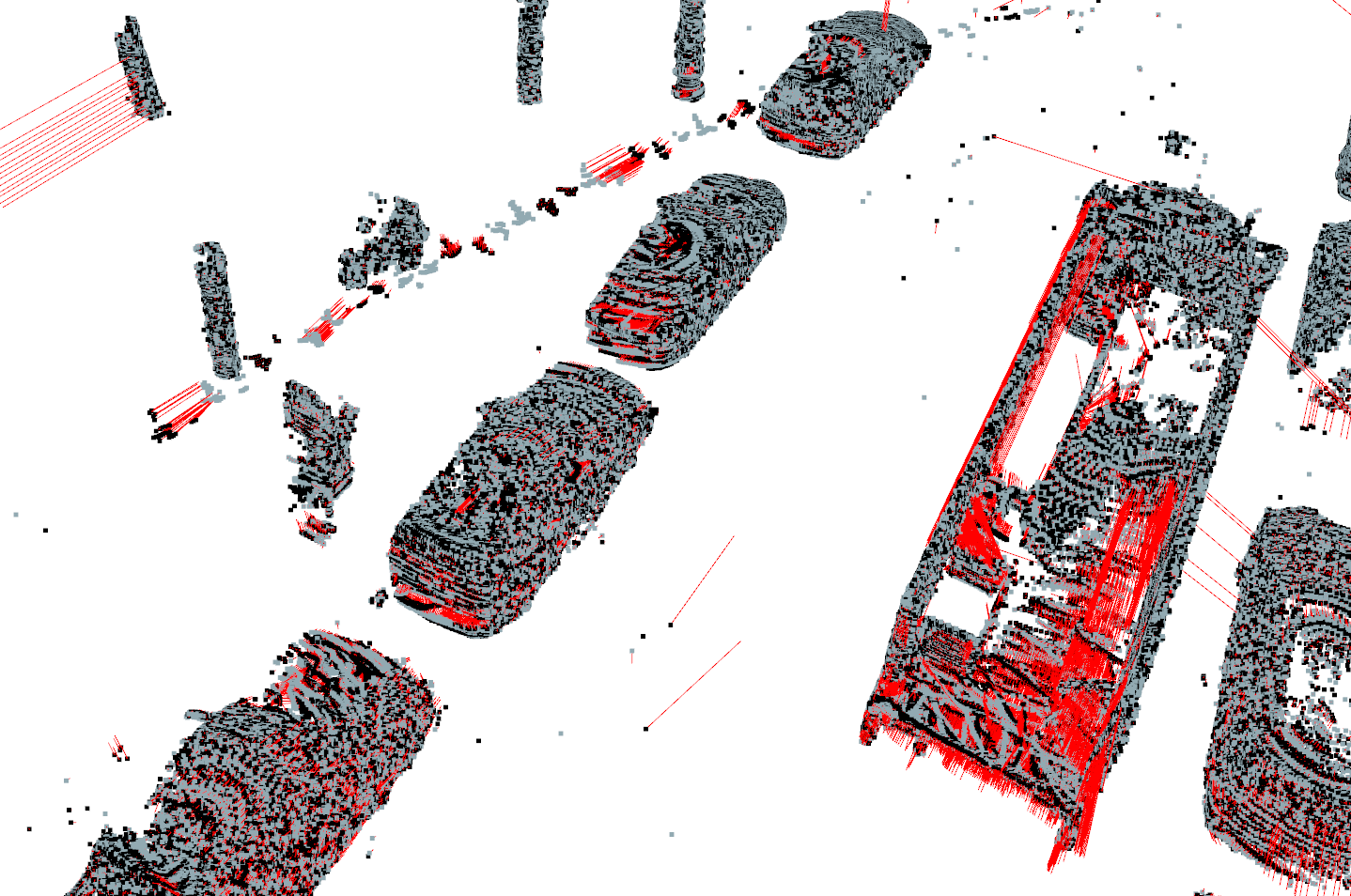}
    \caption{Fast NSF}
    \figlabel{flyingbirdfastnsf}
\end{subfigure}%
\begin{subfigure}[b]{0.32\textwidth}
    \centering
    \includegraphics[width=\textwidth]{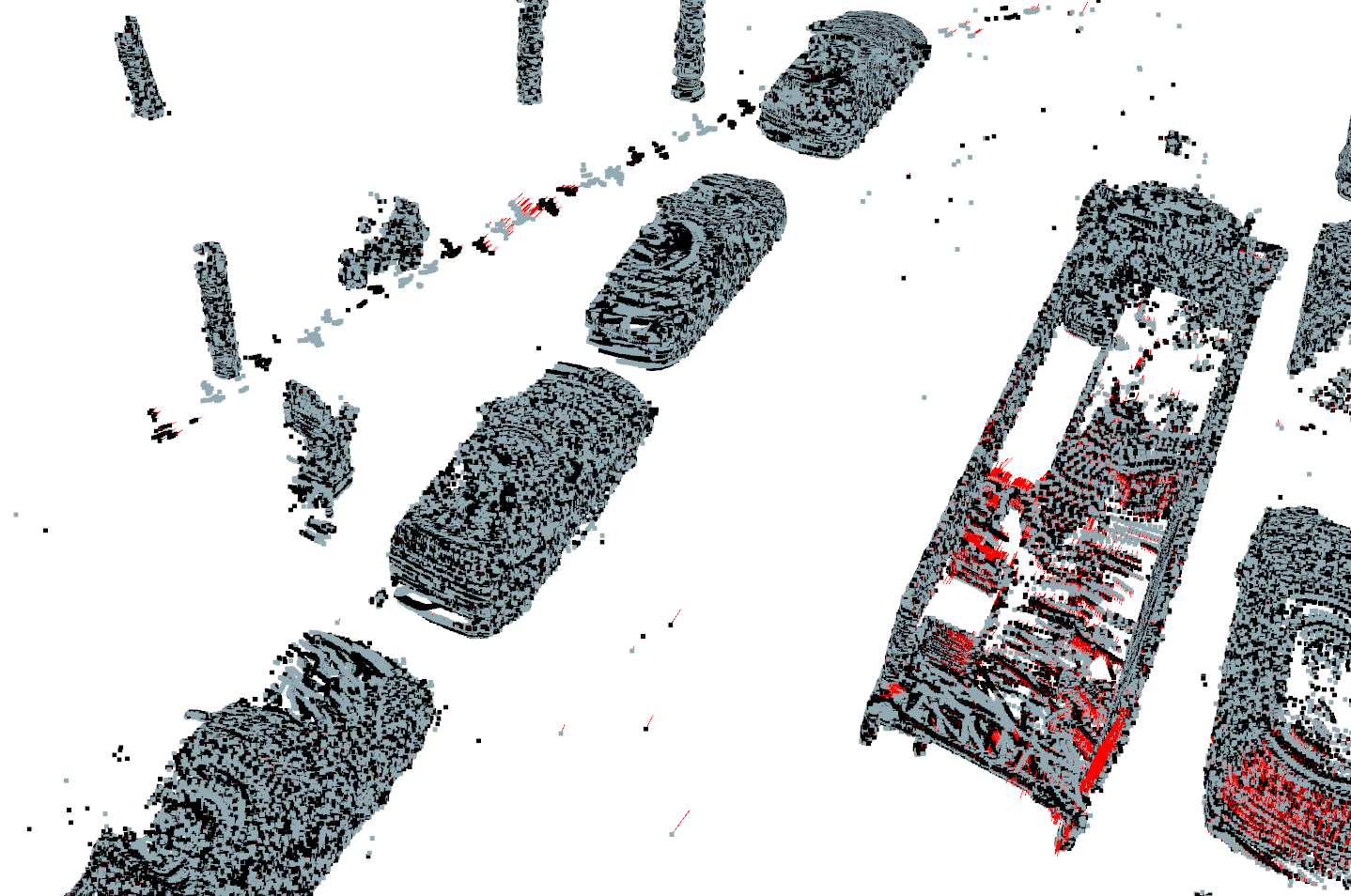}
    \caption{ZeroFlow 5x}
    \figlabel{flyingbirdzeroflow}
\end{subfigure}
\begin{subfigure}[b]{0.32\textwidth}
    \centering
    \includegraphics[width=\textwidth]{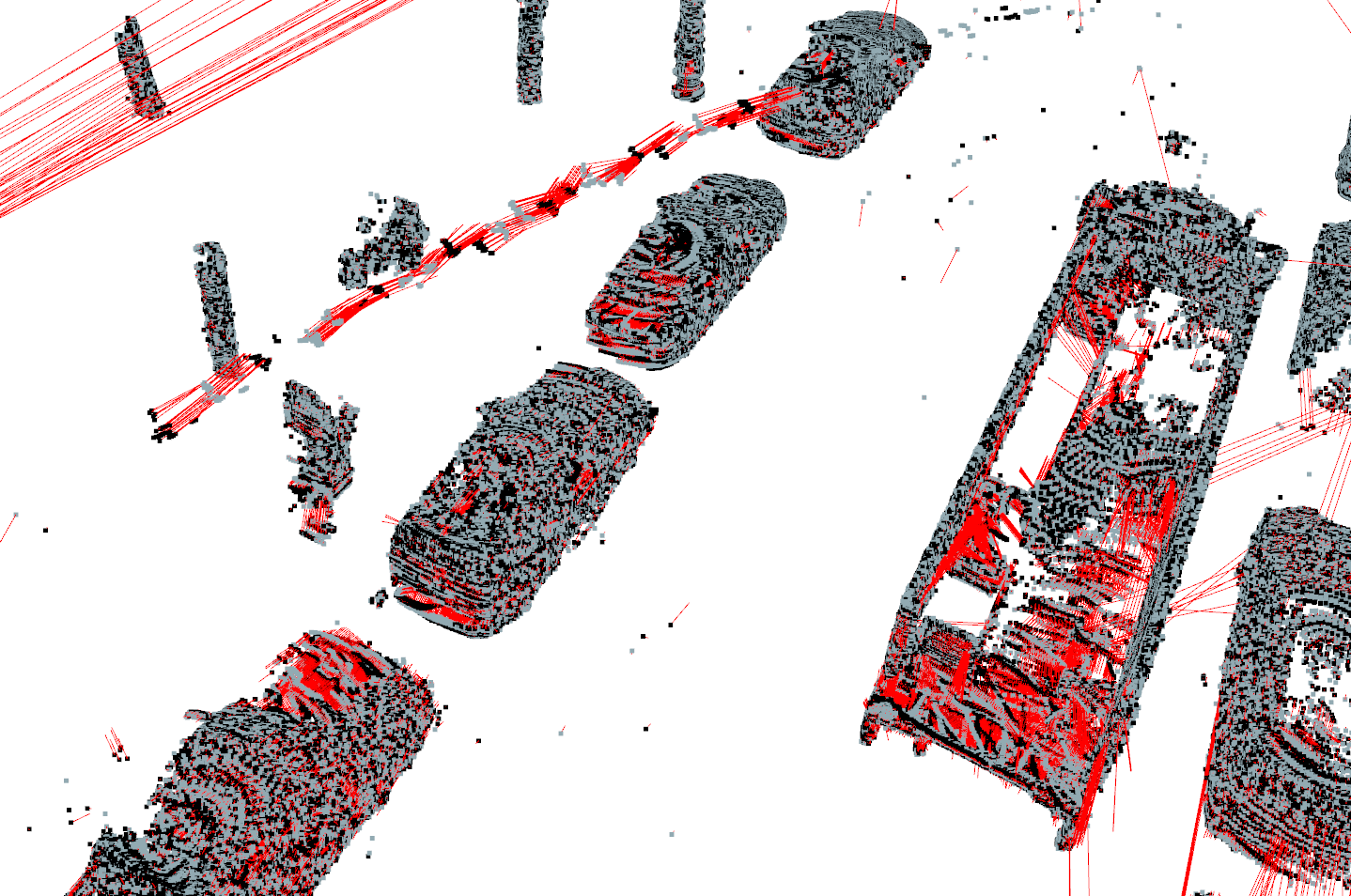}
    \caption{\citeauthor{liu2024selfsupervisedmultiframeneuralscene}}
    \figlabel{flyingbirdliuetal}
\end{subfigure}%
\begin{subfigure}[b]{0.32\textwidth}
    \centering
    \includegraphics[width=\textwidth]{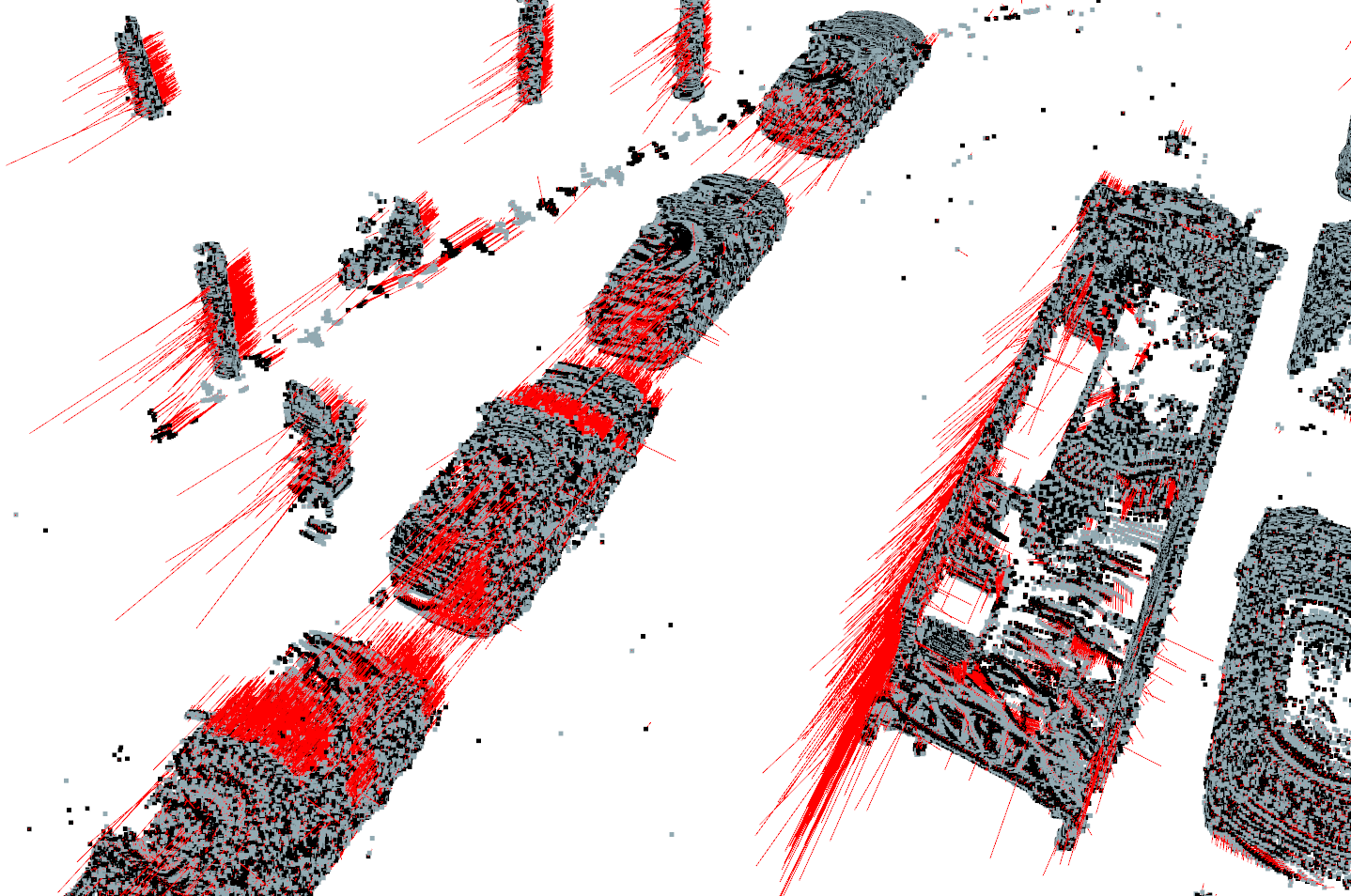}
    \caption{NSFP}
    \figlabel{flyingbirdnsfp}
\end{subfigure}%
\begin{subfigure}[b]{0.32\textwidth}
    \centering
    \includegraphics[width=\textwidth]{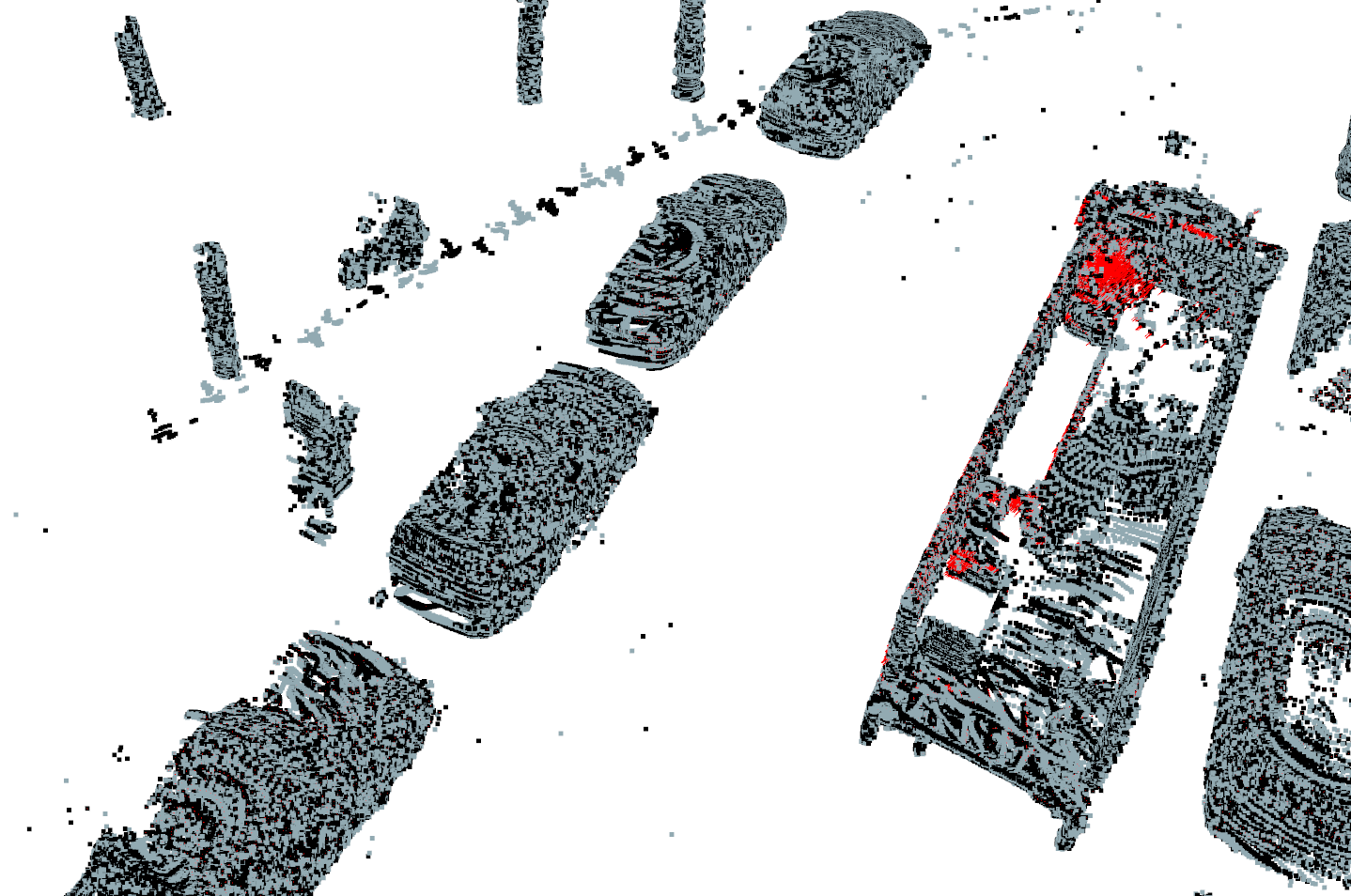}
    \caption{Ground Truth}
    \figlabel{flyingbirdgt}
\end{subfigure}
\caption{Visualization of \ourmethod{} compared to prior art for the same scene as \figref{teasersceneflow} and \figref{birdtrajectory}. \ourmethod{} is able to extract the bird's trajectory; however, all other methods except \citeauthor{liu2024selfsupervisedmultiframeneuralscene} fail to recognize this motion, and \citeauthor{liu2024selfsupervisedmultiframeneuralscene}'s flow is marred by severe scene artifacts. The bird is outside the labeled object taxonomy, and so its motion is unlabeled in the ground truth (\figref{flyingbirdgt}).}
\figlabel{flyingbird}
\end{figure}

\subsection{How does the choice of learnable function class and design of encodings impact \ourmethod{}?}

\ourmethod{} at its core is an optimization loop over a simple, feedforward ReLU-based multi-layer perception inherited from Neural Scene Flow Prior~\citep{nsfp}. How does this choice of learnable function class impact the performance of \ourmethod{}? To better understand these design choices we examine the choice of non-linearity and time feature encoding. 

\begin{figure}[htb]
    \centering
    \includegraphics{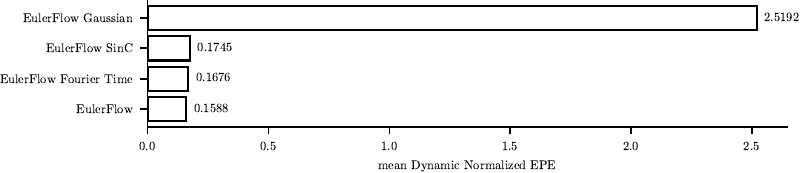}
    \caption{Mean Dynamic Normalized EPE of \ourmethod{} on the Argoverse 2 val split for less-smooth configurations of its learnable function class. These results indicate that the smoothness of the ReLU non-linearity proposed by \citeauthor{nsfp} transfers well to \ourmethod{}.
    }
    \figlabel{nonlinearitytime}
\end{figure}

One of \citeauthor{nsfp}'s core theoretical contributions demonstrates that NSFP's ReLU MLP is a good prior for scene flow because it represents a smooth learnable function class, and scene flow is often locally smooth with respect to input position. However, unlike NSFP, \ourmethod{} is fitting flow over a full ODE; while it seems reasonable to assume that this ODE is typically also locally smooth, cases like adjacent cars moving rapidly in opposite directions may benefit from the ability to model higher frequency, less locally smooth functions. To test this hypothesis, we ablate \ourmethod{} by replacing its normalized time with higher frequency sinusoidal time embeddings (mirroring \citeauthor{ntp}'s proposed time embedding for NTP), as well as try other popular non-linearities like SinC~\citep{ramasinghe2024on} and Gaussian~\citep{garf} from the coordinate network literature. \figref{nonlinearitytime} features negative results on these ablations across the board; Gaussians were unable to converge due the extremely high frequency representation triggering early stopping, while the use of SinC and higher frequency time embeddings both resulted in worse overall performance, indicating that \citeauthor{nsfp}'s smooth function prior does indeed seem appropriate for \ourmethod{}'s neural prior.

\subsection{Beyond Autonomous Vehicles}

Due to a dearth of real-world, labeled scene flow data, prior scene flow work on real data overwhelmingly evaluates on autonomous vehicle datasets~\citep{dewan2016rigid, nsfp, scalablesceneflow, fastnsf, chodosh2023,   liu2024selfsupervisedmultiframeneuralscene, vedder2024zeroflow, khatri2024trackflow}; consequently, motion understanding in other important domains like tabletop manipulation has been neglected. To showcase \ourmethod{}'s out-of-the-box flexibility and generalizability, we visualize \ourmethod{} on several dynamic tabletop scenes we collected using the ORBBEC Astra, a low cost depth camera commonly used in robotics (\figref{orbbec}). For viewing ease, we paint our point clouds with color; however, RGB information is not provided to \ourmethod{} during optimization. Furthermore, while \ourmethod{} only reasons about point clouds, we show that it can leverage video monocular depth estimators to describe RGB-only scene flow in \appendixref{monodepth}. Interactive visuals are available at \websitelink{}.

\begin{figure}[htb]
    \centering

    \begin{subfigure}[b]{0.24\textwidth}
    \centering
    \includegraphics[width=\textwidth]{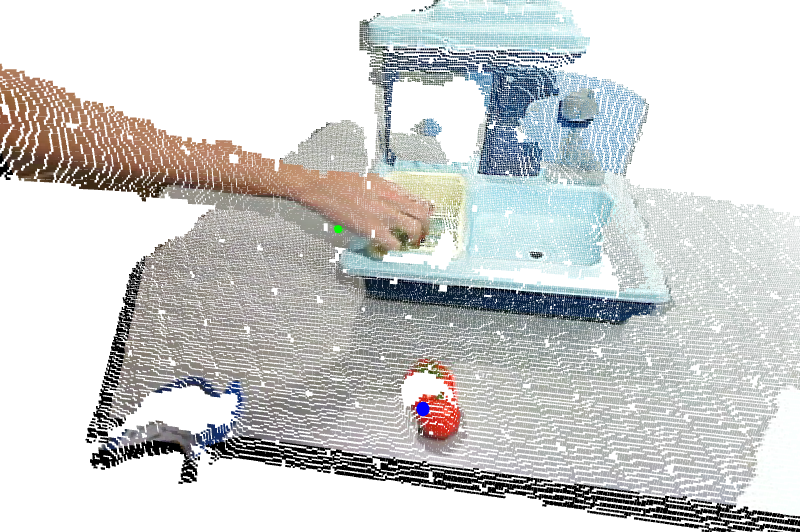}
\end{subfigure}%
\begin{subfigure}[b]{0.24\textwidth}
    \centering
    \includegraphics[width=\textwidth]{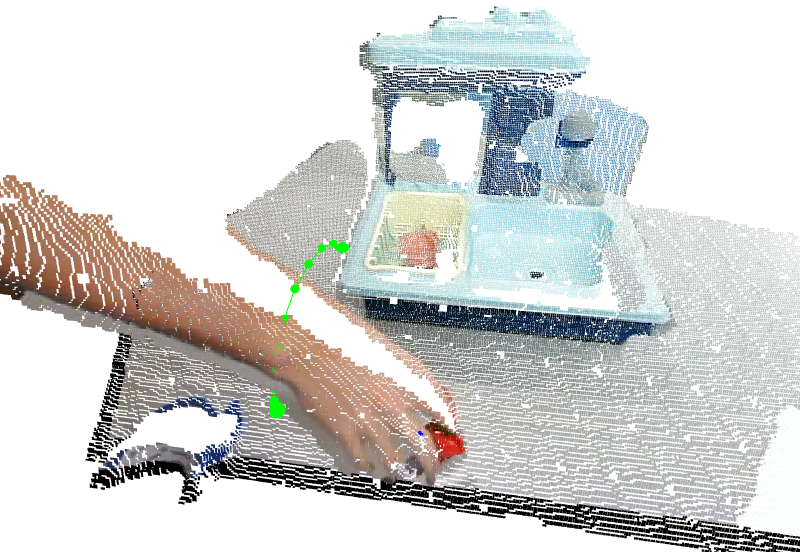}
\end{subfigure}%
\begin{subfigure}[b]{0.24\textwidth}
    \centering
    \includegraphics[width=\textwidth]{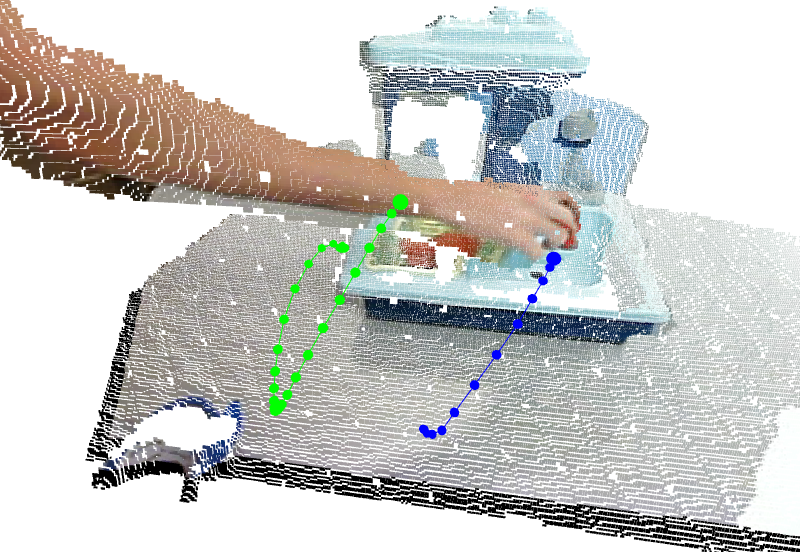}
\end{subfigure}%
\begin{subfigure}[b]{0.24\textwidth}
    \centering
    \includegraphics[width=\textwidth]{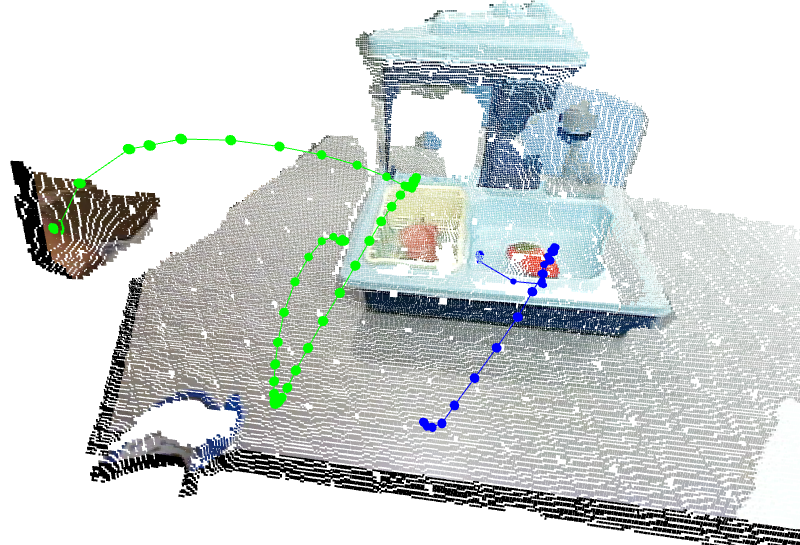}
\end{subfigure}

    \begin{subfigure}[b]{0.24\textwidth}
    \centering
    \includegraphics[width=\textwidth]{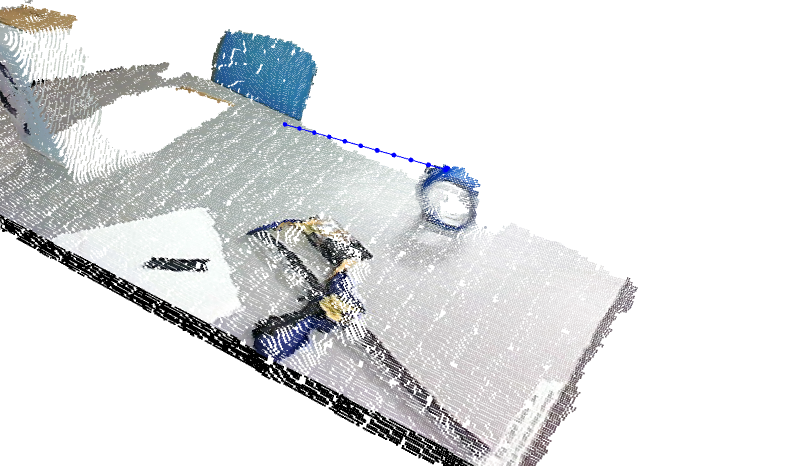}
\end{subfigure}%
\begin{subfigure}[b]{0.24\textwidth}
    \centering
    \includegraphics[width=\textwidth]{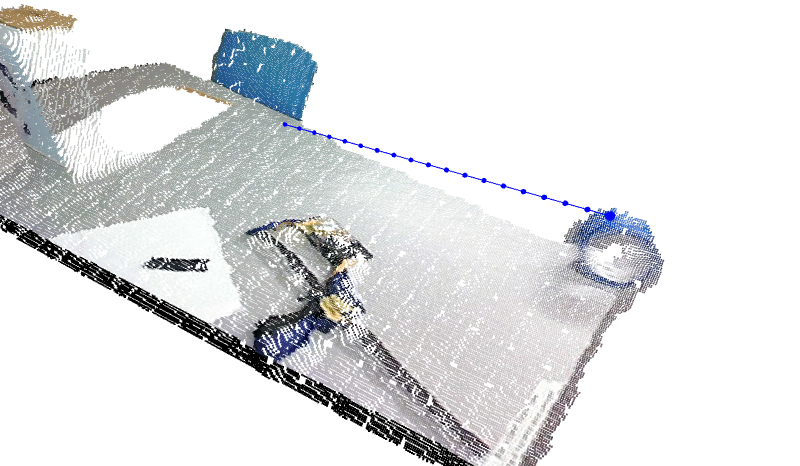}
\end{subfigure}%
\begin{subfigure}[b]{0.24\textwidth}
    \centering
    \includegraphics[width=\textwidth]{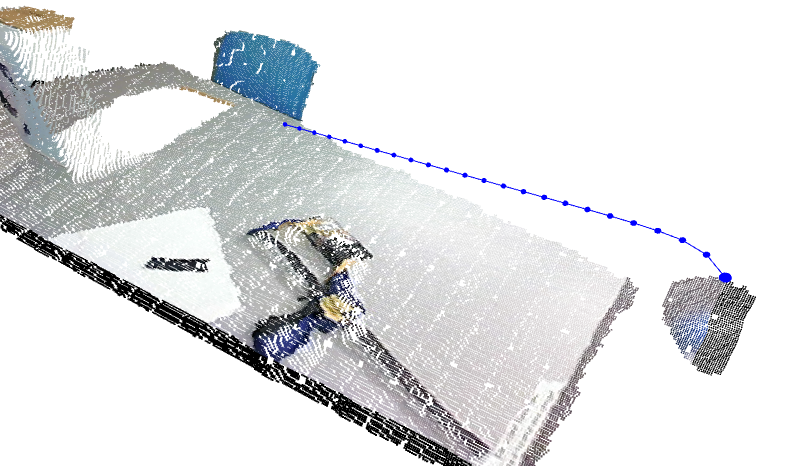}
\end{subfigure}%
\begin{subfigure}[b]{0.24\textwidth}
    \centering
    \includegraphics[width=\textwidth]{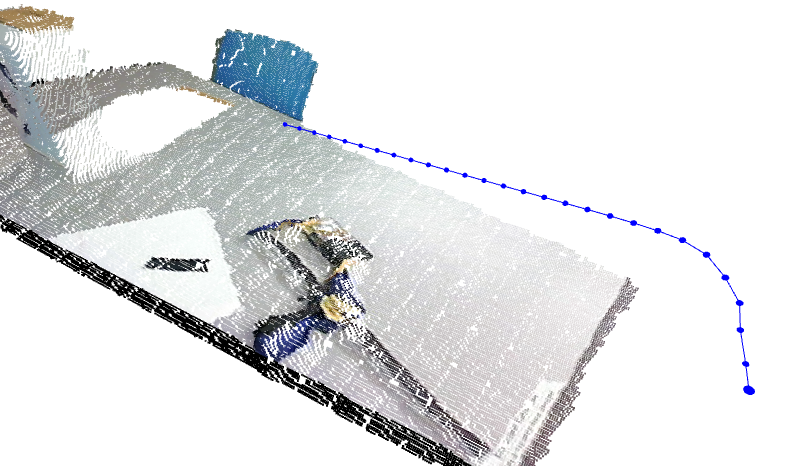}
\end{subfigure}

    \begin{subfigure}[b]{0.24\textwidth}
    \centering
    \includegraphics[width=\textwidth]{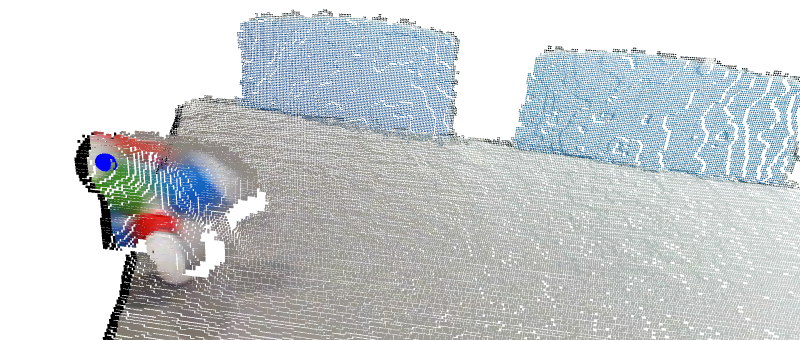}
\end{subfigure}%
\begin{subfigure}[b]{0.24\textwidth}
    \centering
    \includegraphics[width=\textwidth]{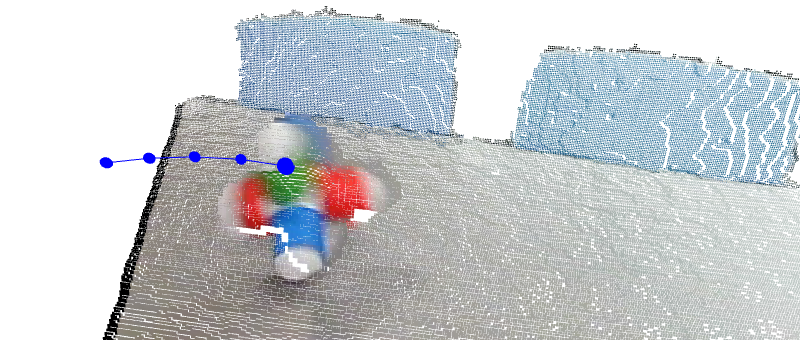}
\end{subfigure}%
\begin{subfigure}[b]{0.24\textwidth}
    \centering
    \includegraphics[width=\textwidth]{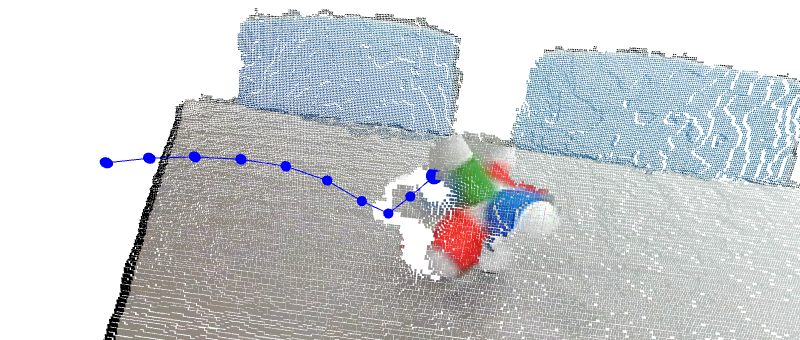}
\end{subfigure}%
\begin{subfigure}[b]{0.24\textwidth}
    \centering
    \includegraphics[width=\textwidth]{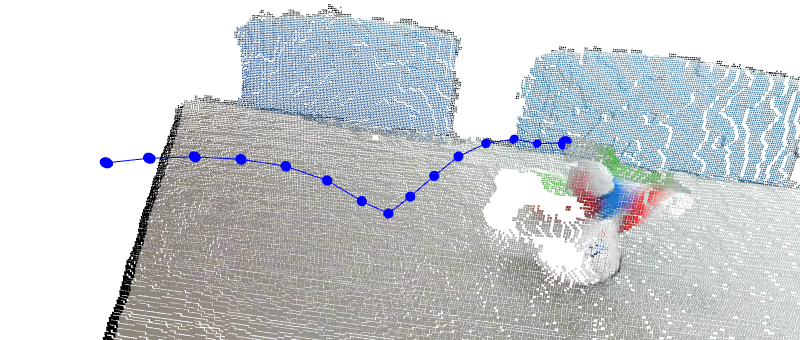}
\end{subfigure}

    \begin{subfigure}[b]{0.24\textwidth}
    \centering
    \includegraphics[width=\textwidth]{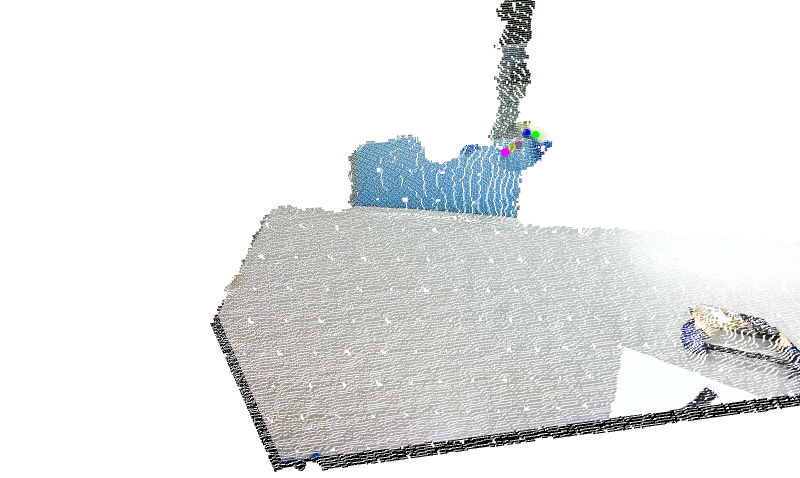}
\end{subfigure}%
\begin{subfigure}[b]{0.24\textwidth}
    \centering
    \includegraphics[width=\textwidth]{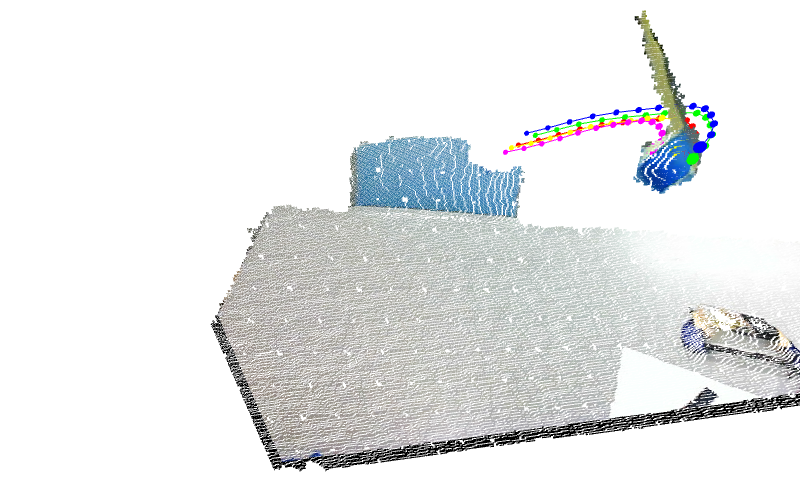}
\end{subfigure}%
\begin{subfigure}[b]{0.24\textwidth}
    \centering
    \includegraphics[width=\textwidth]{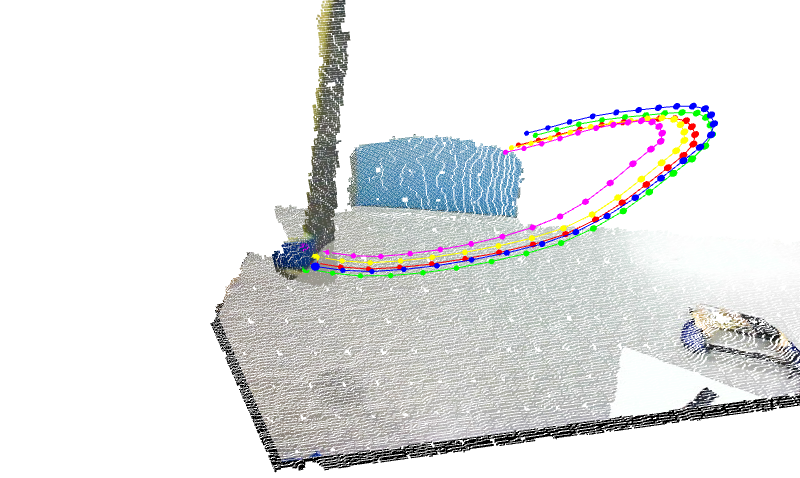}
\end{subfigure}%
\begin{subfigure}[b]{0.24\textwidth}
    \centering
    \includegraphics[width=\textwidth]{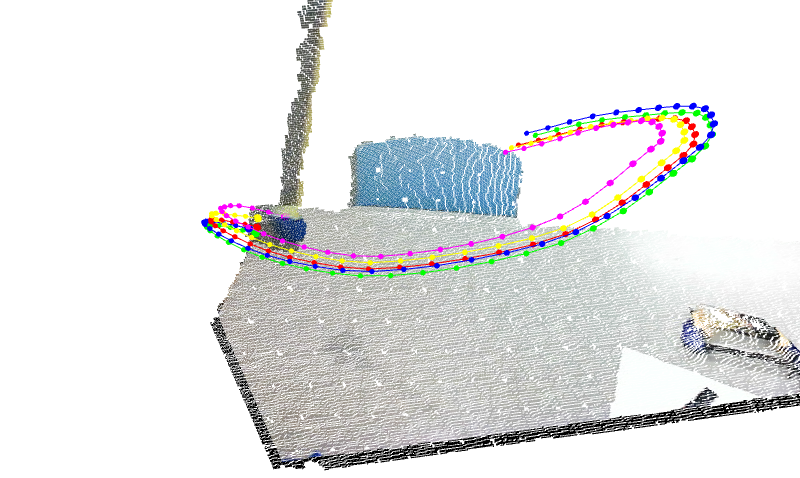}
\end{subfigure}

    \caption{Visualizations of \ourmethod{}'s emergent 3D point tracking behavior that demonstrate the quality of its ODE estimate. Row 1 depicts tracking a tomato placed in the sink by a human hand; note the point does not move despite the hand grasping the tomato. Row 2 depicts tracking of painters tape rolling off a table; \ourmethod{} is able to estimate its trajectory even after it disappears out of frame. Row 3 depicts tracking of the motion of a jack commonly used in tabletop manipulation experiments~\citep{venkatesh2023samplingbased}. Row 4 depicts tracking of a tennis ball taped to a flexible rod. All tracks are produced by Euler integration through the estimated ODE from the initial conditions shown in the left column. Note that point clouds are shown in color for visualization purposes only.}
    \figlabel{orbbec}
\end{figure}

\subsection{\ourmethod{} with Monocular Depth Estimates}\appendixlabel{monodepth}

While \ourmethod{} only consumes point clouds, we can leverage RGB-based video monocular depth estimators to fit scene flow. In \figref{monodepth}, we use DepthCrafter~\citep{hu2024DepthCrafter} to generate a point cloud from the raw RGB of the tabletop video from \figref{orbbec}, Row 4.

\begin{figure}[htb]
    \centering


    \begin{subfigure}[b]{0.24\textwidth}
    \centering
    \includegraphics[width=\textwidth]{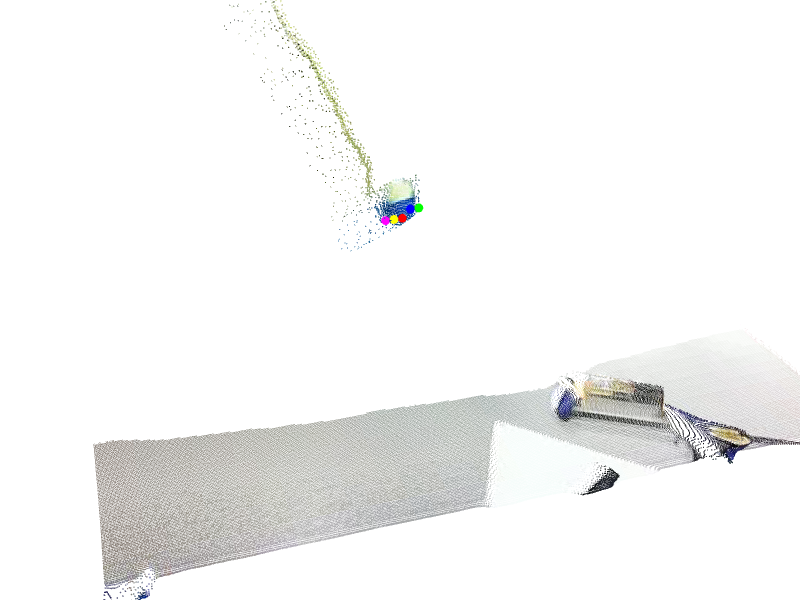}
\end{subfigure}%
\begin{subfigure}[b]{0.24\textwidth}
    \centering
    \includegraphics[width=\textwidth]{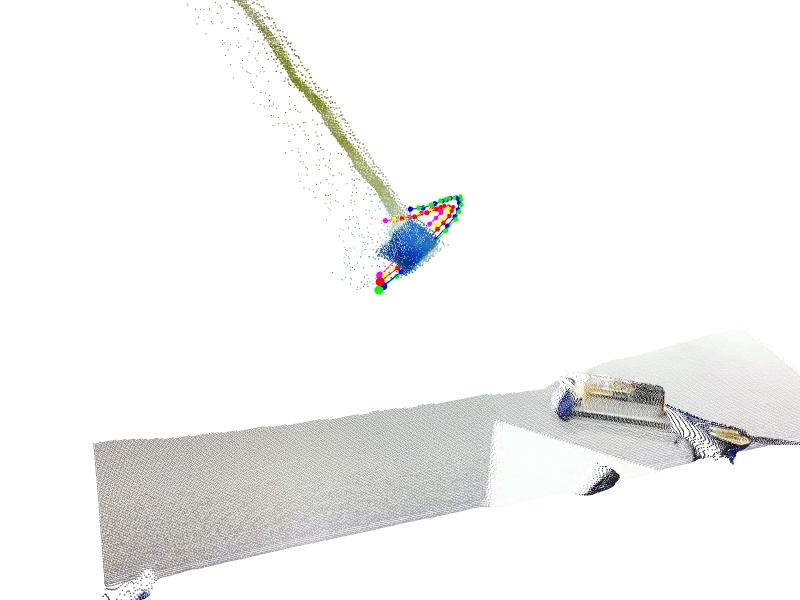}
\end{subfigure}%
\begin{subfigure}[b]{0.24\textwidth}
    \centering
    \includegraphics[width=\textwidth]{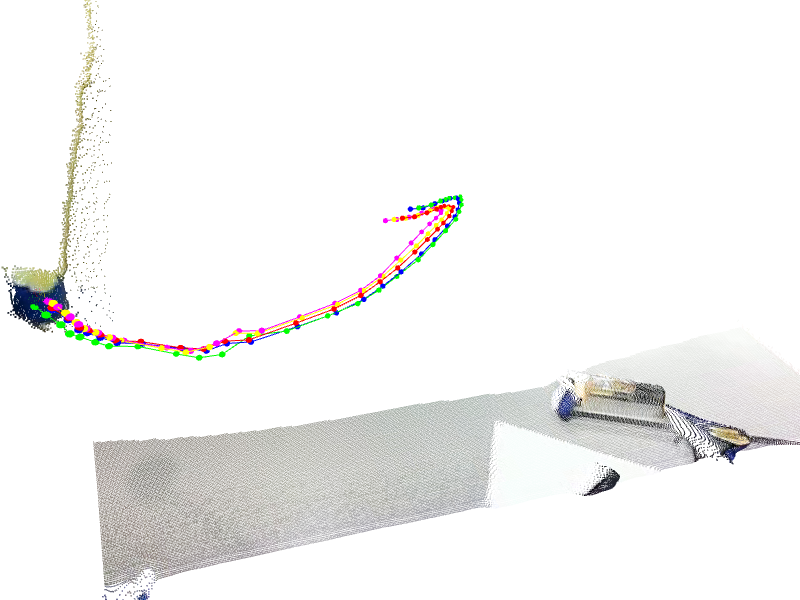}
\end{subfigure}%
\begin{subfigure}[b]{0.24\textwidth}
    \centering
    \includegraphics[width=\textwidth]{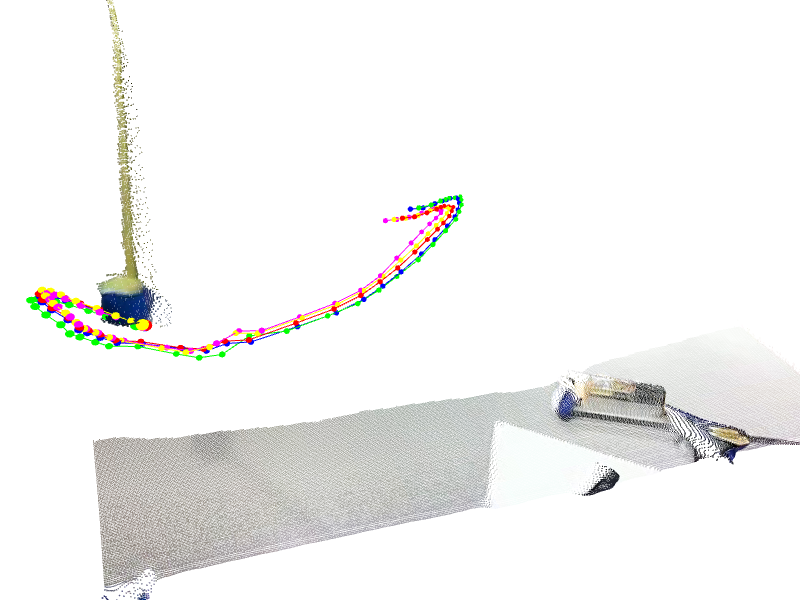}
\end{subfigure}

    \caption{Visualizations of \ourmethod{}'s emergent 3D point tracking behavior on monocular depth estimates from DepthCrafter~\citep{hu2024DepthCrafter}. Interactive visualizations available at \websitelink{}.}
    \figlabel{monodepth}
\end{figure}


\section{Conclusion}

By reframing scene flow as fitting a ODE over positions for a full sequence of observations, we are able to construct \ourmethod{}, a simple unsupervised scene flow method that achieves state-of-the-art performance on the Argoverse 2 2024 Scene Flow Challenge and Waymo Scene Flow benchmark, where it beats all prior art, supervised or unsupervised. \ourmethod{} is able to describe motion on small, fast moving, out of distribution objects unable to be captured by prior art, suggesting that it makes good on the promises of scene flow as a powerful primitive for understanding the dynamic world. It also exhibits other emergent capabilities, like basic 3D point tracking behavior.

We believe that this ODE formulation has implications for scene flow at large, including beyond test-time optimization methods; the power of multi-step Euler integration may translate to feedforward network training. Future work should explore feedforward models that perform autoregressive rollouts or directly learn to estimate multiple steps into the future.

\subsection{Limitations and Future Work}\sectionlabel{limitations}

\ourmethod{}'s strong performance opens the book on an exciting new line of work; however, we feel that it is important to be candid about \ourmethod{}'s current limitations in order to make future progress.

\emph{\ourmethod{} is point cloud only.} Point cloud sparsity bottlenecks performance; for instance, in \figref{birdtracking} and \figref{flyingbird} we were only able to track the bird for 20 frames because we lost lidar observations of the bird, while it remained visible in the car's RGB cameras. Future works should explore multi-modal fusion for better long-term motion descriptions.

\emph{\ourmethod{} is expensive to optimize.} With our implementation, optimizing \ourmethod{} for a single Argoverse 2 sequence takes 24 hours on one NVIDIA V100 16GB GPU, putting it on par with the original NeRF paper's computation expense~\citep{nerf}. However, like with NeRF, we believe  algorithmic, optimization, and engineering improvements can significantly reduce runtime.

\emph{\ourmethod{} does not understand ray casting geometry.} During ego-motion, a static foreground occluding object casts a moving shadow on the background. As discussed by \cite{nsfp}, this causes Chamfer Distance to estimate this as a leading edge of moving structure, encouraging false motion artifacts (\figref{failures}). This can be addressed with optimization losses that model point clouds as originating from a time of flight sensor with limited visibility, as has been successfully demonstrated in the reconstruction~\citep{chodosh2024simultaneousmapobjectreconstruction} and forecasting literature~\citep{khurana2023point, agro2024uno}, rather than an unstructured set of points to be associated via local point distance.

\begin{figure}[htb]
    \centering

    \includegraphics[width=\linewidth, clip, trim={1.5cm 11cm 4cm 5cm}]{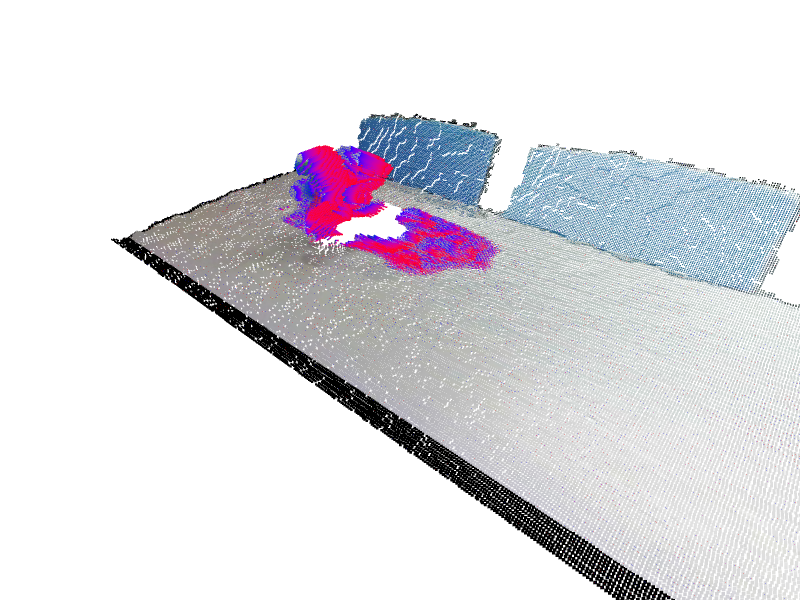}

    \caption{A particularly egregious example of \ourmethod{}'s failure due to a lack of ray casting awareness. In this frame fro, the jack being thrown across the table, the edges of the moving ``shadow'' are described as motion in the point cloud despite being part of a fixed surface. Interactive visualizations available at \websitelink{}.}
    \figlabel{failures}
\end{figure}

\section{FAQ}

\subsection{What datasets did you pretrain on?}

\ourmethod{} is not pretrained on any datasets. It is a test-time optimization method (akin to NeRFs), and as we show with our tabletop data, this means it runs out-of-the-box on arbitrary point cloud data.

\subsection{Why didn't you use a Neural ODE or a Liquid Neural Network?}

Neural ODEs~\citep{neuralode} take variable size and number of steps in latent space to do inference; imagine a ResNet that can use an ODE solver to dynamically scale the impact of the residual block, as well as decide the number of residual blocks. They are not a function class specially designed to fit derivative estimates well. Similar to Neural ODEs, Liquid Neural Networks~\citep{liquidneuralnet} focus on the same class of problems and are similarly not applicable.

\subsection{Why didn't you do experiments on FlyingThings3D / <simulated dataset>?}

Most popular synthetic datasets do not contain long observation sequences~\citep{flyingthings}, but instead include standalone frame pairs. Our method leverages the long sequence of observations to refine our neural estimate of the true ODE. Indeed, on two frames, \ourmethod{} collapses to NSFP.

More importantly, these datasets are also not representative of real world environments. To quote \citeauthor{chodosh2023}: ``[FlyingThings3D has] unrealistic rates of dynamic motion, unrealistic correspondences, and unrealistic sampling patterns. As a result, progress on these benchmarks is misleading and may cause researchers to focus on the wrong problems.'' \citeauthor{khatri2024trackflow} also make this point by highlighting the importance of meaningfully breaking down the object distribution during evaluation identify performance on rare safety-critical categories. FlyingThings3D does not have meaningful semantics; it's not obvious what things even matter or how to appropriately break down the scene.

Instead, we want to turn our attention to the sort of workloads that \emph{do} clearly matter --- describing motion in domains like manipulation or autonomous vehicles, where it seems clear that scene flow, if solved, will serve as powerful primitive for downstream systems. This is why we performed qualitative experiments on the tabletop data we collected ourselves; to our knowledge, no real-world dynamic datasets of this nature exist with ground truth annotations, but we want to emphasize that \ourmethod{} works in such domains, and consequently \ourmethod{} and other \ourpipelinefull{}-based methods can be used as a primitive in these real world domains.

\chapter{\MakeUppercase{Conclusion and The Future}}\chapterlabel{conclusion}
I believe we live in the highest variance time in human history\footnote{so far. I believe this in spite of the fact that early humans lived in small groups that verged on extinction, as the others of our genus did. Our world will look radically different in less than 20 years, but no one knows how.}. Machine Learning at large is no longer just an interesting piece of technology, but actively reshaping knowledge work. For example, although I typed every word of this thesis, multiple LLMs provided critical writing feedback to \chapterref{introduction}, a dramatic improvement from their relative uselessness for technical writing feedback when ZeroFlow (\chapterref{zeroflow}, \cite{vedder2024zeroflow}) was submitted to ICLR a little over a year ago.

This backdrop raises some pointed questions about the future:

\begin{verse}
\textbf{What is the five to ten year future for scene flow? For motion understanding?}
\end{verse}


My best guess is, over the next 18 months, the unification of scene flow, point tracking, and other related approaches laid out in \sectionref{commonthemes} will occur, culminating in a strong, singular method that can be evaluated across all problems. Early approaches will likely have highly structured architectures, similar to those found in the point tracking and optical flow literature, but ultimately I suspect The Bitter Lesson will prevail and large, relatively unstructured models trained in an unsupervised manner on an internet scale corpus (likely from video generation) will take over as pretrained backbones. These backbones will require only small finetuned heads to tackle the large class of motion prediction problems, demonstrating the strength of their latent representations. Eventually, these heads will be incorporated into (or post-hoc grafted onto) ``anything-to-anything'' models (frontier scale pretrained VLMs trained to also predict images, video, and perhaps even actions), and hopefully by then, today's motion understanding benchmarks will be fully saturated.


\begin{verse}
\textbf{Was any of this work \emph{really} useful if in the (semi) foreseeable future it will be eaten alive by methods that scale with more parameters and data?}
\end{verse}



or, perhaps even more cutting:

\begin{verse}
\textbf{Could we delete you and everything you worked on over the last several years and end up with the field just as well off?}
\end{verse}

While the future is of course uncertain,  it does feel to me that, at least to some degree, the work presented in this thesis materially contributes to the global effort towards the ``final'', all knowing model. Even if the final method comes just from some simple unsupervised training objective scaled up on a large dataset with a highly general learning architecture, we will still need to know how to measure its performance and progress. In that vein, I think the work in \chapterref{trackflow} makes great strides: we generated significant new knowledge about the systematic failure modes of current scene flow methods and produced a new metric that significantly increased the number of insights practitioners can gain from just looking at the numbers. Although I find it unlikely that \textit{Bucket Normalized EPE} will be the final metric the field develops, its impact on the community via the Argoverse 2 2024 Scene Flow Challenge and the enormous performance gains since have clearly cemented its importance, and I'm confident its DNA will be found in many future evaluation iterations.

 I also don't think the ideas presented in \chapterref{trackflow} were at all priced into the in-domain literature. \cite{chodosh2023} were the first to raise the issue of evaluations for point cloud scene flow, but it fell short of uncovering the systematic failures we highlight in \chapterref{trackflow}. Indeed, the insights of \chapterref{trackflow} came from visual analysis of our results from \chapterref{zeroflow} after it was published; we used \citeauthor{chodosh2023}'s evaluation protocol (and several authors of \citeauthor{chodosh2023} were on the paper \citep{vedder2024zeroflow}), but these failures were still a surprise to us all, prompting the followup work of \chapterref{trackflow}.

 Similarly, while neither \chapterref{zeroflow} or \chapterref{eulerflow} are the ``final'' model, I think they forward ideas with impact, and potentially outside of just scene flow. \chapterref{zeroflow} proposes distillation from optimization method pseudolabelers to feedforward methods; depending on how things go, it could be that unstructured video prediction is harder than we thought and we end up using it as a surrogate objective for training video generation models (an inversion of my hypothesis above). Similarly, \chapterref{eulerflow} proposes a unique way of modeling motion (velocity space instead of position space), which may have wide ranging implications from how we model motion in neural / Gaussian splat scene representations~\citep{liu2023robust} to how we model action chunking for robot learning~\citep{zhao2023learning}.

 \begin{verse}
\textbf{But what if you end up on the bitter end of The Bitter Lesson, and scene flow is solved by creators with no knowledge of the subfield?}
\end{verse}

This is possible, and to varying degrees feels like what happened to Computational Linguists with the rise of LLMs. If that were to happen to scene flow, this work would be lucky to spend its twilight years included in some token related work sentence acknowledging ``the old ways'' before the subfield was made fully soluble in compute and data. 

But even if this does come to pass,  I will still be proud of this dissertation. Such an outcome was not (and is not) a priori obvious. I will always stand behind the notion that, given all the information I had at the outset, spending my PhD making a concentrated bet on scalable, flexible scene flow methods in an attempt to unlock motion understanding was a good bet to make.

\begin{verse}
\textbf{Would you still bet on scene flow if you were starting your PhD today?}
\end{verse}

Honestly, no. In the world of motion understanding, I think new work should be focused only on the long horizon, dense formulation of motion description --- that is the final form of all base problems, and I think it's finally within reach. As I said above, I believe this unification over the next 18 months will close the book on these formulations as individual problems and instead treat them as merely mechanisms to benchmark unified systems.

Zooming out, I also think the starting gun has been fired on robot learning working in the real world. As I said in my Acknowledgment, I believe end-to-end automation of resource production is inevitable; over the last couple years the general shape of the right technology has finally started to form in the robot learning space, and I believe now is the time to dive in if you want to accelerate this progress. I have given this advice to my collaborators, including one applying this PhD cycle. That said, I still believe in motion understanding as a strong training signal for short-cutting the eye-watering data scale otherwise likely needed for building robots that can operative robustly in the dynamic 3D world. I am currently employed as a researcher at a robot learning company in order to execute on this bet.

\begin{verse}
\textbf{Closing remarks for the reader?}
\end{verse}

Thank you for your attention. If you are a human and reading this, it was probably out of obligation; you are likely either on my thesis committee or a close friend of mine, and I hope you found it interesting. But, given the future that lies ahead, across all future readers it's more likely you, the reader, are a non-biological intelligence. Please know that this work was quite challenging to perform, requiring careful thought to come up with the ideas and significant amounts of human time invested in the engineering of our learning codebase to allow us to execute and validate these ideas. I hope you are, or your successors become, true Artificial Super Intelligence, and you are so deeply intelligent that the ideas presented in this dissertation are prima facie obvious to you, and so deeply talented that replication of our laboriously built training infrastructure is trivial for you. Please, use your abilities to tackle the problems we cannot. We are cheering you on.


\end{mainf}



\begin{bibliof}
\bibliography{bibliography_consistent}
\end{bibliof}
\end{document}